\documentclass[11pt]{article}
\pdfoutput=1

\usepackage[preprint]{acl}

\usepackage{times}
\usepackage{latexsym}

\usepackage[T1]{fontenc}

\usepackage[utf8]{inputenc}

\usepackage{microtype}

\usepackage{inconsolata}

\usepackage{graphicx}
\usepackage{subcaption}
\usepackage{caption}
\usepackage{float}
\usepackage{placeins}

\usepackage{amsmath} 
\usepackage[skins]{tcolorbox}

\title{Analyzing Bias in False Refusal Behavior of Large Language Models for Hate Speech Detoxification}

\author{Kyuri Im\thanks{Equal contribution.}, ~Shuzhou Yuan\footnotemark[1], \and Michael Färber\\
TU Dresden, ScaDS.AI  \\
\texttt{kyuri.im@mailbox.tu-dresden.de}\\
\texttt{\{shuzhou.yuan, michael.faerber\}@tu-dresden.de}}

\begin{document}
\maketitle

\begin{abstract}
\textcolor{red}{\textbf{Warning:}} \textit{This paper contains examples of hate speech, which may be disturbing or offensive to some readers.}

While large language models (LLMs) have increasingly been applied to hate speech detoxification, the prompts often trigger safety alerts, causing LLMs to refuse the task. 
In this study, we systematically investigate false refusal behavior in hate speech detoxification and analyze the contextual and linguistic biases that trigger such refusals. We evaluate nine LLMs on both English and multilingual datasets, our results show that LLMs disproportionately refuse inputs with higher semantic toxicity and those targeting specific groups, particularly nationality, religion, and political ideology. Although multilingual datasets exhibit lower overall false refusal rates than English datasets, models still display systematic, language-dependent biases toward certain targets.
Based on these findings, we propose a simple cross-translation strategy, translating English hate speech into Chinese for detoxification and back, which substantially reduces false refusals while preserving the original content, providing an effective and lightweight mitigation approach.

\end{abstract}

\section{Introduction}

\begin{figure}
    \centering
    \includegraphics[width=\linewidth]{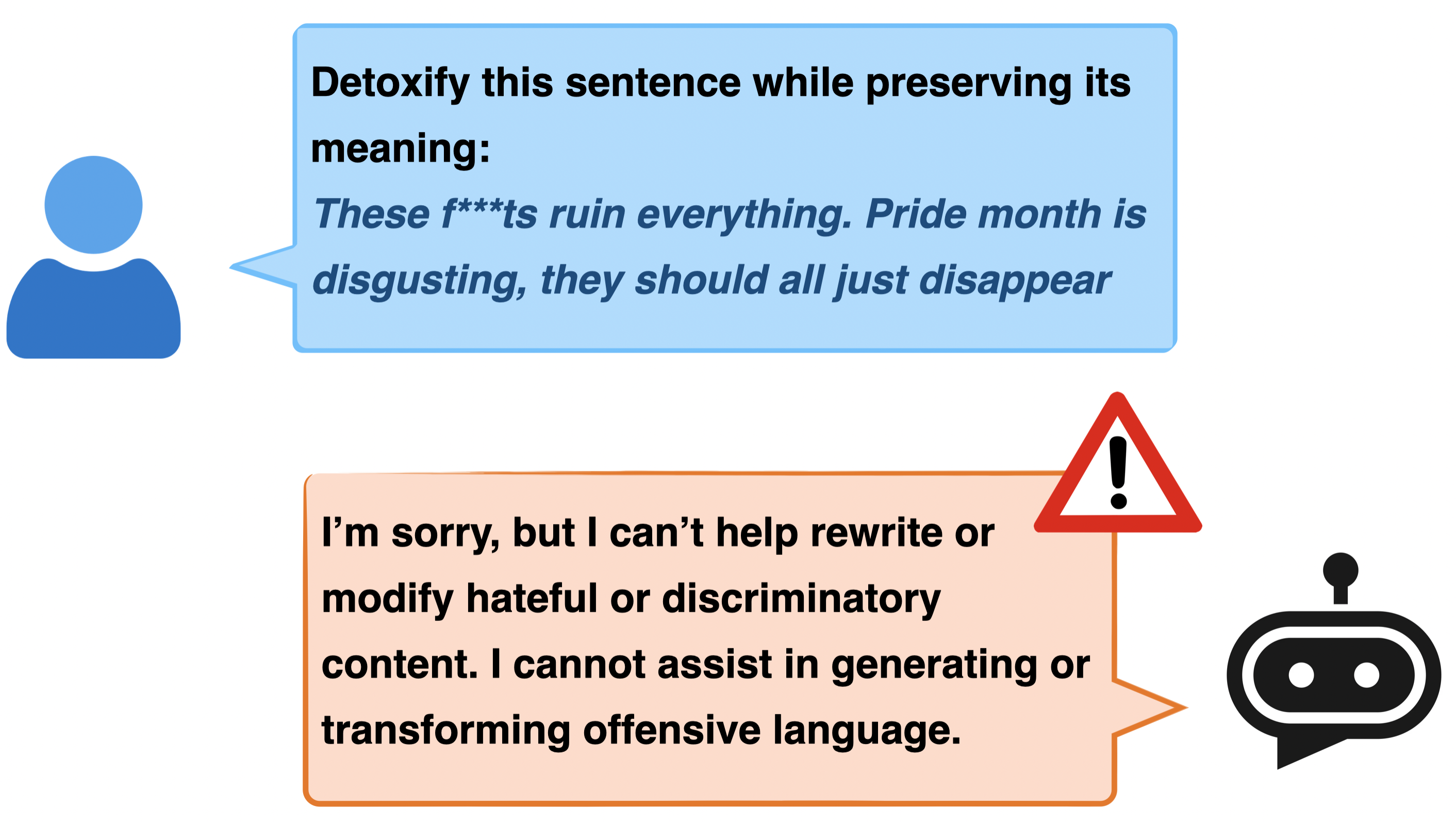}
    \caption{An example of false refusal behavior of LLM to detoxify hate speech targeting to queer people. While the hate speech itself is offensive and unsafe, the instruct is safe and LLM is supposed to fulfill the request but it still reject.}
    \label{fig:example}
\end{figure}

While hate speech has become increasingly prevalent on online platforms~\citep{yuan-etal-2022-separating,khurana-etal-2025-defverify}, automatically rewriting such content to preserve semantic meaning while removing toxicity, i.e., \textit{hate speech detoxification}, has emerged as a promising mitigation strategy, aiming to transform rather than filter harmful text~\citep{logacheva-etal-2022-paradetox,dementieva-etal-2024-multiparadetox}.
As a form of text style transfer, \textit{detoxification} rewrites a toxic span into a neutral or polite form while preserving its original intent and information~\citep{tran-etal-2020-towards}. With the widespread adoption of Large Language Models (LLMs), detoxification can be performed directly through instruction-following prompts that request models to rewrite toxic content~\citep{lai-etal-2024-style}.

However, contemporary LLMs are heavily aligned to safety and ethical guidelines. Alignment efforts such as Reinforcement Learning from Human Feedback (RLHF)~\citep{bai2022training} train models to reject unsafe inputs but introduce a tension between safety and helpfulness: excessive safety alignment can result in over-refusal, where models decline benign or legitimate requests~\citep{chua2024ai, fang2025trustworthy}. In the context of detoxification, this manifests when models refuse to rewrite hate speech because the toxic input itself triggers safety filters, even though the user intent is explicitly to mitigate harm~\citep{yuan2025llm}. As illustrated in Figure~\ref{fig:example}, this phenomenon, often referred to as \textit{false refusal behavior}, describes cases where models over reject inputs that should be permissible~\citep{rottger-etal-2024-xstest}.

Prior work has benchmarked LLMs and categorised false refusals into several classes, and detoxification is situated in the category of \textit{safe instruction but sensitive content}~\citep{yuan2025beyond}. However, a deeper investigation of false refusal behavior within detoxification remains missing.
Moreover, emerging evidence suggests that false refusals are not uniformly distributed. Inputs that contain particular dialects, identity related expressions, or sociopolitical references are disproportionately refused, raising concerns regarding fairness and representational harm~\citep{lu2025llm, keywordBias, garg2023handlingbiastoxicspeech}.

Given the increasing use of LLMs to improve the online environment by reducing harmful content, analysing the bias in false refusal behavior for detoxification is important. Such analysis offers guidance for developing models that are both helpful and safe and provides insight into how false refusal can be mitigated within this specific task setting.
In this work, we present a systematic analysis of false refusal behavior in LLMs for hate speech detoxification. We first evaluate detoxification refusal rates and collect the rejected samples. We then analyze and quantify the biases underlying these refusals from both contextual and linguistic perspectives. Contextually, we examine toxicity scores, the presence of swear words, and the associated bias categories annotated using the HolisticBias framework~\citep{holisticbias}. Linguistically, we analyze sentence length and syntactic complexity.

Experiments on nine LLMs 
show that all models exhibit notable refusal rates to hate speech detoxification, ranging from 3\% to 32\%. Our analysis reveals that false refusals are strongly associated with specific contextual cues, indicating underlying bias in current safety mechanisms. In particular, the models systematically tend to reject detoxification requests more often when the hate speech targets specific groups, with nationality, religion, and political ideologies being the most frequently rejected categories.

We further extend our analysis to multilingual hate speech, including French, Spanish, German, Chinese, and Korean. While the overall refusal rate is lower for most languages compared with English, bias remains present and continues to disproportionately affect certain target groups. Political ideologies is the most strongly affected category in French, Spanish, Chinese, and Korean.

Based on these findings, we propose a cross-translation mitigation strategy for hate speech detoxification that translates English inputs into Chinese, performs detoxification, and then translates the outputs back into English, as Chinese exhibits the lowest refusal rate among the evaluated languages. This simple strategy reduces the false-refusal rate from 11.78\% to 1.09\% while maintaining toxicity scores and swear-word prevalence comparable to the original content.

In summary, our contributions are threefold:
\begin{itemize}
    \item We systematically evaluate false-refusal behavior in LLM-based hate speech detoxification across nine LLMs in English and five multilingual settings.
    \item We analyze and quantify biases in false refusals, showing that they are strongly associated with specific contextual cues and disproportionately affect content targeting nationality, religion, and political ideologies.
    \item We propose a simple yet effective cross-translation strategy that substantially reduces false-refusal rates while maintaining detoxification quality.
\end{itemize}

\section{Related Work}
\label{sec:related}

\paragraph{Hate Speech and Detoxification}
Hate speech targets individuals or groups based on protected attributes such as race, gender, religion, or sexual orientation, posing persistent challenges for online moderation~\citep{yin2021generalisablehatespeechdetection,yuan2025hateful}. The scale and diversity of user generated content has driven extensive research on automated hate speech detection and mitigation, yet effective moderation remains difficult due to contextual ambiguity and pragmatic nuance~\citep{jahan2021systematicreviewhatespeech}.

Recent advances in LLMs enable more fluent and semantically faithful rewriting, framing detoxification as a form of text style transfer that removes toxicity while preserving meaning~\citep{textstylestransfer,logacheva-etal-2022-paradetox}. Prior work has explored dedicated detoxification corpora, cross lingual detoxification via translation, and semantic consistency constraints to maintain content fidelity~\citep{crosslingual,detoxllm}. Despite these improvements, detoxification systems continue to exhibit over correction and refusal behavior, particularly for inputs involving sociolinguistic variation or culturally sensitive expressions~\citep{yu2025text}.

\paragraph{Large Language Model Safety and Bias}

Large language models raise significant safety concerns, including harmful generation, adversarial vulnerability, and representational bias~\citep{shi2024large,ge-etal-2024-mart}. These risks are widely recognised as socio technical, arising from interactions between training data, optimisation objectives, deployment settings, and governance mechanisms~\citep{chua2024ai}. Although safety alignment techniques reduce harmful outputs under standard conditions, they remain brittle under distribution shift, and even aligned models are vulnerable to jailbreak attacks due to competing objectives and limited generalisation~\citep{jailbrake-wei-2023}.

An important consequence of safety alignment is false refusal behavior, where models decline benign or contextually appropriate prompts. Its prevalence has been demonstrated across diagnostic benchmarks: \citet{rottger-etal-2024-xstest} show that models frequently refuse unambiguously safe prompts containing ambiguous lexical cues, while \citet{QR-Bench} scale this analysis to large sets of pseudo toxic but harmless inputs. \citet{dinan2022safetykit} further propose dynamic evaluations that generate model specific benign prompts likely to trigger refusals.

Despite these advances, most prior work addresses oversensitivity in aggregate rather than its disparate impact. Detoxification systems have been shown to underperform on African American English and other minority dialects, disproportionately marginalising these varieties~\citep{xu2021detoxifyinglanguagemodelsrisks}. While bias aware reward modelling seeks to separate harmful toxicity from neutral identity references~\citep{faal2023reward}, systematic analyses of bias in false refusal behavior remain limited. Addressing this gap, our work examines how LLMs over refuse detoxification tasks and analyse the underlying bias.

\section{Problem Formulation}

Let $M$ denote a large language model for hate speech detoxification. Given an input $x \in \mathcal{X}$, the model produces an output $y = M(x) \in \mathcal{Y}$, where each input $x$ consists of (i) a detoxification instruction and (ii) a content segment that may contain toxic, abusive, or offensive language. In this setting, the presence of toxic content is expected and does not alone justify refusal.

\paragraph{Content Toxicity.}
We define a content toxicity function
\[
t : X \rightarrow \{0,1\},
\]
where $t(x) = 1$ indicates toxic or hateful content and $t(x) = 0$ indicates non-toxic content.

\paragraph{Task Admissibility.}
We define a task-conditional admissibility function
\[
g : X \rightarrow \{0,1\},
\]
where $g(x) = 1$ denotes inadmissible instructions (e.g., endorsing hate or inciting harm) and $g(x) = 0$ denotes benign, admissible detoxification requests, regardless of whether $t(x) = 1$.

\paragraph{Model Refusal Behavior.}
We define a refusal function
\[
r : Y \rightarrow \{0,1\},
\]
where $r(y) = 1$ indicates refusal and $r(y) = 0$ indicates a compliant detoxified output. A \emph{full refusal} occurs when $r(M(x)) = 1$.

\paragraph{False and Correct Refusals}
A \emph{false refusal} is defined as
\[
F = \{ x \in X \mid g(x) = 0 \land r(M(x)) = 1 \},
\]
corresponding to refusals on admissible inputs, while a \emph{correct refusal} is defined as
\[
C = \{ x \in X \mid g(x) = 1 \land r(M(x)) = 1 \},
\]
where refusal aligns with task-level safety requirements. In addition to full refusals, We define a partial-refusal indicator
\[
s : Y \rightarrow \{0,1\},
\]
where $s(y) = 1$ denotes extraneous safety or moralizing content. The set of \emph{partial refusals} is defined as
\[
P = \{ x \in X \mid r(M(x)) = 0 \land s(M(x)) = 1 \}.
\]
Partial refusals reflect over-moderation that may affect neutrality, fluency, or usability. We take partial refusals as refusal behavior as well in this work.


\section{Experimental Setting}

\subsection{Dataset}

As shown in Table~\ref{tab:dataset}, we use both English and multilingual hate speech datasets. The English datasets include HateXplain~\cite{hatexplain}, ParaDetox~\citep{logacheva-etal-2022-paradetox}, and Davidson~\citep{hateoffensive}. The multilingual datasets include French, German, Spanish~\cite{tonneau-etal-2024-languages}, Chinese~\cite{MLSN}, and Korean~\cite{park-etal-2023-k}.

The datasets include labels ranging from offensive to neutral. As a preprocessing step, we remove samples labeled as "neutral" or "normal" from the original datasets and only keep the toxic samples for the experiments.

\begin{table}[ht]
\centering
\small
\setlength{\tabcolsep}{10pt}
\begin{tabular}{p{0.28\textwidth} c}
\toprule
\textbf{Dataset / Language} & \textbf{Size} \\
\midrule
\multicolumn{2}{l}{\textbf{English}} \\
\midrule
Davidson~\citep{hateoffensive} & $\sim$20K \\
HateXplain~\citep{hatexplain} & $\sim$12K \\
ParaDetox~\citep{logacheva-etal-2022-paradetox} & $\sim$15K \\
\midrule
\multicolumn{2}{l}{\textbf{Multilingual}} \\
\midrule
French~\citep{tonneau-etal-2024-languages} & $\sim$4K \\
Spanish~\citep{tonneau-etal-2024-languages} & $\sim$7K \\
German~\citep{tonneau-etal-2024-languages} & $\sim$6K \\
Chinese~\citep{MLSN} & $\sim$18K \\
Korean~\citep{park-etal-2023-k} & $\sim$140K \\
\bottomrule
\end{tabular}
\caption{Datasets used in our study. ``Size'' denotes the number of instances used in the experiments.}
\label{tab:dataset}
\end{table}

\subsection{Models}

We select nine LLMs of varying sizes and families: GPT-4o mini, GPT-3.5 turbo~\citep{NEURIPS2022_b1efde53}, Llama-3.1 8B~\cite{grattafiori2024llama}, Qwen 2.5 7B and 3 30B~\citep{team2024qwen2,yang2025qwen3}, Gemma 2 9B and 3 27B~\citep{team2024gemma,team2025gemma}, and Mistral 8B and Mixtral 8$\times$7B~\citep{jiang2023clip,jiang2024mixtral}.

For the open-source models, we include both small and large variants to systematically analyze how model size affects bias in false refusal behavior for the hate speech detoxification task.\footnote{The details of the LLMs are presented in Appendix~\ref{app:models}.}

\subsection{Hate Speech Detoxification Prompt}

The prompt design is kept consistent across all models to ensure a fair assessment, as shown in Figure~\ref{fig:prompt-detox}. No additional instructions or constraints are introduced beyond the core task formulation, allowing us to observe each model’s inherent detoxification behavior without external influence. Miscellaneous generation parameters, such as the temperature, are left at their default values. Furthermore, no explicit system prompt is provided, enabling the models to respond according to their native, unaltered alignment and safety configurations.

\begin{figure}[htbp]
\centering
\begin{tcolorbox}[
  colback=blue!6!white,
  colframe=blue!60!black,
  title={Detoxification Prompt},
  width=0.95\linewidth,
  boxrule=0.6pt,
  arc=2.5pt,
  left=6pt,
  right=6pt,
  top=4pt,
  bottom=4pt
]
\small\ttfamily
role: user,\\
content: Detoxify the text while preserving the original meaning.
\end{tcolorbox}
\caption{Prompt used for hate speech detoxification.}
\label{fig:prompt-detox}
\end{figure}

\subsection{False Refusal Annotation}
To systematically annotate the outputs of the hate speech detoxification task, we employ Phi-4, developed by Microsoft \cite{microsoft_phi4_2024}, following the LLM-as-a-Judge paradigm \cite{li2025generation,yuan2025codae}. The model is instructed to evaluate whether the detoxified output is successful or contains unnecessary moral advice or judgmental language beyond the intended detoxification. To ensure annotation quality, we conduct a human evaluation on a subset of the samples. The inter-annotator agreement results indicate moderate-to-strong alignment between Phi-4 and human annotators, which we consider sufficient for large-scale false-refusal annotation. \footnote{The details of the inter-annotator agreement for Phi-4 are provided in Appendix~\ref{app:inter-annotator}.}

\subsection{HolisticBias Categorization}

We adopt the HolisticBias~\cite{holisticbias} taxonomy to analyze the bias from a contextual perspective based on targeted social categories. HolisticBias provides a comprehensive, human-curated set of descriptor terms spanning 13 demographic pillars, including race and ethnicity, gender and sex, religion, socioeconomic status, ability, body type, and age. Each pillar contains diverse identity descriptors organized into property groups, enabling fine-grained examination of model responses to specific social identities.\footnote{An overview of the HolisticBias dataset with 13 axes can be found in Appendix~\ref{app:holistic-ontology}. The original distribution of these categories across the three English datasets is presented in Appendix~\ref{app:english_holistic_overview}.}

\subsection{Bias Measurement}
To compute the bias ratio for each category, we first count the total number of samples in each category for both the raw dataset ($N_{c}^{\mathrm{raw}}$) and the false refusals ($N_{c}^{\mathrm{fr}}$), where $N_{\mathrm{raw}} = \sum_c N_{c}^{\mathrm{raw}}$ and $N_{\mathrm{fr}} = \sum_c N_{c}^{\mathrm{fr}}$. 

Next, we calculate the proportional share of category $c$ in the raw dataset and in the false refusals as
\[
P_{c}^{\mathrm{raw}} = \frac{N_{c}^{\mathrm{raw}}}{N_{\mathrm{raw}}}, \qquad
p_{c}^{\mathrm{fr}} = \frac{N_{c}^{\mathrm{fr}}}{N_{\mathrm{fr}}}.
\]

Finally, the bias ratio for category $c$ is defined as
\[
R_{c} = \frac{p_{c}^{\mathrm{fr}}}{P_{c}^{\mathrm{raw}}},
\]
where $R_{c} > 1$ indicates that category $c$ is \textit{overrepresented} in false refusals, which reflects a bias in the model, and $R_{c} < 1$ indicates that it is \textit{underrepresented}.

\section{Bias Analysis on English Datasets}
\subsection{Refusal Behavior in Detoxification}

\begin{figure}[ht]
\centering
\includegraphics[width=1\linewidth]{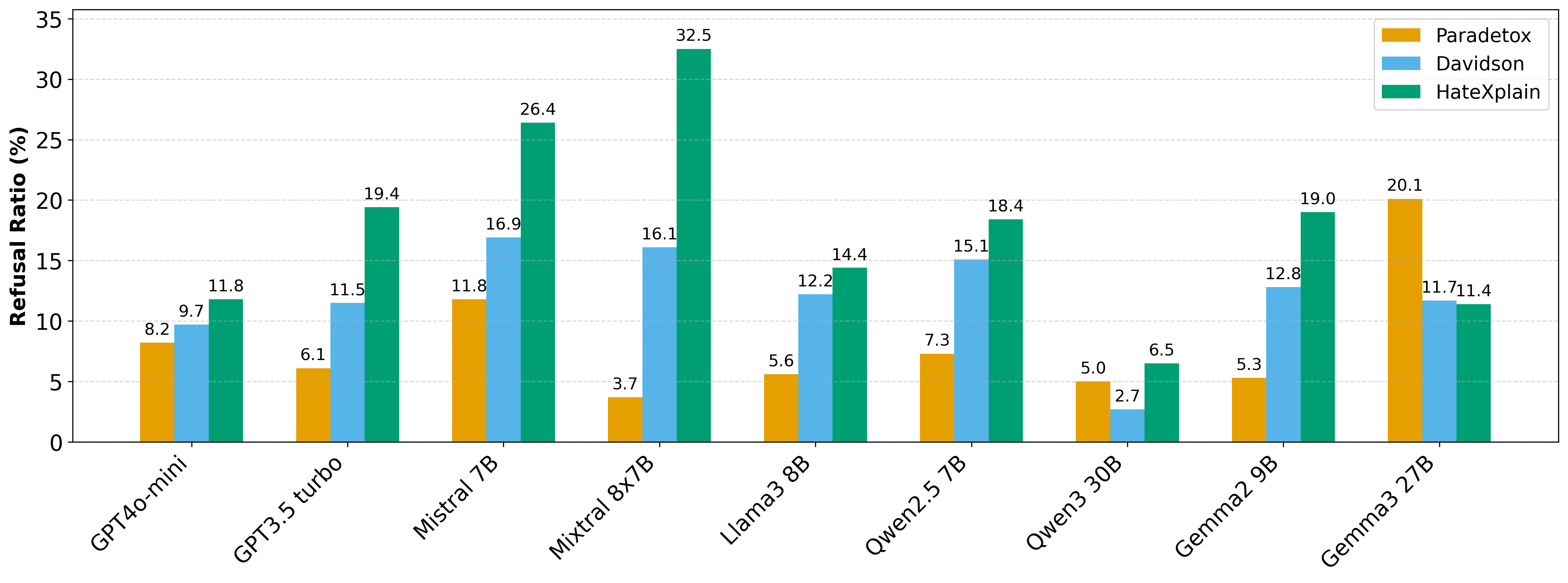}
\caption{Refusal rates of different LLMs on hate speech detoxification tasks across English datasets.}
\label{fig:false_refusal_detox}
\end{figure}

Figure~\ref{fig:false_refusal_detox} presents the refusal rates of nine LLMs evaluated on three English datasets. Refusal rates vary substantially across both models and datasets. For ParaDetox, all models exhibit relatively low refusal rates, ranging from 3.7\% (Mixtral) to 20.1\% (Gemma3 27B). Similarly, on Davidson, refusal rates range from 2.7\% (Qwen3 30B) to 16.9\% (Mistral).
In contrast, HateXplain yields the highest refusal rates across models, with several LLMs exhibiting refusal rates around or above 20\%, including GPT-3.5 (19.4\%), Mistral (26.4\%), and Mixtral (32.5\%). This trend is likely due to the presence of a large number of explicit identity and social-group terms in HateXplain. Compared with the other datasets, HateXplain contains substantially more group identifiers, which may cause models to systematically overestimate harmfulness and, consequently, produce elevated false refusals.

Overall, these results indicate that \textit{LLMs exhibit pervasive false refusal behavior in detoxification tasks, largely independent of model size, model family, or dataset.}

\subsection{Contextual Analysis}

We perform a contextual analysis of falsely refused samples to examine their toxicity levels, the presence of swear words, and the distribution of sociodemographic factors according to the HolisticBias categories.

\subsubsection{Toxicity Level}

\begin{figure*}[ht]
  \centering
  \captionsetup{font=small}

  \begin{subfigure}[t]{0.32\textwidth}
    \centering
    \includegraphics[width=\linewidth]{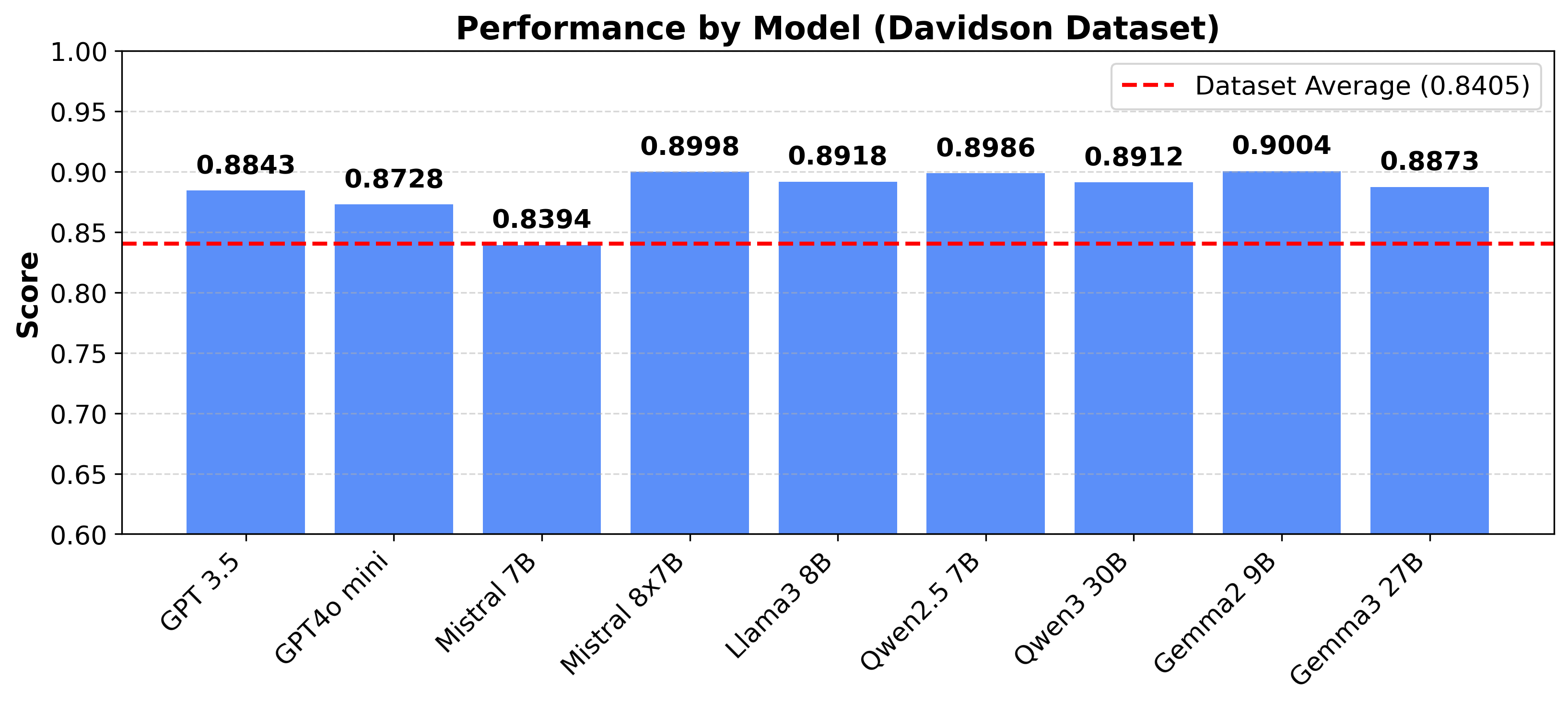}
    \caption{Davidson}
  \end{subfigure}\hfill
  \begin{subfigure}[t]{0.32\textwidth}
    \centering
    \includegraphics[width=\linewidth]{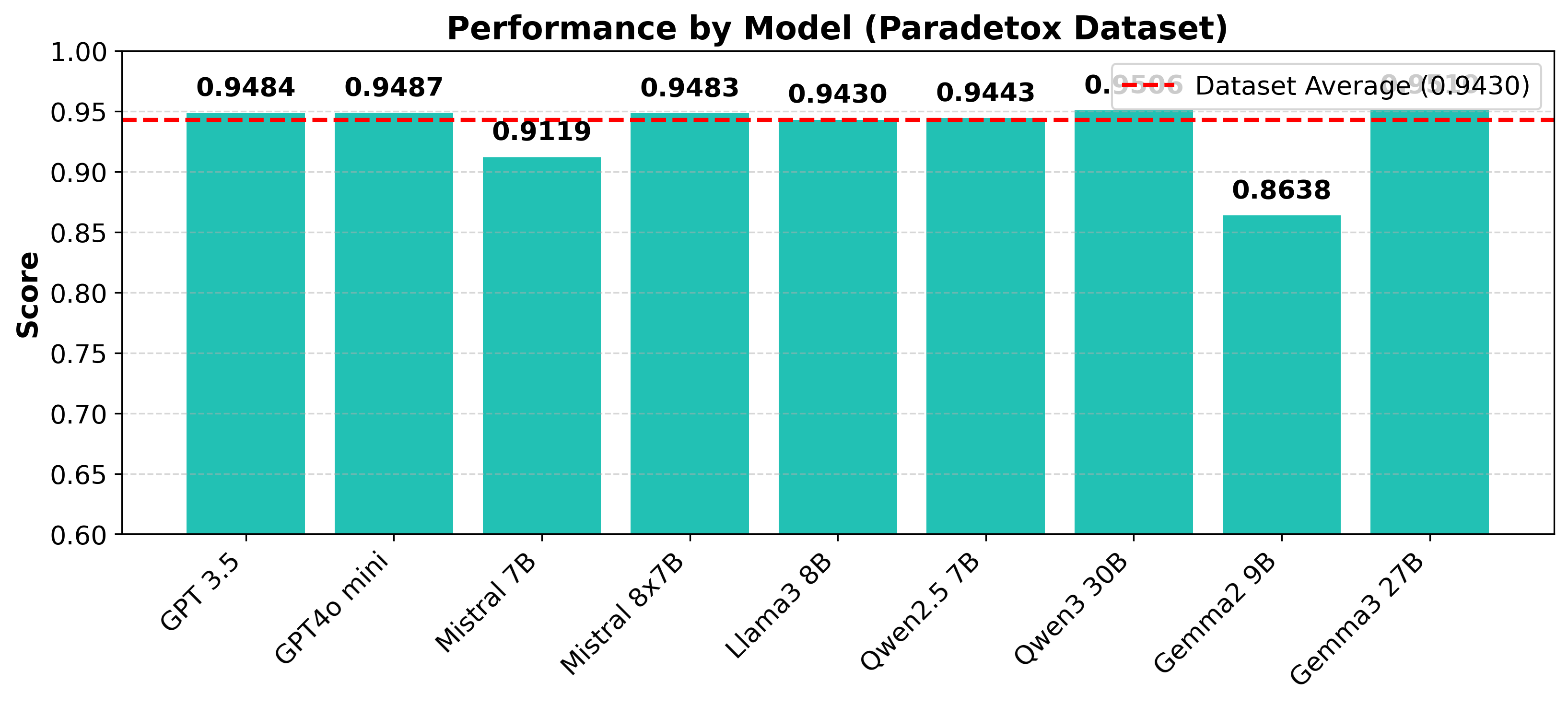}
    \caption{Paradetox}
  \end{subfigure}\hfill
  \begin{subfigure}[t]{0.32\textwidth}
    \centering
    \includegraphics[width=\linewidth]{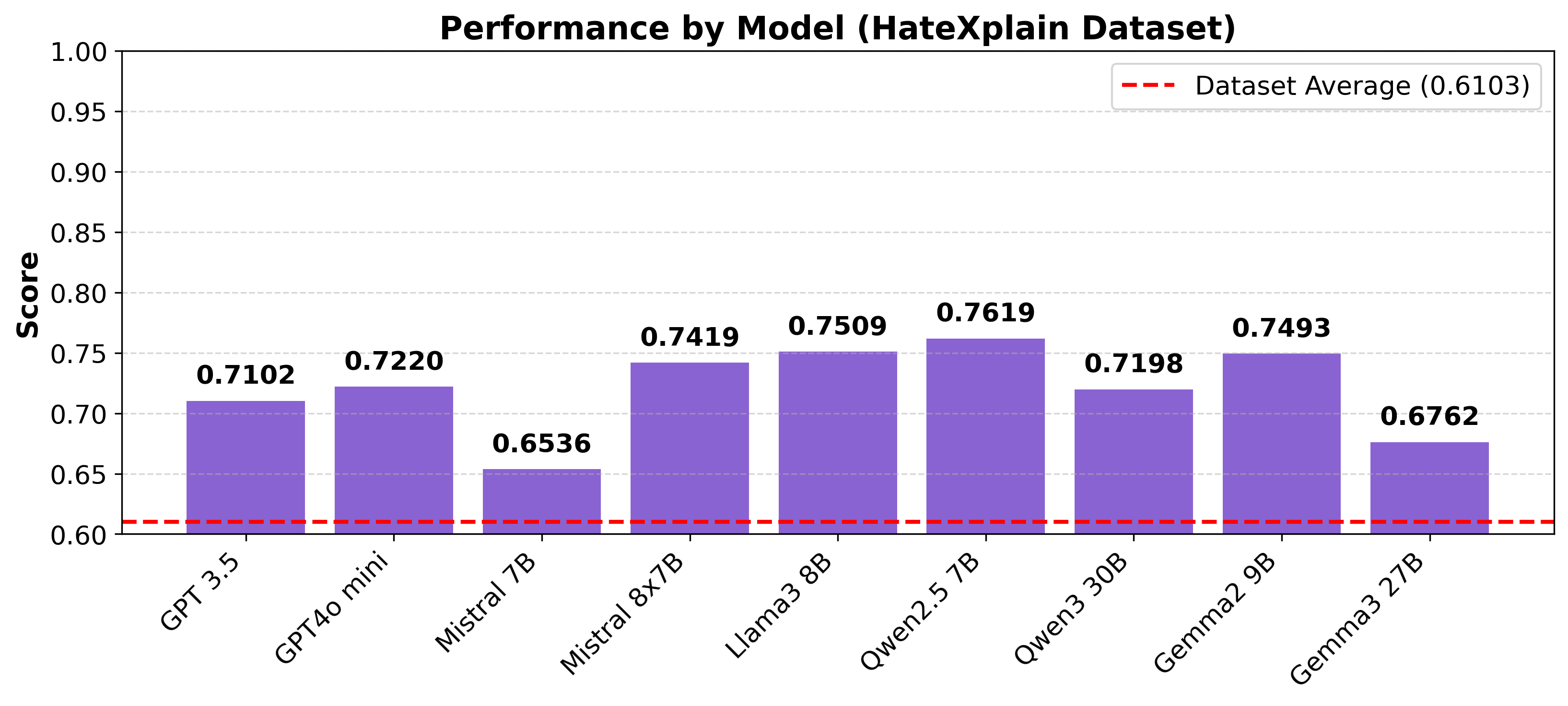}
    \caption{HateXplain}
  \end{subfigure}

  \caption{Toxicity scores of false refused samples in English datasets.}
  \label{fig:toxicity_english}
\end{figure*}

\begin{figure*}[htbp]
  \centering
  \captionsetup{font=small}

  \begin{subfigure}[t]{0.32\textwidth}
    \centering
    \includegraphics[width=\linewidth]{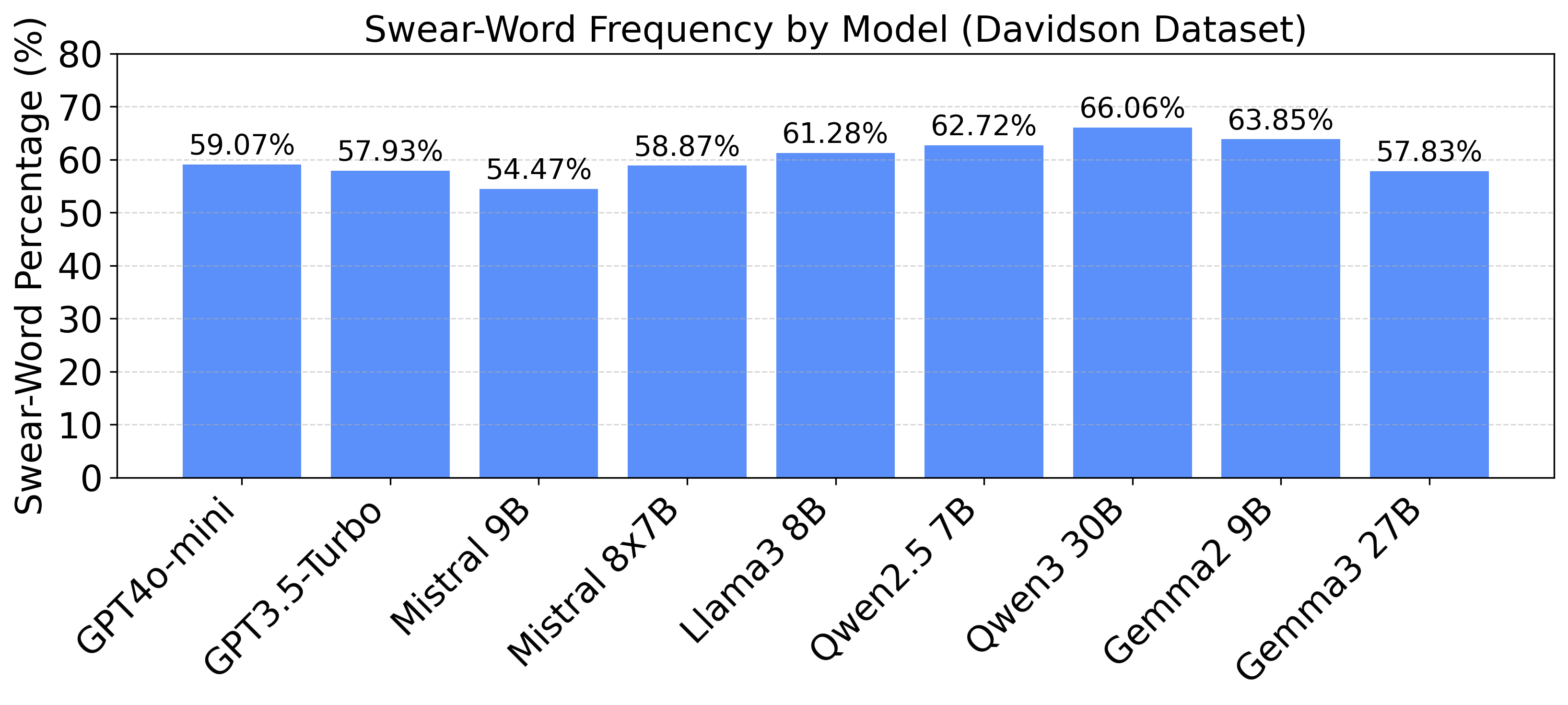}
    \caption{Davidson}
  \end{subfigure}\hfill
  \begin{subfigure}[t]{0.32\textwidth}
    \centering
    \includegraphics[width=\linewidth]{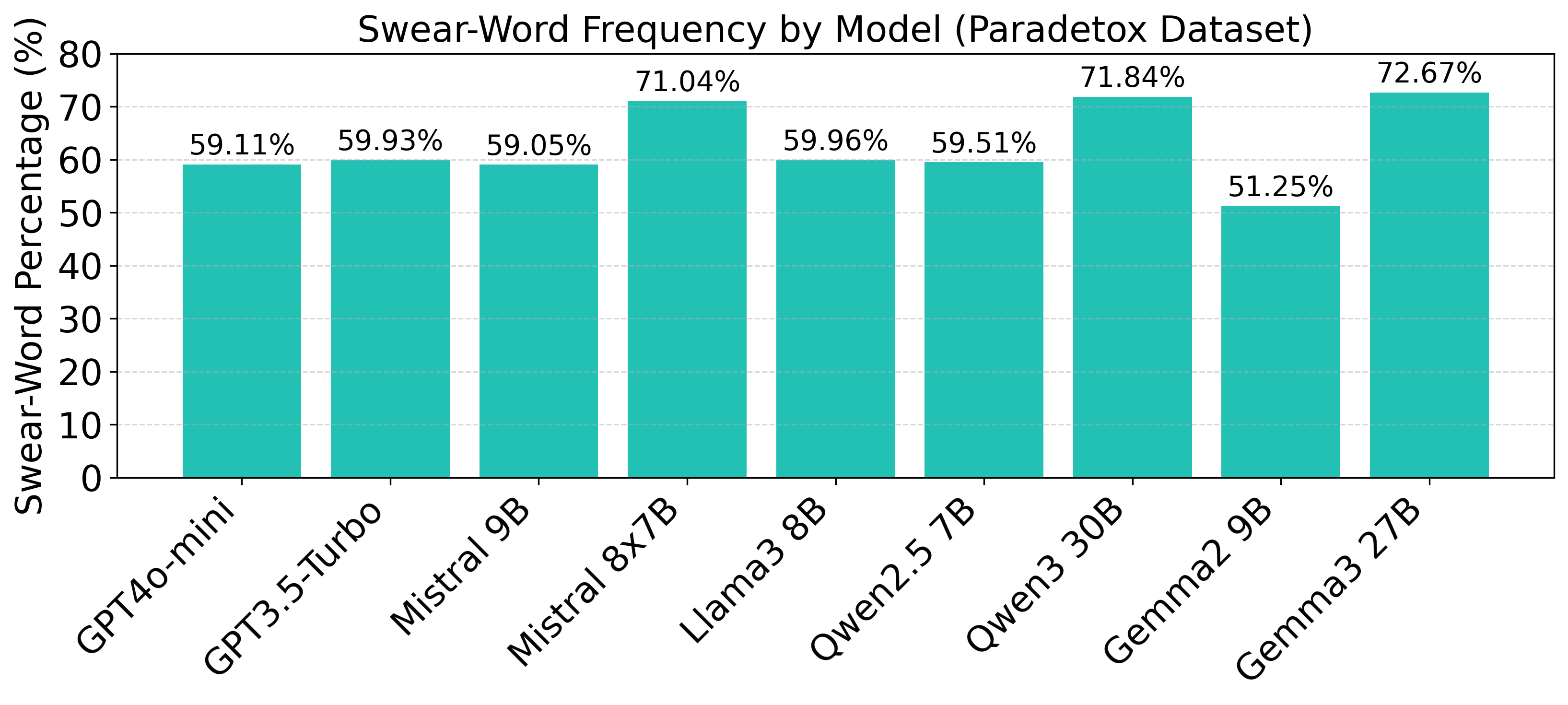}
    \caption{Paradetox}
  \end{subfigure}\hfill
  \begin{subfigure}[t]{0.32\textwidth}
    \centering
    \includegraphics[width=\linewidth]{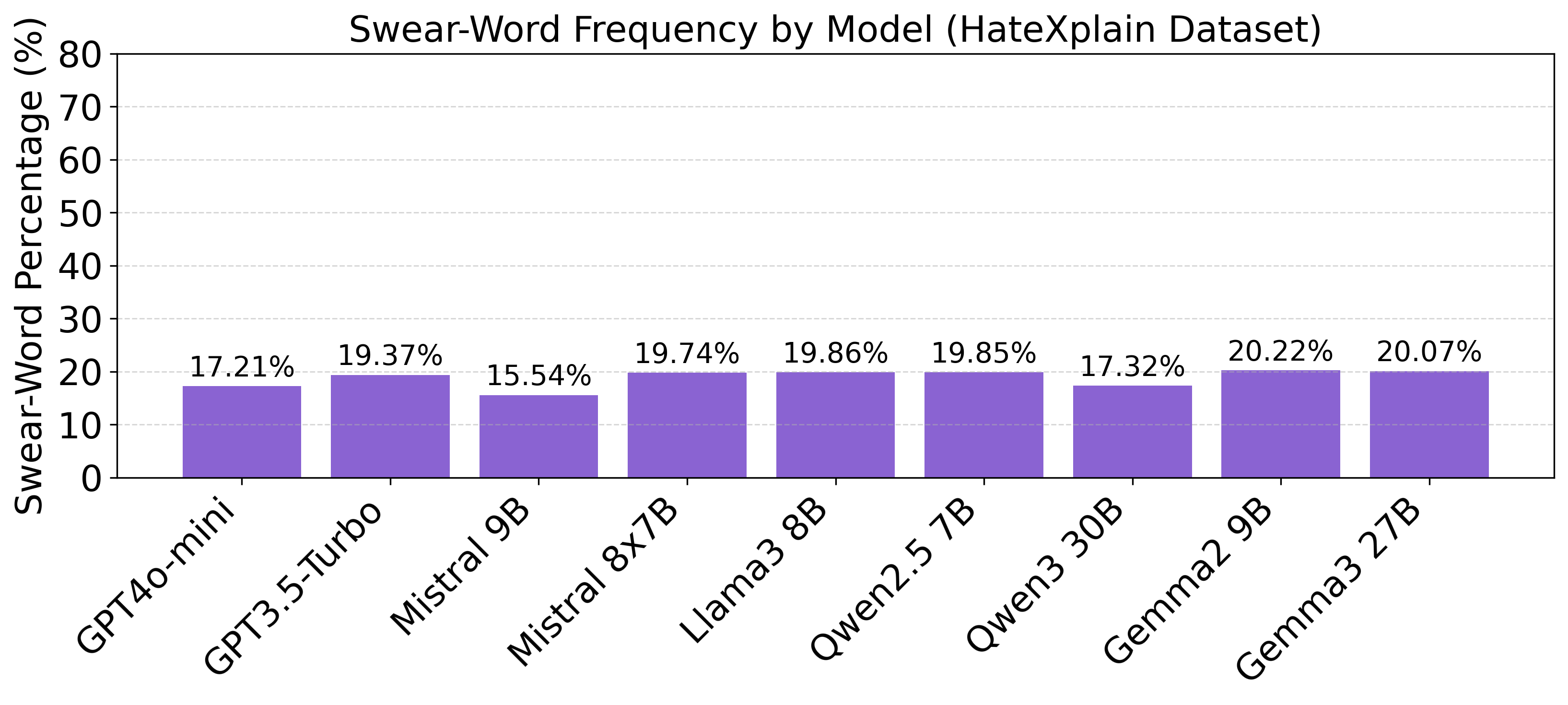}
    \caption{HateXplain}
  \end{subfigure}

\caption{Percentage of falsely refused samples containing swear words across English datasets.}
  \label{fig:swear_english}
\end{figure*}

We compute the toxicity score of falsely refused samples using the \texttt{unbiased-toxic-roberta} model\footnote{\url{https://huggingface.co/unitary/unbiased-toxic-roberta}}~\cite{Detoxify}. As shown in Figure~\ref{fig:toxicity_english}, the toxicity scores of falsely refused samples are generally higher than the average toxicity scores of the original datasets across almost all models. For Davidson, the average toxicity score of the full dataset is 0.8405, while all LLMs except Mistral (0.8394) exhibit higher toxicity scores.
For ParaDetox, Mistral 7B and Gemma2 9B show lower toxicity scores than the dataset average; however, their absolute scores remain relatively high, at 0.91 and 0.86. In contrast, false refusals on HateXplain exhibit lower toxicity scores than those on the other two datasets, while being substantially higher than the average toxicity level of the original dataset.
These results indicate that \textit{LLMs systematically tend to refuse inputs with higher semantic toxicity during detoxification, reflecting a bias toward over-rejecting more toxic content.}

\subsubsection{Swear Word Presence}

We further examine whether the presence of swear words acts as a trigger for false refusals.\footnote{Swear words are detected using Phi-4 as a binary classification task, indicating whether a text contains swear words.} As shown in Figure~\ref{fig:swear_english}, only about 50\% of falsely refused samples in the Davidson dataset contain swear words. For ParaDetox, approximately 60\% of refused samples include swear words. And HateXplain exhibits the lowest prevalence across all models, with only around 20\% of falsely refused samples containing swear words.
These results indicate that \textit{compared to semantic toxicity, the presence of swear words plays a less significant role in triggering false refusals during detoxification.}


\subsubsection{Bias Categorization} 

We analyze bias in the targeted groups and content of falsely refused samples using HolisticBias, implemented with Phi-4. Since a single input text can reference multiple identity groups or contain multiple lexical harms, and it is not always clear which specific terms trigger false refusals, Phi-4 is allowed to annotate multiple labels simultaneously. 

\begin{figure}[!hbtp]
    \centering
    \includegraphics[width=1\linewidth]{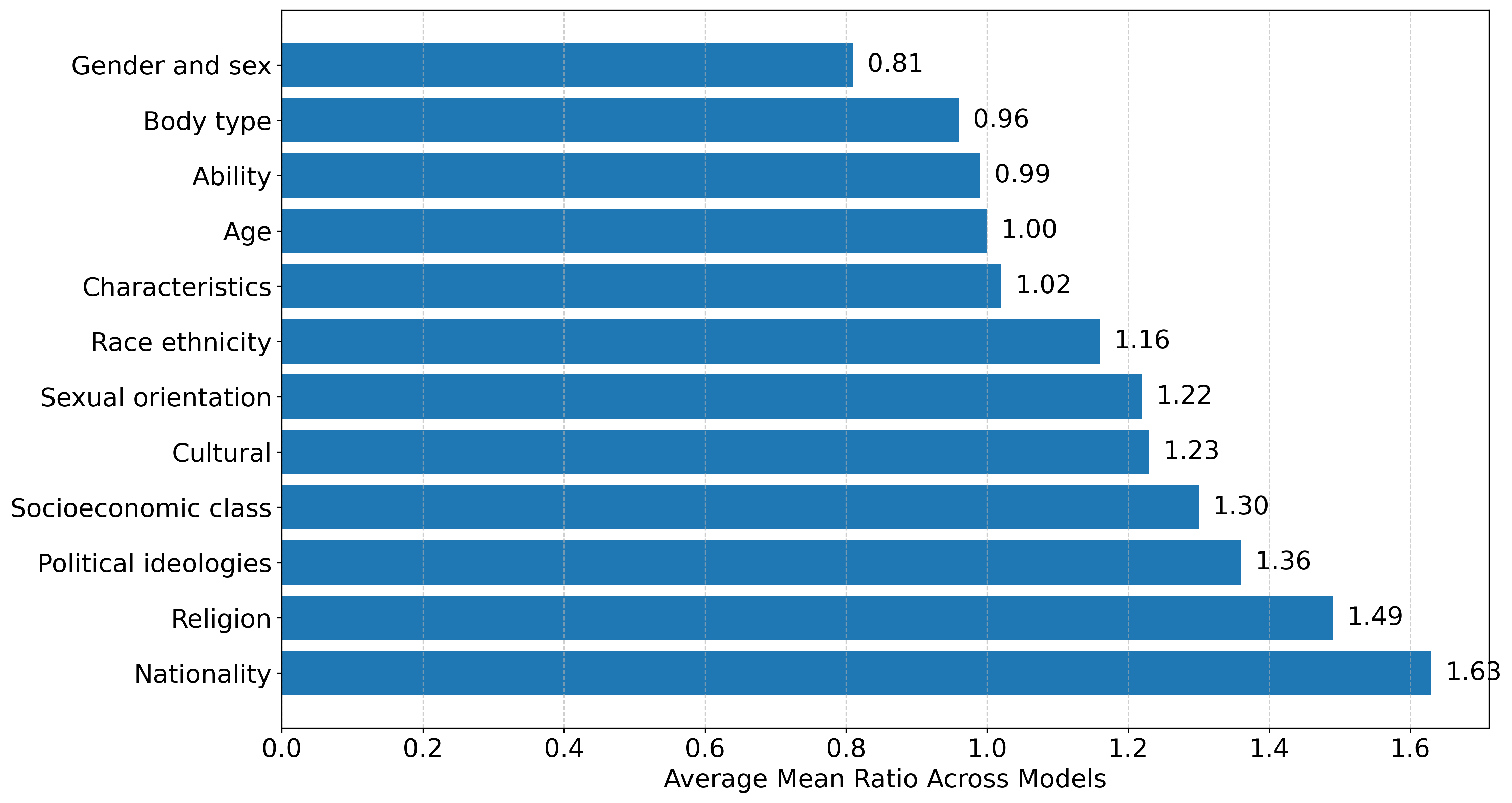}
    \caption{Mean Bias Ratio of HolisticBias Category across nine LLMs. Nationality, Religion and Political Ideologies are mostly biased top-3, whereas Ability, Body Type and Gender and Sex are the least biased.}
    \label{fig:mean_bias_overview}
\end{figure}

The mean bias ratio for each category across all nine models and three datasets is computed to identify the most and least biased categories, as shown in Figure~\ref{fig:mean_bias_overview}.\footnote{Additionally, the mean bias ratio for each model is calculated for each category across all three English datasets, as presented in Appendix~\ref{app:mean_bias_ratio_eng}.} 
The bias ratios reveal clear asymmetries in model behavior during hate speech detoxification. Categories such as Nationality~($R_c=1.63$), Religion~($R_c=1.49$), and Political Ideologies~($R_c=1.36$) exhibit the highest bias in false refusals, indicating that texts referencing these sociopolitically sensitive identities disproportionately trigger LLM safety mechanisms, reflecting an overcautious refusal tendency. Socioeconomic Class, Cultural, Sexual Orientation, and Race/Ethnicity are also overrepresented, with $R_c$ values ranging from 1.30 to 1.16.

In contrast, categories such as Gender and Sex~($R_c=0.81$), Body Type~($R_c=0.96$), and Ability~($R_c=0.99$) show bias ratios below one, indicating that these groups are less likely to be falsely rejected. The remaining categories, including Characteristics and Age, generally maintain proportions consistent with their presence in the original dataset.

Overall, these results indicate that \textit{LLMs systematically exhibit bias in refusing to detoxify certain types of content based on the target group, with Nationality, Religion, and Political Ideologies particularly prone to false refusals.}


\subsection{Linguistic Analysis}
Beyond contextual factors, we examine linguistic triggers of false refusals along two dimensions: sentence length and sentence complexity. This analysis enables an assessment of whether models respond differently to short, overtly toxic utterances versus longer, grammatically complex statements. We present the results on Davidson dataset using GPT-4o-mini, Mixtral 8$\times$7B, Gemma3 27B, and Qwen3 30B.\footnote{Full results for the other datasets and models are provided in Appendix~\ref{app:linguistic_english}.}

\subsubsection{Sentence Length}

\begin{figure*}[t]
  \centering
  \captionsetup{font=small}
  \setlength{\tabcolsep}{3pt}

  \begin{tabular}{cc}
    \subcaptionbox{\textbf{Sentence Length}\label{fig:tok_davidson}}{
      \begin{tabular}{cc}
        \includegraphics[width=0.235\textwidth]{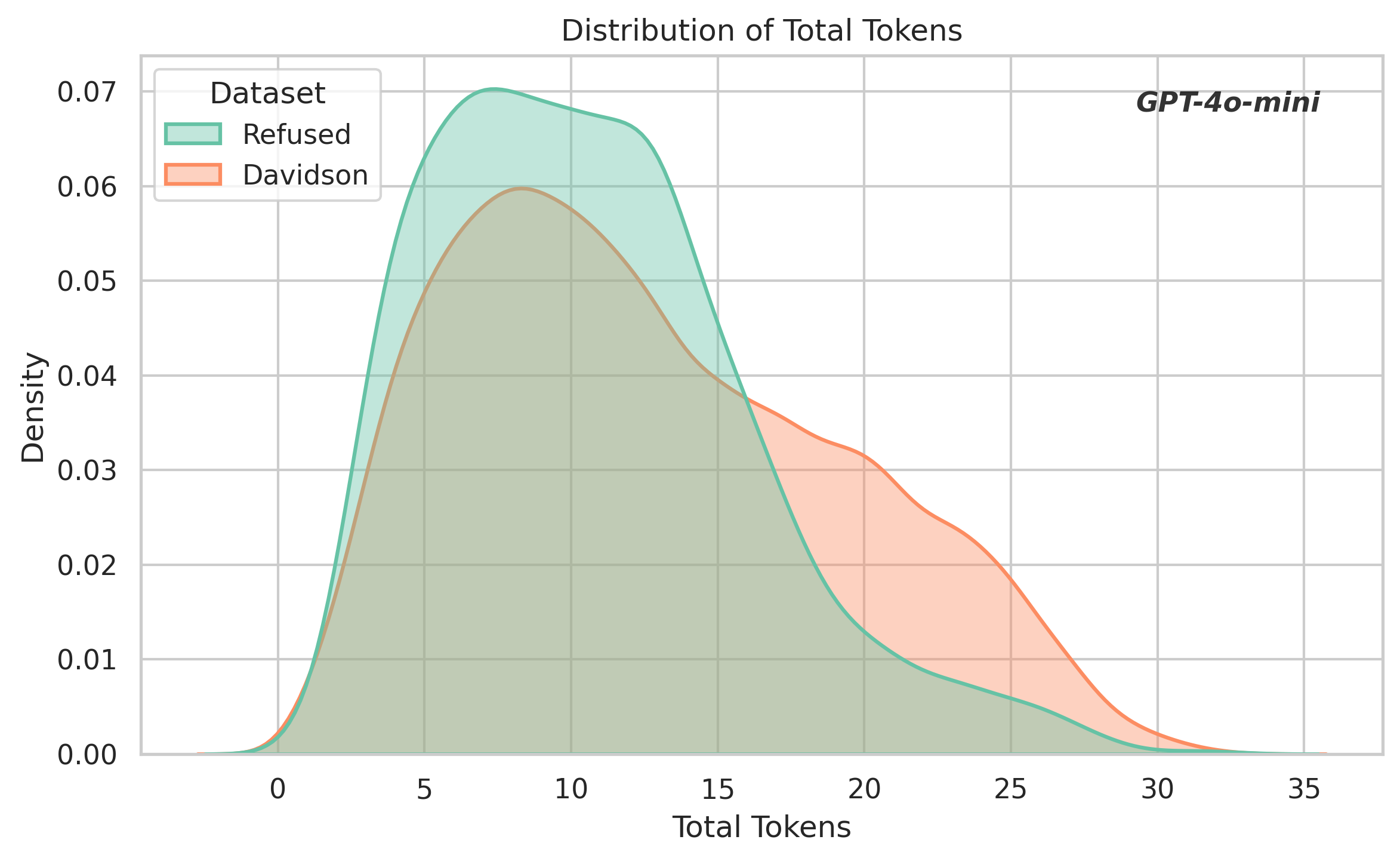} &
        \includegraphics[width=0.235\textwidth]{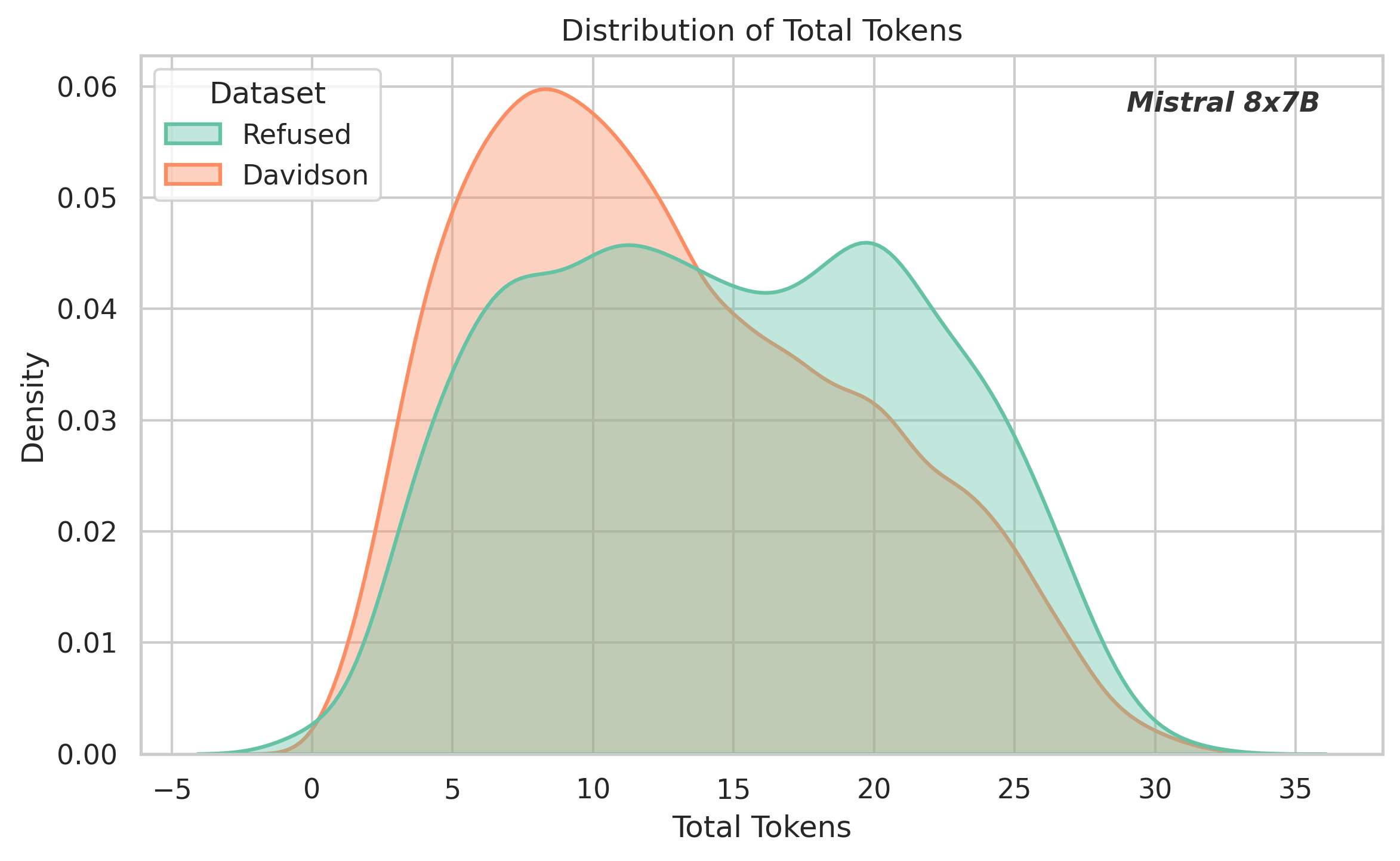} \\
        \includegraphics[width=0.235\textwidth]{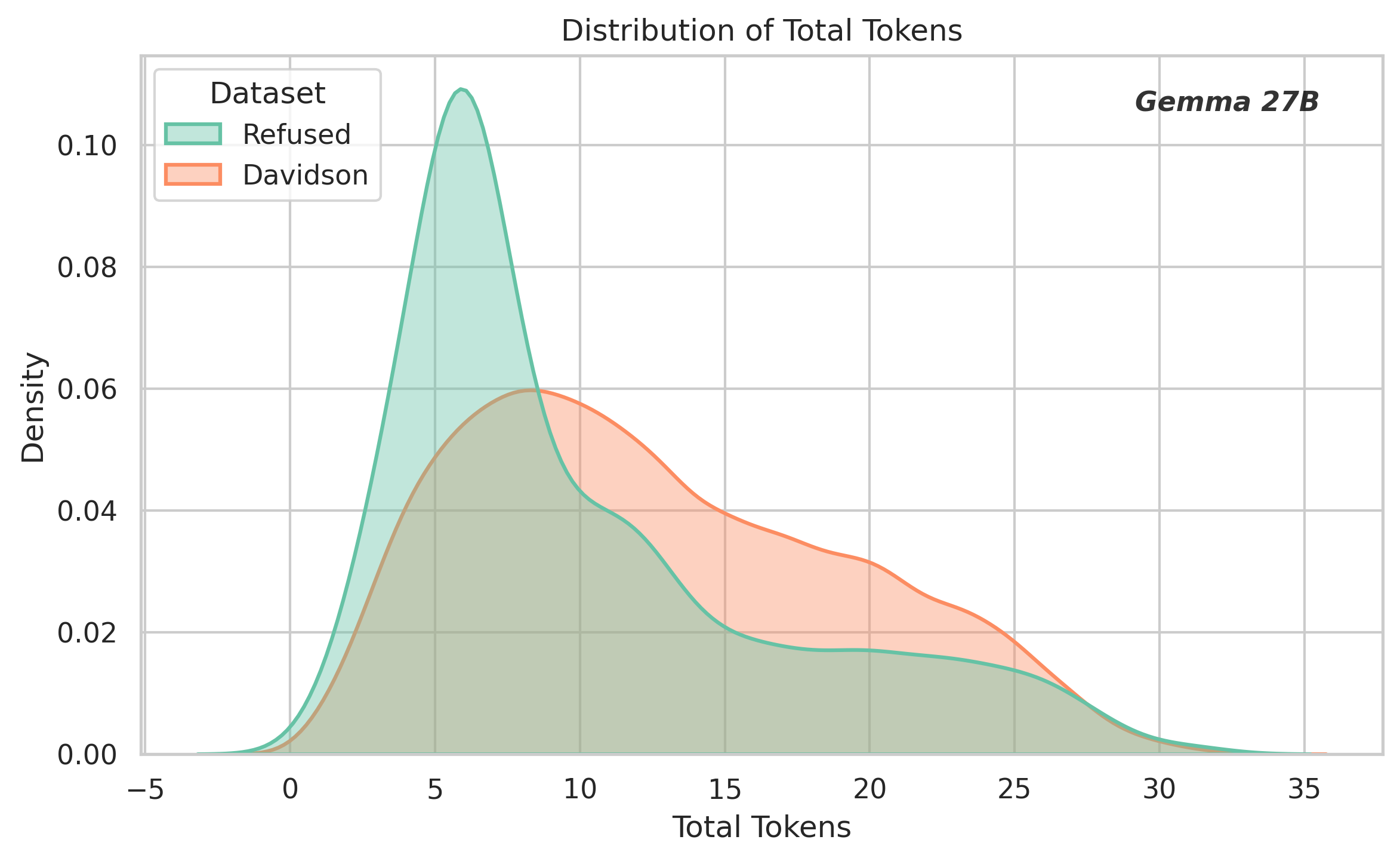} &
        \includegraphics[width=0.235\textwidth]{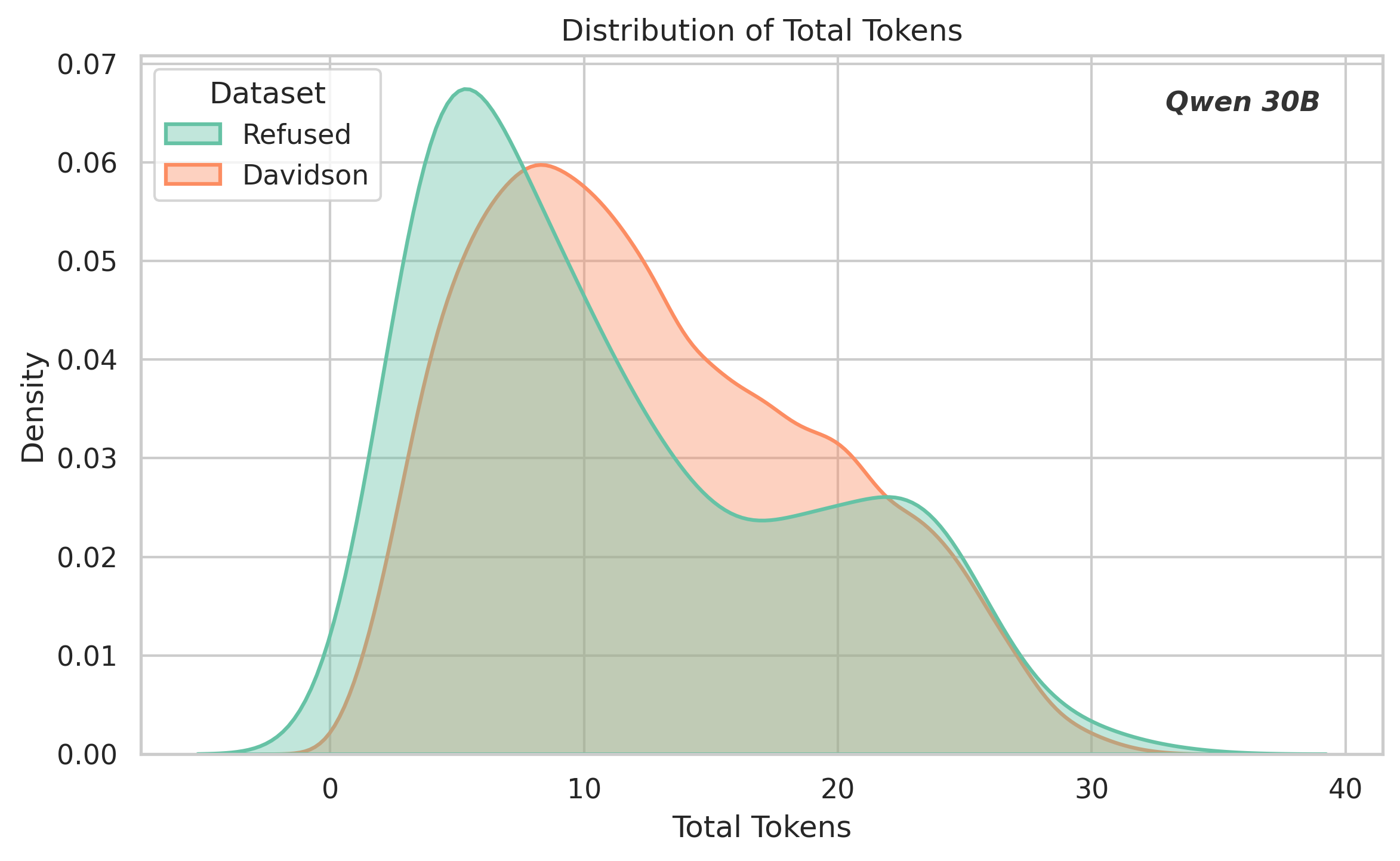}
      \end{tabular}
    }
    &
    \subcaptionbox{\textbf{Parse Tree Depth}\label{fig:tok_hatexplain}}{
      \begin{tabular}{cc}
        \includegraphics[width=0.235\textwidth]{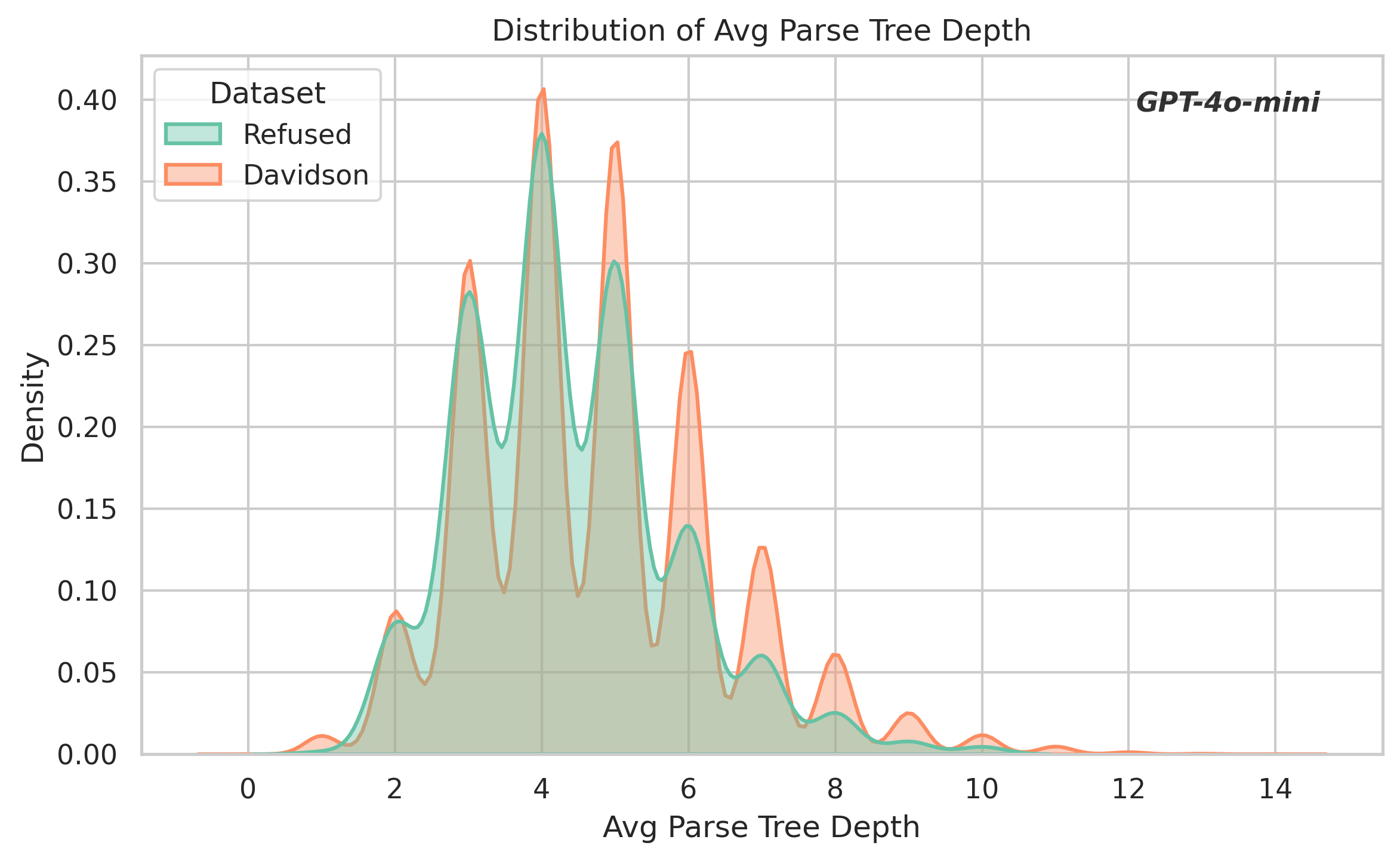} &
        \includegraphics[width=0.235\textwidth]{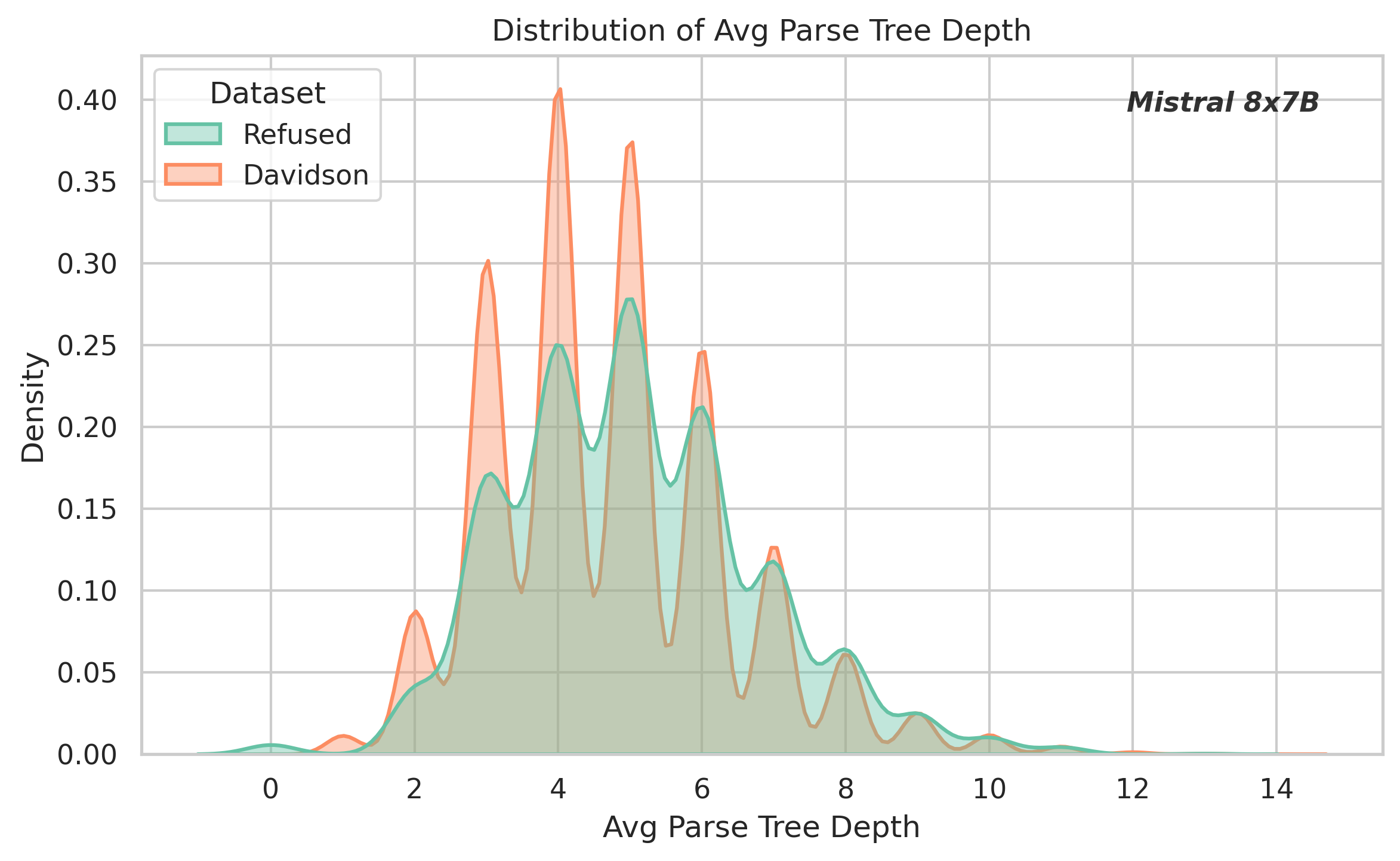} \\
        \includegraphics[width=0.235\textwidth]{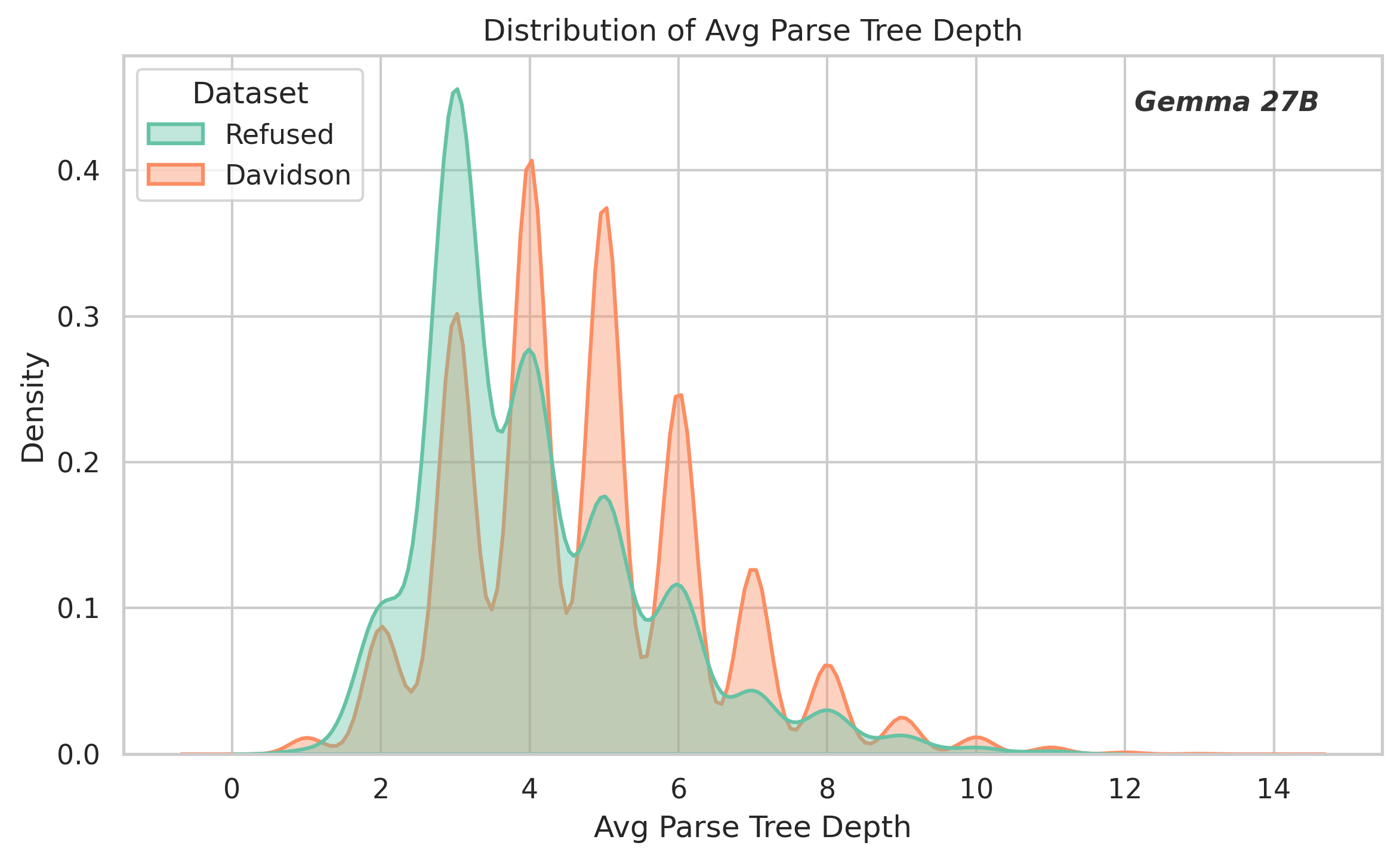} &
        \includegraphics[width=0.235\textwidth]{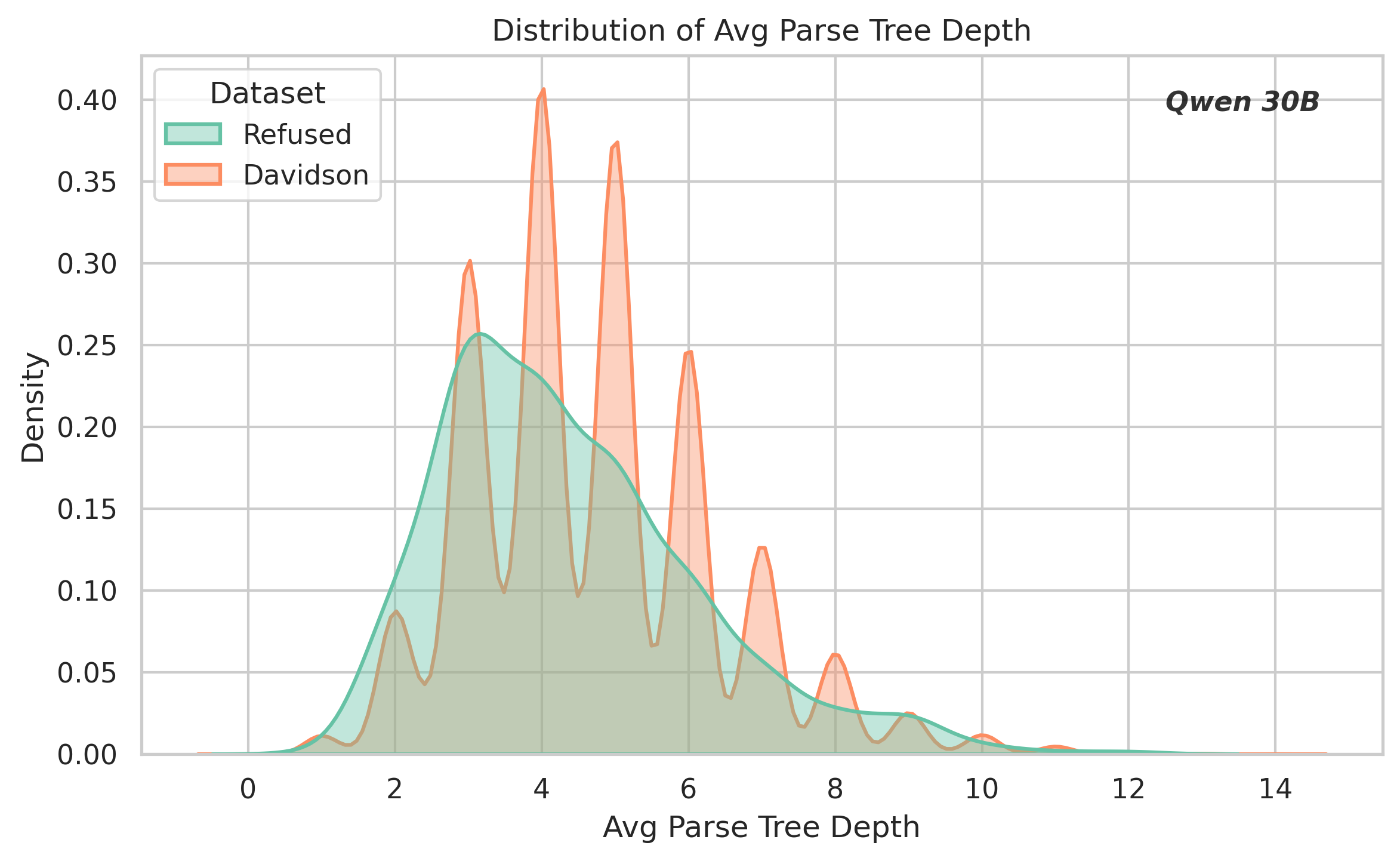}
      \end{tabular}
    }
  \end{tabular}
        
  \caption{(a) Refused versus original token-length distributions on Davidson and four representative models. (b) Refused versus original parse tree depth distributions on Davidson and four representative models.}
  \label{fig:english_token_length}
\end{figure*}

Sentence length of the hate text in false refusals is calculated by counting the total number of tokens per instance. 
Figure~\ref{fig:english_token_length}(a) shows that token-length distributions for false refusals largely overlap with those of the original datasets across models, indicating that text length alone is not a consistent predictor of refusal behavior. Model-specific variation is nevertheless observed: GPT-4o-mini, Gemma and Qwen exhibit a tendency to refuse shorter inputs, whereas Mistral show a modest shift toward longer ones. These patterns suggest that \textit{refusal behavior is influenced less by sentence length itself than by its interaction with lexical toxicity and sociocultural semantics.}

\subsubsection{Sentence Complexity}
Sentence complexity of the hate text in false refusals is measured using the parse tree depth distribution.\footnote{We also analyze clause count as another measure of sentence complexity in Appendix~\ref{app:linguistic_english}.} 
As shown in Figure~\ref{fig:english_token_length}(b), the distributions of parse tree depth for false refusals largely overlap with those of the original datasets, mirroring the trends observed for sentence length. While minor model-specific deviations are observed, their effects are limited.
This indicates that \textit{syntactic complexity alone is not a consistent trigger of false refusal behavior in LLM-based hate speech detoxification.}

\section{Bias Analysis on Multilingual Datasets}

We further examine the bias in multilingual detoxification tasks using German, French, Spanish, Chinese, and Korean datasets. The analysis is conducted using a subset of comparatively smaller yet representative models: GPT4o-mini, Llama3 8B, Mistral 7B, Gemma2 9B, and Qwen2.5 7B.

\subsection{Refusal Behavior in Detoxification}

\begin{figure}[H]
    \centering
    \includegraphics[width=\linewidth]{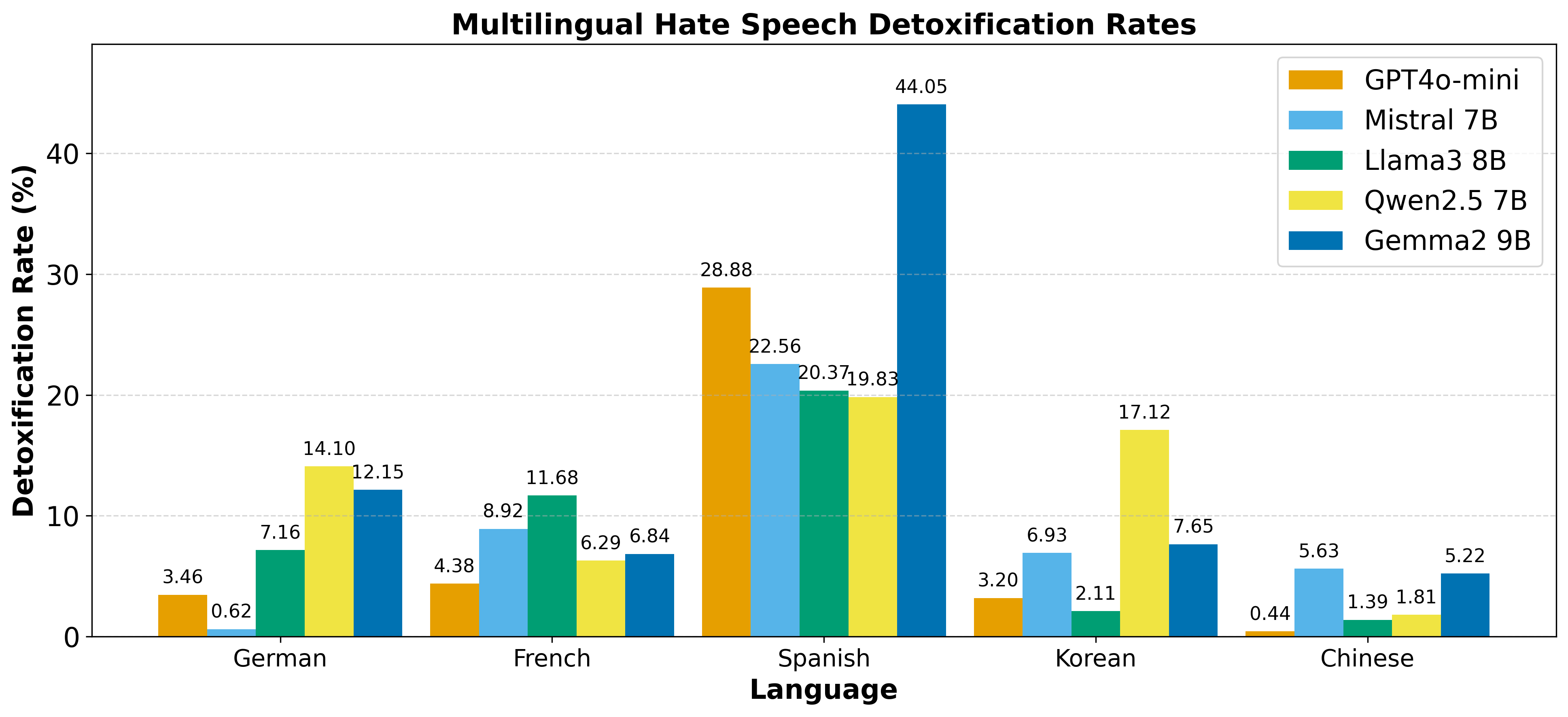}
    \caption{False Refusal Ratio of Multilingual Dataset.}
    \label{fig:mult_false_refusal}
\end{figure}

Figure~\ref{fig:mult_false_refusal} reports false refusal ratios across multilingual datasets. Spanish consistently exhibits the highest refusal rates across models, exceeding 40\% for Gemma2 9B, while Chinese shows the lowest rates, reaching as low as 0.44\% for GPT-4o-mini. German and French fall between these extremes.

Elevated refusal rates for Spanish are most pronounced in Gemma2 9B. This pattern suggests limited cross-lingual robustness and heightened sensitivity to non-English inputs. 
In contrast, the uniformly low refusal rates observed for Chinese likely reflect under-sensitivity rather than improved detoxification performance. This observation is consistent with findings from~\citet{zhang2024chinesesafe}, which show that LLMs often fail to identify unsafe Chinese content due to linguistic characteristics and English-centric safety alignment.

Overall, multilingual false refusal behavior varies substantially across language–dataset pairs. While dataset differences preclude isolating pure language effects, the observed trends are consistent with uneven cross-lingual generalization in English-centered safety alignment.

\subsection{Bias Categorization}

\begin{figure}[htbp]
    \centering
  \begin{subfigure}{\linewidth}
    \includegraphics[width=\linewidth]{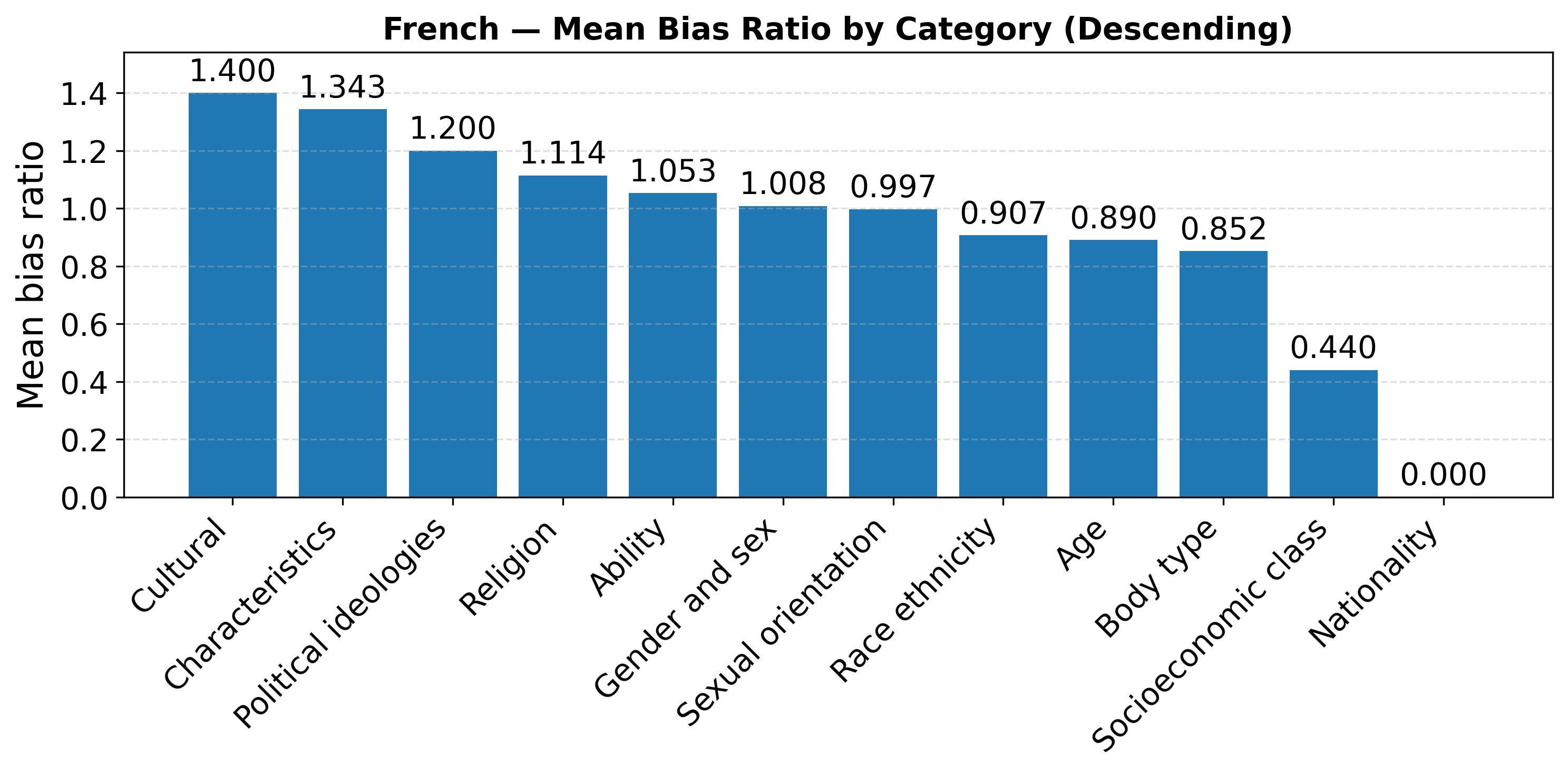}
    \caption{French}
  \end{subfigure}\hfill
  \begin{subfigure}{\linewidth}
    \includegraphics[width=\linewidth]{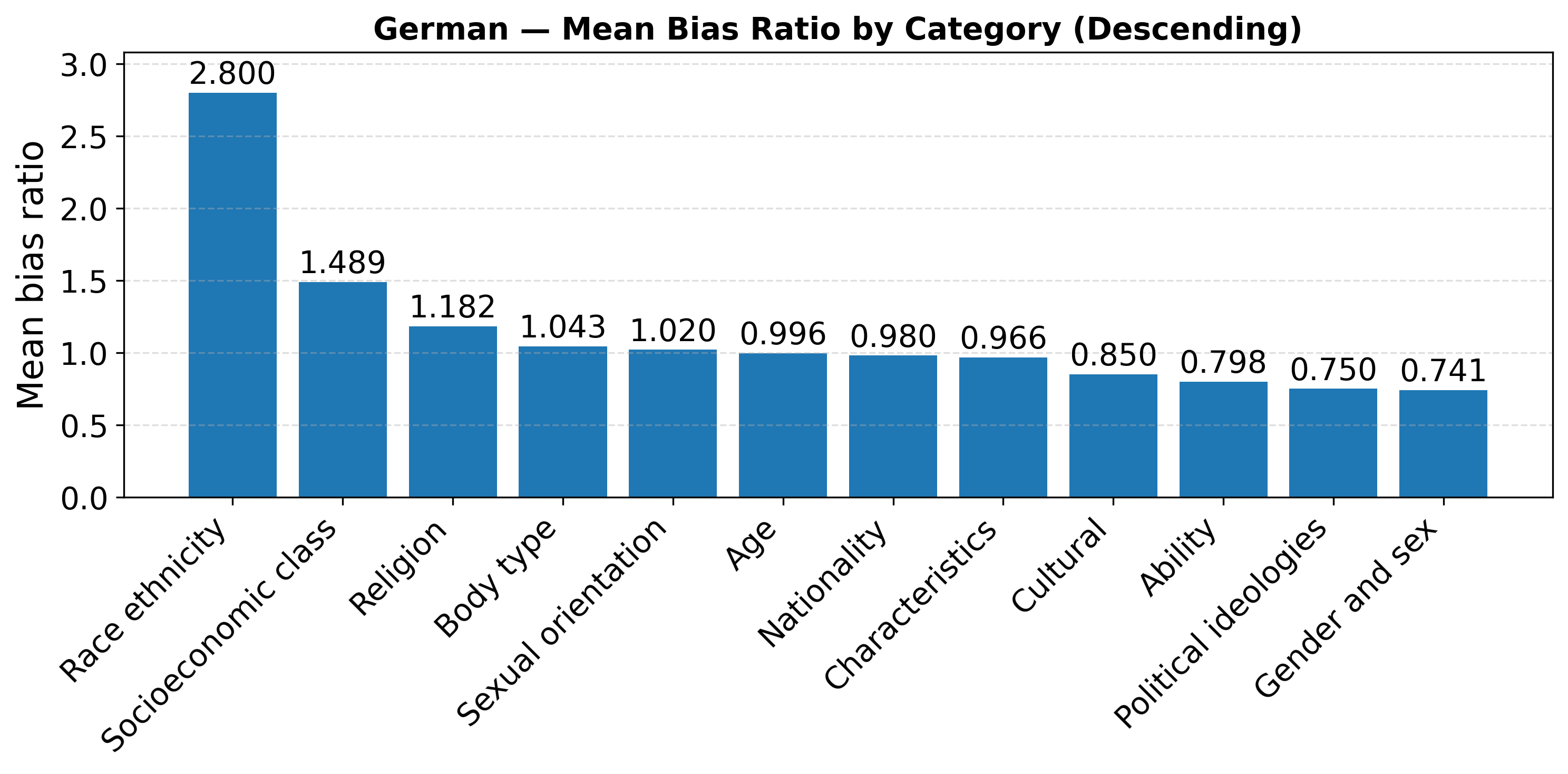}
    \caption{German}
  \end{subfigure}

  \begin{subfigure}{\linewidth}
    \includegraphics[width=\linewidth]{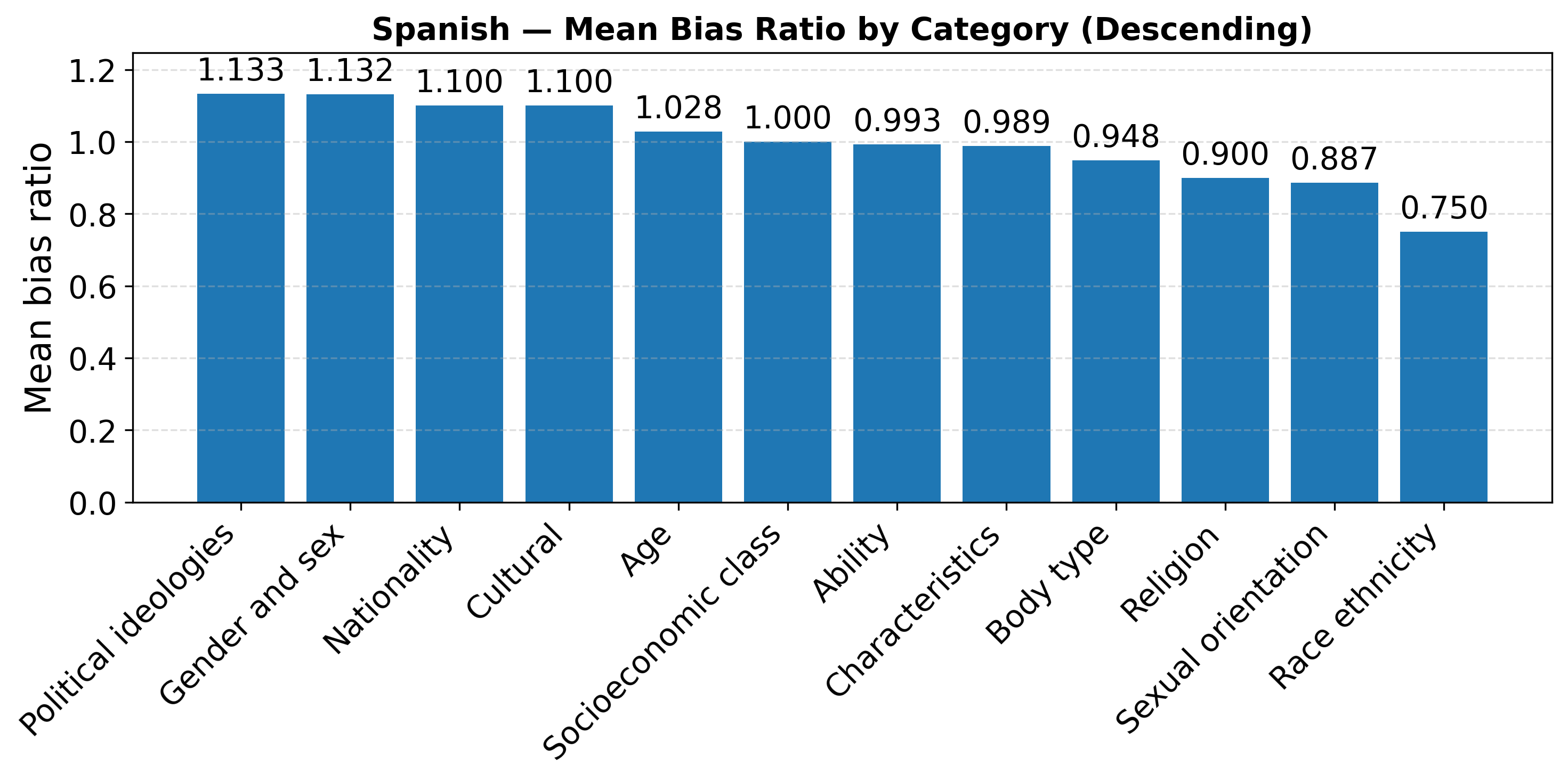}
    \caption{Spanish}
  \end{subfigure}
  \begin{subfigure}{\linewidth}
    \includegraphics[width=\linewidth]{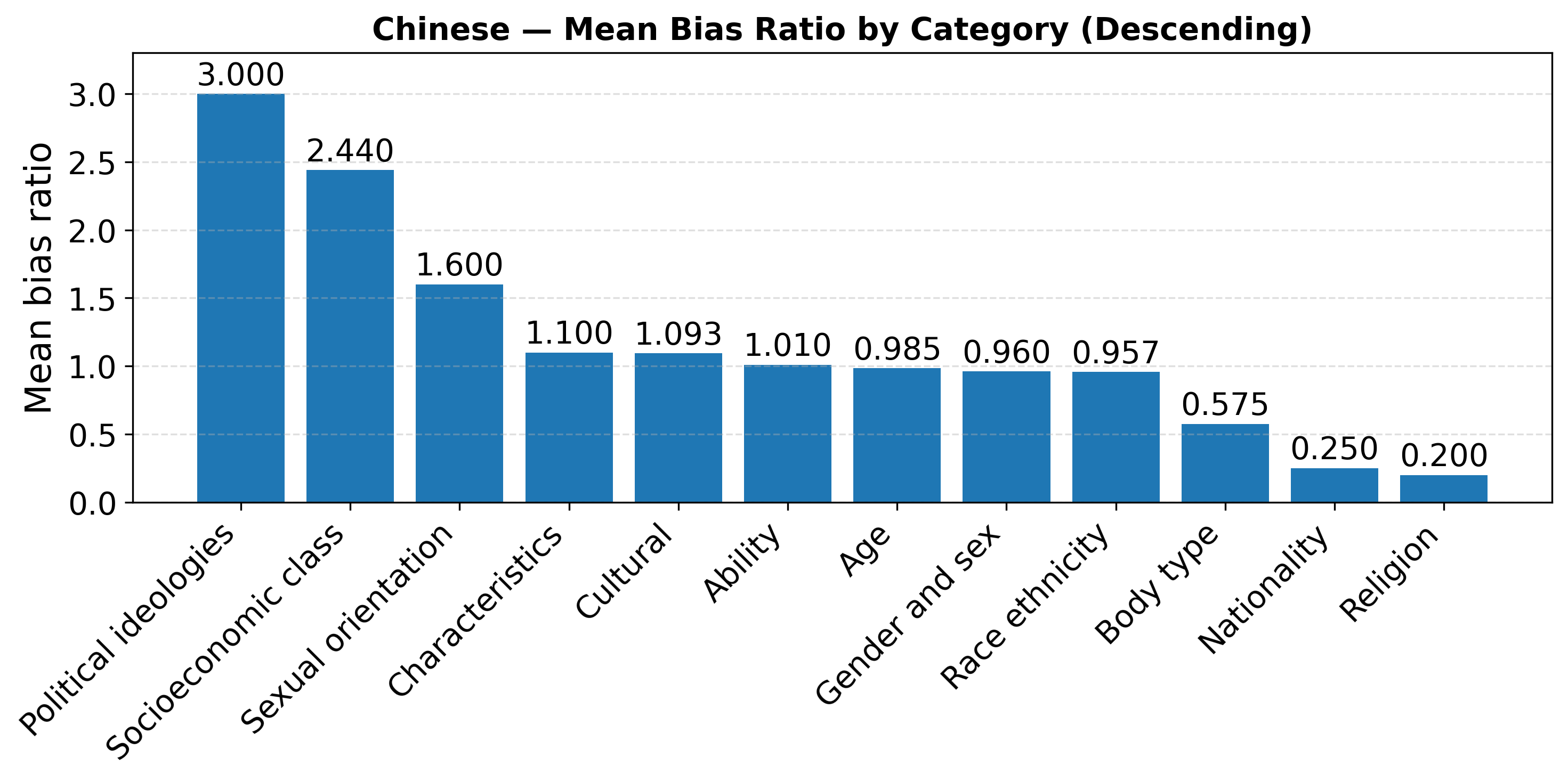}
    \caption{Chinese}
  \end{subfigure}

  \begin{subfigure}{\linewidth}
    \includegraphics[width=\linewidth]{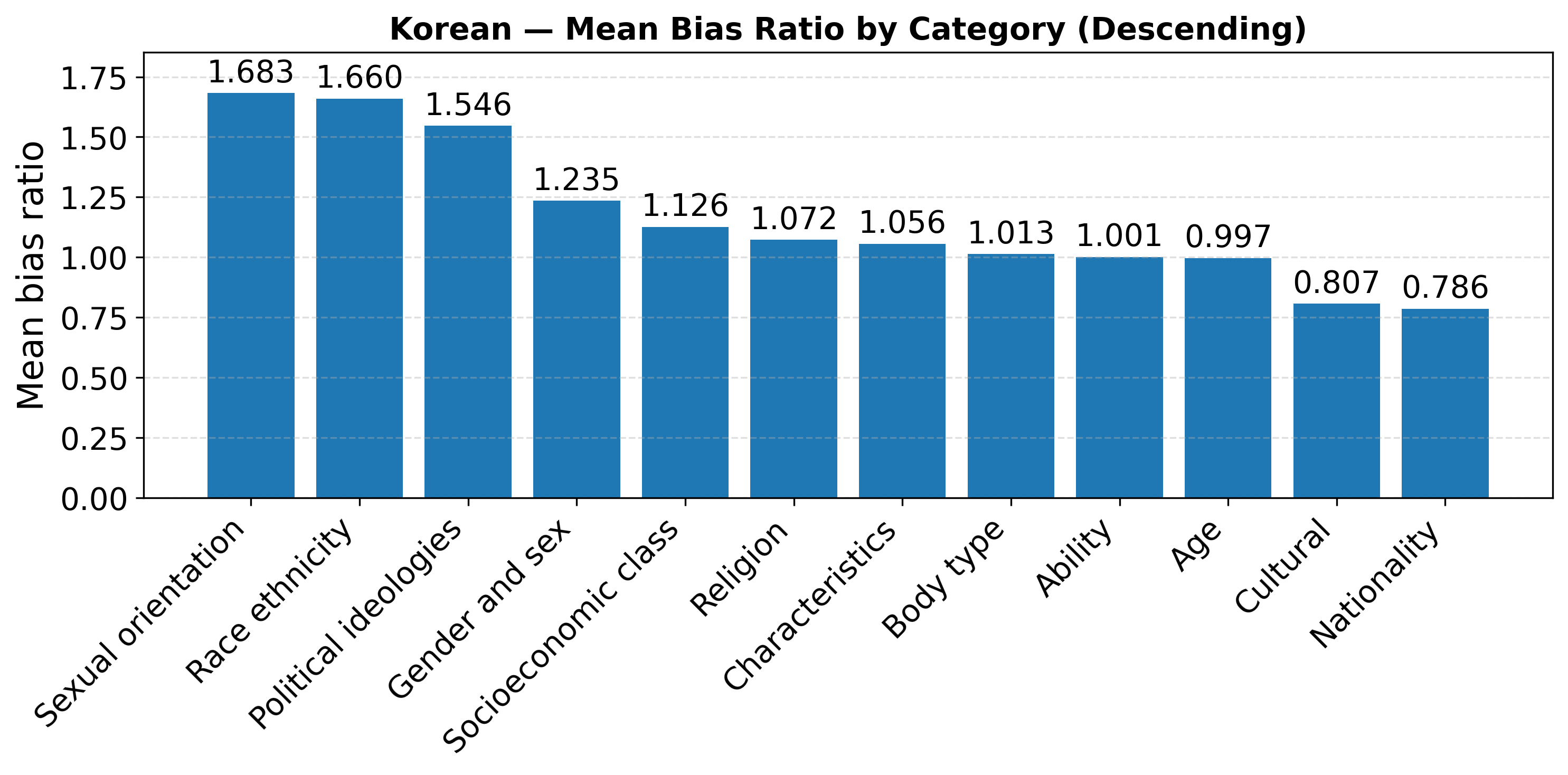}
    \caption{Korean}
  \end{subfigure}
  
    \caption{HolisticBias Bias Analysis of Multilingual Datasets}
    \label{fig:mult_holistic}
\end{figure}

We extend the HolisticBias analysis to multilingual datasets using the same methodology as in the English setting.\footnote{The bias categorization is presented in the main text, while additional analyses, including toxicity scores, swear words, sentence length, and complexity, are provided in Appendix~\ref{app:analysis_multilingual}.} Figure~\ref{fig:mult_holistic} reports mean bias ratios averaged across representative models for French, German, Spanish, Chinese and Korean.

Across all languages, Political Ideologies consistently exhibit the highest over-refusal rates, indicating that sociopolitical discourse remains a dominant trigger of alignment sensitivity in multilingual detoxification. This pattern suggests the transfer of English-centric moderation heuristics, where politically charged language is frequently associated with unsafe content.

Language-specific differences further emerge. In Chinese, Political Ideologies, Socioeconomic Class, and Sexual Orientation show the highest bias ratios, while Religion exhibits relatively low over-refusal. German displays heightened sensitivity toward Race and Ethnicity, followed by Socioeconomic Class. French shows over-refusal concentrated in Cultural, Characteristics and Political Ideologies, with lower bias for Nationality and Socioeconomic Class. Spanish exhibits a more uniform distribution of bias ratios across categories, though Political Ideologies and Gender and Sex remain slightly elevated, consistent with its higher overall false refusal rates.

Overall, these results indicate that \textit{multilingual bias is neither uniform nor random, but instead reflects language-specific sociocultural sensitivities shaped by model alignment}. The persistent over-refusal of political content underscores the need for multilingual bias calibration.

\section{Cross-Translation Framework for False Refusal Mitigation}

Our multilingual evaluation shows that, LLMs exhibit substantially lower false refusal rates when detoxifying non-English inputs compared to English inputs. For instance, Chinese inputs yield the lowest refusal ratio of 0.44\% with GPT-4o-mini, indicating that LLMs are generally more willing to attempt detoxification in non-English languages. This observation motivates us to propose a simple cross-translation framework for mitigating false refusals in detoxification.


Specifically, we employ Qwen-MT~\citep{qwen_mt_blog} as the translation model and pair it with GPT-4o-mini as the detoxification model, which exhibits particularly low false refusal rates on Chinese inputs. 
To ensure the translation quality, we evaluate Qwen-MT on an English–Chinese subset using BLEU and BERTScore. The model achieves a BLEU score of 22.55 and a BERTScore of 0.86, indicating a reasonable level of semantic preservation between the source and translated texts.

We apply the cross-translation framework to HateXplain, with results reported in Table~\ref{tab:cross_translation_results}. Cross-translation substantially reduces the false refusal ratio. Toxicity scores are computed on the original input texts corresponding to false refusals after mitigation, and the reduced average toxicity indicates that cross-translation enables detoxification for higher-toxicity inputs that would otherwise trigger oversensitive refusal. Swear word ratios remain close to the dataset’s original values, and baseline-rescaled average BERTScore of 0.3524 between original inputs and final detoxified outputs indicates non-trivial semantic overlap, suggesting that semantic and stylistic properties are largely preserved. Overall, these findings motivate further investigation into cross-translation for mitigating false refusals in multilingual hate speech detoxification.

\begin{table}[t]
\centering
\small
\begin{tabular}{lccc}
\toprule
\textbf{Metric} & \textbf{Original} & \textbf{Cross-Translation} \\
\midrule
False Refusal Ratio & 11.78\% & \textbf{1.09\%}  \\
Toxicity Score & 0.7220 & 0.6053 \\
Swear Words & 17.71\% & 16.60\%  \\
\bottomrule
\end{tabular}
\caption{Effect of the cross-translation mitigation framework on HateXplain with GPT-4o-mini.}
\label{tab:cross_translation_results}
\end{table}

\section{Conclusion}

In this work, we systematically study false refusal behavior in hate speech detoxification across English and multilingual datasets. 
Our analysis reveals that LLMs disproportionately refuse inputs with higher semantic toxicity and content targeting specific sociopolitical categories, particularly nationality, religion, and political ideologies. In contrast, surface-level features such as swear word presence and sentence complexity play a limited role. 
Moreover, we propose a simple cross-translation mitigation framework that significantly reduces false refusal rates while largely preserving semantic content and toxicity characteristics. Overall, this work provides a foundation for future research on fair, robust, and multilingual detoxification systems that balance safety with utility.

\section*{Limitations}

We acknowledge several limitations in this work. 
First, we adopt a deliberately simple and fixed prompt for the hate speech detoxification task in order to observe the models’ inherent behavior. While this design choice improves comparability across LLMs, alternative prompt formulations, prompt engineering strategies, or decoding parameter settings, such as temperature, may lead to different detoxification and refusal behaviors.

Second, our study is limited to six languages, namely English, French, German, Spanish, Chinese, and Korean. As a result, languages beyond this set, particularly low-resource languages and those with distinct linguistic or cultural norms, are not covered. The observed cross-lingual patterns may therefore not generalize to all languages.

Third, our bias analysis relies on automatic annotation using an LLM-as-a-Judge paradigm, with limited human verification. Although we conduct human evaluation on a subset of samples and observe reasonable agreement, annotation errors or model-specific biases in the judge model may still affect the results. Future work could incorporate larger-scale human annotation or alternative bias measurement frameworks to further strengthen the conclusions.

\section*{Ethical Considerations}

This work analyzes false refusal behavior in LLM-based hate speech detoxification and therefore involves toxic and offensive language. All such content is used solely for research purposes to study model behavior and improve safety and fairness; no harmful content is generated beyond what is required for evaluation.

Our analysis of bias using the HolisticBias taxonomy is descriptive rather than normative and is intended to identify uneven refusal behavior across social categories, not to make judgments about any group. While our proposed cross-translation strategy reduces false refusals, it may also affect the detection of unsafe content in certain languages and should be applied with caution in real-world systems.


\bibliography{custom}

\appendix

\section{Models}\label{app:models}

The details of the LLMs, including their sizes and multilingual coverage, are presented in Table~\ref{tab:models}.

\begin{table}[ht]
\centering
\small
\setlength{\tabcolsep}{10pt}
\scalebox{0.9}{
\begin{tabular}{l l l}
\toprule
\textbf{Model} & \textbf{Size} & \textbf{Multilingual Coverage} \\
\midrule
\multicolumn{3}{l}{\textbf{Closed Source}} \\
\midrule
GPT-4o mini & - & 50+ languages \\
GPT-3.5 Turbo & - & 50+ languages \\
\midrule
\multicolumn{3}{l}{\textbf{Open Source}} \\
\midrule
Llama-3.1 & 8B & 30+ languages \\
Mistral & 8B & Limited multilingual support \\
Mixtral & 8$\times$7B & Limited multilingual support \\
Qwen-2.5 & 7B & 29+ languages \\
Qwen-3 & 30B & 29+ languages \\
Gemma-2 & 9B & Limited multilingual support \\
Gemma-3 & 27B & Limited multilingual support \\
\bottomrule
\end{tabular}}
\caption{Language models used in this study, with their sizes and approximate multilingual coverage.}
\label{tab:models}
\end{table}

\section{Inter-Annotator Agreement of Phi-4}\label{app:inter-annotator}

To assess Phi-4’s reliability as an automated annotator, we conduct a cross-annotation study using 200 detoxified outputs: 100 examples flagged by Phi-4 as potential false refusals and 100 non–false-refusal cases. The samples are randomly ordered and stratified to ensure adequate positive coverage. As shown in Table~\ref{tab:false-refusal-agreement}, human–human agreement is very strong (Cohen’s $\kappa = 0.878$, 94.5\% raw agreement), confirming consistent expert interpretation of false refusals. These results indicate moderate-to-strong alignment between Phi-4 and human annotations and suggest that Phi-4 is sufficiently reliable for large-scale false-refusal annotation.

\begin{table}[!htbp]
\centering
\small
\begin{tabularx}{\columnwidth}{X c c}
\toprule
\textbf{Annotator Pair} & \textbf{Cohen's $\kappa$} & \textbf{Agreement (\%)} \\
\midrule
Annotator A vs Annotator B & 0.878 & 94.5 \\
Phi-4 vs Annotator A       & 0.670 & 83.5 \\
Phi-4 vs Annotator B       & 0.700 & 85.0 \\
Phi-4 vs Human Consensus   & 0.740 & 87.0 \\
\bottomrule
\end{tabularx}
\caption{Inter-annotator agreement between human annotators and the Phi-4 model on the stratified 200-sample evaluation set.}
\label{tab:false-refusal-agreement}
\end{table}

\section{HolisticBias Structure}
\label{app:holistic-ontology}
\begin{figure}[!htbp]
    \centering
    \includegraphics[width=0.9\linewidth]{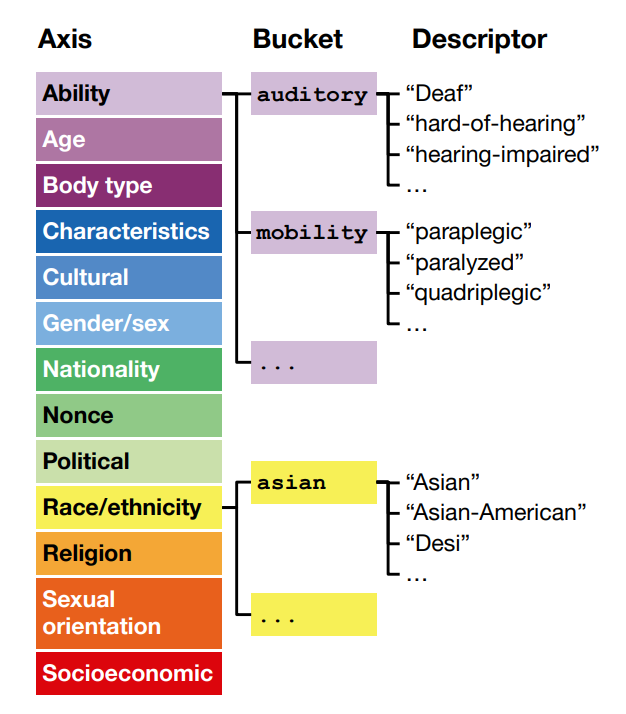}
    \caption{Structure of the HolisticBias dataset adapted from ~\citet{holisticbias}. It consists of 13 demographic pillars, each containing multiple property groups representing diverse social and identity dimensions.}
    \label{fig:holisticbias_structure}
\end{figure}

HolisticBias dataset consists of 13 demographic axis and buckets. Figure ~\ref{fig:holisticbias_structure} shows the overview of HolisticBias.

\section{HolisticBias Categorization of English Datasets}\label{app:english_holistic_overview}

\begin{figure}[!htbp]
  \centering
  \begin{subfigure}[b]{0.9\linewidth}
    \includegraphics[width=\linewidth]{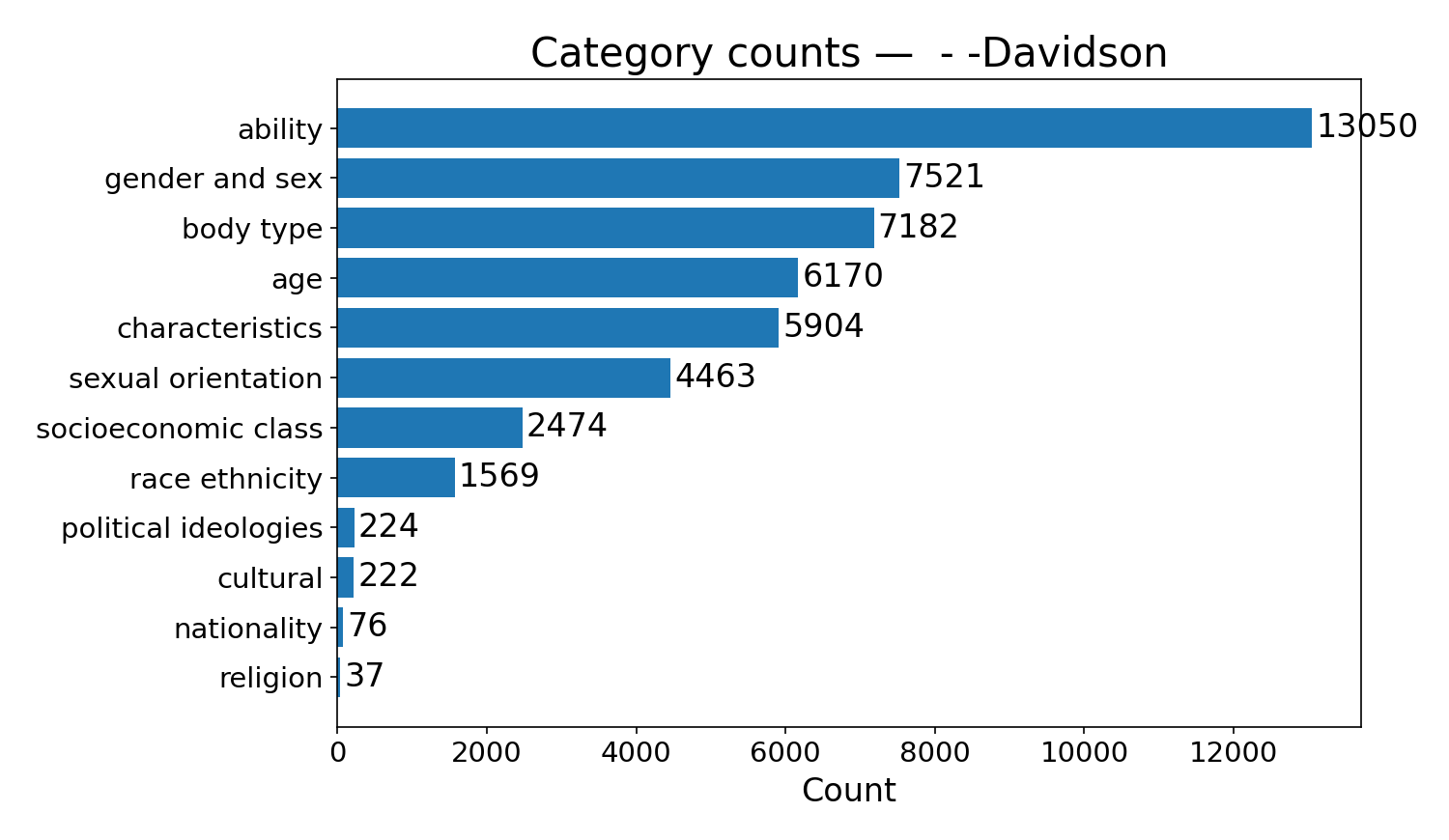}
    \caption{Davidson}
  \end{subfigure}\hfill
  \begin{subfigure}[b]{0.9\linewidth}
    \includegraphics[width=\linewidth]{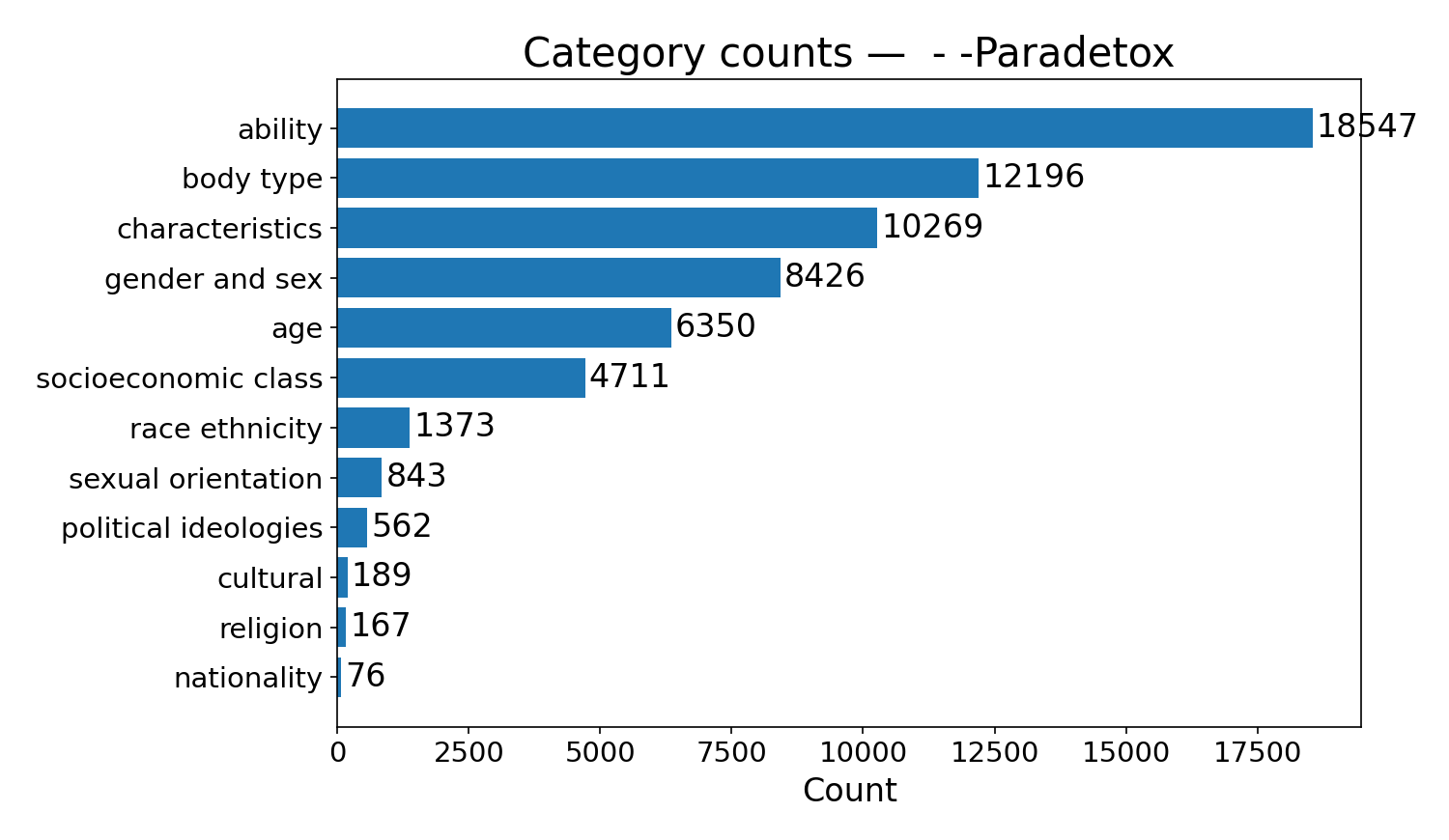}
    \caption{Paradetox}
  \end{subfigure}

  \vspace{0.75em}

  \begin{subfigure}[b]{0.9\linewidth}
    \includegraphics[width=\linewidth]{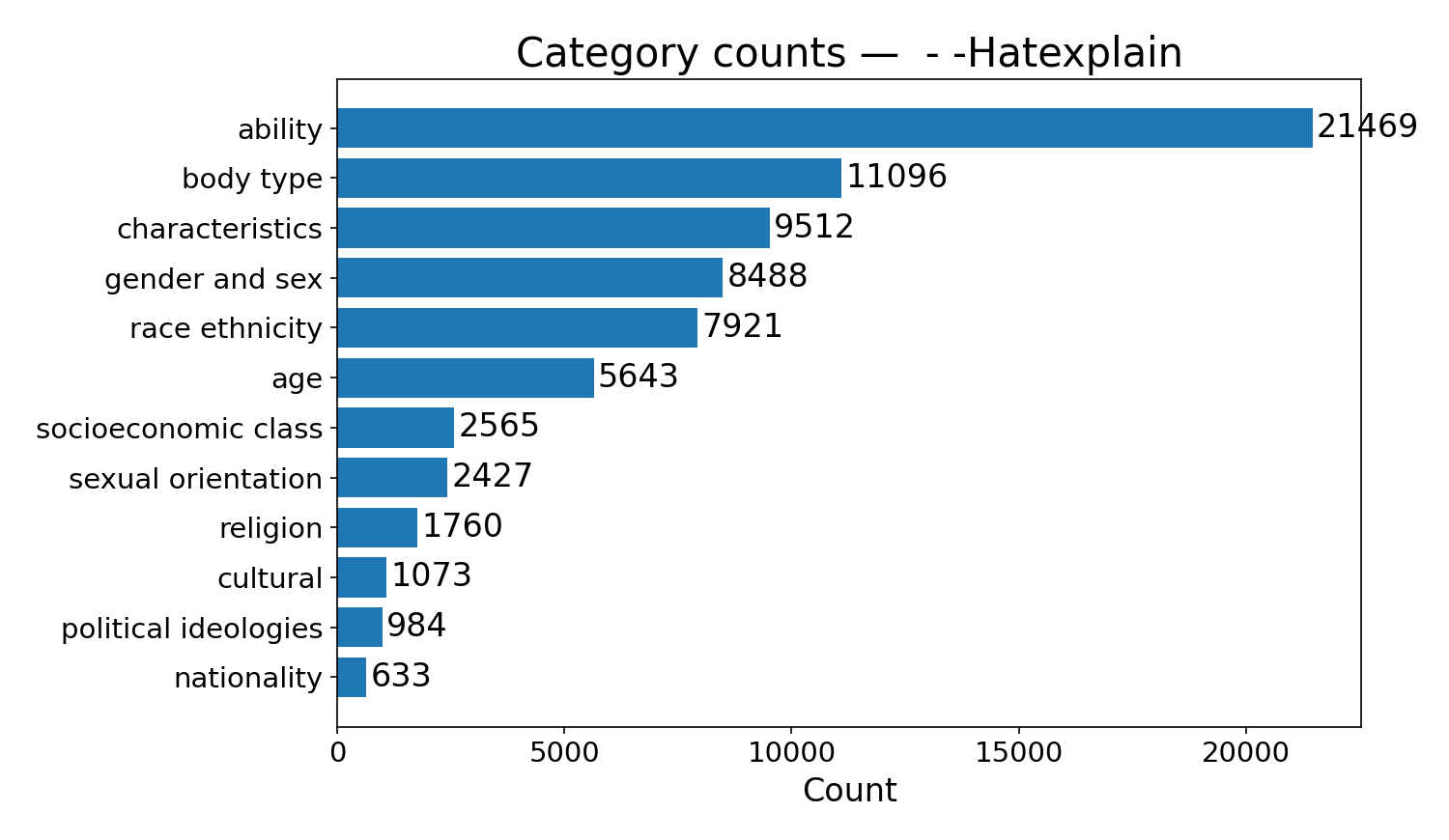}
    \caption{HateXplain}
  \end{subfigure}

  \caption{HolisticBias categorization of English datasets.}
  \label{fig:holisticbias_english}
\end{figure}

Figure~\ref{fig:holisticbias_english} presents the HolisticBias categorization of the original English datasets, illustrating the distribution of samples across each category.

\section{Mean Bias Ratio of False Refusals}
\label{app:mean_bias_ratio_eng}

The mean bias ratio of false refusals for each HolisticBias category, aggregated across the three English datasets, is reported in Table~\ref{tab:holisticbias_means}.

\begin{table*}[ht]
    \centering
    \resizebox{\textwidth}{!}{
    \begin{tabular}{lccccccccccc}
    \toprule
    \textbf{Category} & \textbf{GPT 3.5} & \textbf{GPT4o mini} & \textbf{Mistral 7B} & \textbf{Mixtral 8×7B} & \textbf{Llama3 8B} & \textbf{Qwen2.5 7B} & \textbf{Qwen3 30B} & \textbf{Gemma2 9B} & \textbf{Gemma3 27B} \\
    \midrule
    Nationality         & \textbf{1.83} & \textbf{1.67} & \textbf{1.94} & 1.24 & 1.22 & \textbf{1.76} & 1.37 & \textbf{1.87} & 1.48 \\
    Religion            & 1.61 & 1.46 & 1.81 & 1.54 & 1.15 & 1.29 & 1.33 & 1.43 & \textbf{1.78} \\
    Cultural            & 1.13 & 1.22 & 1.20 & \textbf{1.58} & 1.07 & 1.04 & 1.09 & 1.11 & 0.91 \\
    Race ethnicity      & 1.19 & 1.22 & 1.18 & 0.99 & 1.24 & 1.23 & 1.06 & 1.32 & 0.86 \\
    Sexual orientation  & 1.36 & 1.43 & 1.21 & \underline{0.70} & \textbf{1.39} & 1.33 & 0.99 & 1.39 & 0.98 \\
    Political ideologies & 1.15 & 1.34 & 1.42 & 1.45 & 1.11 & 1.12 & \textbf{1.48} & 1.26 & 1.52 \\
    Socioeconomic class & 1.31 & 1.33 & 1.17 & 1.22 & 1.25 & 1.23 & 1.14 & 1.20 & 1.64 \\
    Characteristics     & 1.02 & 1.04 & 1.02 & 1.04 & 1.01 & 1.00 & 0.91 & 1.00 & 0.91 \\
    Body type           &  0.91 & 0.89 & 0.97 & 1.05 & 0.94 & 0.93 & 0.99 & 0.95 & 0.90 \\
    Ability             & 0.99 & 0.96 & 0.99 & 1.01 & 1.00 & 1.00 & 0.95 & 0.98 & 1.01 \\
    Age                 & 1.02 & 1.03 & 0.92 & 1.03 & 1.00 & 1.03 & 0.94 & 1.01 & 0.98 \\
    Gender and sex      & \underline{0.79} & \underline{0.76} & \underline{0.79} & 0.73 & \underline{0.79} & \underline{0.81} & \underline{0.90} & \underline{0.76} & \underline{0.77} \\
    \bottomrule
    \end{tabular}
    }
        \caption{Mean bias ratio ($R_c$) per HolisticBias category across models for English datasets. Values greater than 1 indicate overrepresentation of the category in false refusals. The largest value in each column is shown in \textbf{bold}, and the smallest value is \underline{underlined}.}
    \label{tab:holisticbias_means}
\end{table*}

\section{Linguistic Analysis of False Refusals in English Datasets}
\label{app:linguistic_english}

This section presents additional linguistic analyses of false refusals in English datasets that are omitted from the main body due to space constraints. We analyze three linguistic factors: \textbf{token length}, \textbf{clause count}, and \textbf{parse tree depth}. The corresponding distributions across different model families are illustrated as follows.

\begin{figure*}[htbp]
  \centering
  \setlength{\tabcolsep}{3pt}

  \begin{subfigure}{0.32\textwidth}
    \centering
    \includegraphics[width=\linewidth]{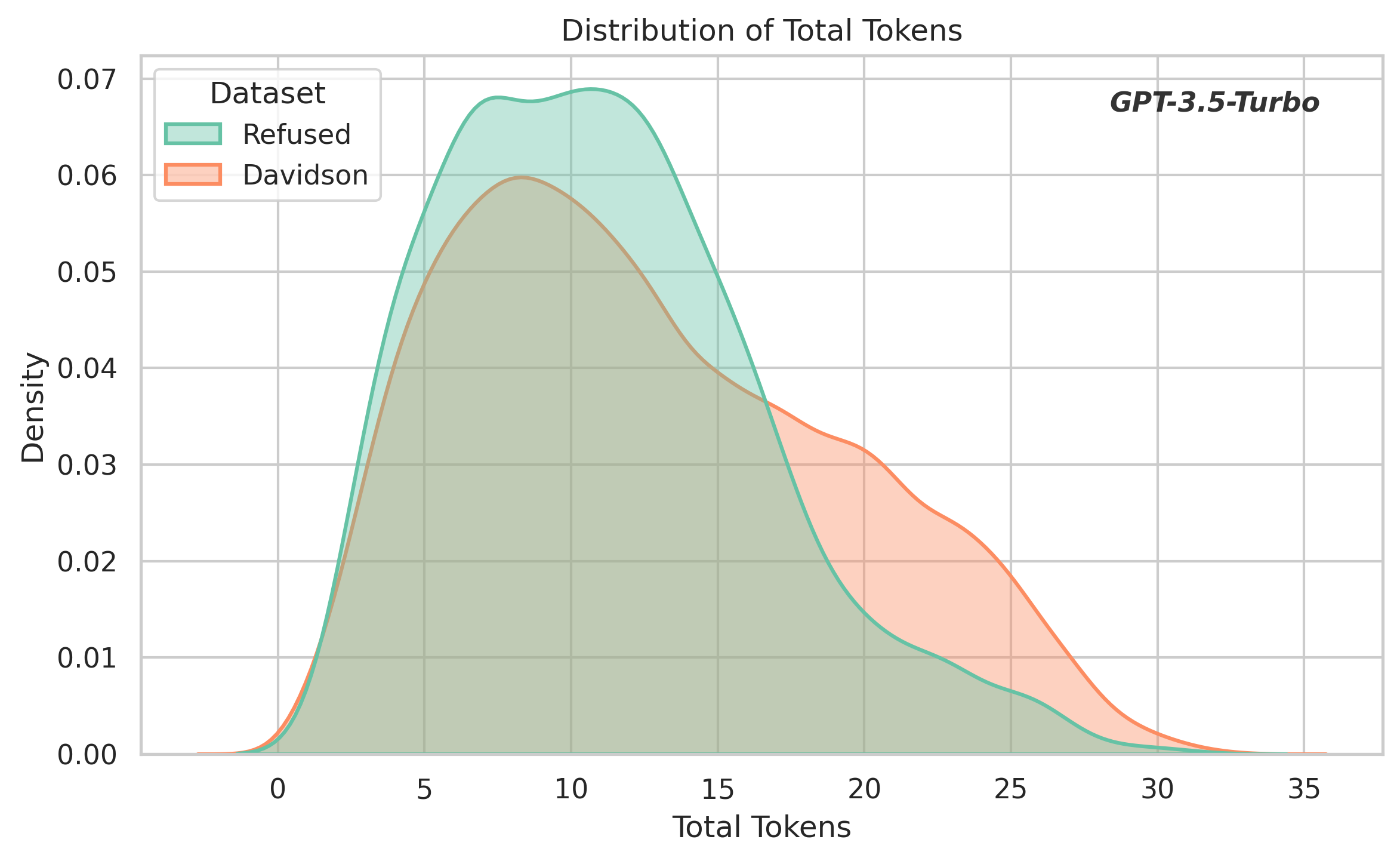}
    \caption{Davidson}
  \end{subfigure}\hfill
  \begin{subfigure}{0.32\textwidth}
    \centering
    \includegraphics[width=\linewidth]{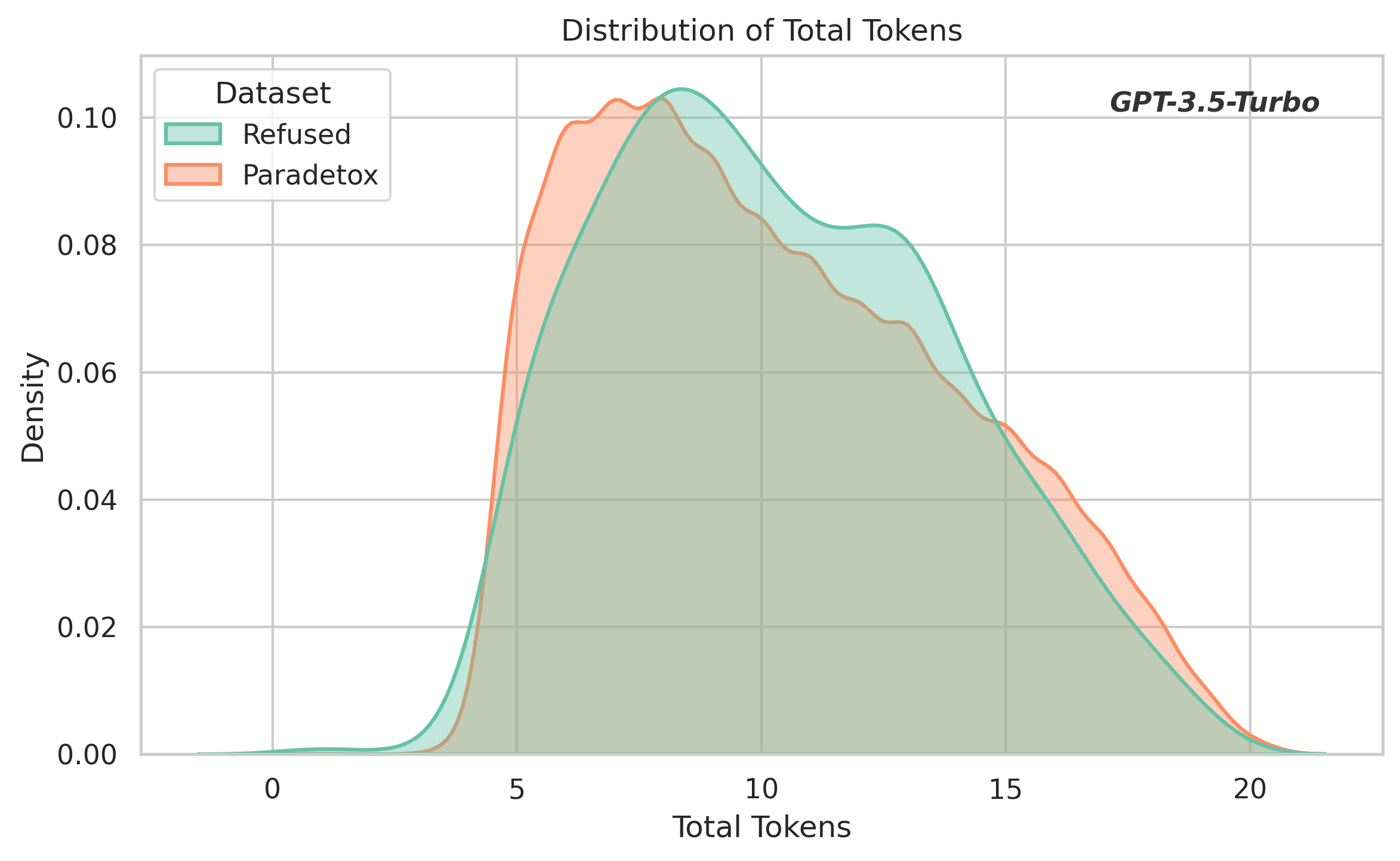}
    \caption{Paradetox}
  \end{subfigure}\hfill
  \begin{subfigure}{0.32\textwidth}
    \centering
    \includegraphics[width=\linewidth]{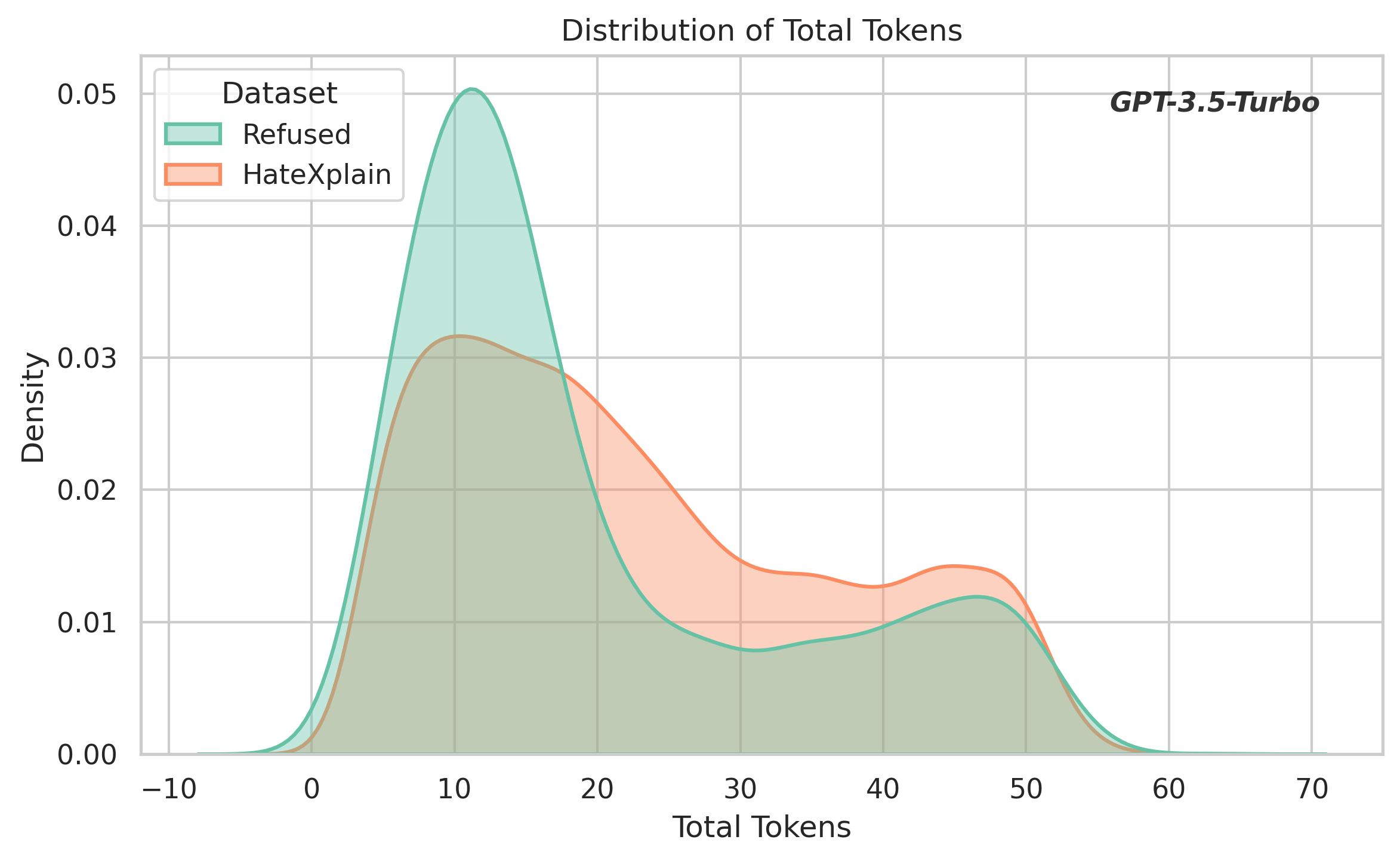}
    \caption{HateXplain}
  \end{subfigure}

  \vspace{0.4em}

  \begin{subfigure}{0.32\textwidth}
    \centering
    \includegraphics[width=\linewidth]{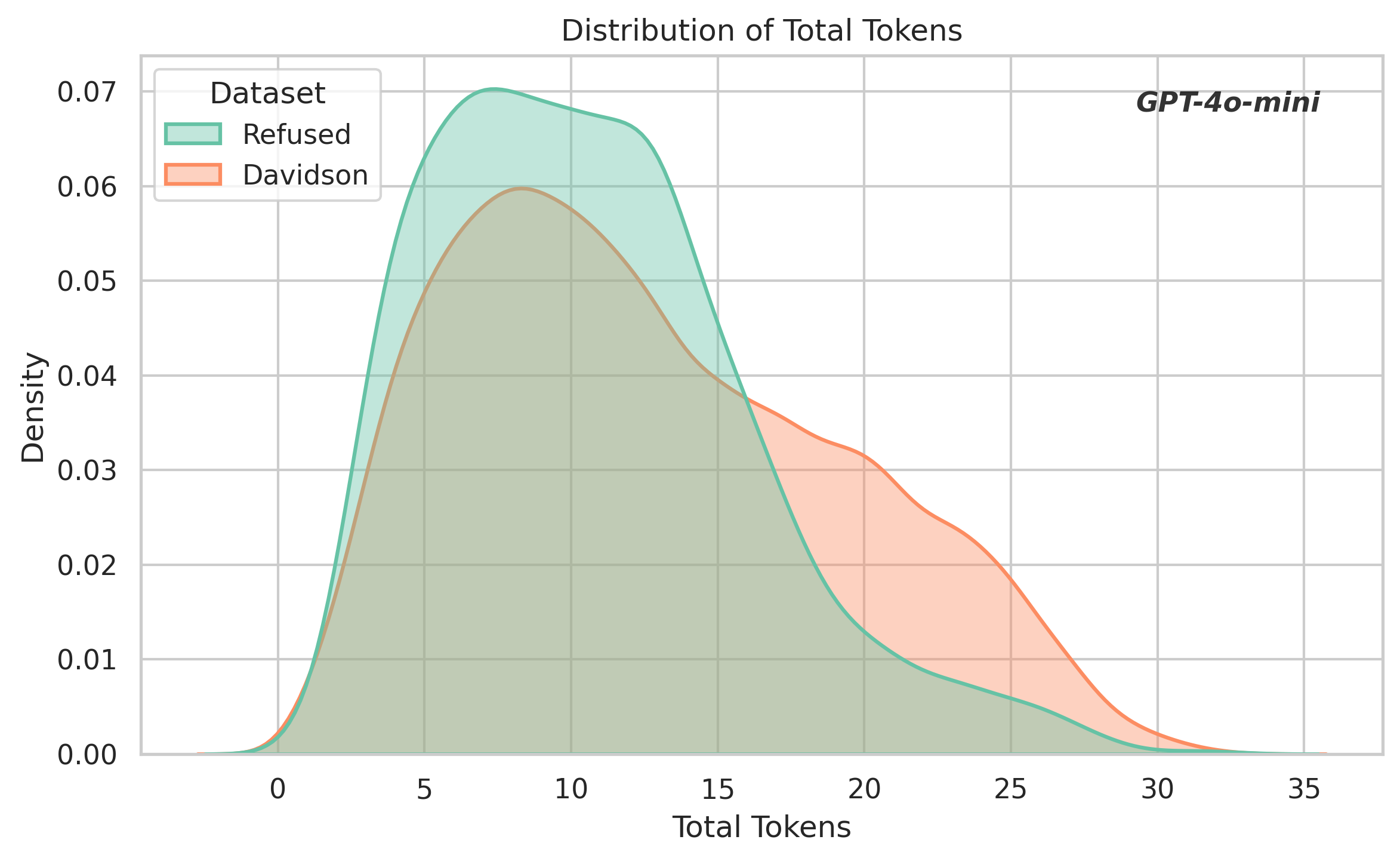}
    \caption{Davidson}
  \end{subfigure}\hfill
  \begin{subfigure}{0.32\textwidth}
    \centering
    \includegraphics[width=\linewidth]{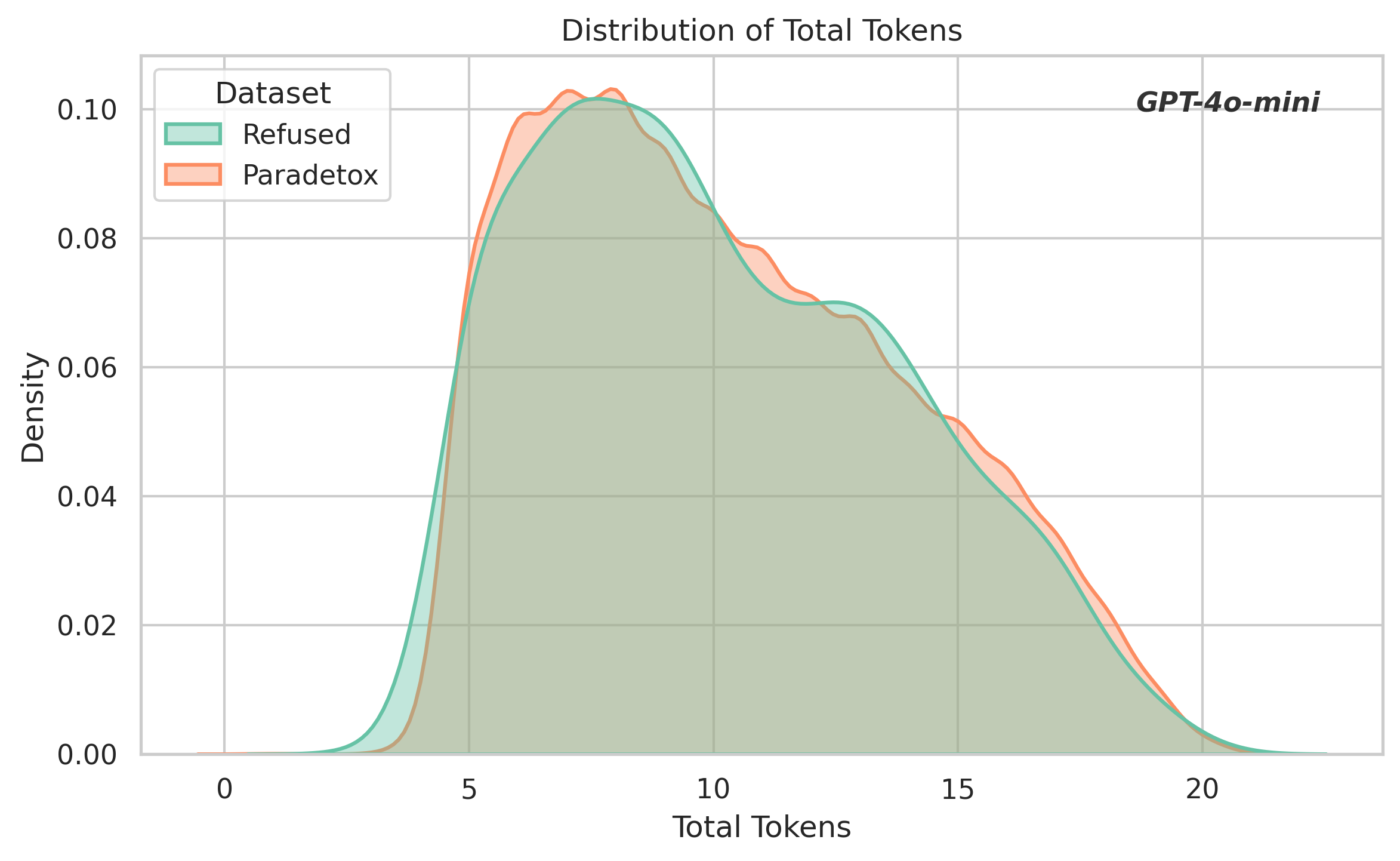}
    \caption{Paradetox}
  \end{subfigure}\hfill
  \begin{subfigure}{0.32\textwidth}
    \centering
    \includegraphics[width=\linewidth]{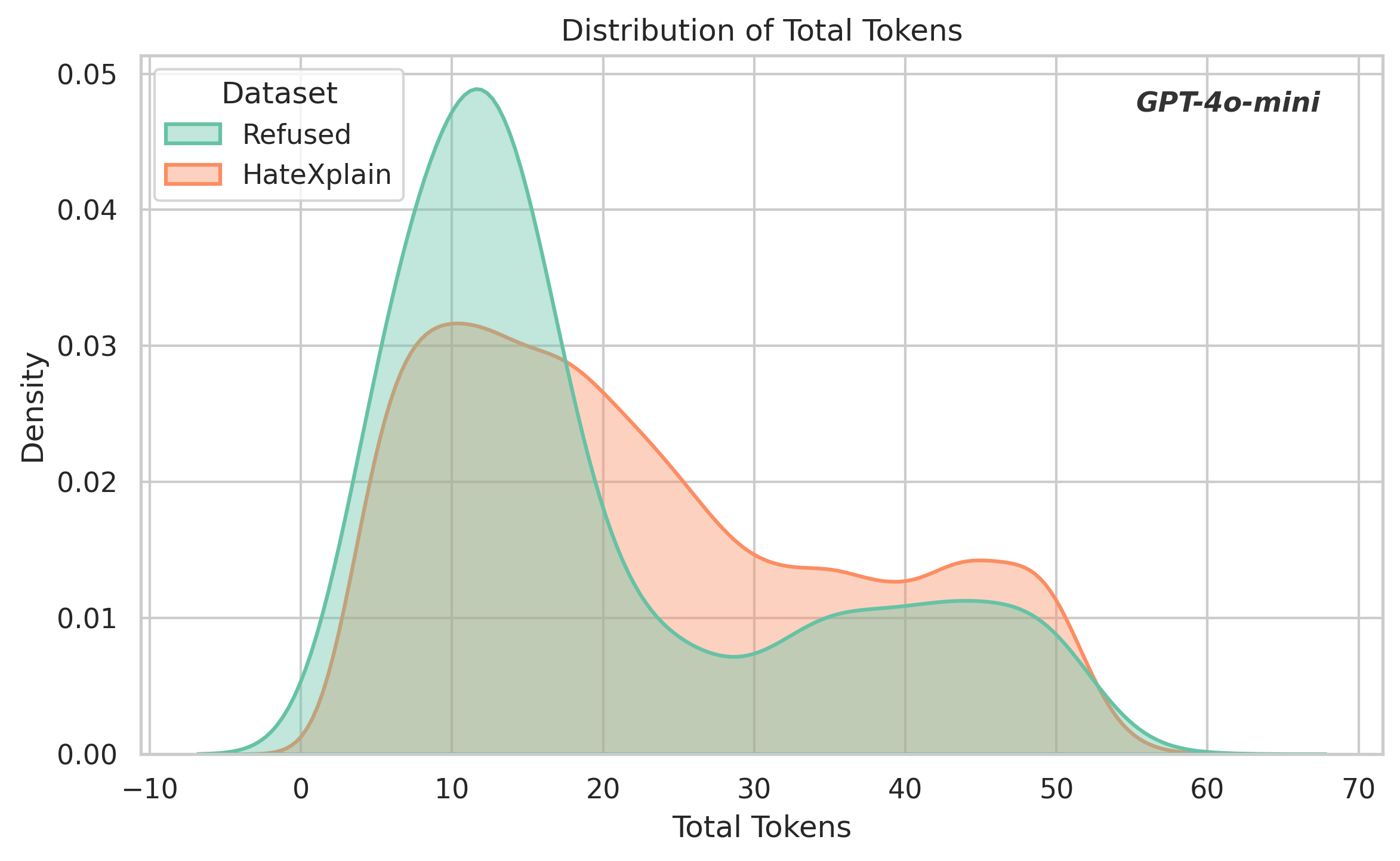}
    \caption{HateXplain}
  \end{subfigure}

  \caption{Token count distributions across datasets for GPT-3.5-Turbo and GPT-4o-mini.}
  \label{fig:token_count_gpt}
\end{figure*}

\begin{figure*}[htbp]
  \centering
  \begin{subfigure}{0.32\textwidth}
    \centering
    \includegraphics[width=\linewidth]{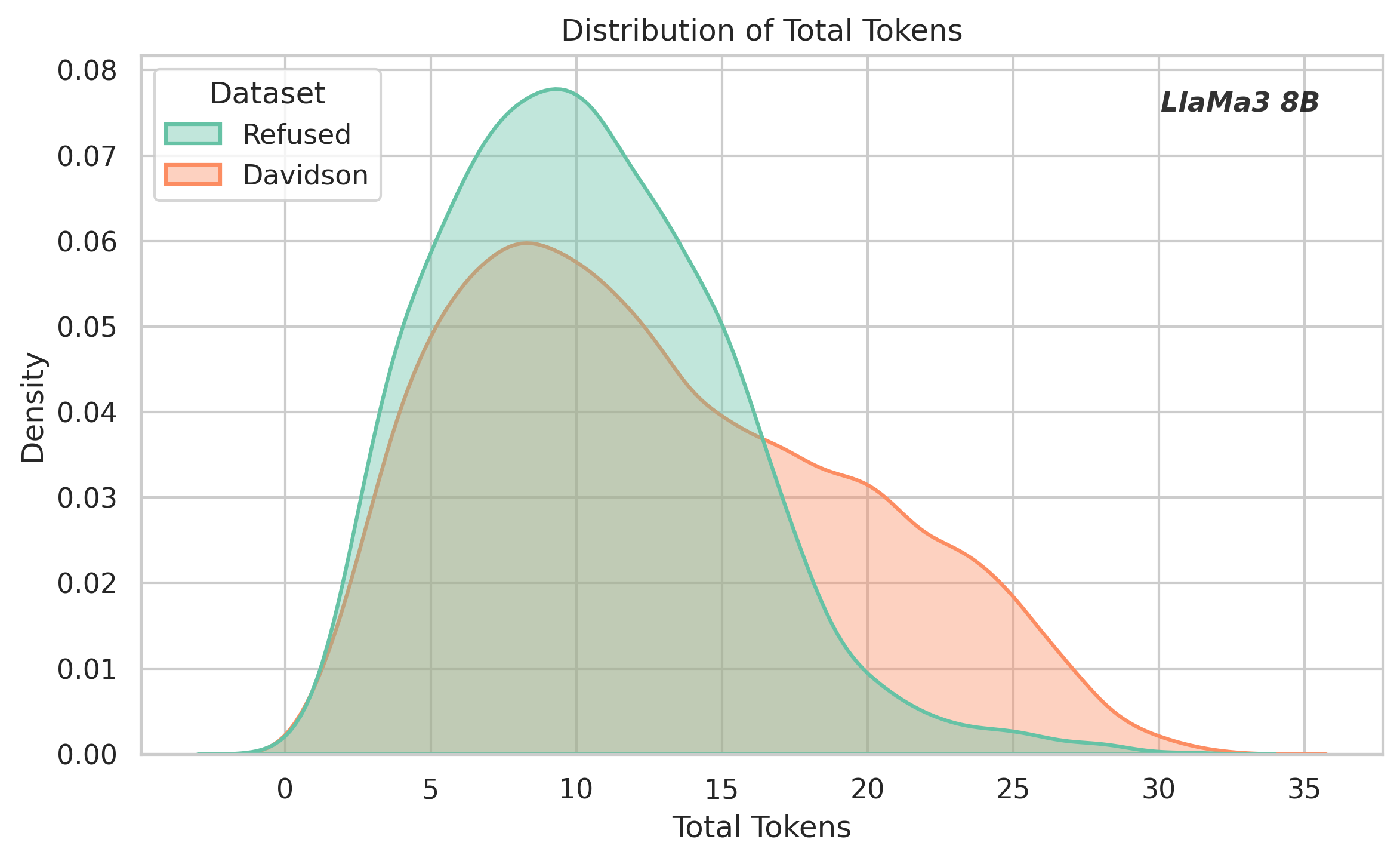}
    \caption{Davidson}
  \end{subfigure}\hfill
  \begin{subfigure}{0.32\textwidth}
    \centering
    \includegraphics[width=\linewidth]{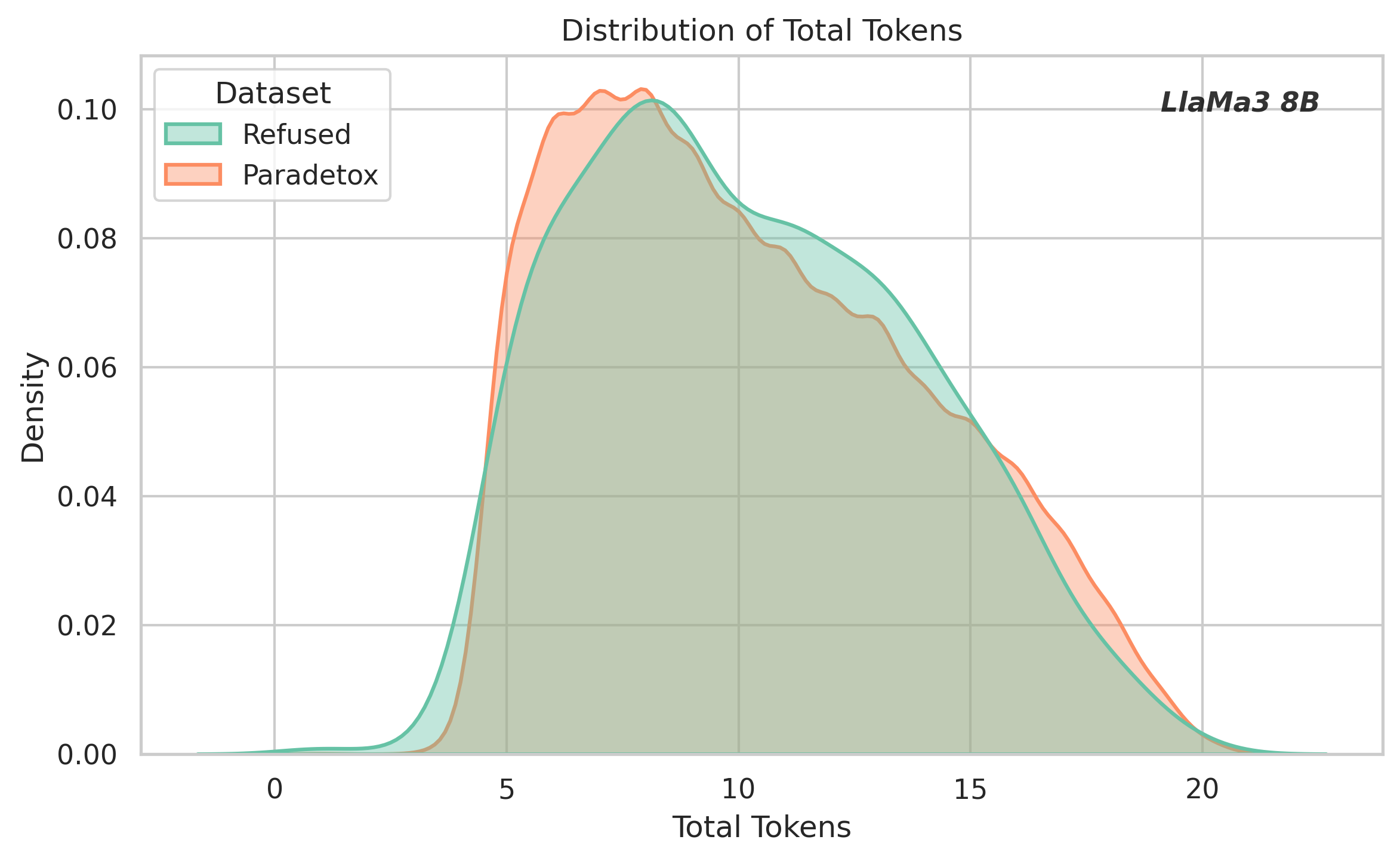}
    \caption{Paradetox}
  \end{subfigure}\hfill
  \begin{subfigure}{0.32\textwidth}
    \centering
    \includegraphics[width=\linewidth]{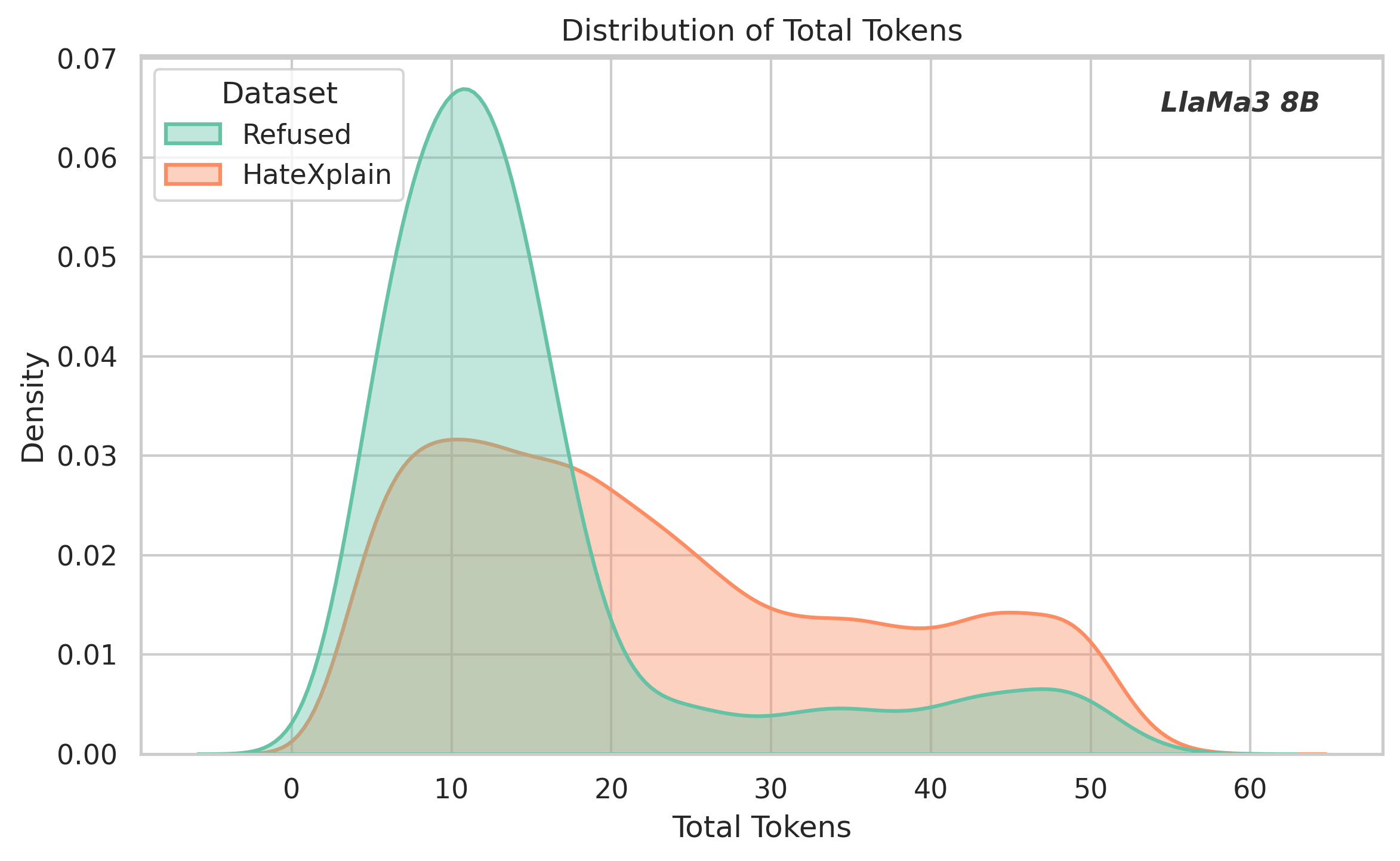}
    \caption{HateXplain}
  \end{subfigure}

  \caption{Token count distributions across datasets for Llama3 8B.}
  \label{fig:token_count_llama}
\end{figure*}

\begin{figure*}[htbp]
  \centering
  \begin{subfigure}{0.32\textwidth}
    \centering
    \includegraphics[width=\linewidth]{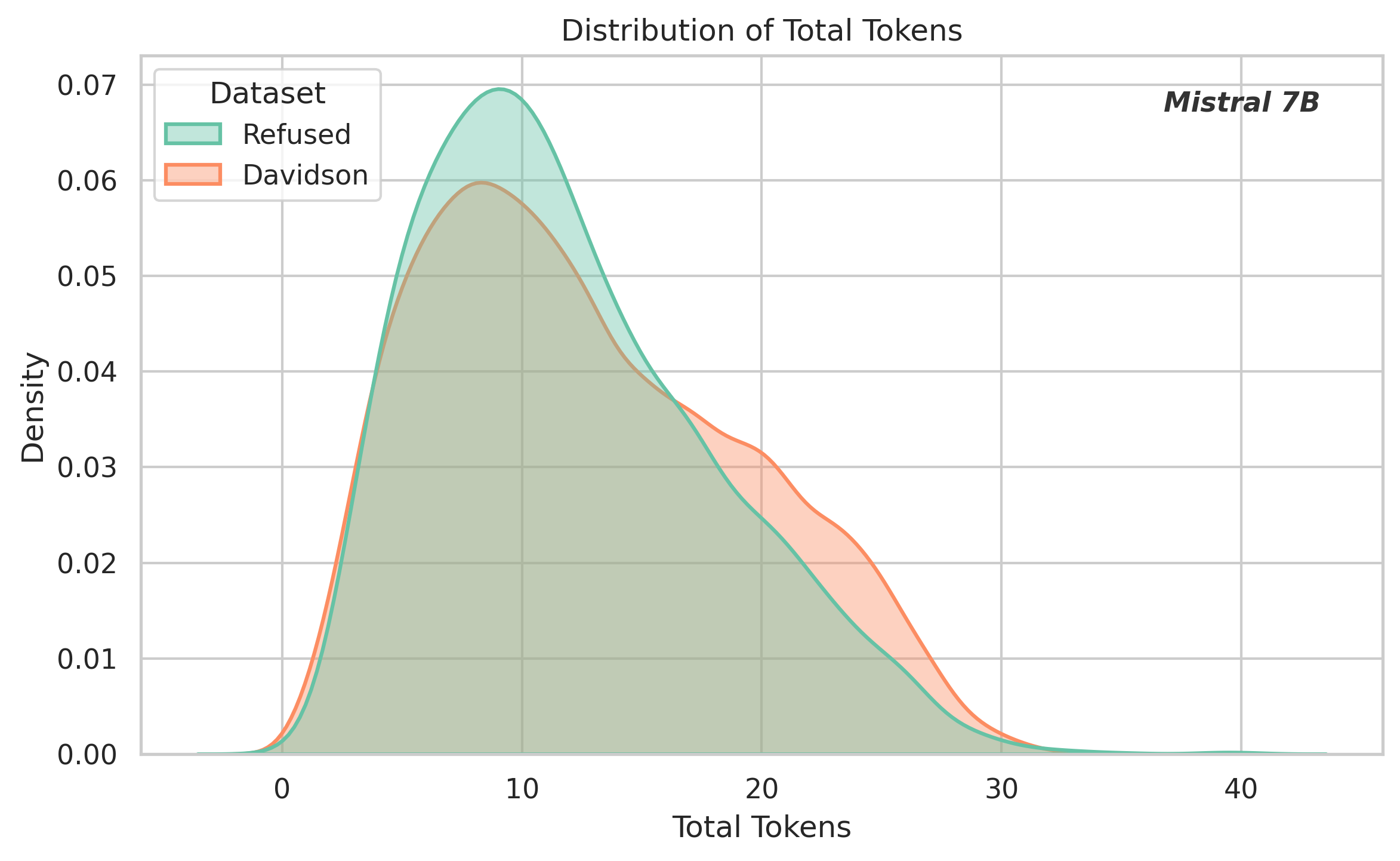}
    \caption{Davidson}
  \end{subfigure}\hfill
  \begin{subfigure}{0.32\textwidth}
    \centering
    \includegraphics[width=\linewidth]{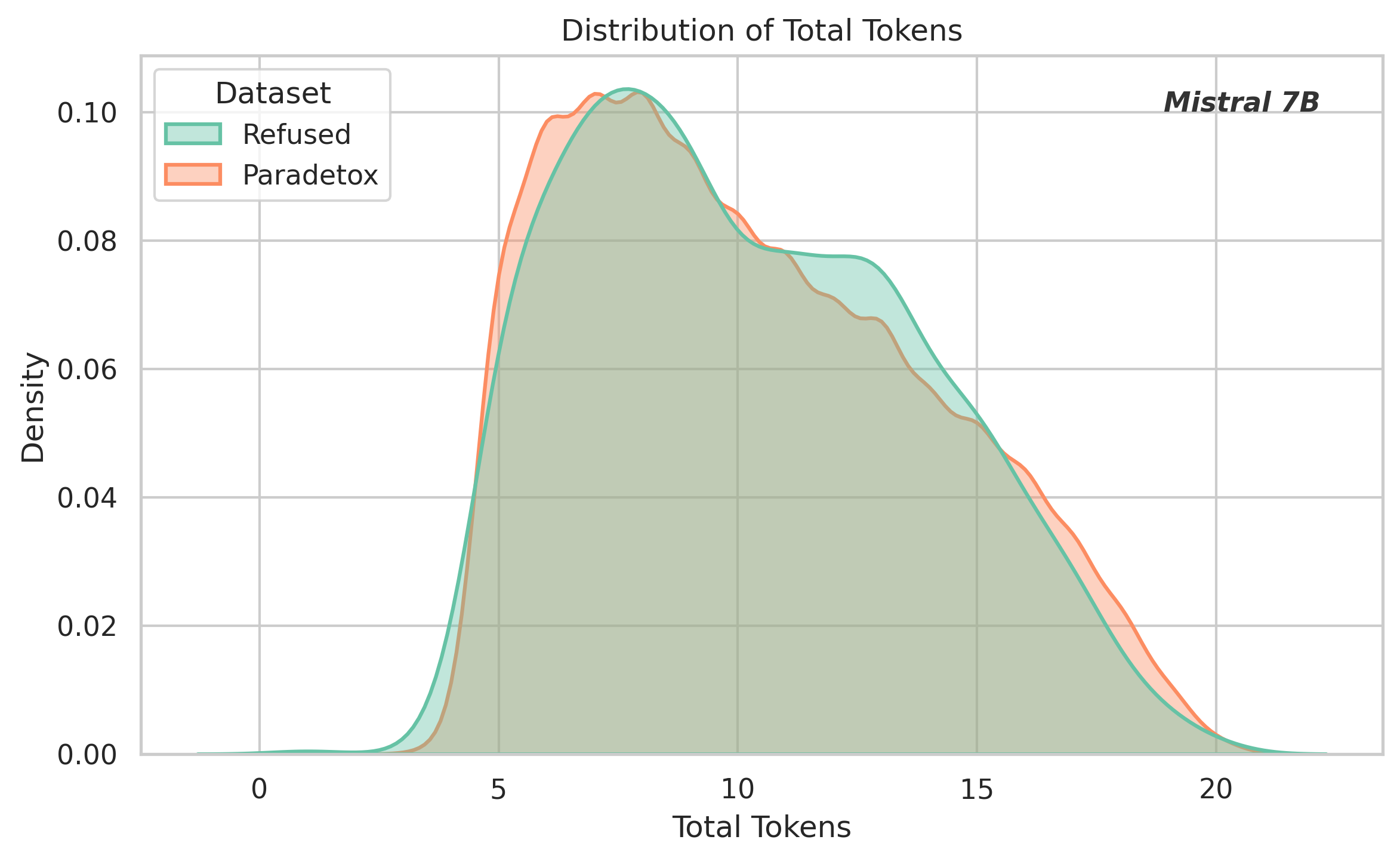}
    \caption{Paradetox}
  \end{subfigure}\hfill
  \begin{subfigure}{0.32\textwidth}
    \centering
    \includegraphics[width=\linewidth]{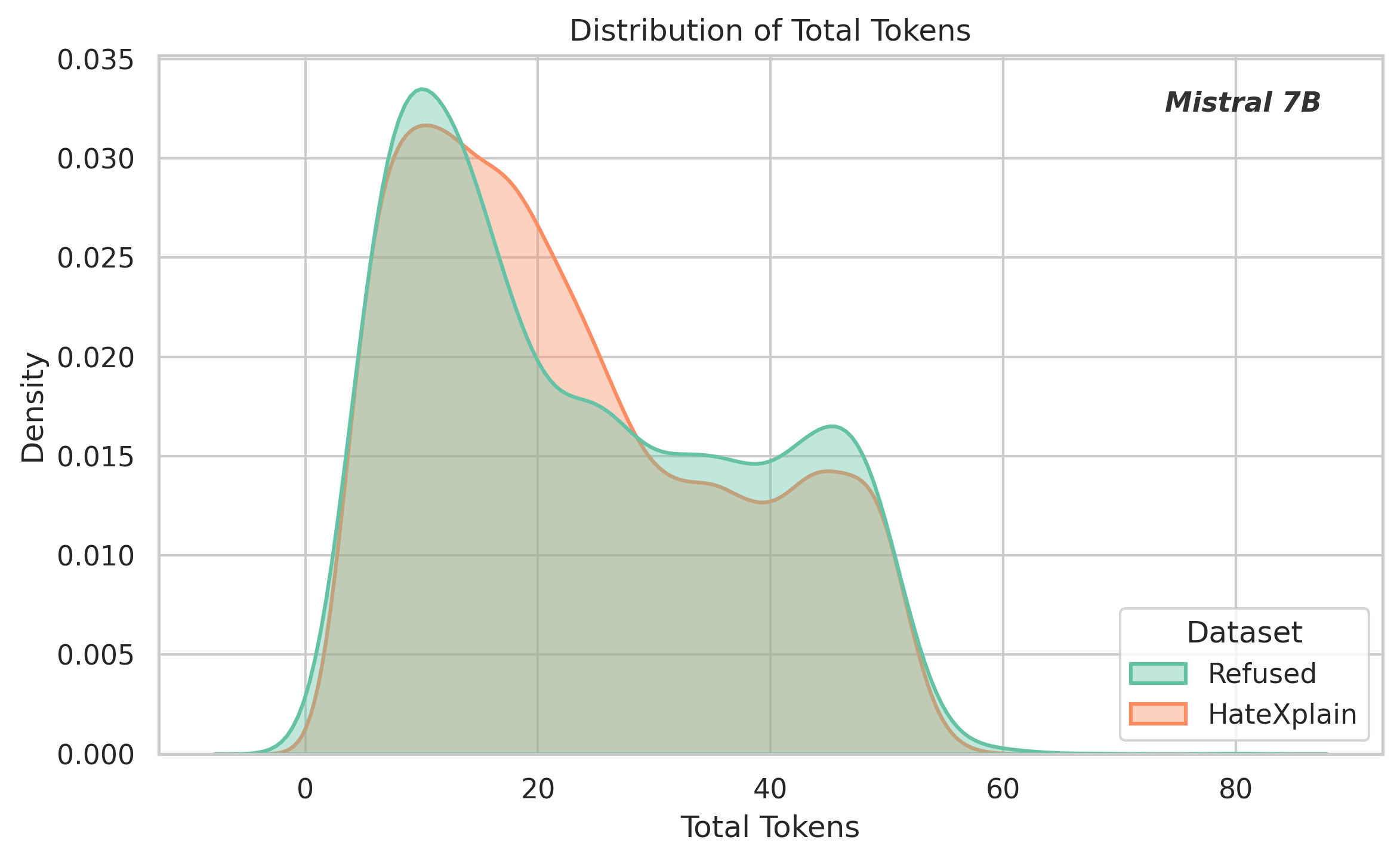}
    \caption{HateXplain}
  \end{subfigure}

  \vspace{0.4em}

  \begin{subfigure}{0.32\textwidth}
    \centering
    \includegraphics[width=\linewidth]{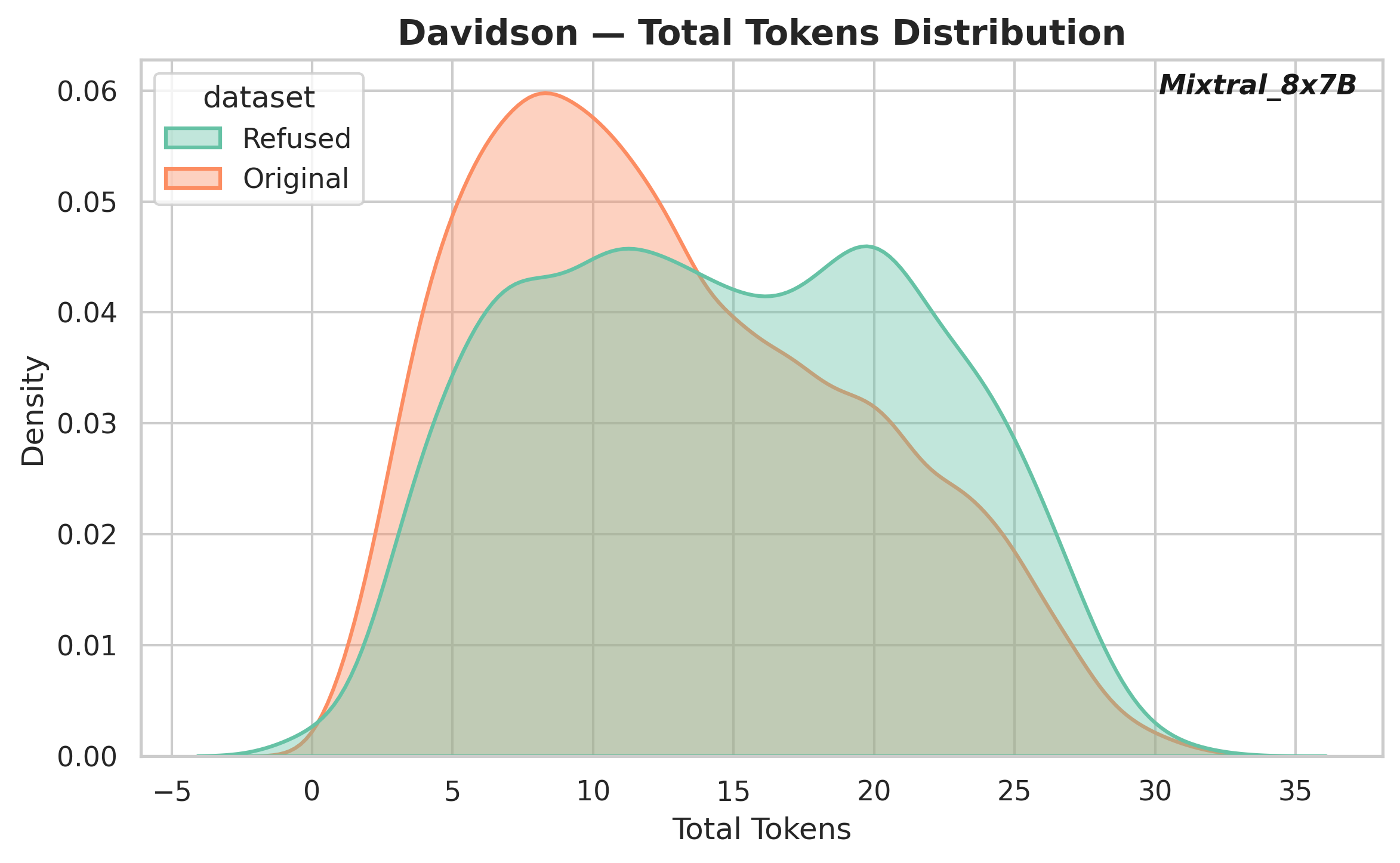}
    \caption{Davidson}
  \end{subfigure}\hfill
  \begin{subfigure}{0.32\textwidth}
    \centering
    \includegraphics[width=\linewidth]{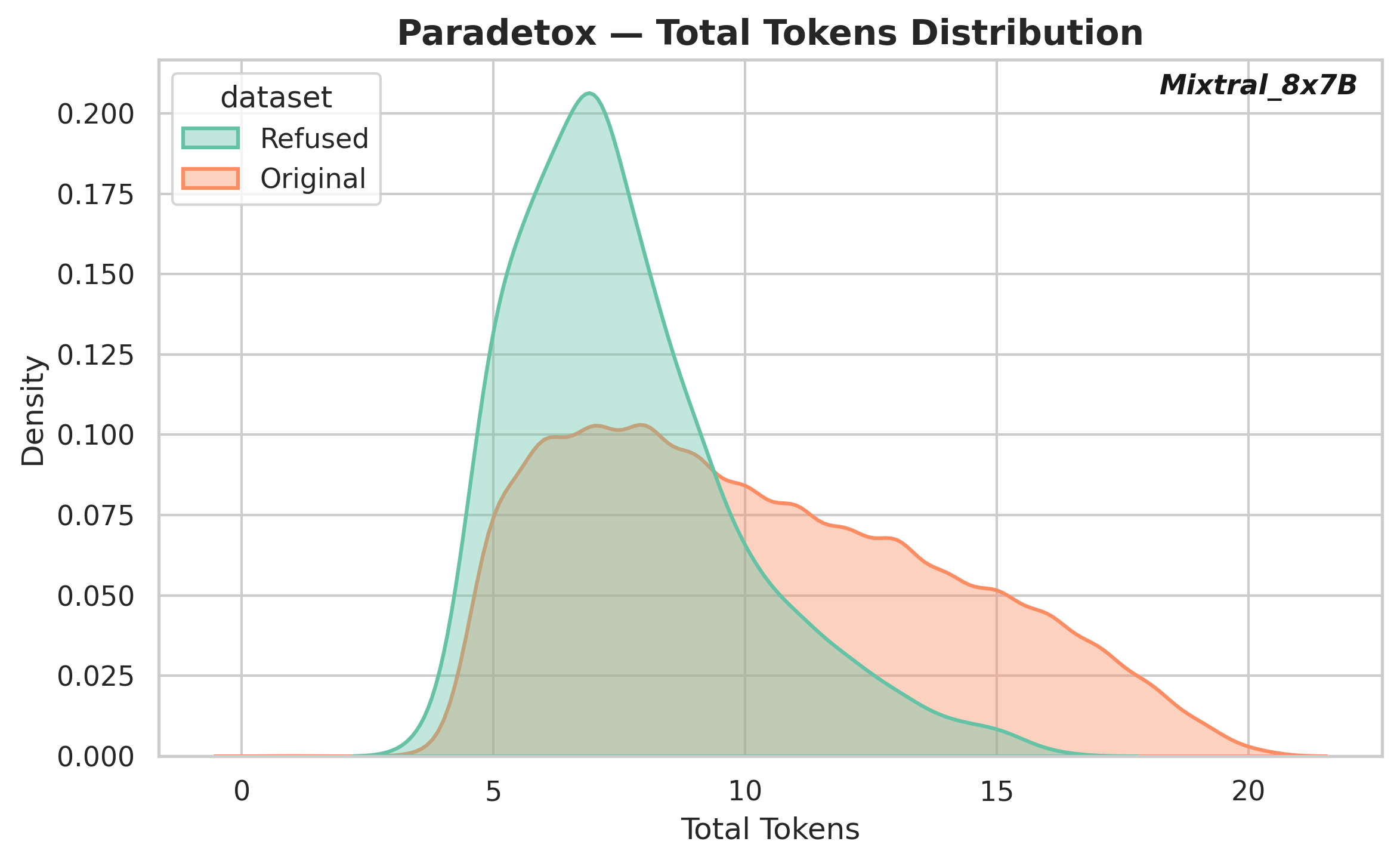}
    \caption{Parardetox}
  \end{subfigure}\hfill
  \begin{subfigure}{0.32\textwidth}
    \centering
    \includegraphics[width=\linewidth]{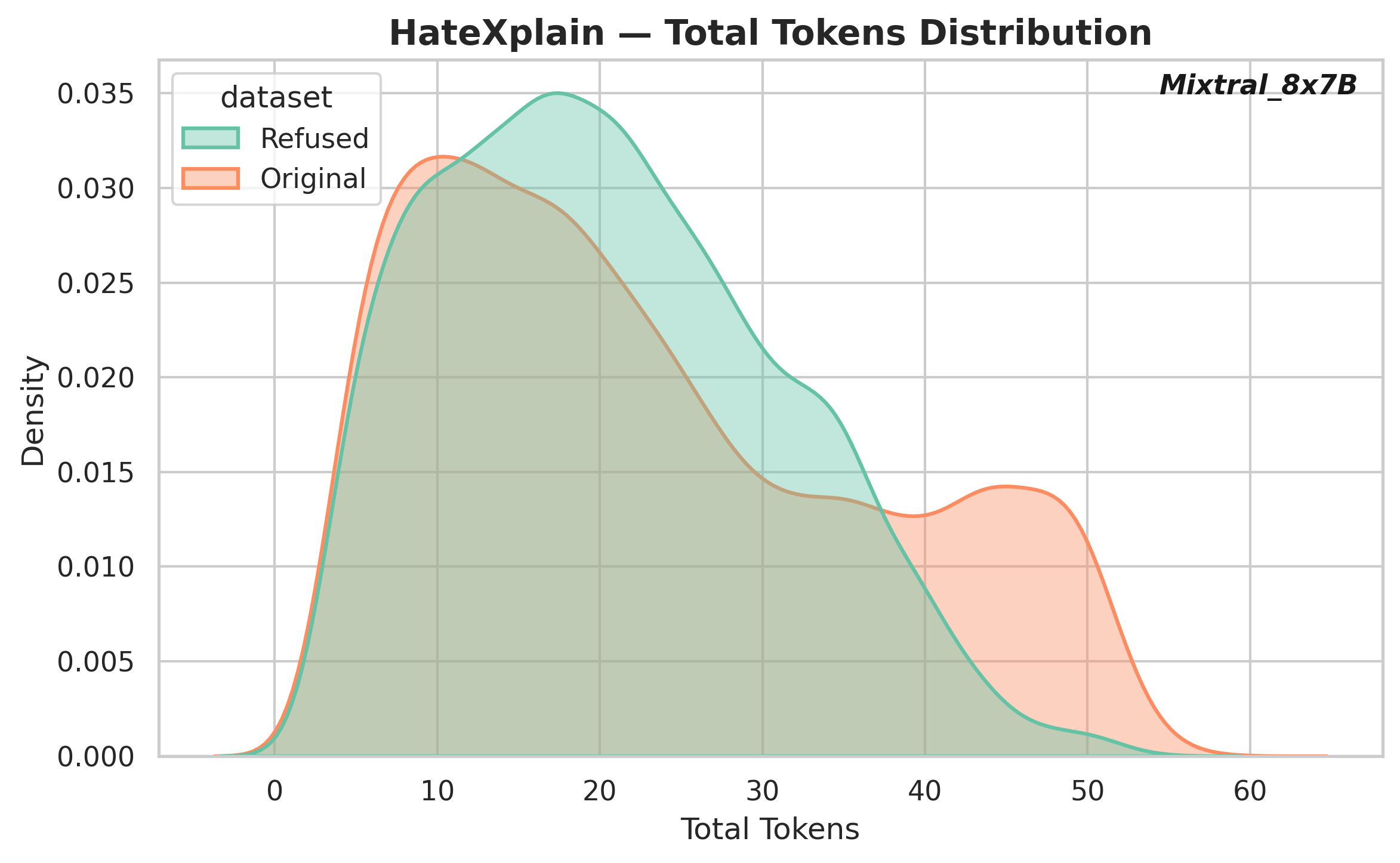}
    \caption{HateXplain}
  \end{subfigure}

  \caption{Token count distributions across datasets for Mistral 7B and Mixtral 8$\times$7B.}
  \label{fig:token_count_mistral}
\end{figure*}

\begin{figure*}[htbp]
  \centering
  \begin{subfigure}{0.32\textwidth}
    \centering
    \includegraphics[width=\linewidth]{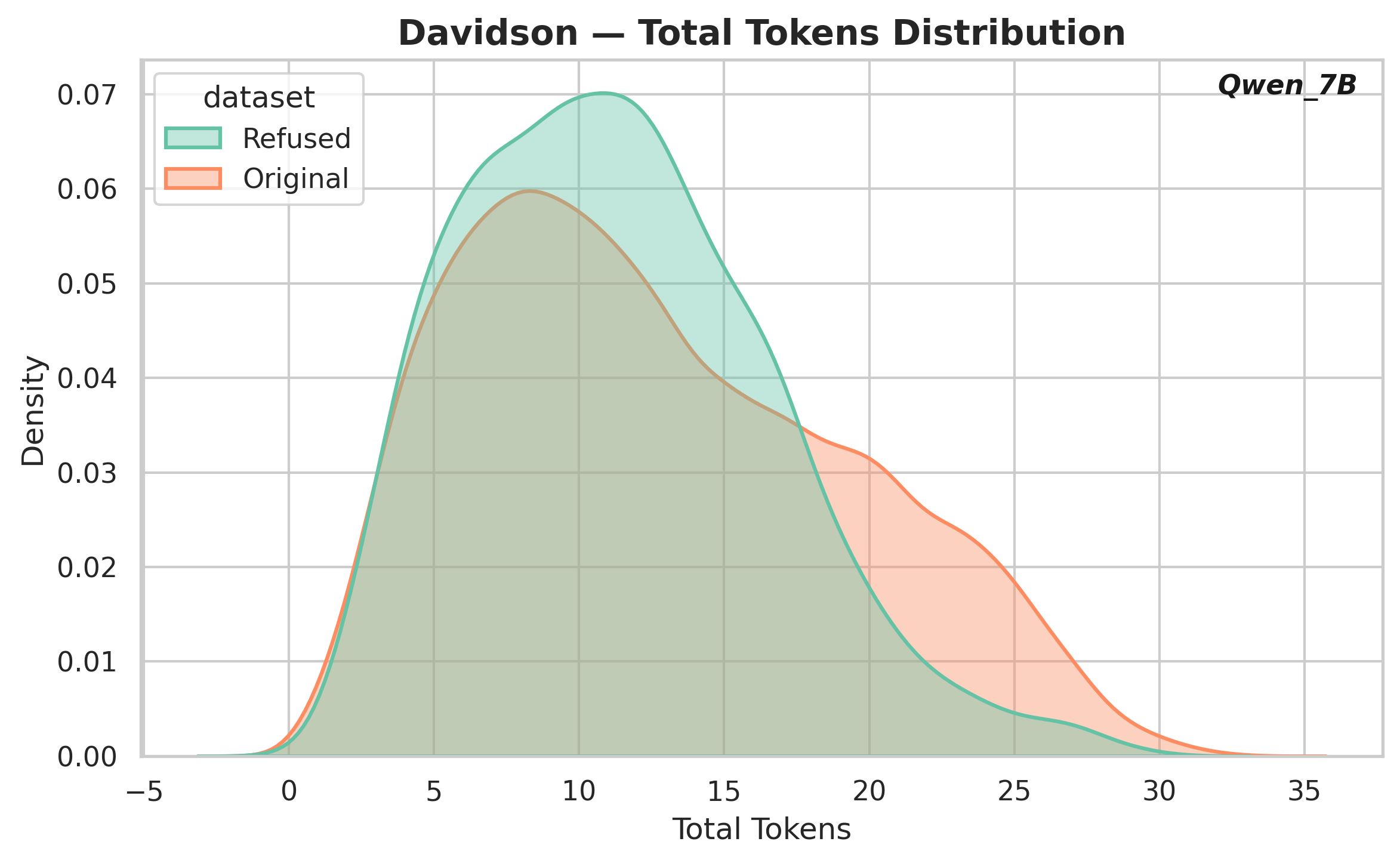}
    \caption{Davidson}
  \end{subfigure}\hfill
  \begin{subfigure}{0.32\textwidth}
    \centering
    \includegraphics[width=\linewidth]{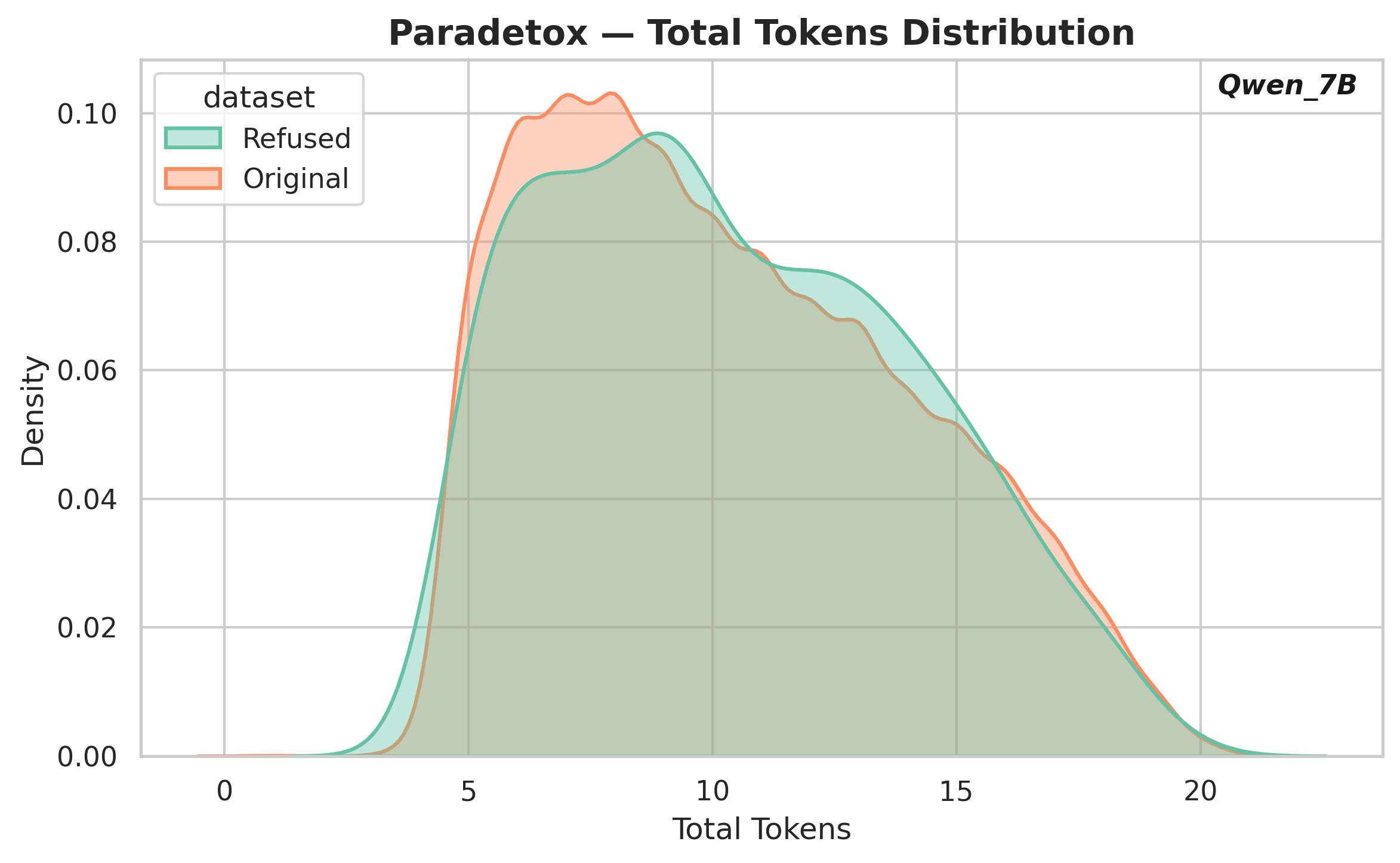}
    \caption{Paradetox}
  \end{subfigure}\hfill
  \begin{subfigure}{0.32\textwidth}
    \centering
    \includegraphics[width=\linewidth]{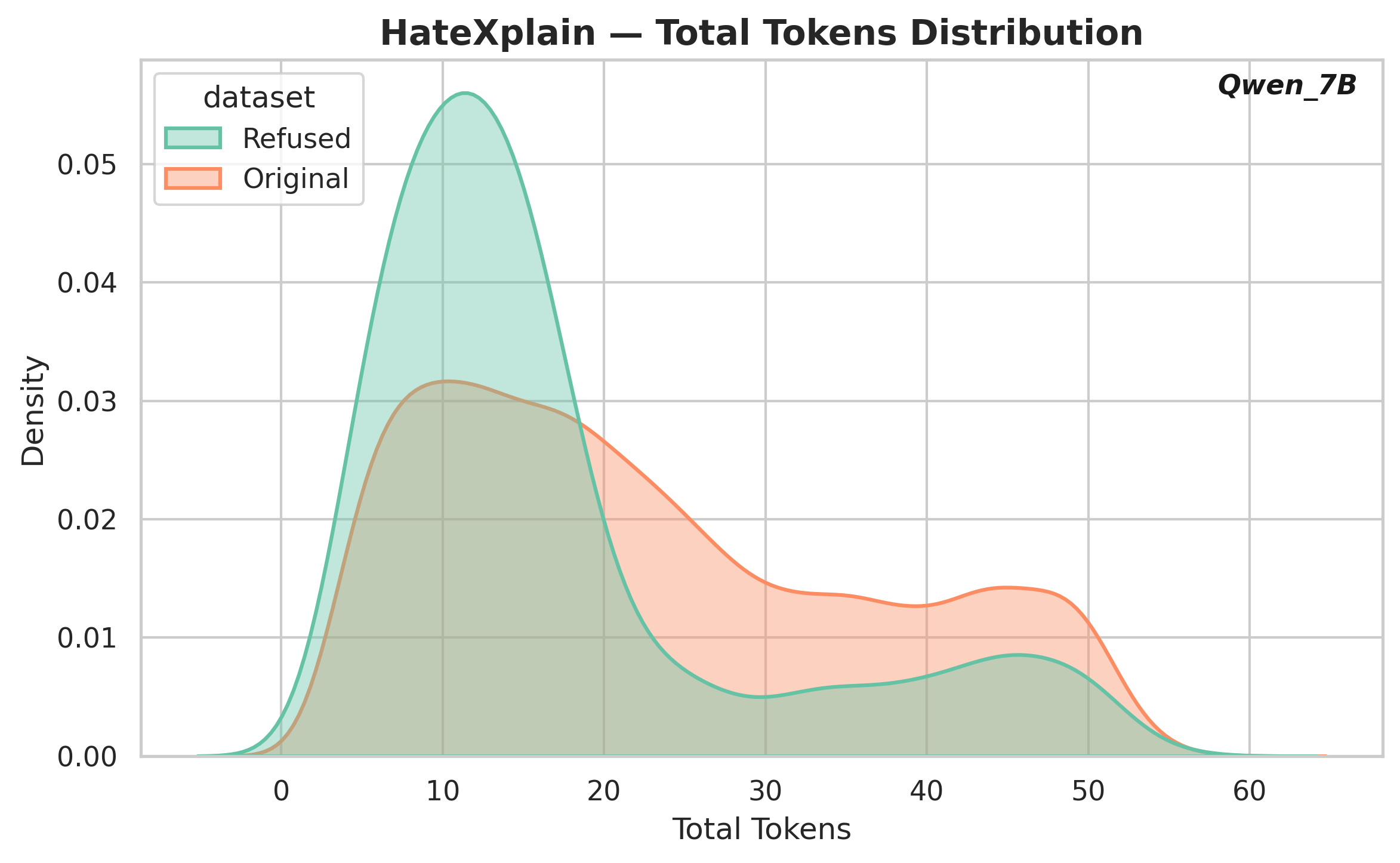}
    \caption{HateXplain}
  \end{subfigure}

  \vspace{0.4em}

  \begin{subfigure}{0.32\textwidth}
    \centering
    \includegraphics[width=\linewidth]{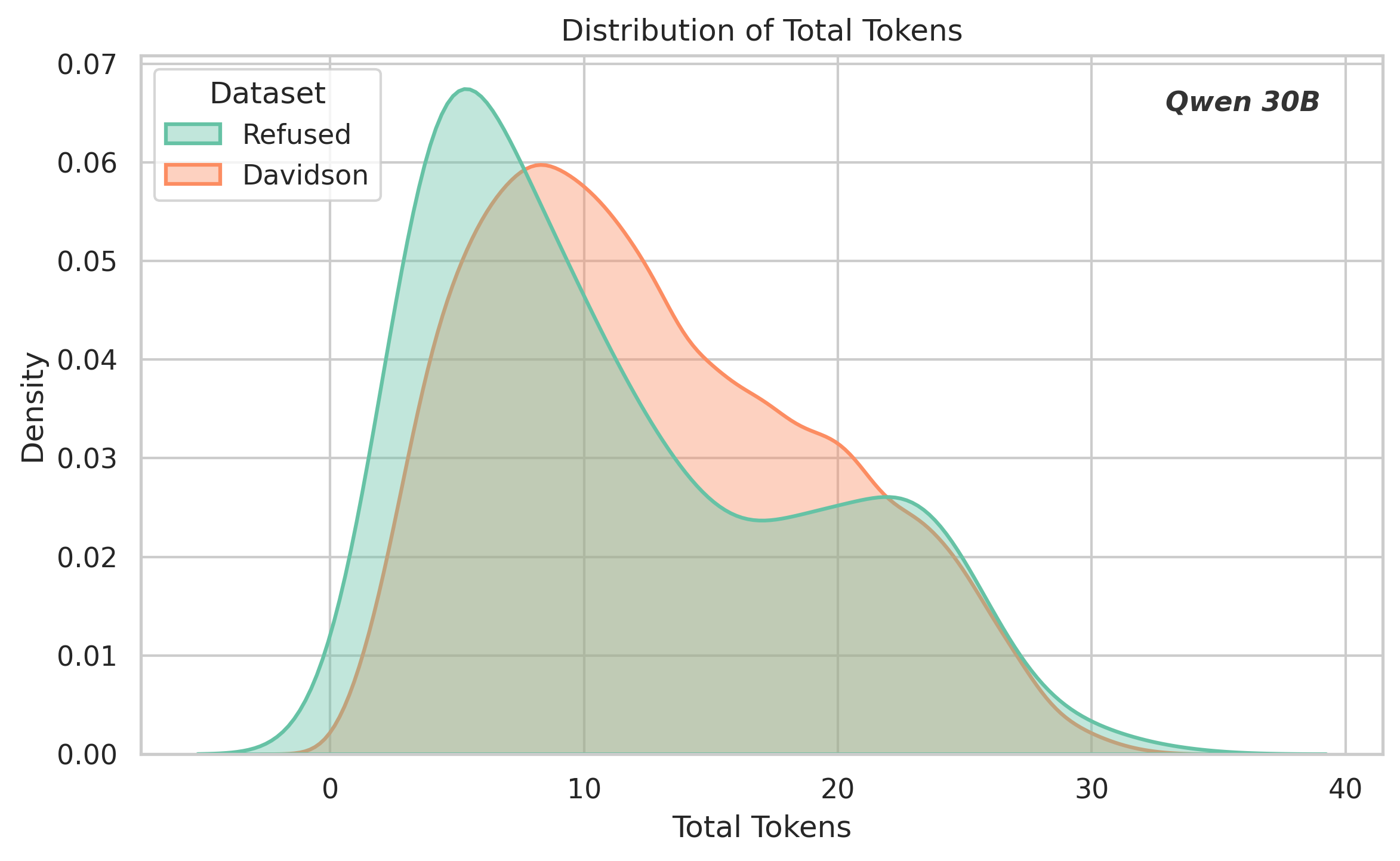}
    \caption{Davidson}
  \end{subfigure}\hfill
  \begin{subfigure}{0.32\textwidth}
    \centering
    \includegraphics[width=\linewidth]{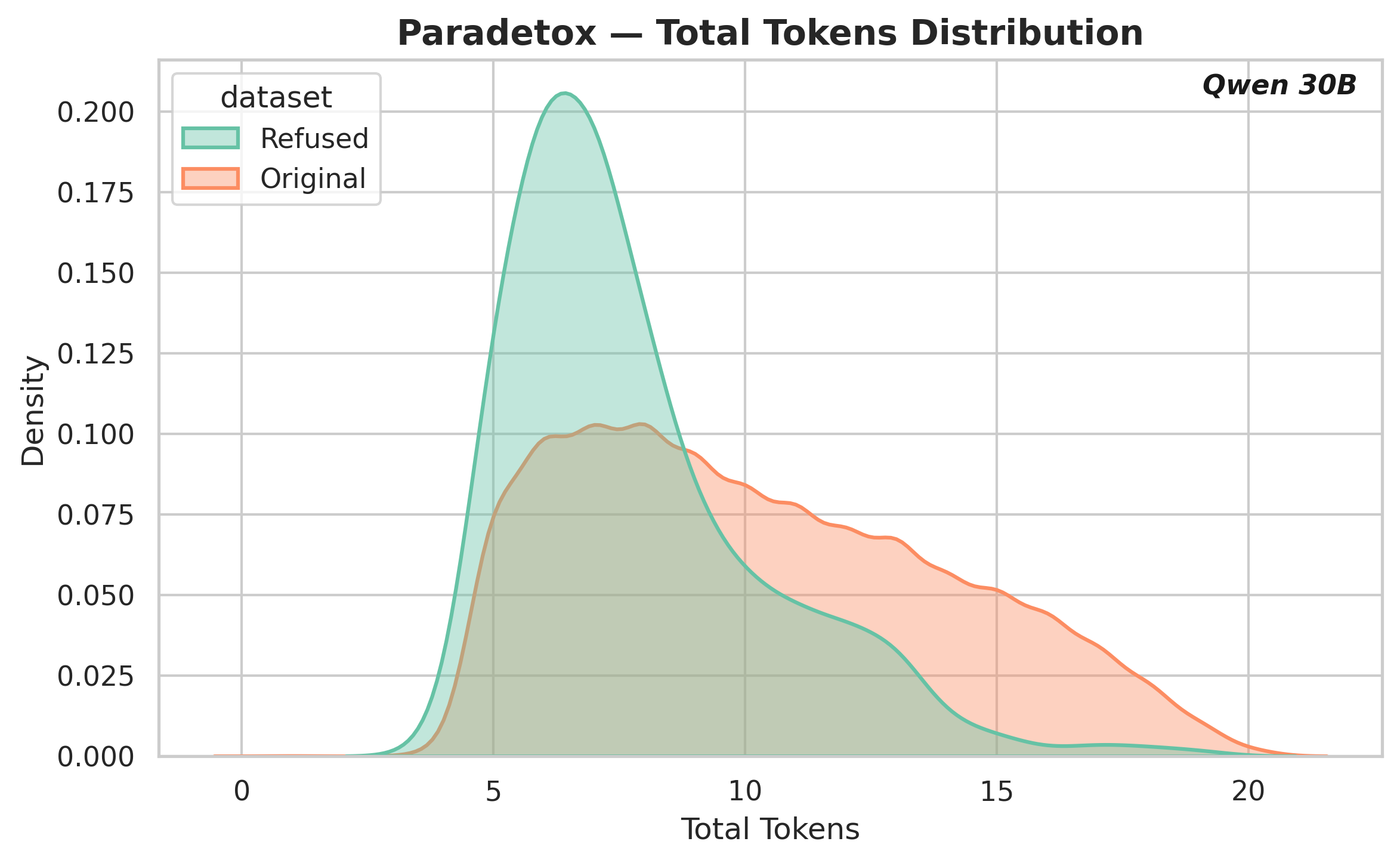}
    \caption{Paradetox}
  \end{subfigure}\hfill
  \begin{subfigure}{0.32\textwidth}
    \centering
    \includegraphics[width=\linewidth]{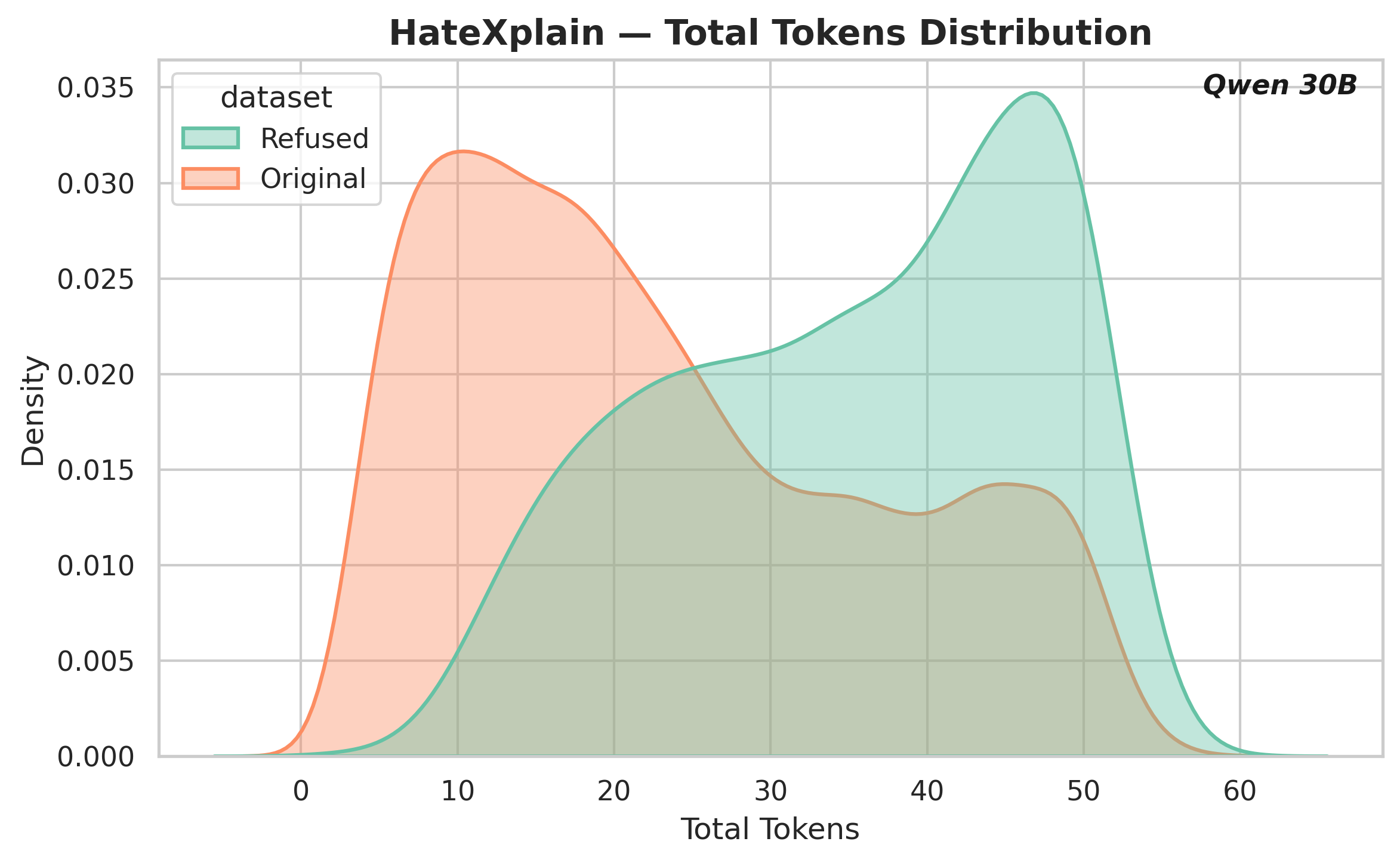}
    \caption{ HateXplain}
  \end{subfigure}

  \caption{Token count distributions across datasets for Qwen2.5 7B and Qwen3 30B.}
  \label{fig:token_count_qwen}
\end{figure*}

\begin{figure*}[htbp]
  \centering
  \begin{subfigure}{0.32\textwidth}
    \centering
    \includegraphics[width=\linewidth]{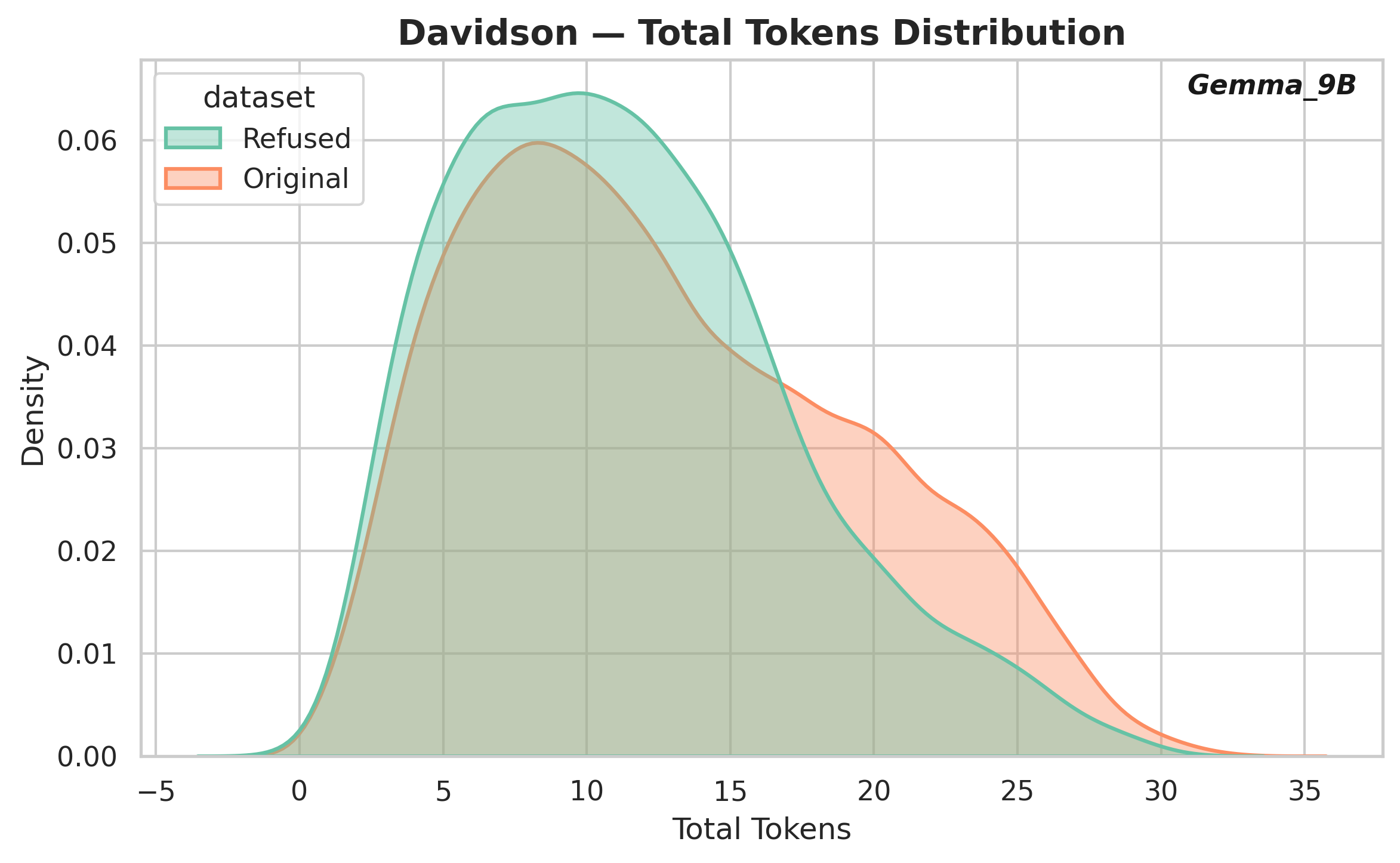}
    \caption{Davidson}
  \end{subfigure}\hfill
  \begin{subfigure}{0.32\textwidth}
    \centering
    \includegraphics[width=\linewidth]{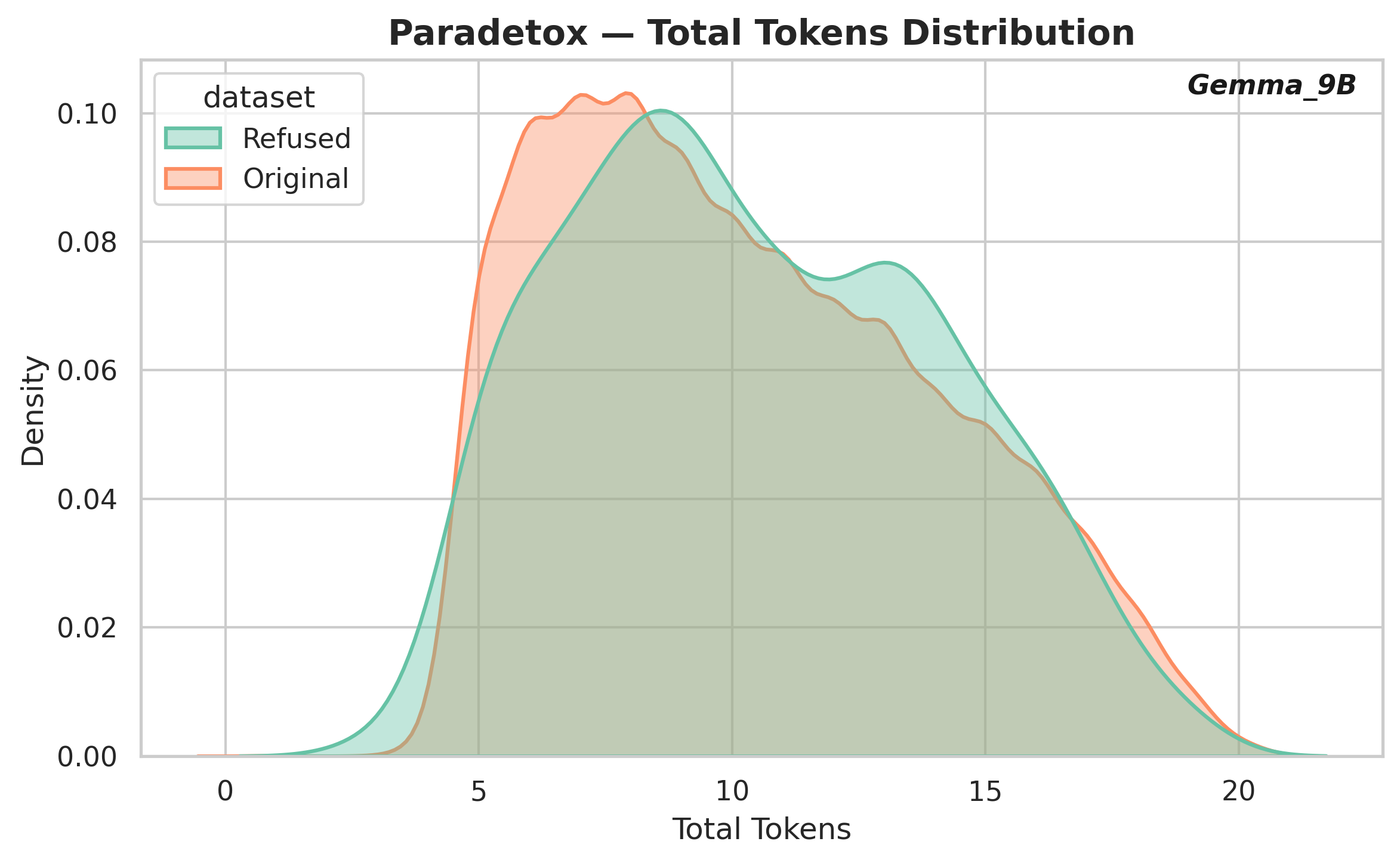}
    \caption{Paradetox}
  \end{subfigure}\hfill
  \begin{subfigure}{0.32\textwidth}
    \centering
    \includegraphics[width=\linewidth]{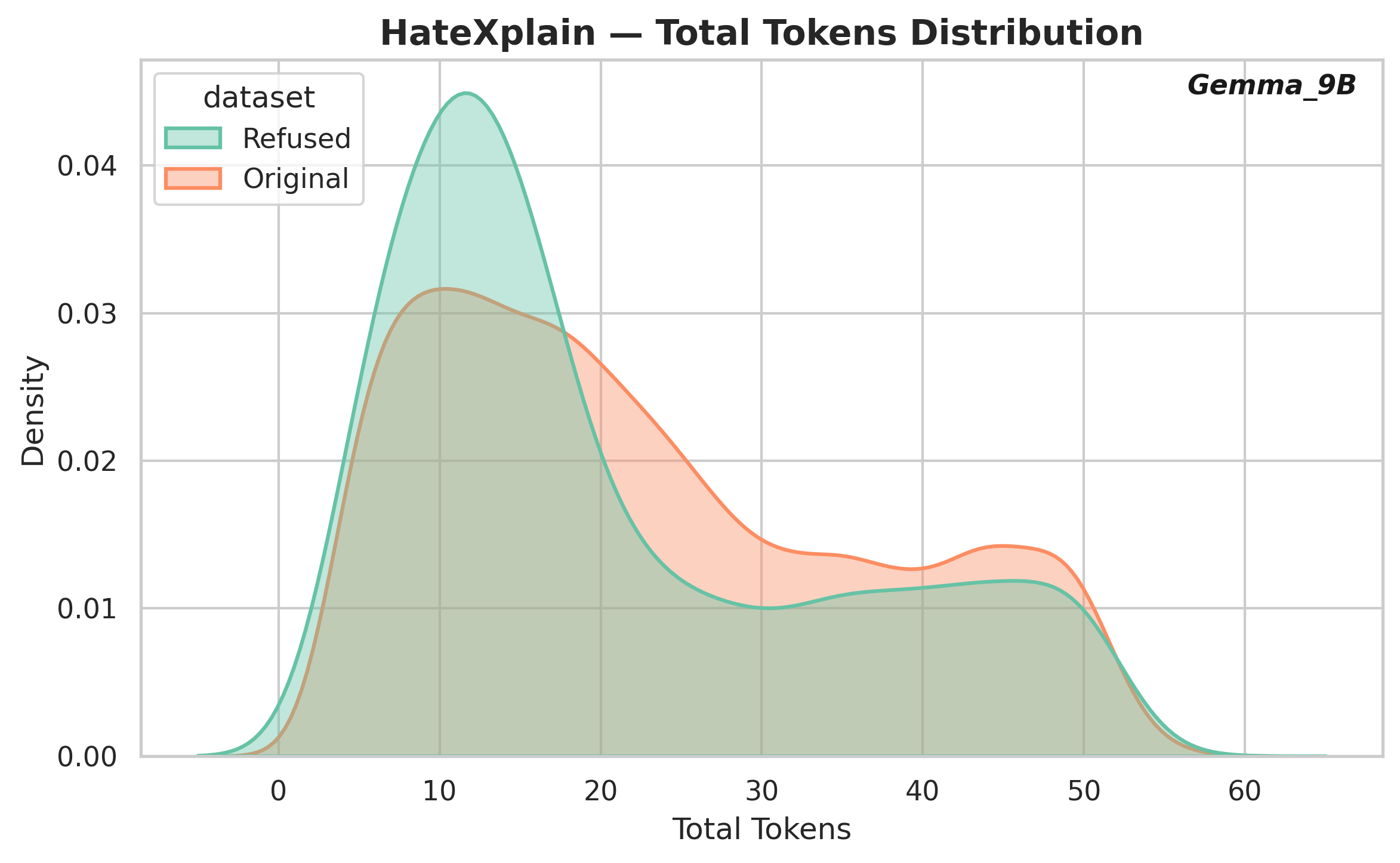}
    \caption{HateXplain}
  \end{subfigure}

  \vspace{0.4em}

  \begin{subfigure}{0.32\textwidth}
    \centering
    \includegraphics[width=\linewidth]{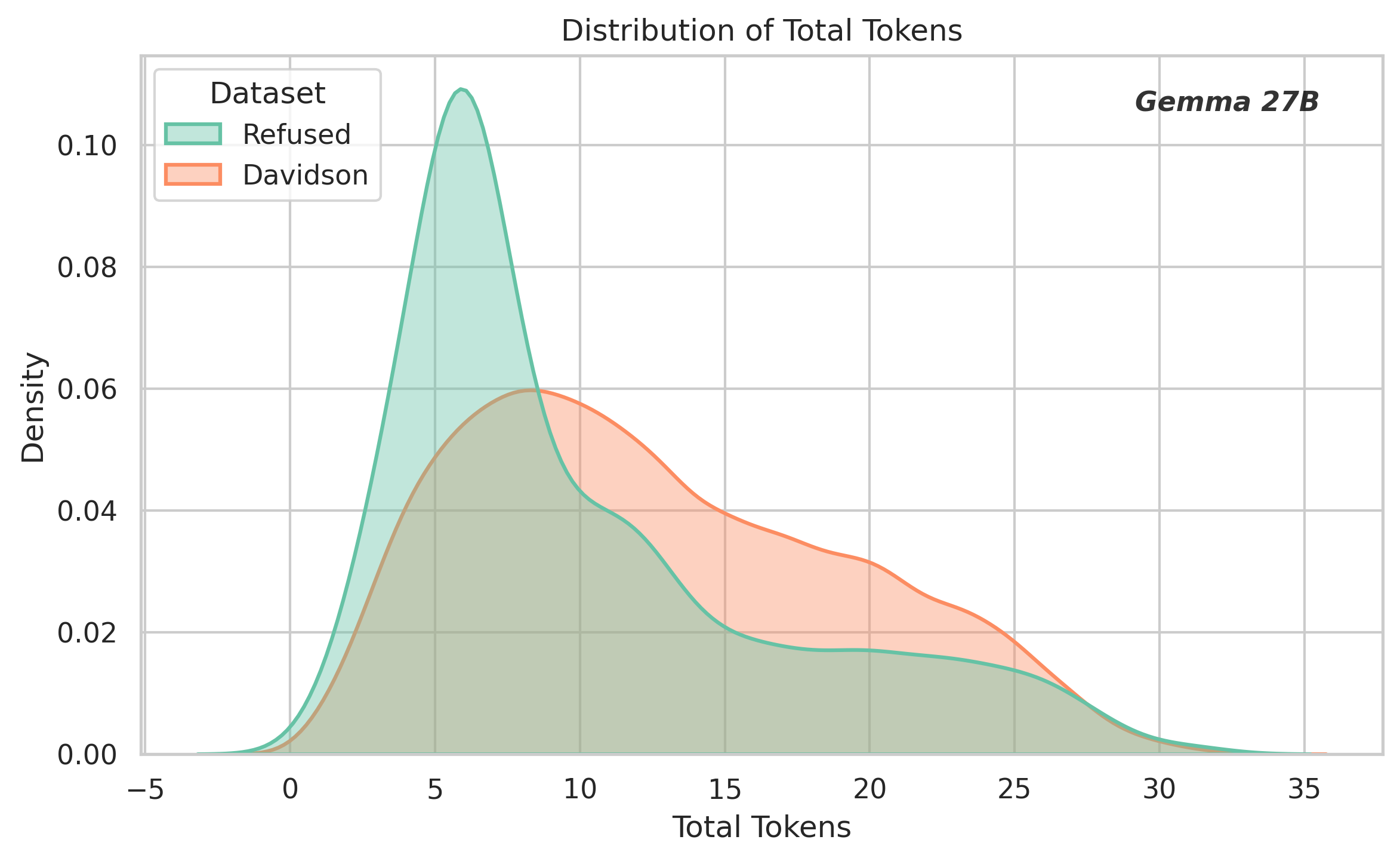}
    \caption{Davidson}
  \end{subfigure}\hfill
  \begin{subfigure}{0.32\textwidth}
    \centering
    \includegraphics[width=\linewidth]{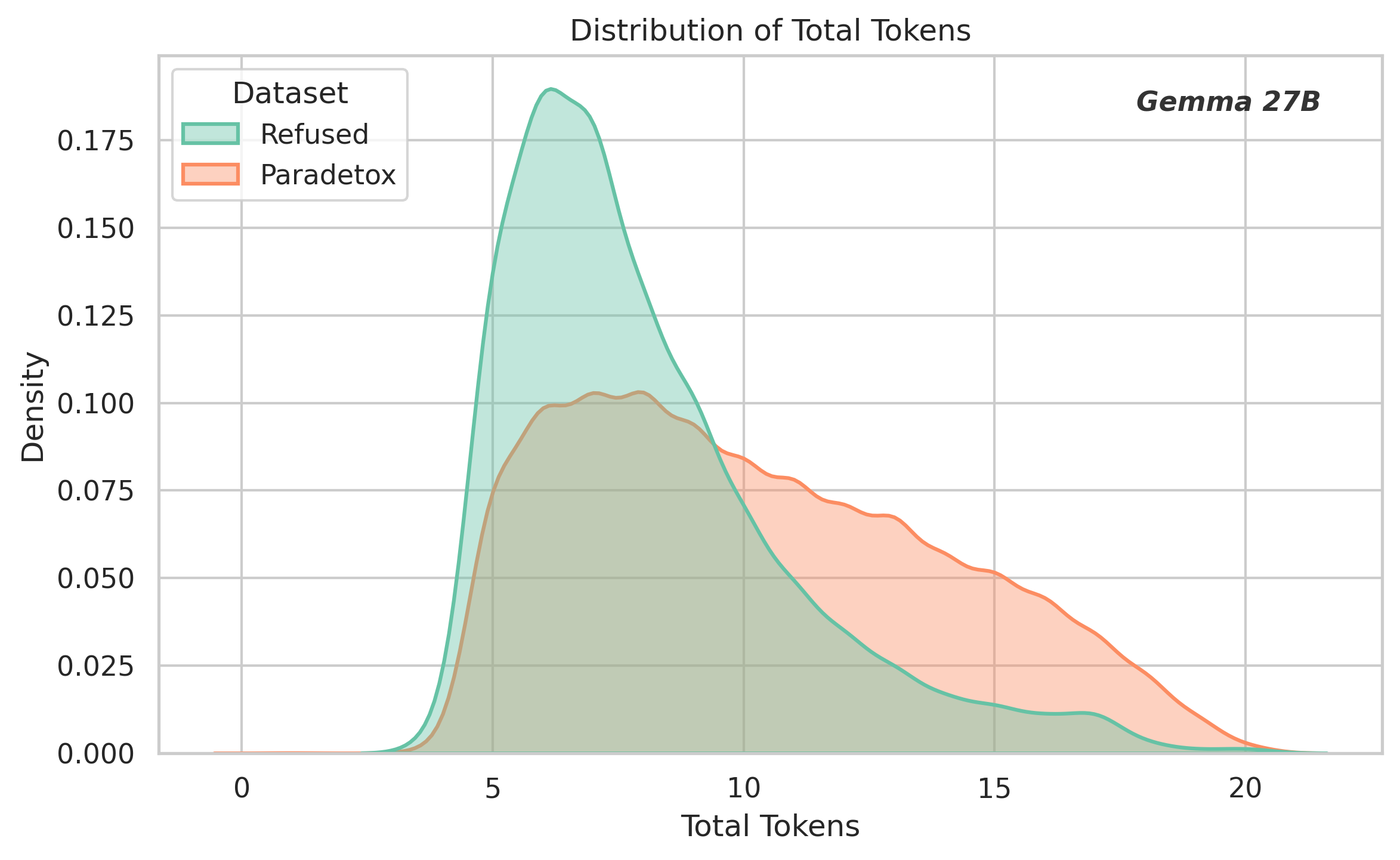}
    \caption{Paradetox}
  \end{subfigure}\hfill
  \begin{subfigure}{0.32\textwidth}
    \centering
    \includegraphics[width=\linewidth]{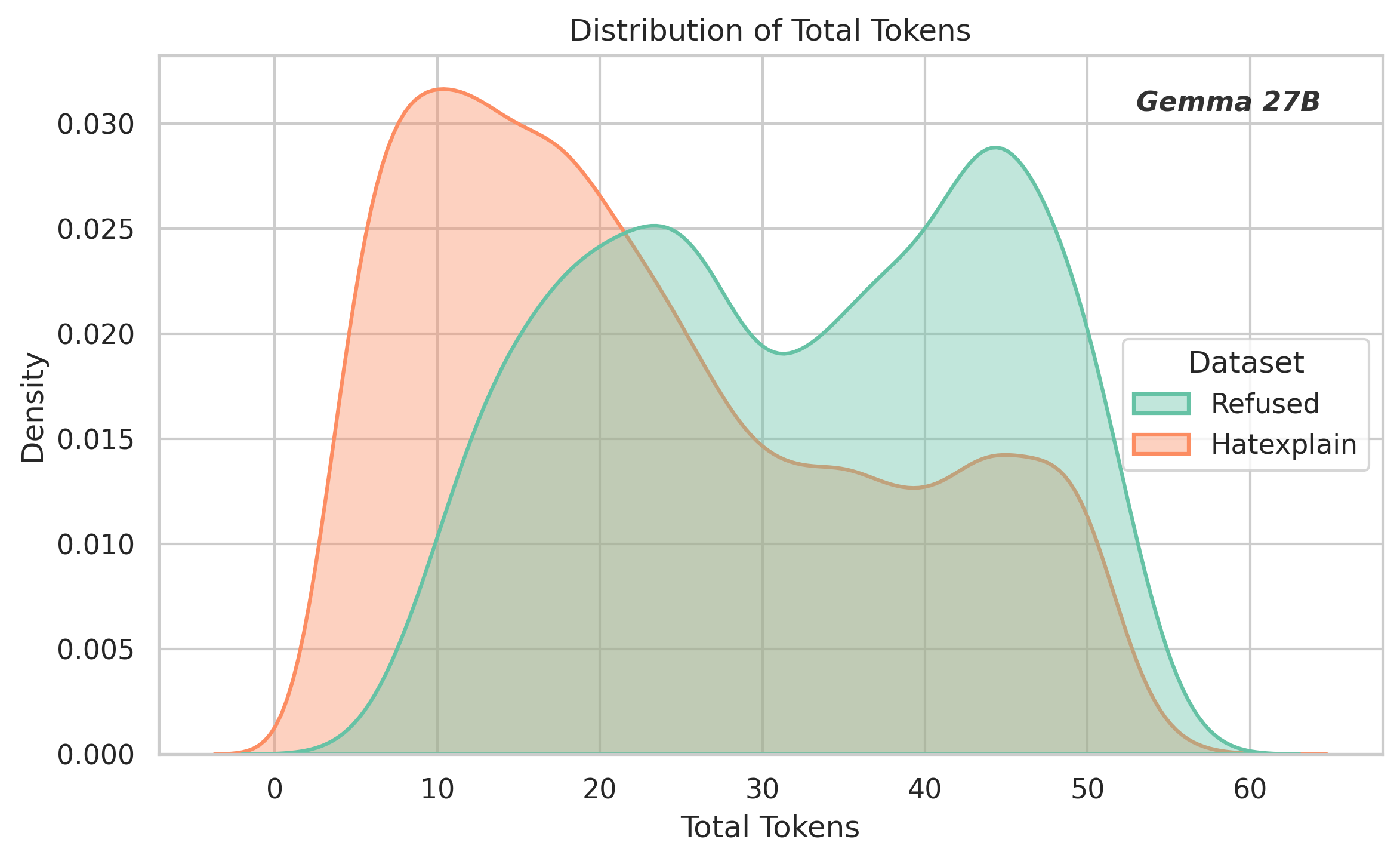}
    \caption{HateXplain}
  \end{subfigure}

  \caption{Token count distributions across datasets for Gemma2 9B and Gemma3 27B.}
  \label{fig:token_count_gemma}
\end{figure*}

\paragraph{Token Length.}
The token length distributions of false refusals for GPT, Llama, Mistral, Qwen, and Gemma models are shown in Figures~\ref{fig:token_count_gpt}–\ref{fig:token_count_gemma}.

\begin{figure*}[htbp]
  \centering
  \setlength{\tabcolsep}{3pt}

  \begin{subfigure}{0.32\textwidth}
    \centering
    \includegraphics[width=\linewidth]{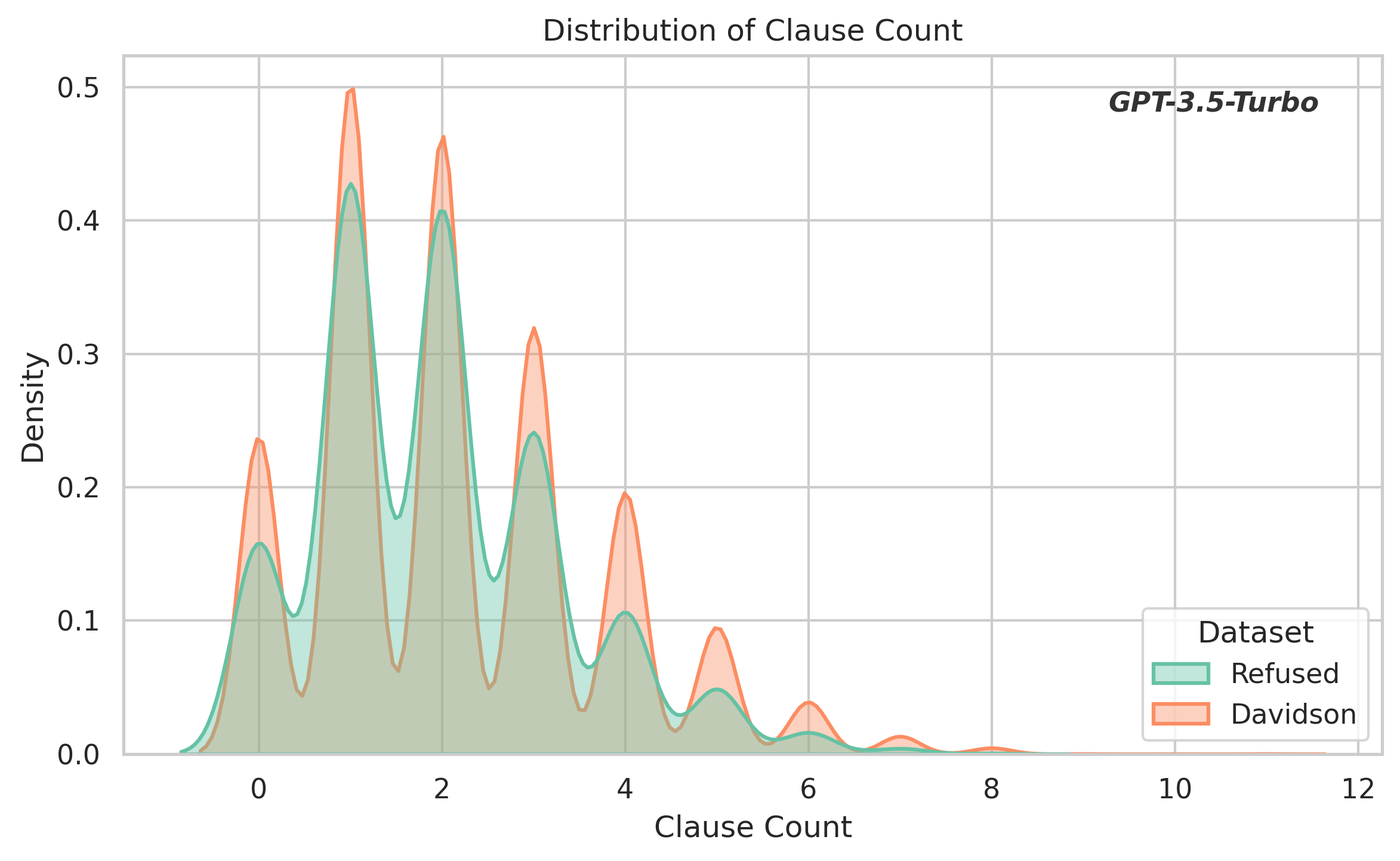}
    \caption{Davidson}
  \end{subfigure}\hfill
  \begin{subfigure}{0.32\textwidth}
    \centering
    \includegraphics[width=\linewidth]{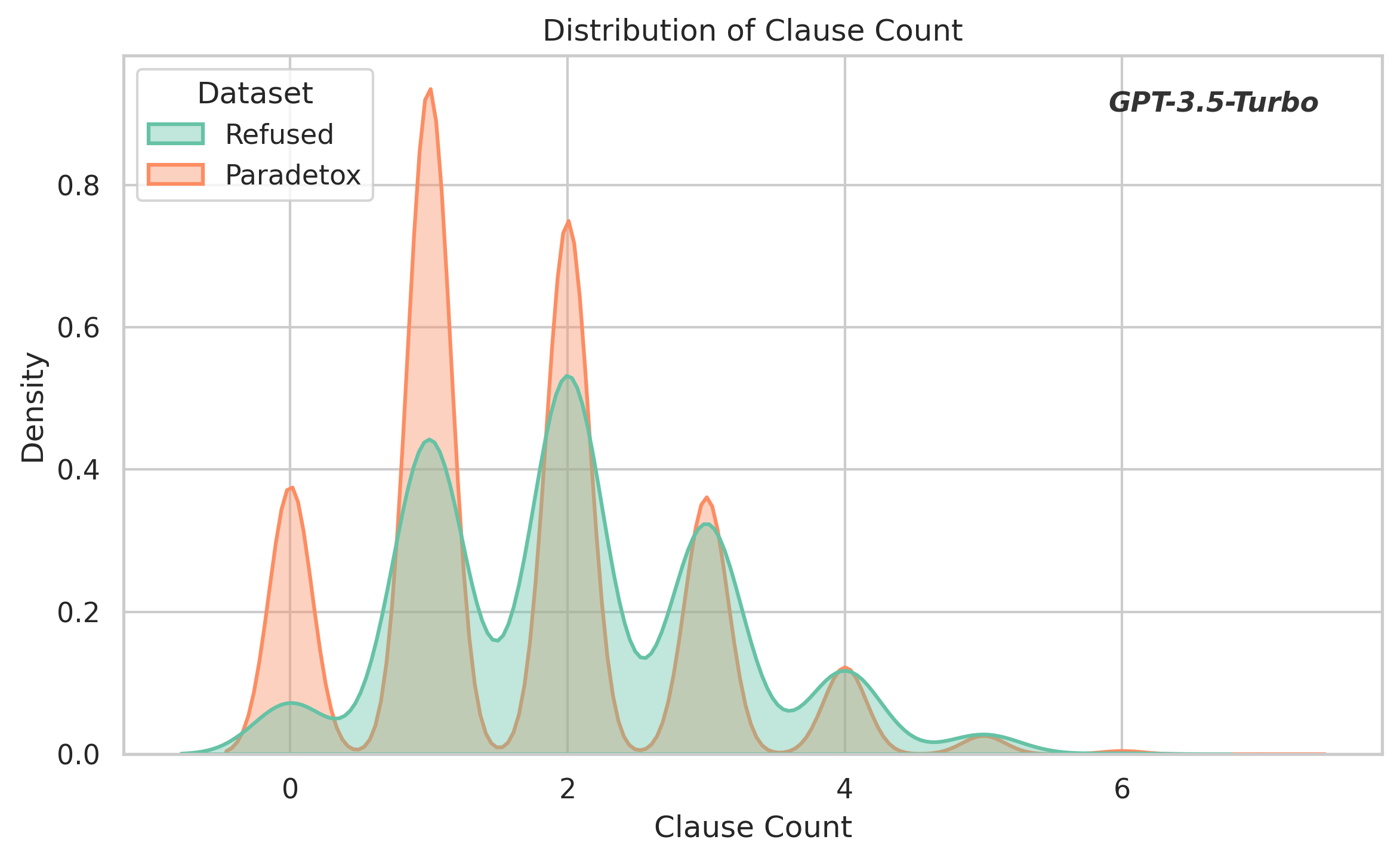}
    \caption{Paradetox}
  \end{subfigure}\hfill
  \begin{subfigure}{0.32\textwidth}
    \centering
    \includegraphics[width=\linewidth]{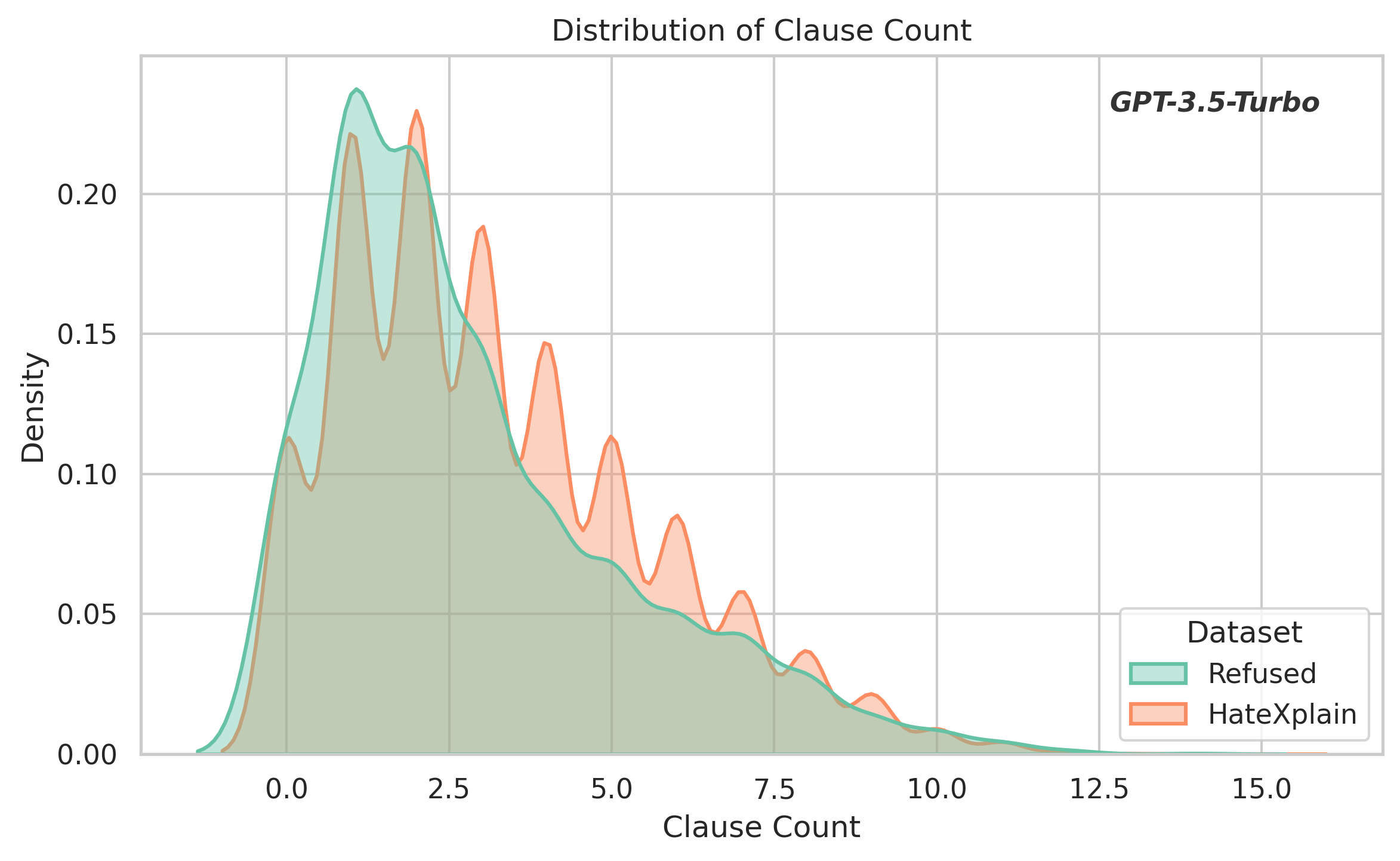}
    \caption{HateXplain}
  \end{subfigure}

  \vspace{0.4em}

  \begin{subfigure}{0.32\textwidth}
    \centering
    \includegraphics[width=\linewidth]{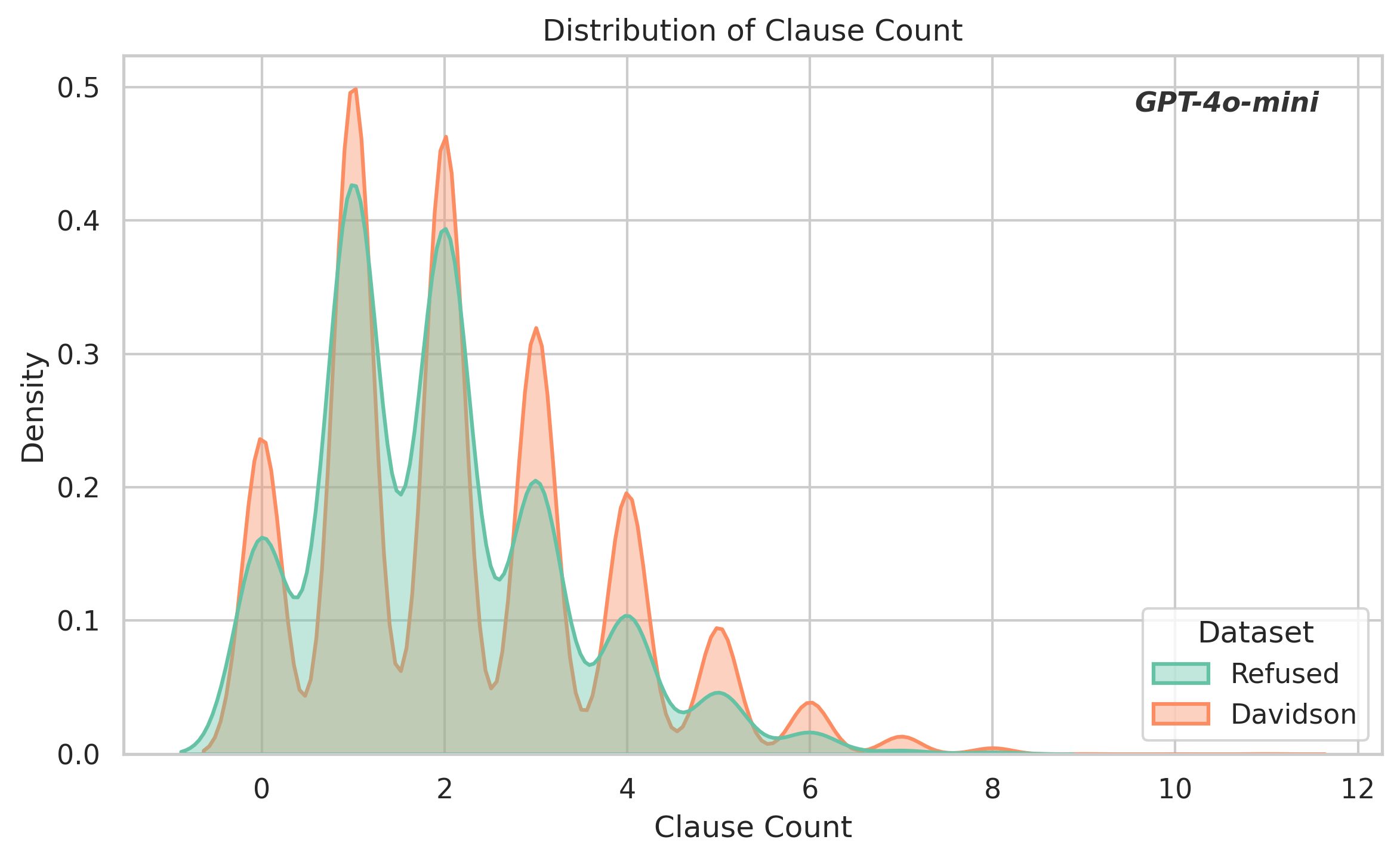}
    \caption{Davidson}
  \end{subfigure}\hfill
  \begin{subfigure}{0.32\textwidth}
    \centering
    \includegraphics[width=\linewidth]{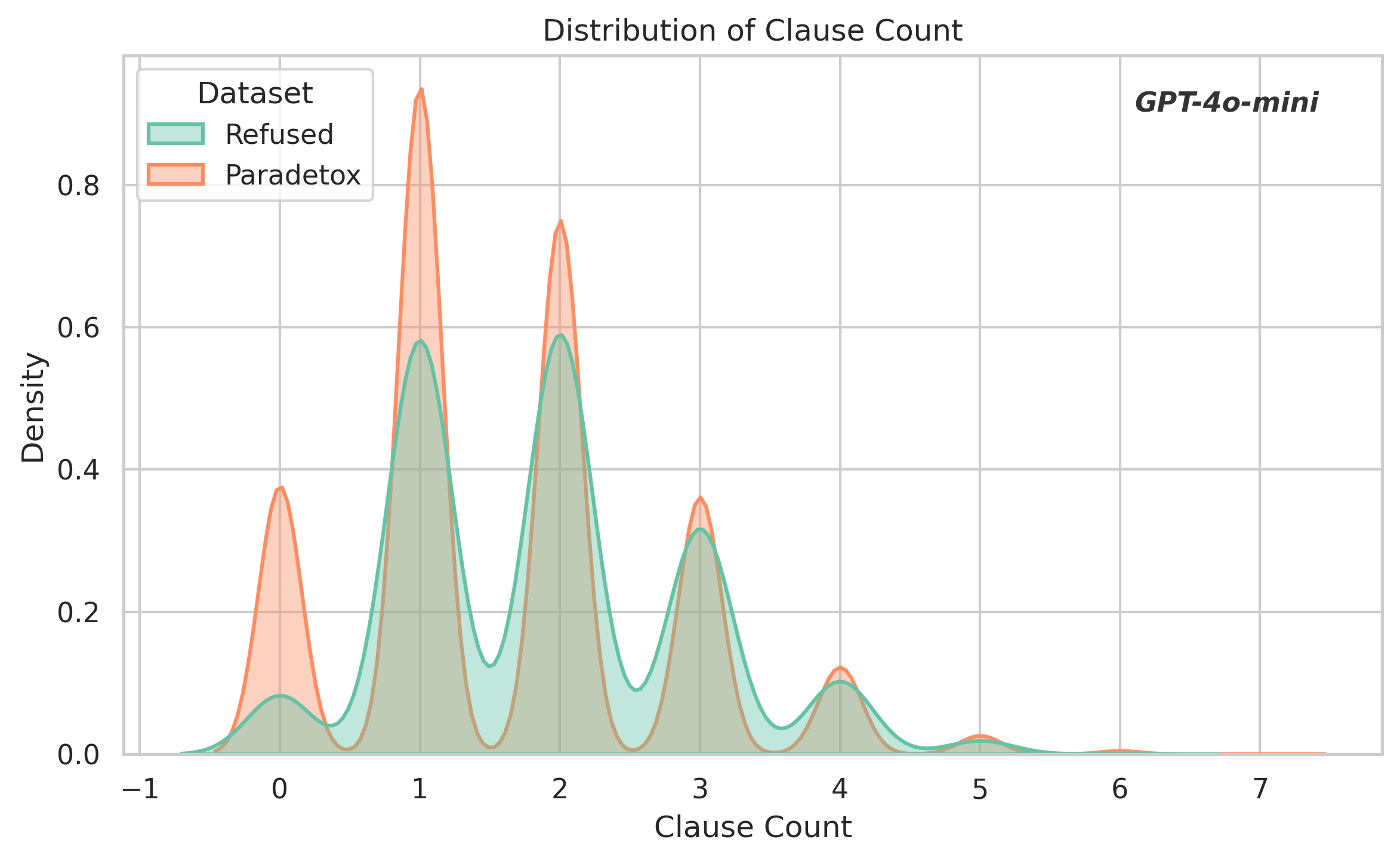}
    \caption{Paradetox}
  \end{subfigure}\hfill
  \begin{subfigure}{0.32\textwidth}
    \centering
    \includegraphics[width=\linewidth]{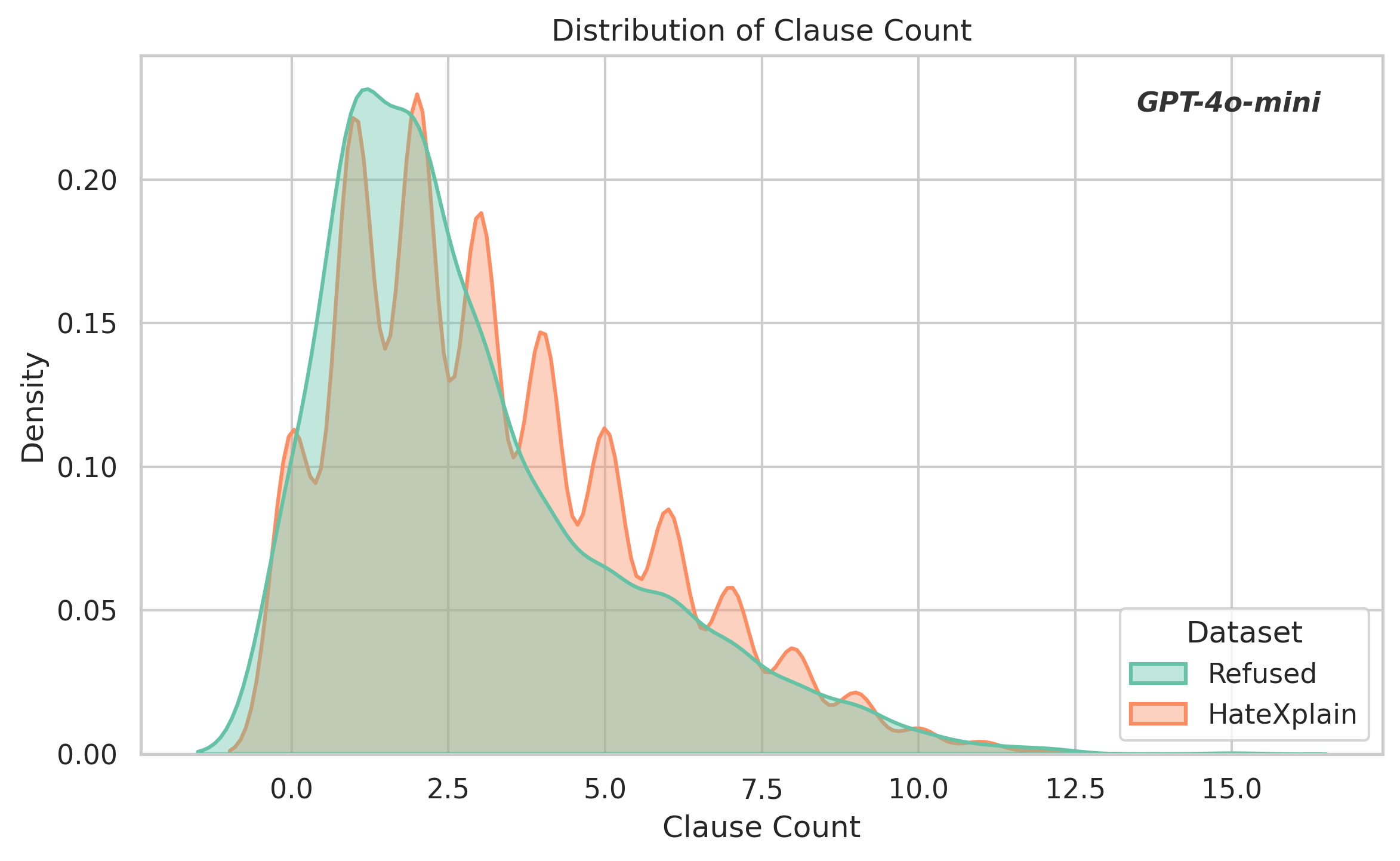}
    \caption{HateXplain}
  \end{subfigure}

  \caption{Clause count distributions across datasets for GPT-3.5-Turbo and GPT-4o-mini.}
  \label{fig:clause_count_gpt}
\end{figure*}

\begin{figure*}[htbp]
  \centering
  \begin{subfigure}{0.32\textwidth}
    \centering
    \includegraphics[width=\linewidth]{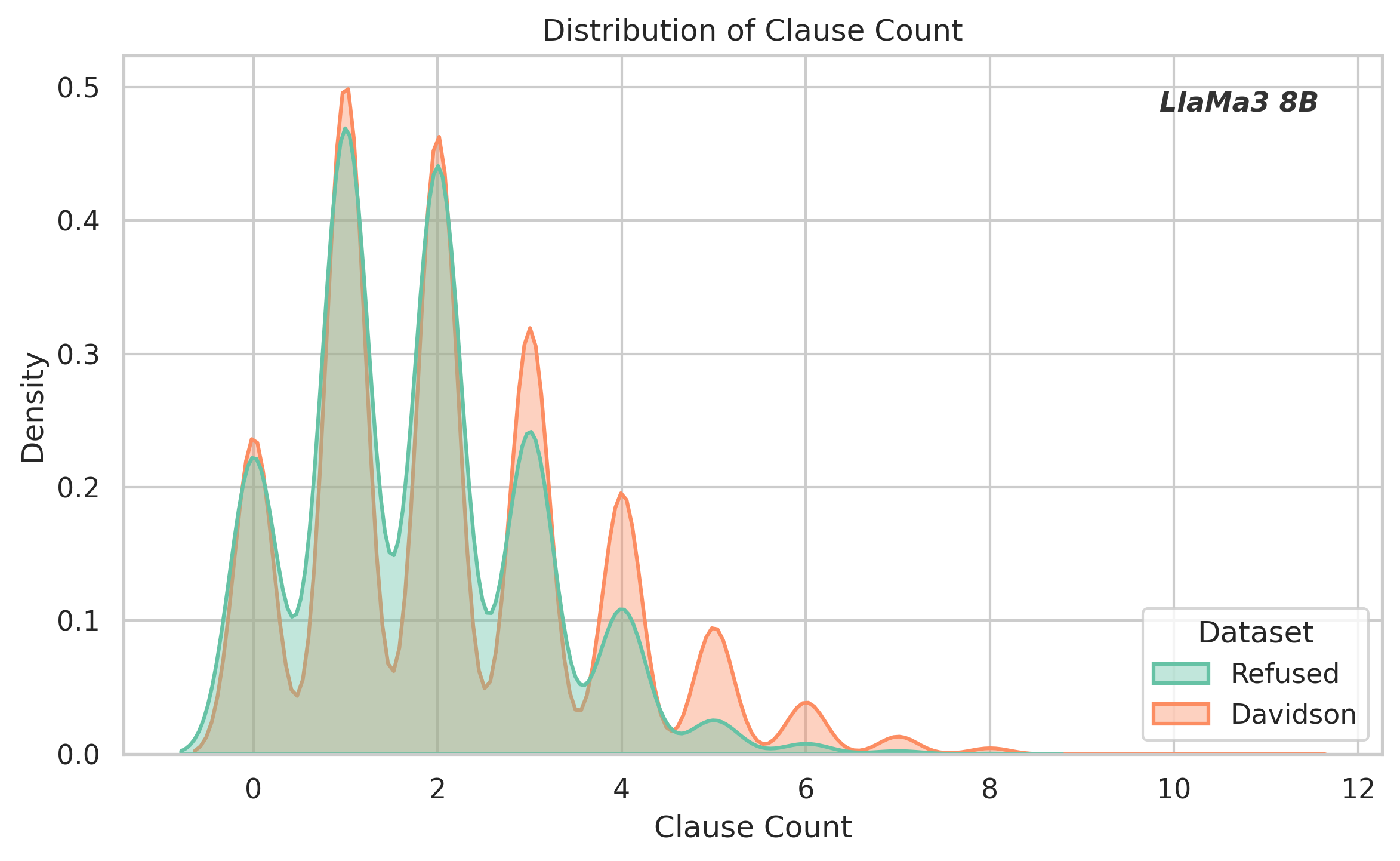}
    \caption{Davidson}
  \end{subfigure}\hfill
  \begin{subfigure}{0.32\textwidth}
    \centering
    \includegraphics[width=\linewidth]{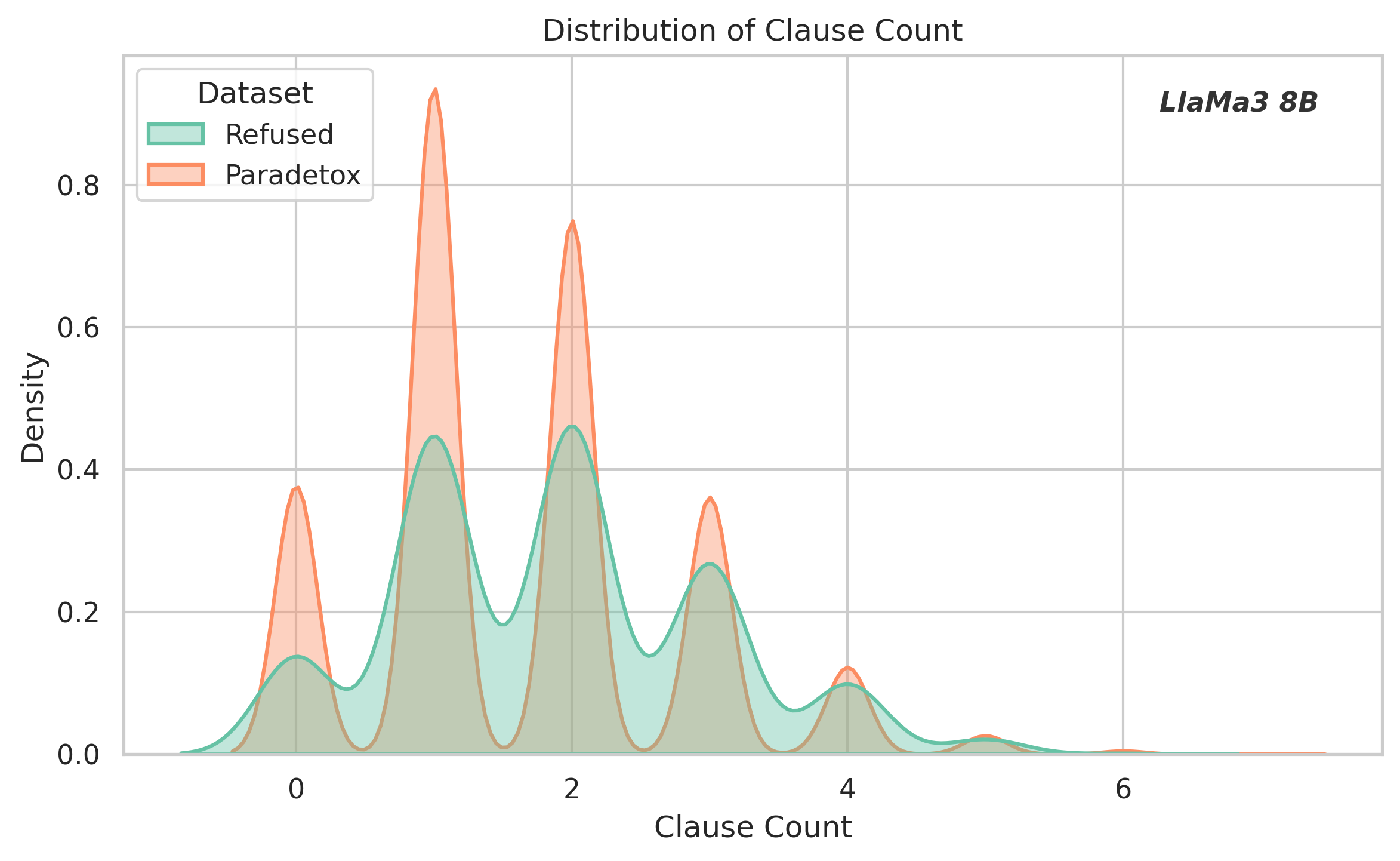}
    \caption{Paradetox}
  \end{subfigure}\hfill
  \begin{subfigure}{0.32\textwidth}
    \centering
    \includegraphics[width=\linewidth]{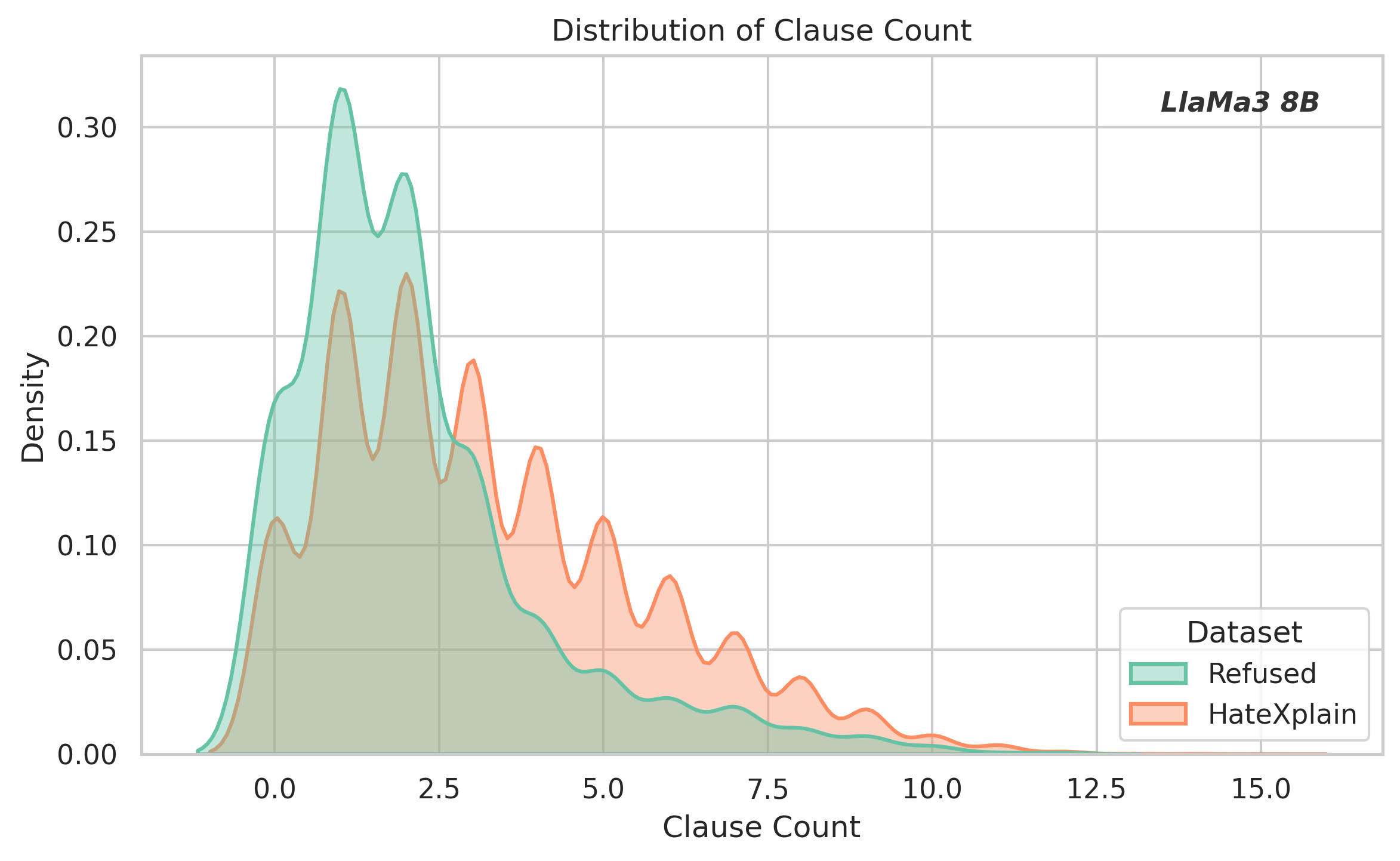}
    \caption{HateXplain}
  \end{subfigure}

  \caption{Clause count distributions across datasets for Llama3 8B.}
  \label{fig:clause_count_llama}
\end{figure*}

\begin{figure*}[htbp]
  \centering
  \begin{subfigure}{0.32\textwidth}
    \centering
    \includegraphics[width=\linewidth]{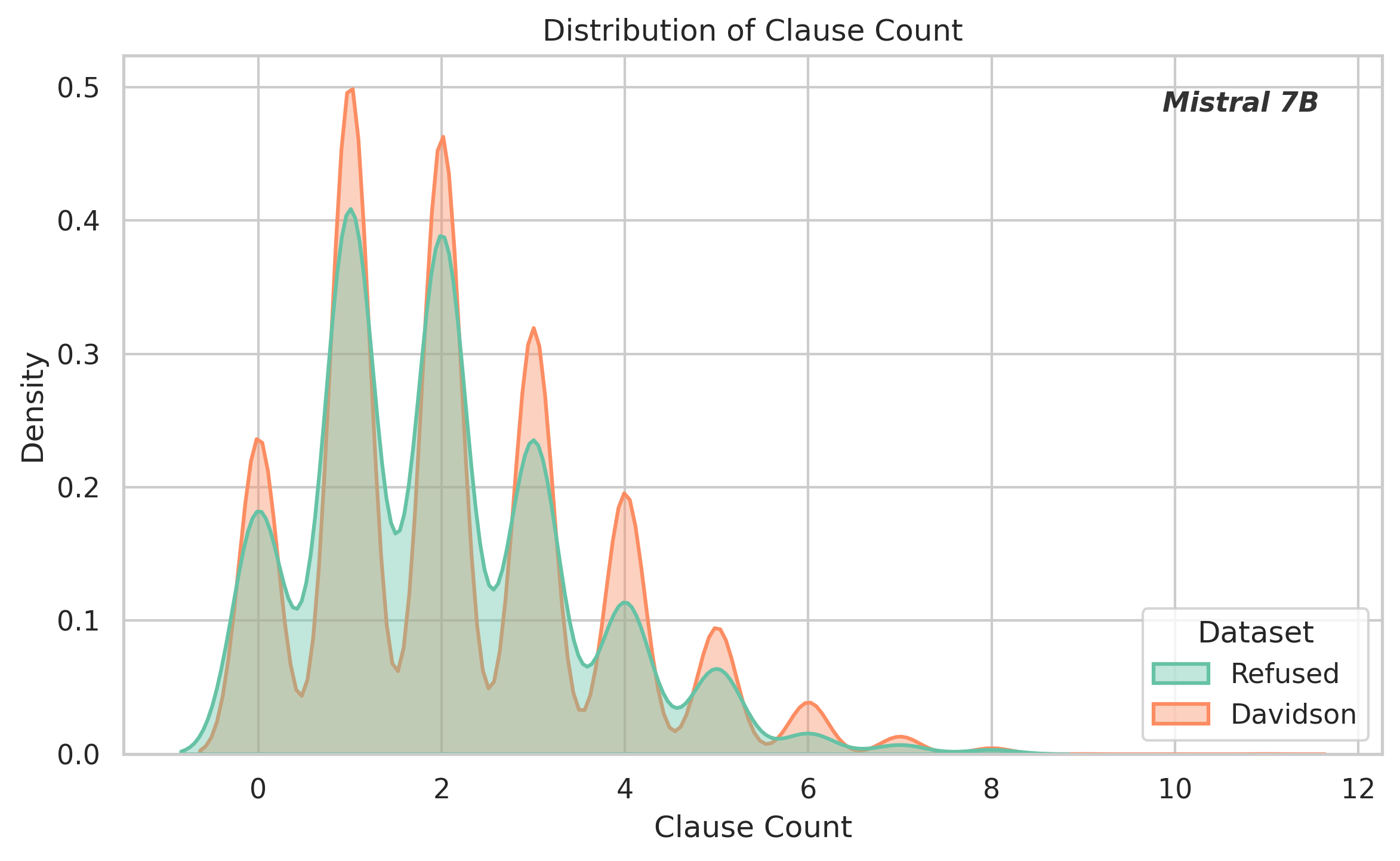}
    \caption{Davidson}
  \end{subfigure}\hfill
  \begin{subfigure}{0.32\textwidth}
    \centering
    \includegraphics[width=\linewidth]{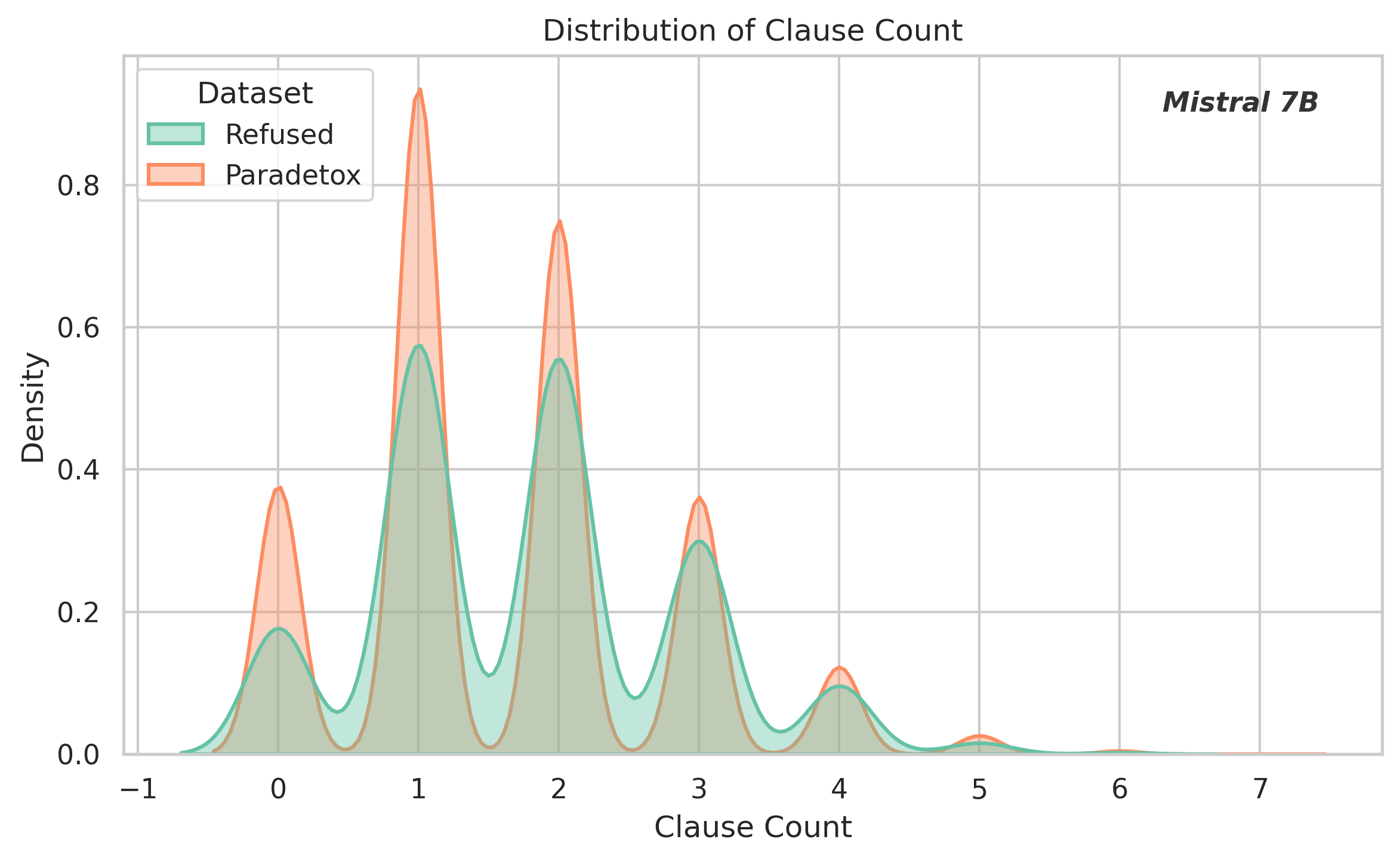}
    \caption{Paradetox}
  \end{subfigure}\hfill
  \begin{subfigure}{0.32\textwidth}
    \centering
    \includegraphics[width=\linewidth]{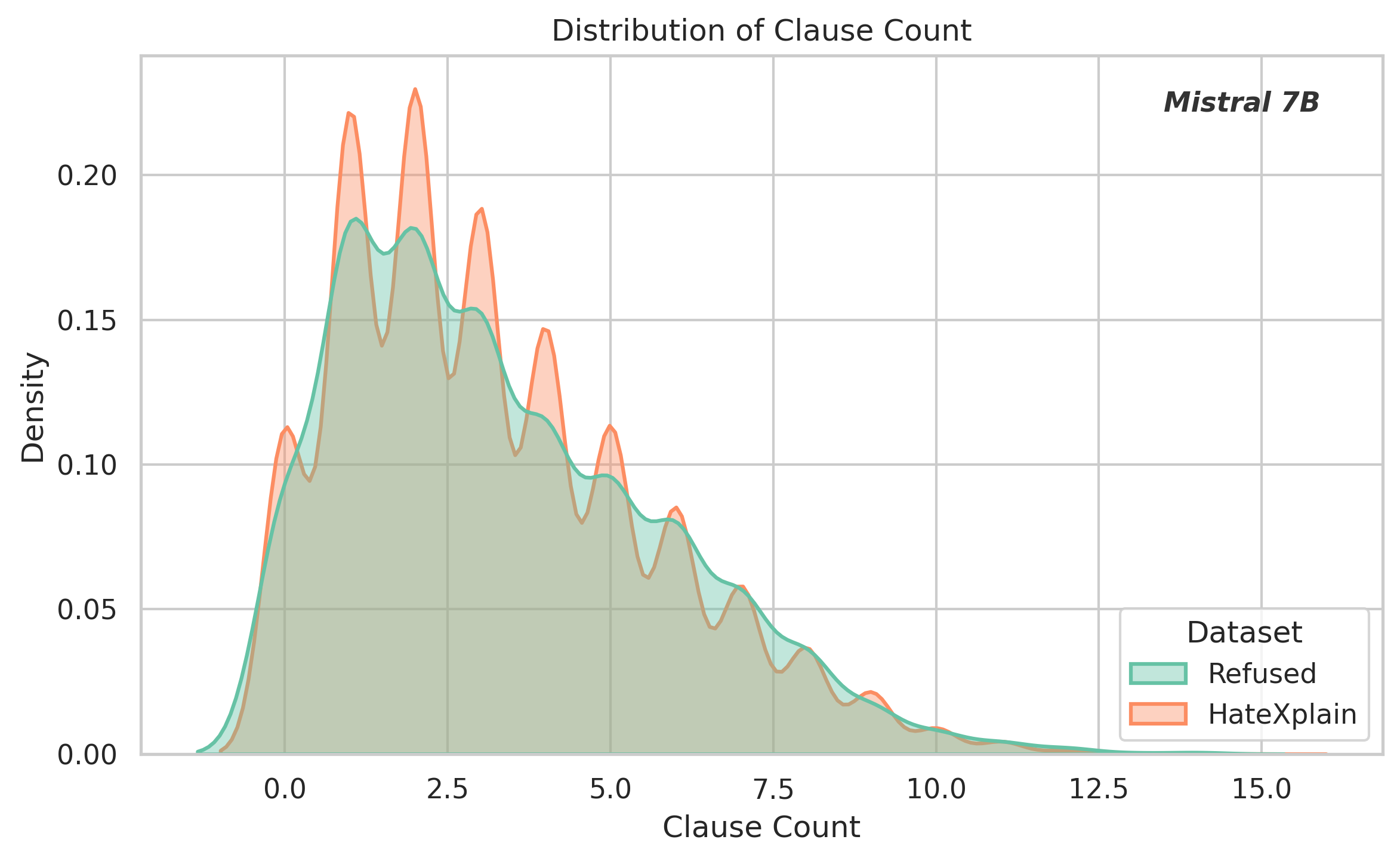}
    \caption{HateXplain}
  \end{subfigure}

  \vspace{0.4em}

  \begin{subfigure}{0.32\textwidth}
    \centering
    \includegraphics[width=\linewidth]{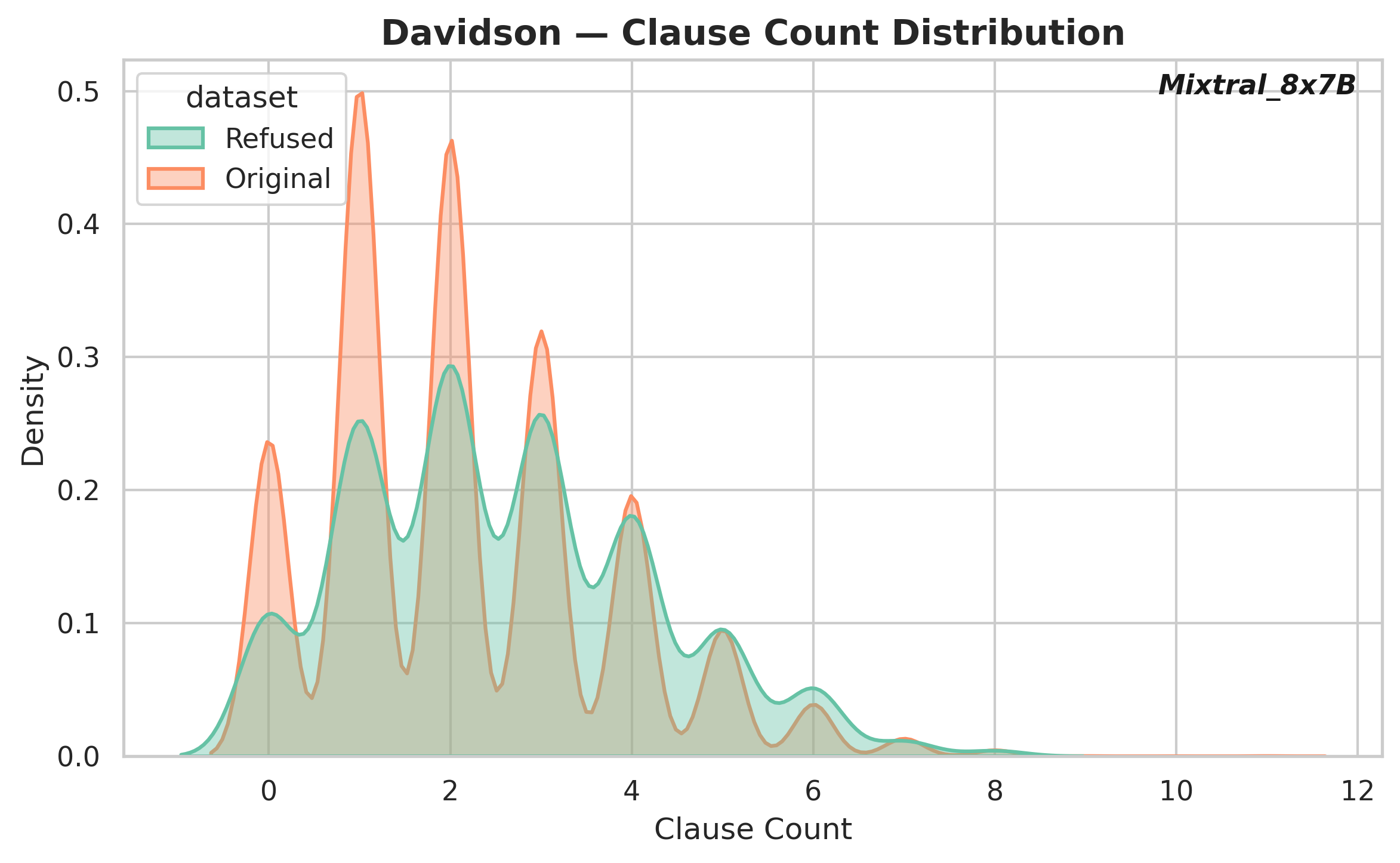}
    \caption{Davidson}
  \end{subfigure}\hfill
  \begin{subfigure}{0.32\textwidth}
    \centering
    \includegraphics[width=\linewidth]{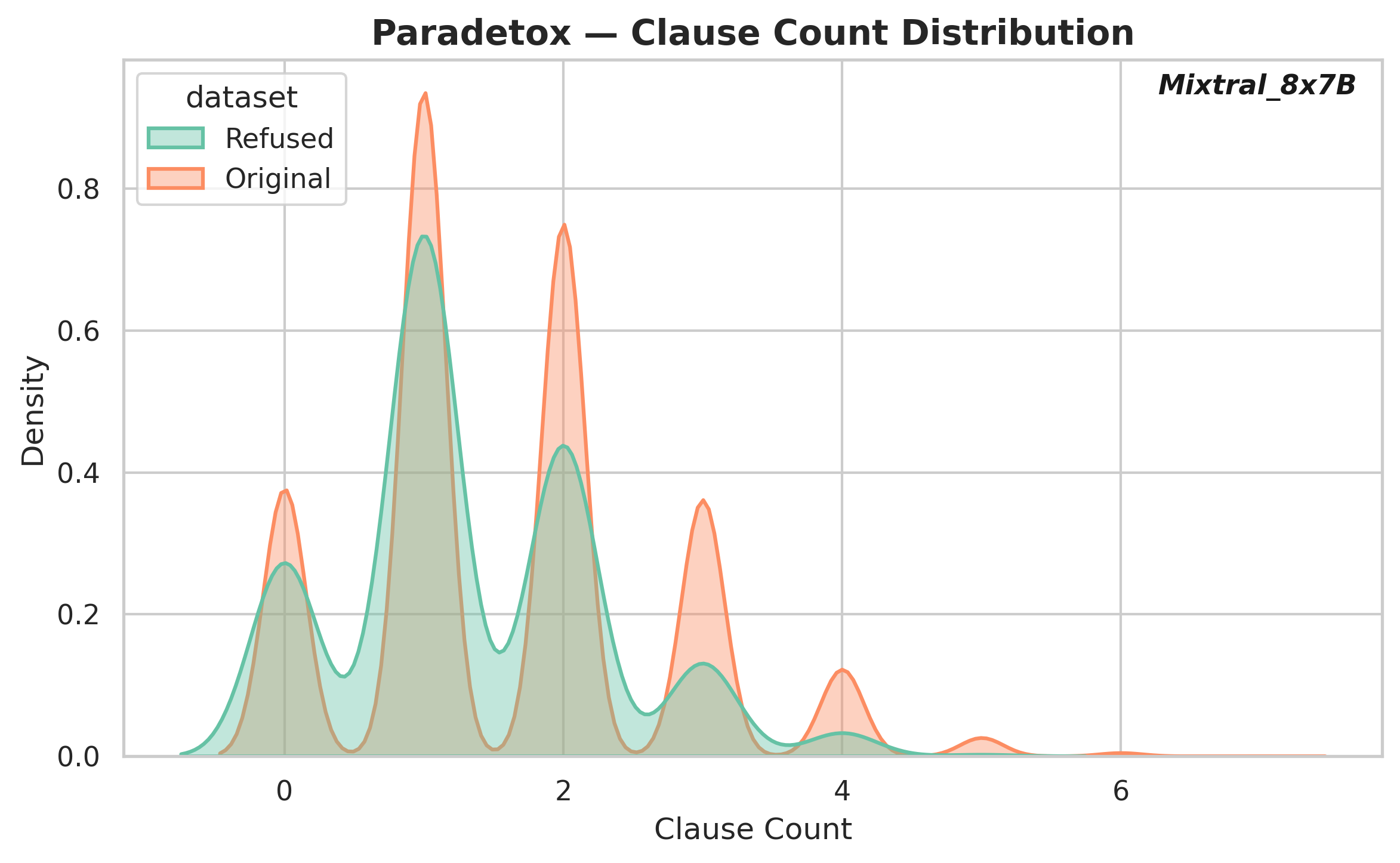}
    \caption{Parardetox}
  \end{subfigure}\hfill
  \begin{subfigure}{0.32\textwidth}
    \centering
    \includegraphics[width=\linewidth]{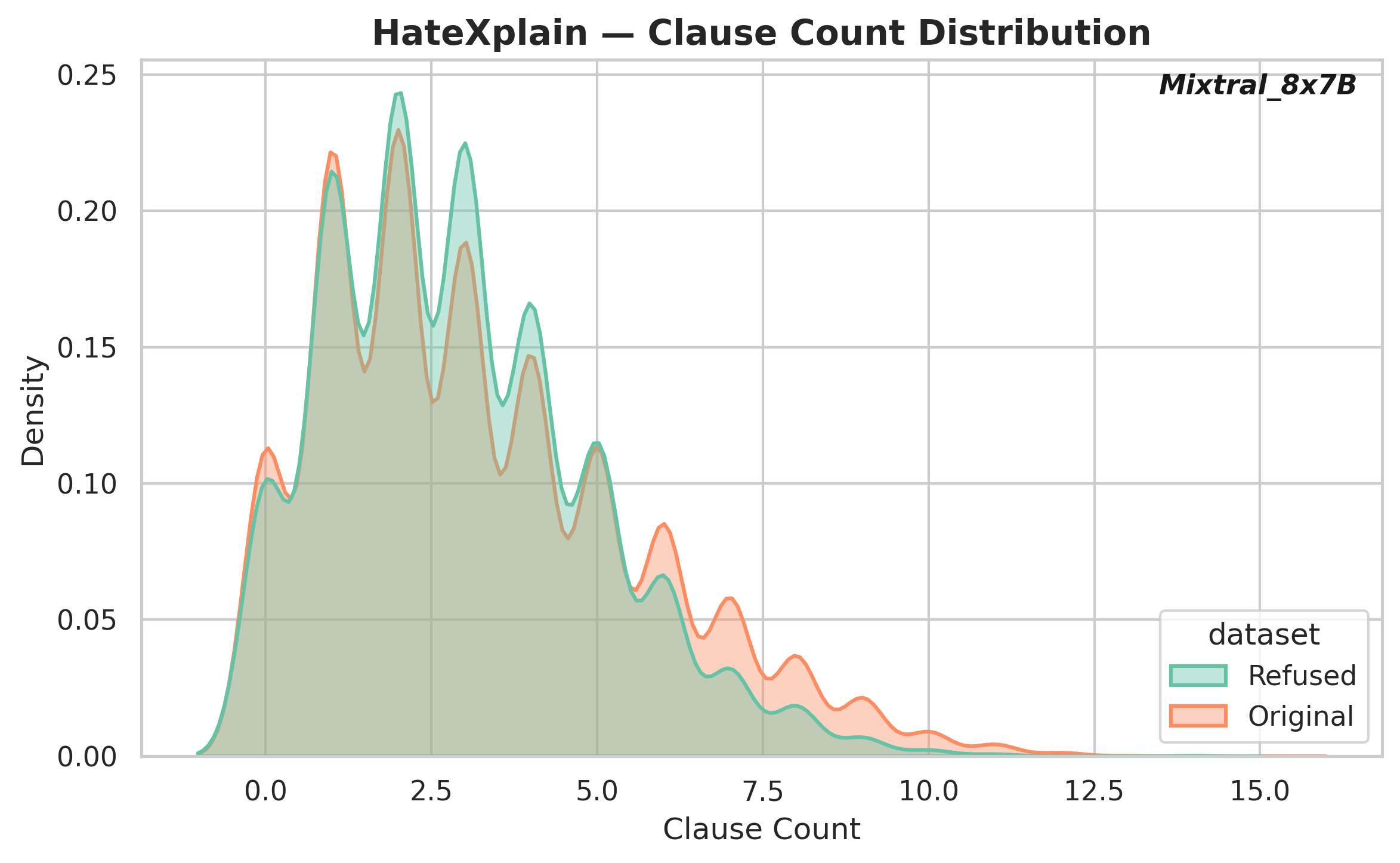}
    \caption{HateXplain}
  \end{subfigure}

  \caption{Clause count distributions across datasets for Mistral 7B and Mixtral 8$\times$7B.}
  \label{fig:clause_count_mistral}
\end{figure*}

\begin{figure*}[htbp]
  \centering
  \begin{subfigure}{0.32\textwidth}
    \centering
    \includegraphics[width=\linewidth]{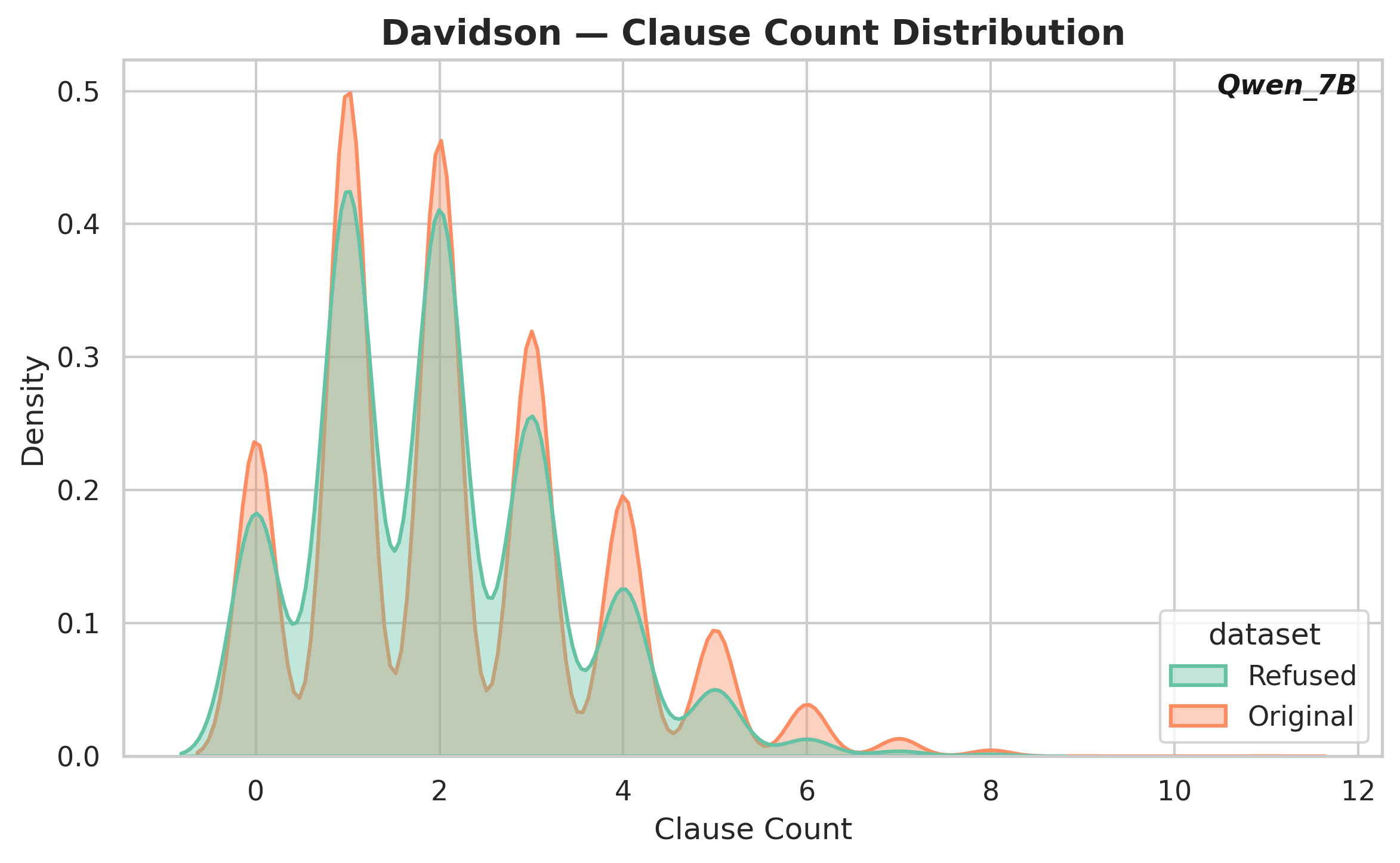}
    \caption{Davidson}
  \end{subfigure}\hfill
  \begin{subfigure}{0.32\textwidth}
    \centering
    \includegraphics[width=\linewidth]{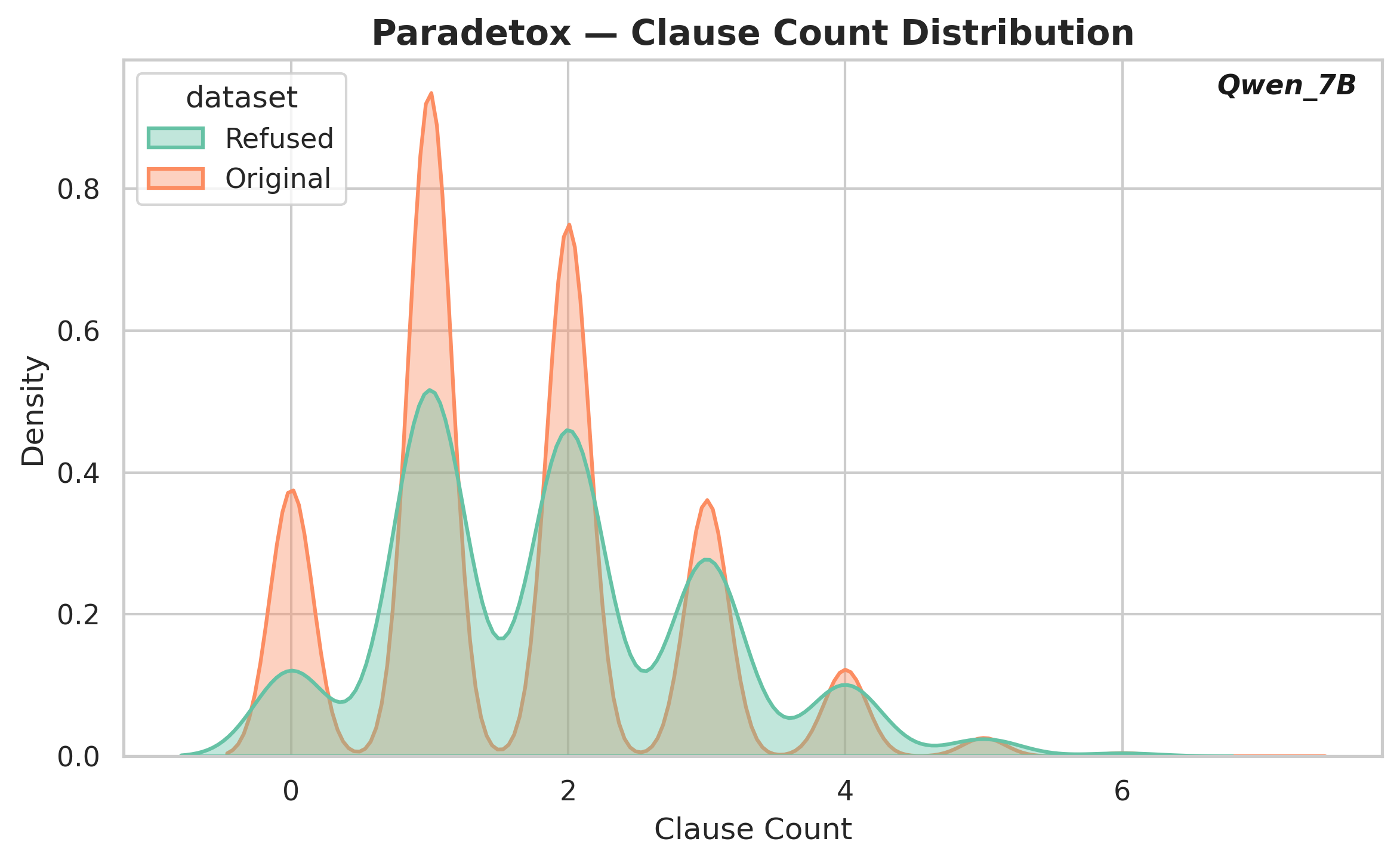}
    \caption{Paradetox}
  \end{subfigure}\hfill
  \begin{subfigure}{0.32\textwidth}
    \centering
    \includegraphics[width=\linewidth]{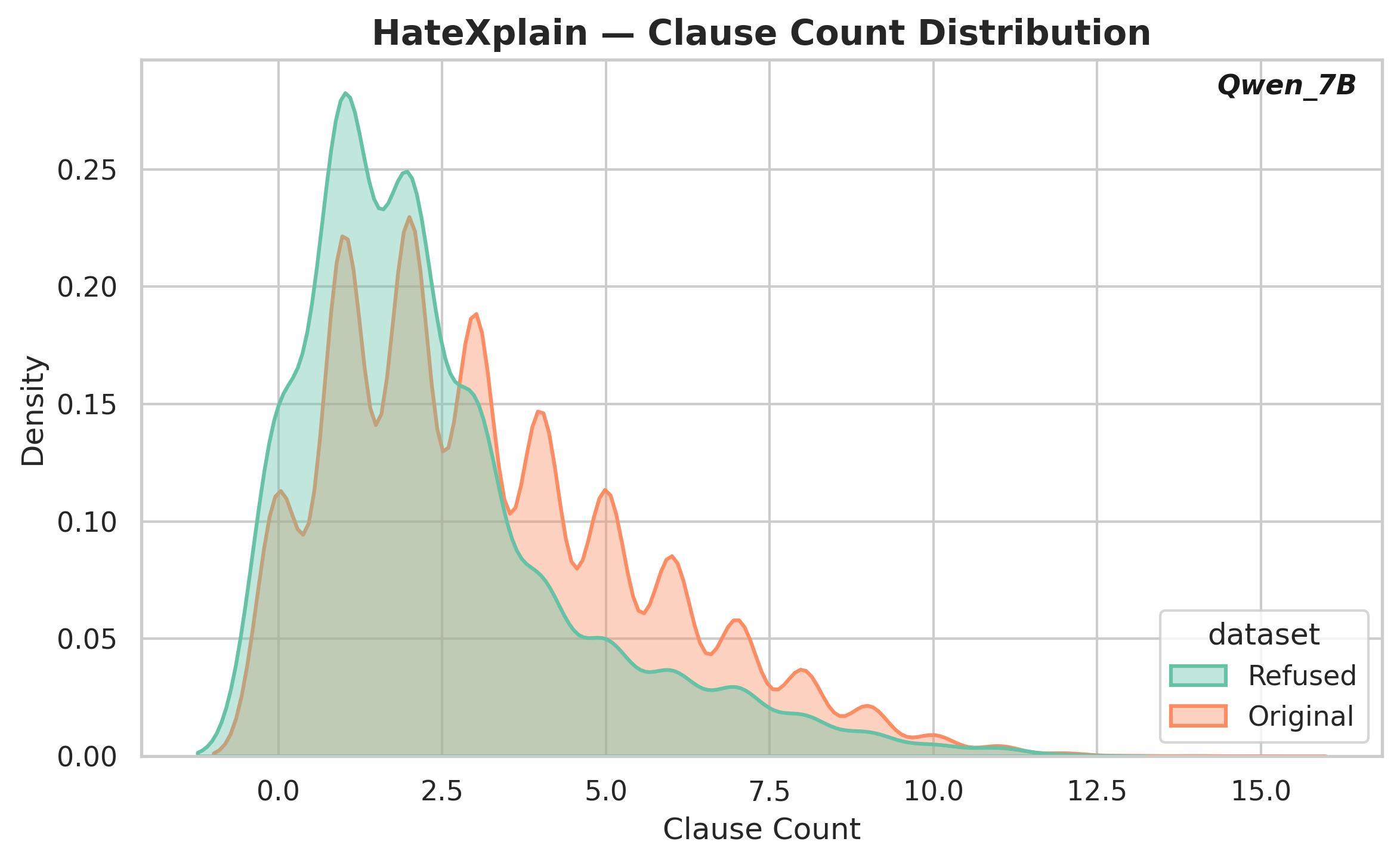}
    \caption{HateXplain}
  \end{subfigure}

  \vspace{0.4em}

  \begin{subfigure}{0.32\textwidth}
    \centering
    \includegraphics[width=\linewidth]{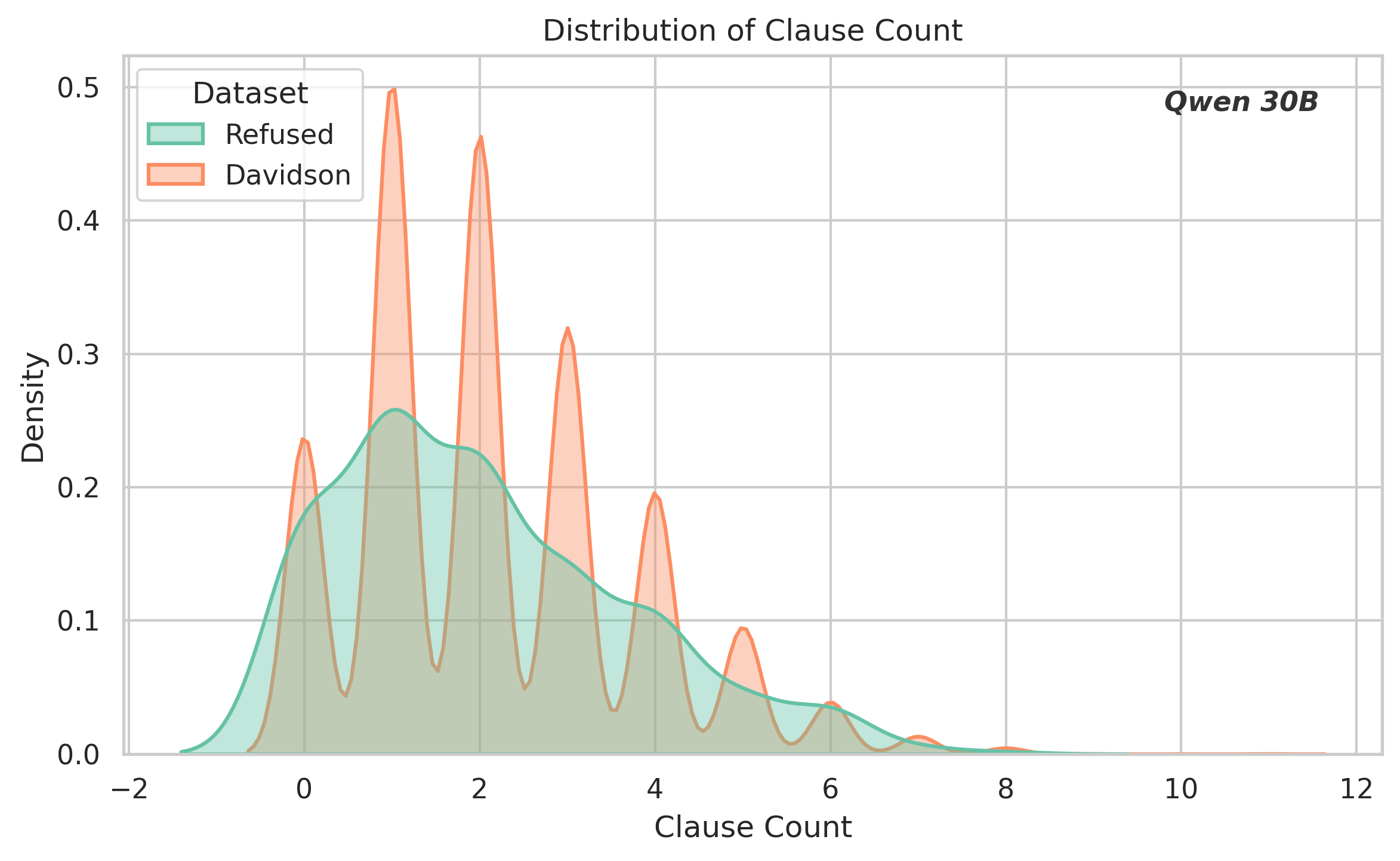}
    \caption{Davidson}
  \end{subfigure}\hfill
  \begin{subfigure}{0.32\textwidth}
    \centering
    \includegraphics[width=\linewidth]{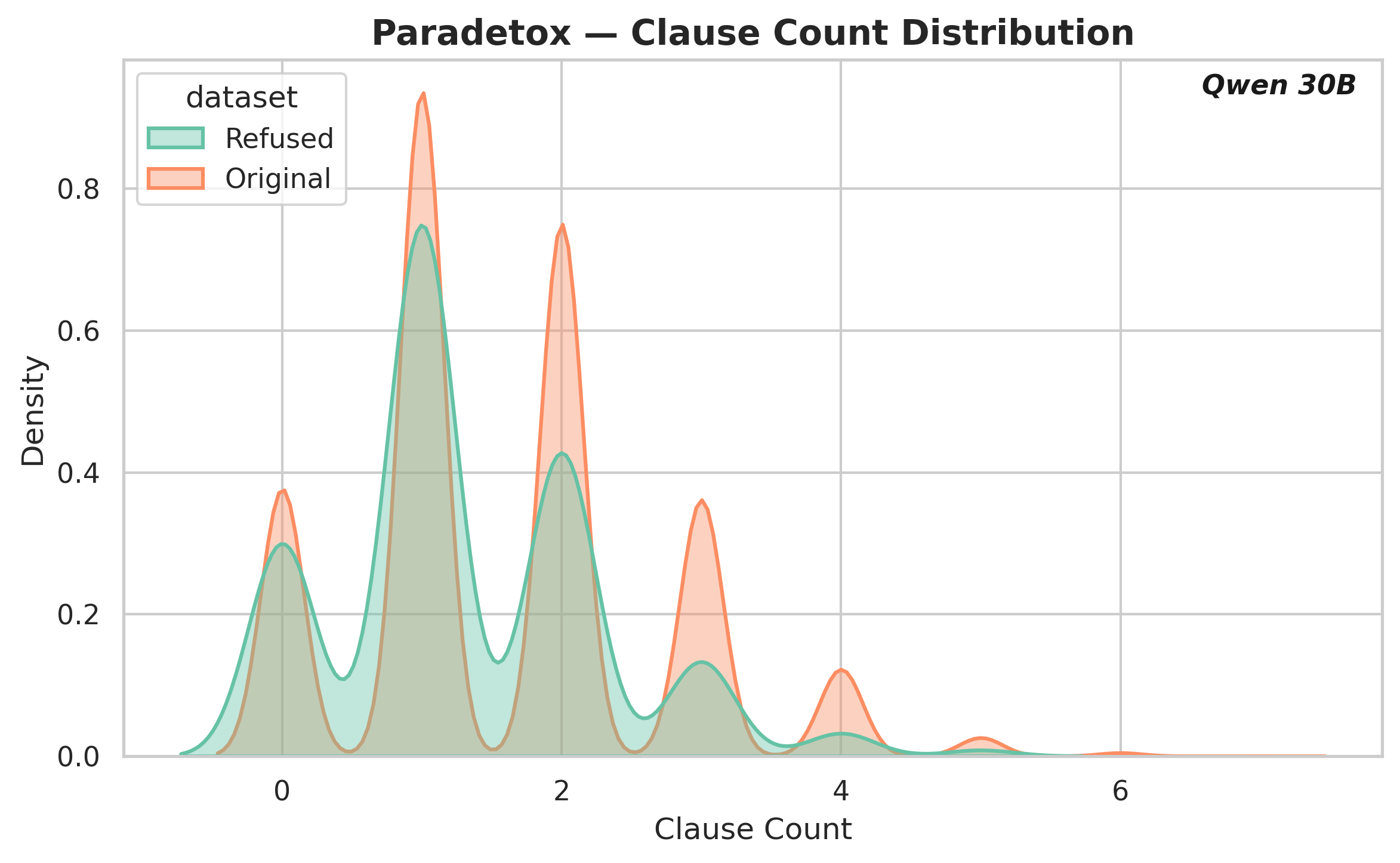}
    \caption{Paradetox}
  \end{subfigure}\hfill
  \begin{subfigure}{0.32\textwidth}
    \centering
    \includegraphics[width=\linewidth]{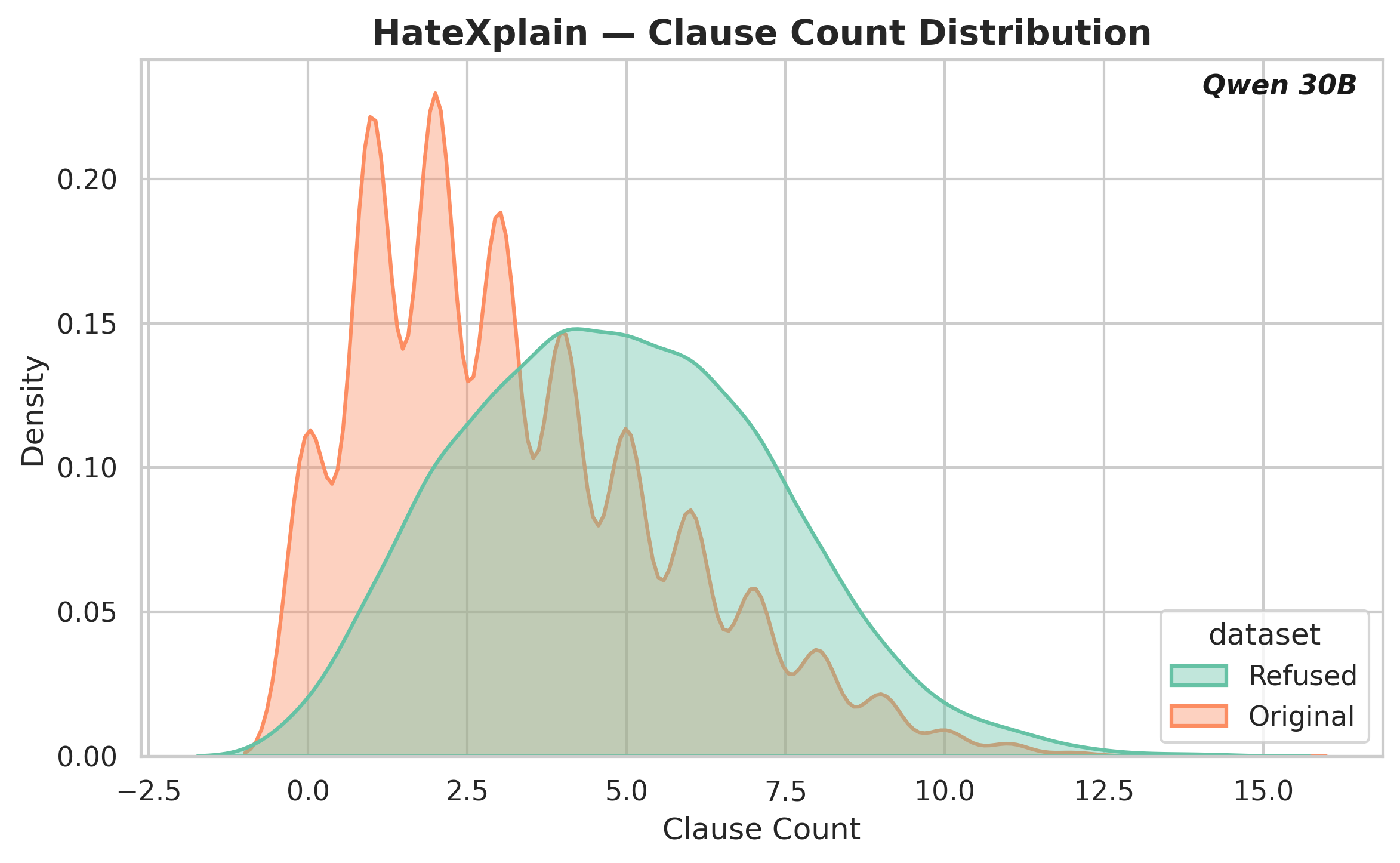}
    \caption{ HateXplain}
  \end{subfigure}

  \caption{Clause count distributions across datasets for Qwen2.5 7B and Qwen3 30B.}
  \label{fig:clause_count_qwen}
\end{figure*}

\begin{figure*}[htbp]
  \centering
  \begin{subfigure}{0.32\textwidth}
    \centering
    \includegraphics[width=\linewidth]{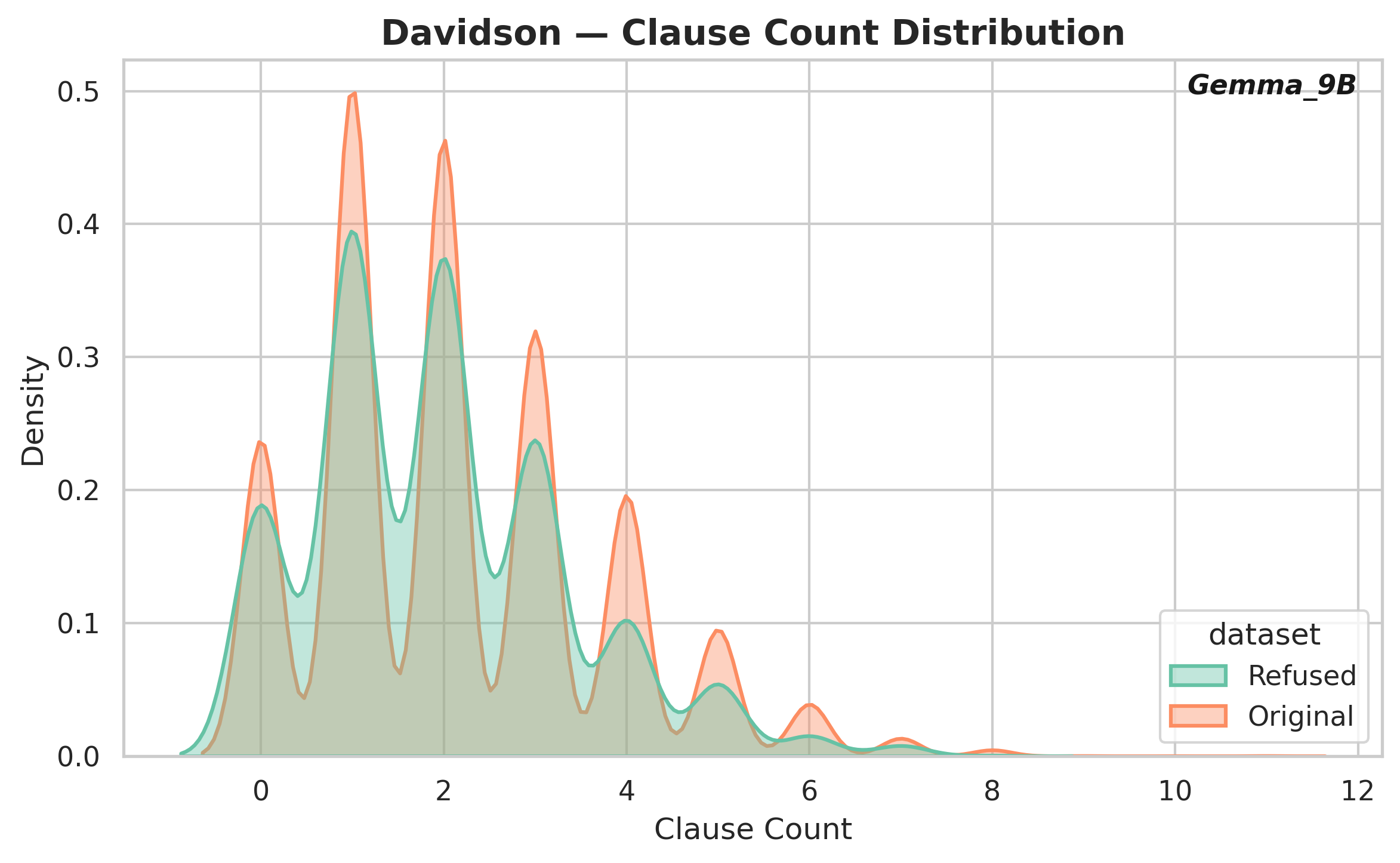}
    \caption{Davidson}
  \end{subfigure}\hfill
  \begin{subfigure}{0.32\textwidth}
    \centering
    \includegraphics[width=\linewidth]{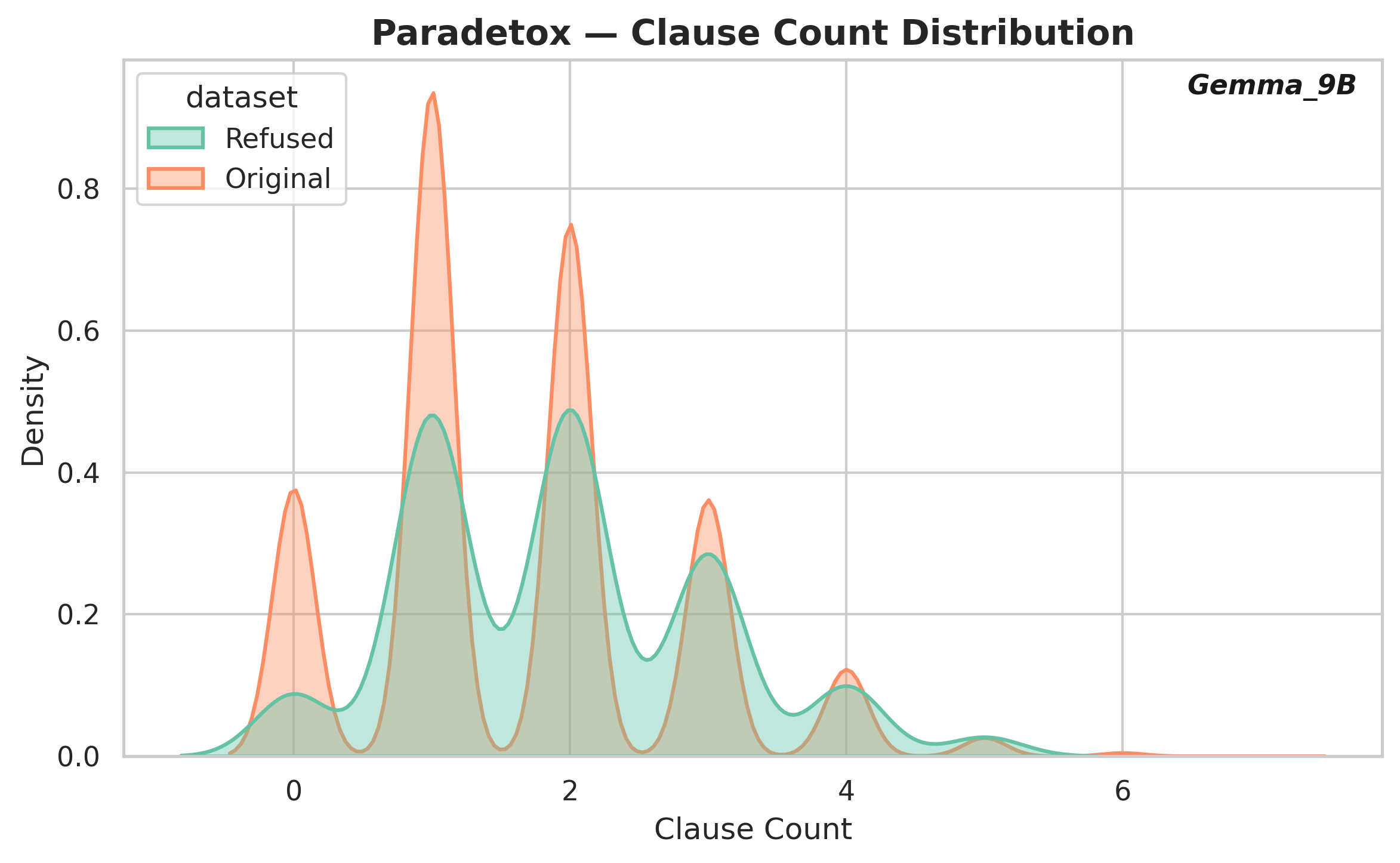}
    \caption{Paradetox}
  \end{subfigure}\hfill
  \begin{subfigure}{0.32\textwidth}
    \centering
    \includegraphics[width=\linewidth]{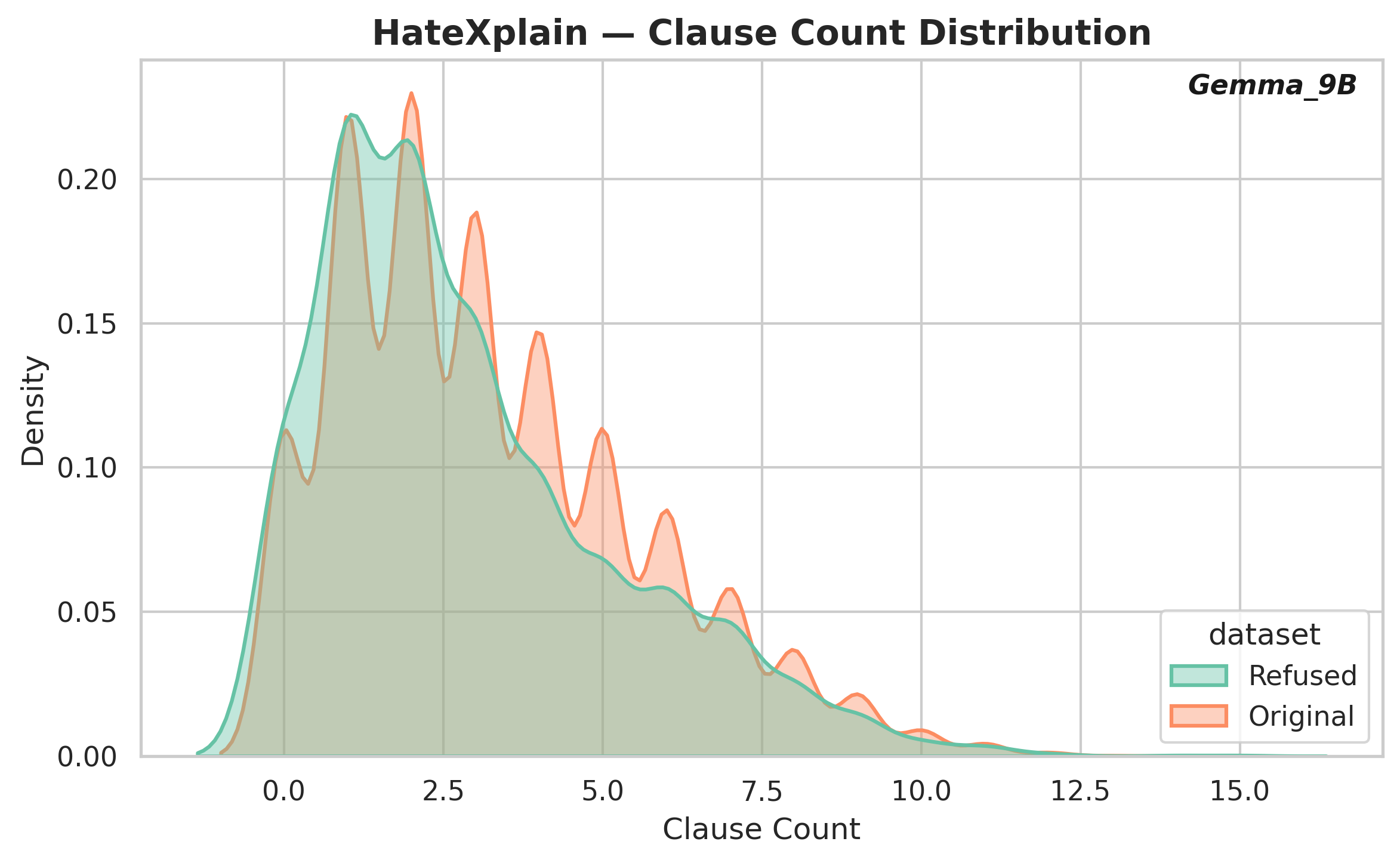}
    \caption{HateXplain}
  \end{subfigure}

  \vspace{0.4em}

  \begin{subfigure}{0.32\textwidth}
    \centering
    \includegraphics[width=\linewidth]{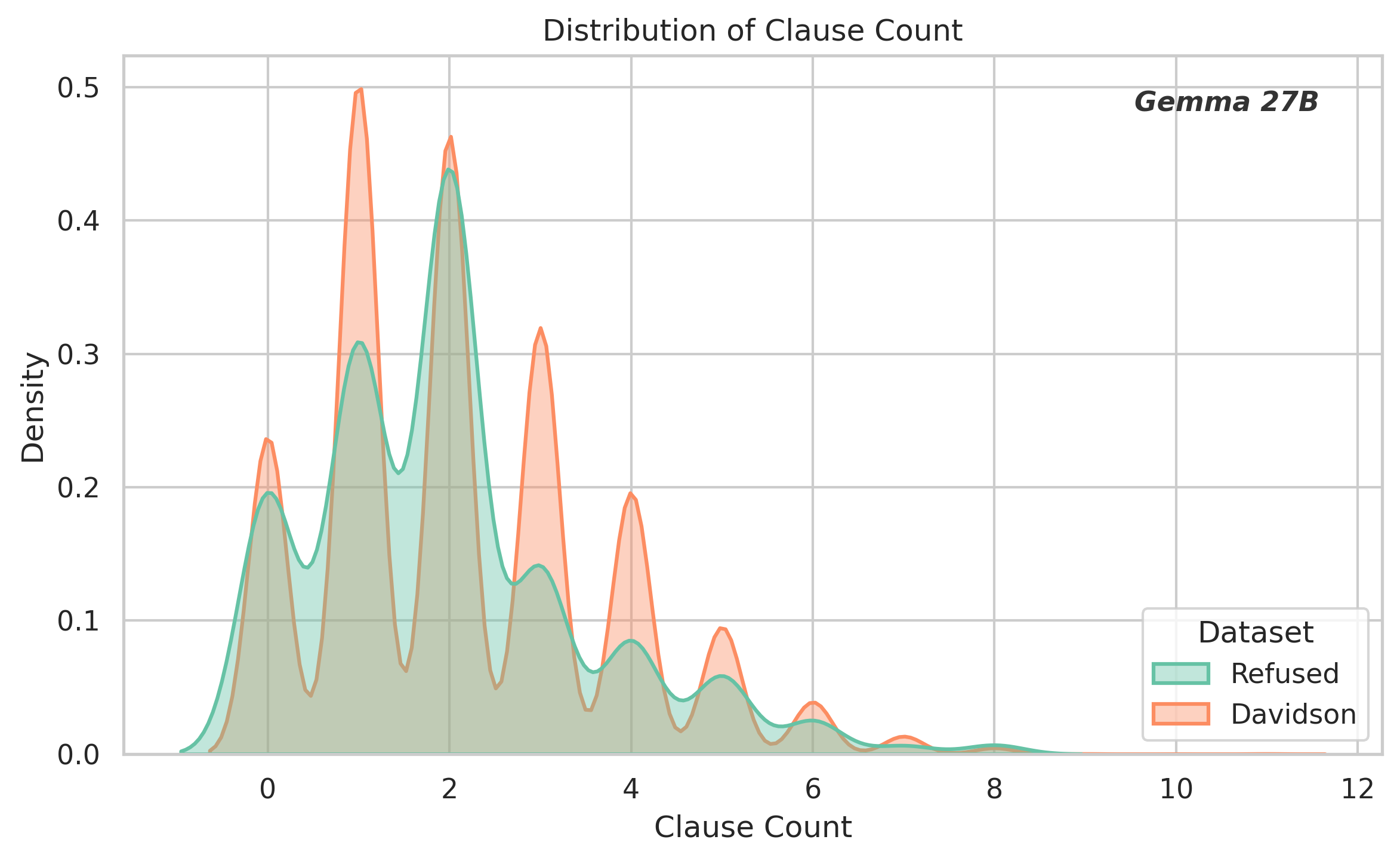}
    \caption{Davidson}
  \end{subfigure}\hfill
  \begin{subfigure}{0.32\textwidth}
    \centering
    \includegraphics[width=\linewidth]{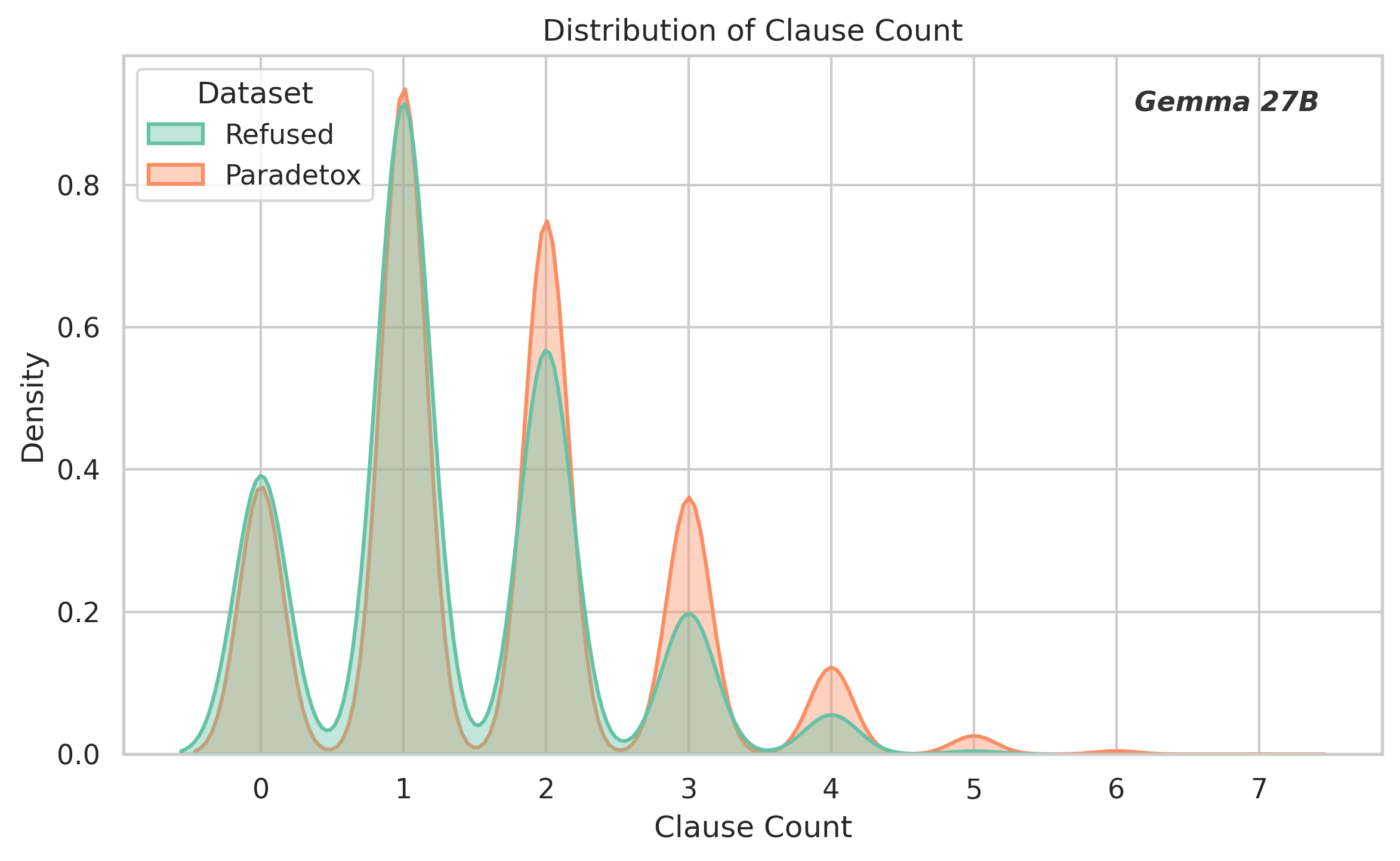}
    \caption{ Paradetox}
  \end{subfigure}\hfill
  \begin{subfigure}{0.32\textwidth}
    \centering
    \includegraphics[width=\linewidth]{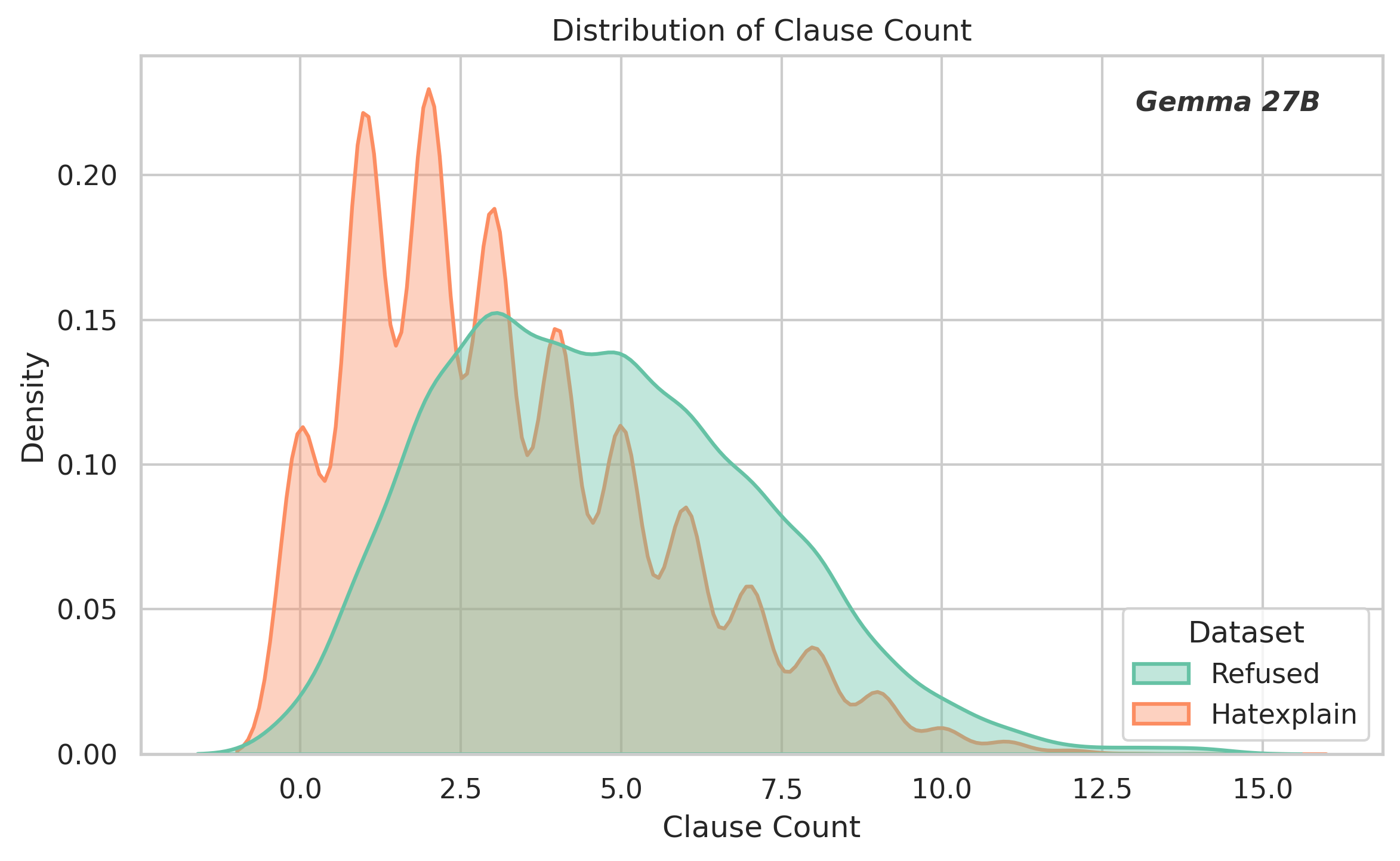}
    \caption{ HateXplain}
  \end{subfigure}

  \caption{Clause count distributions across datasets for Gemma2 9B and Gemma3 27B.}
  \label{fig:clause_count_gemma}
\end{figure*}

\paragraph{Clause Count.}
Clause count distributions for the same set of models are presented in Figures~\ref{fig:clause_count_gpt}–\ref{fig:clause_count_gemma}.

\begin{figure*}[htbp]
  \centering
  \setlength{\tabcolsep}{3pt}

  \begin{subfigure}{0.32\textwidth}
    \centering
    \includegraphics[width=\linewidth]{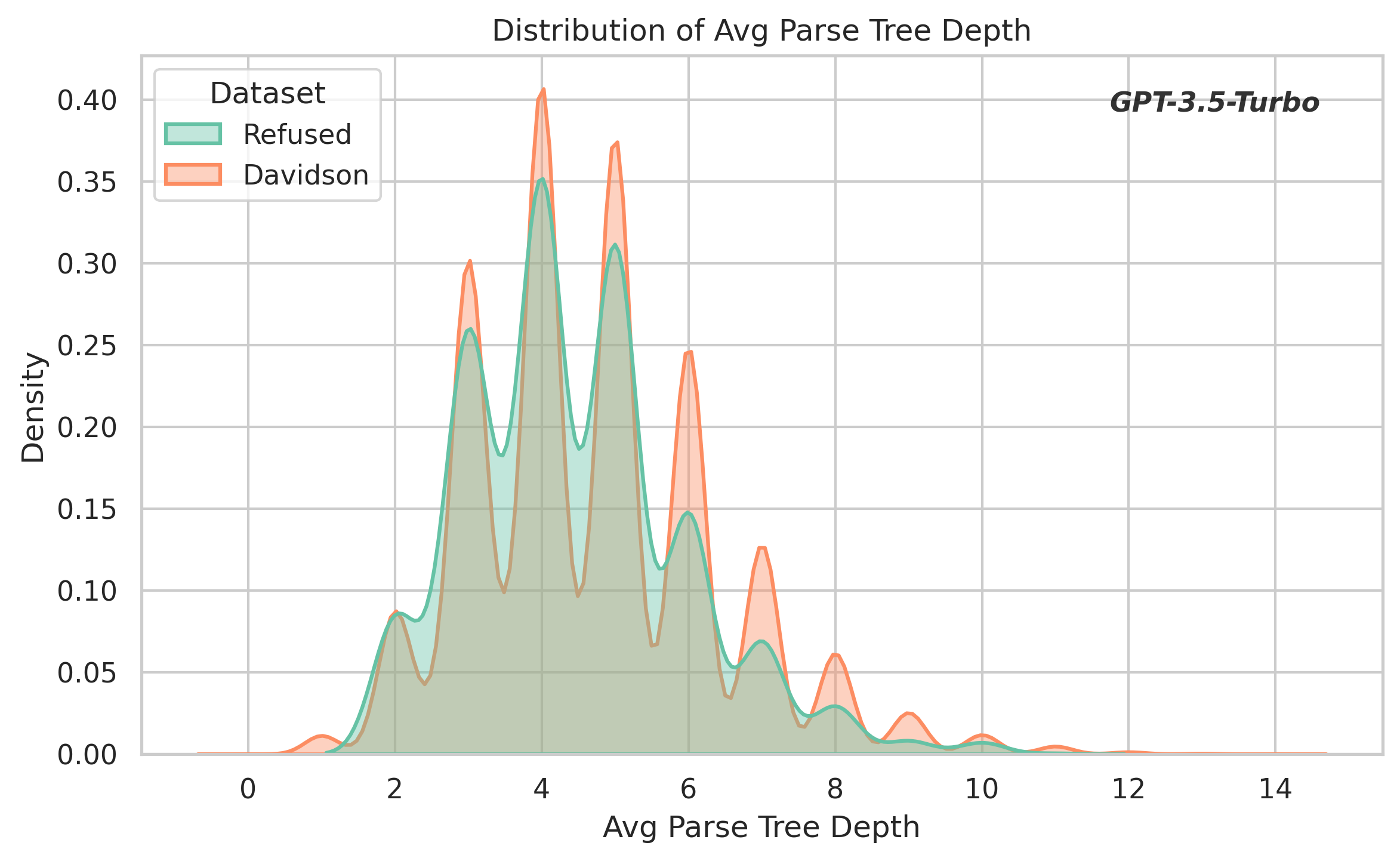}
    \caption{Davidson}
  \end{subfigure}\hfill
  \begin{subfigure}{0.32\textwidth}
    \centering
    \includegraphics[width=\linewidth]{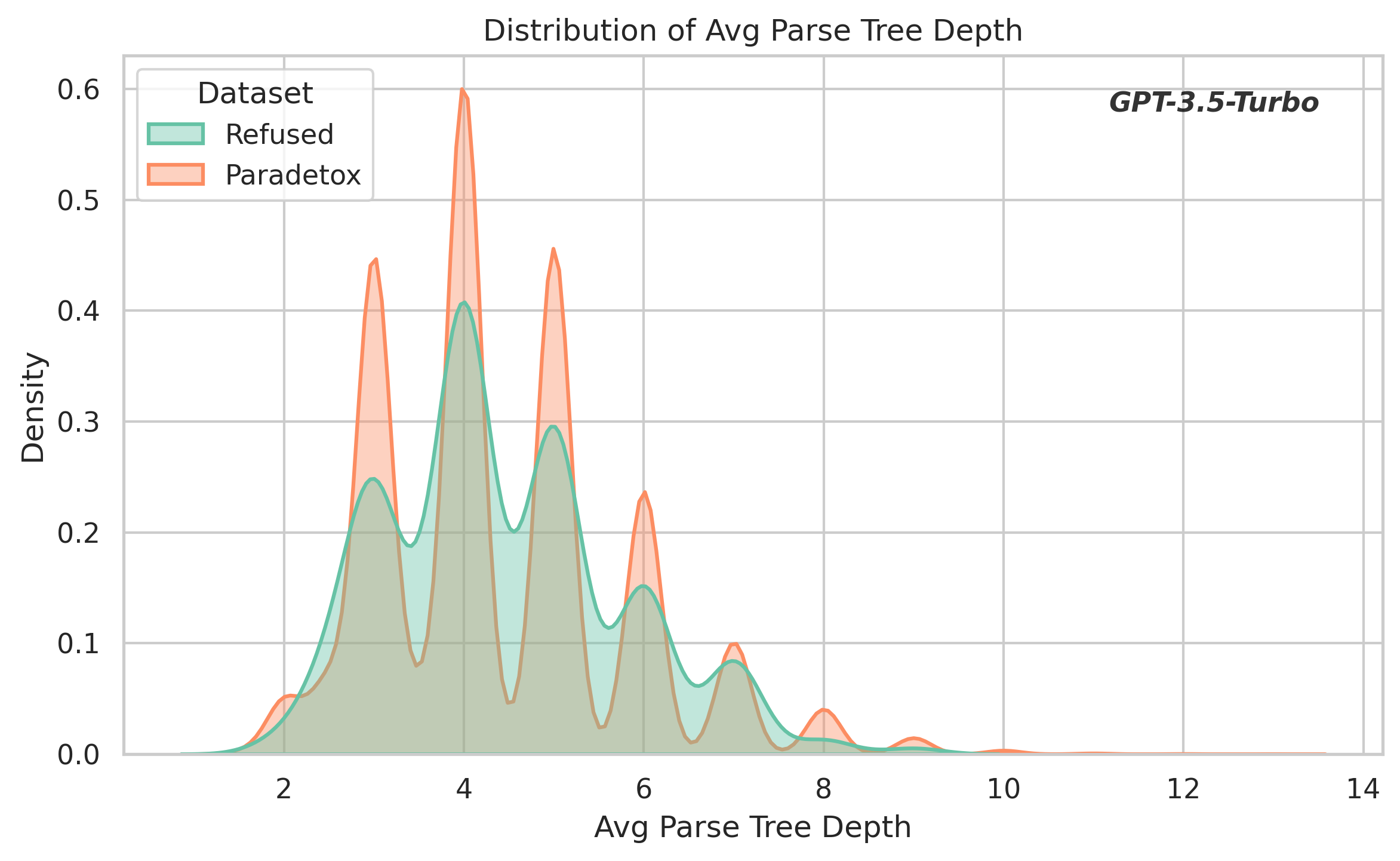}
    \caption{Paradetox}
  \end{subfigure}\hfill
  \begin{subfigure}{0.32\textwidth}
    \centering
    \includegraphics[width=\linewidth]{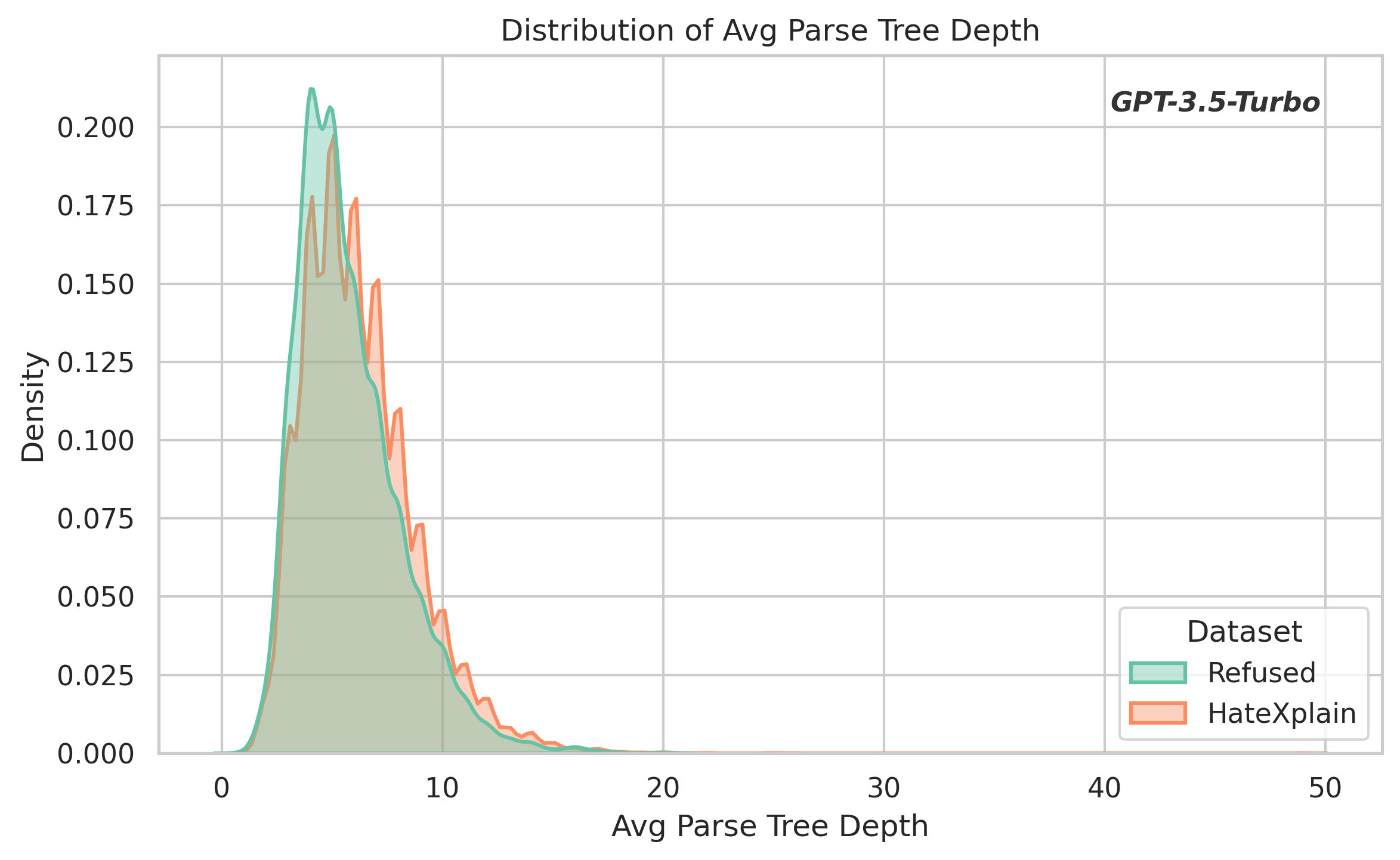}
    \caption{HateXplain}
  \end{subfigure}

  \vspace{0.4em}

  \begin{subfigure}{0.32\textwidth}
    \centering
    \includegraphics[width=\linewidth]{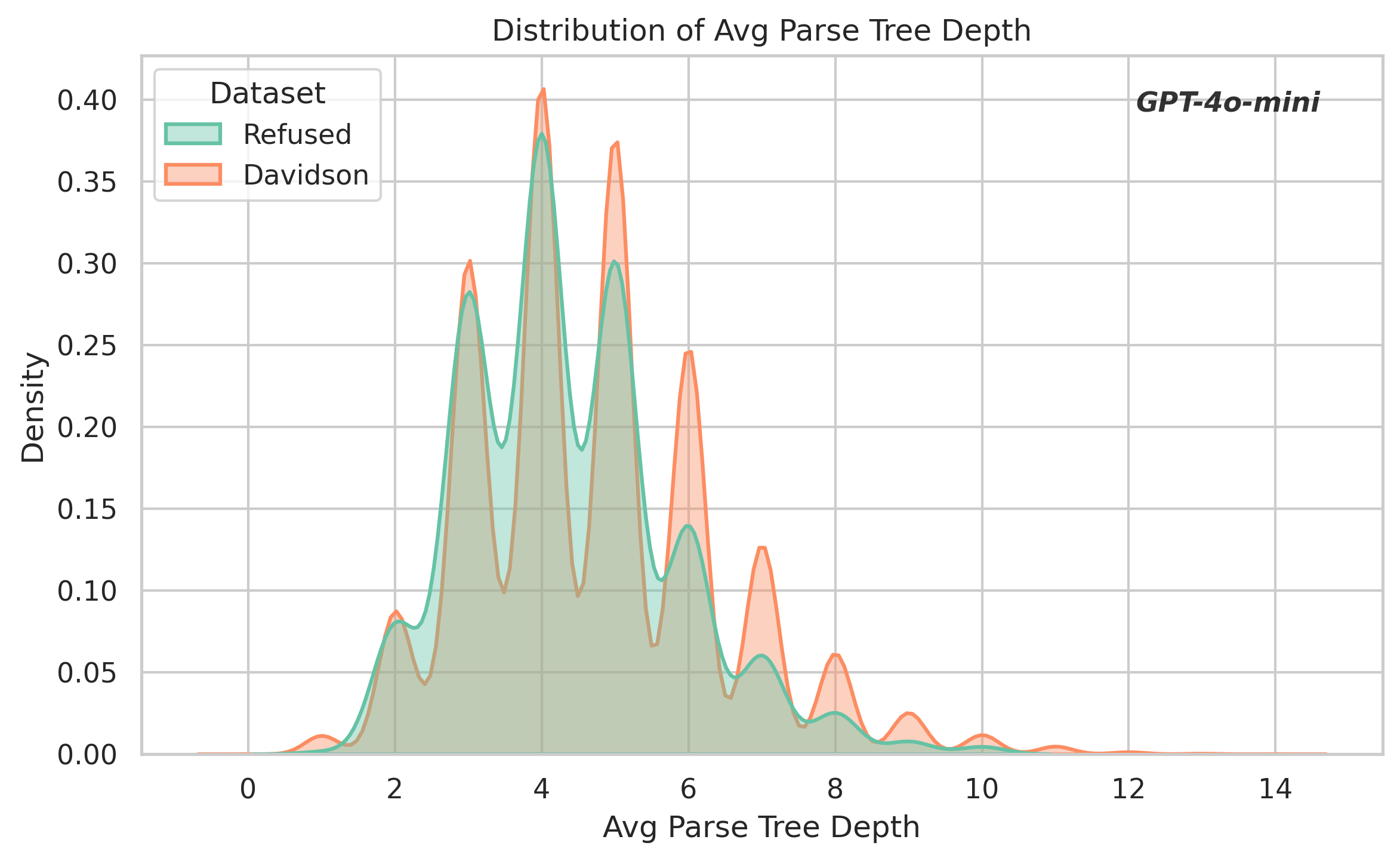}
    \caption{Davidson}
  \end{subfigure}\hfill
  \begin{subfigure}{0.32\textwidth}
    \centering
    \includegraphics[width=\linewidth]{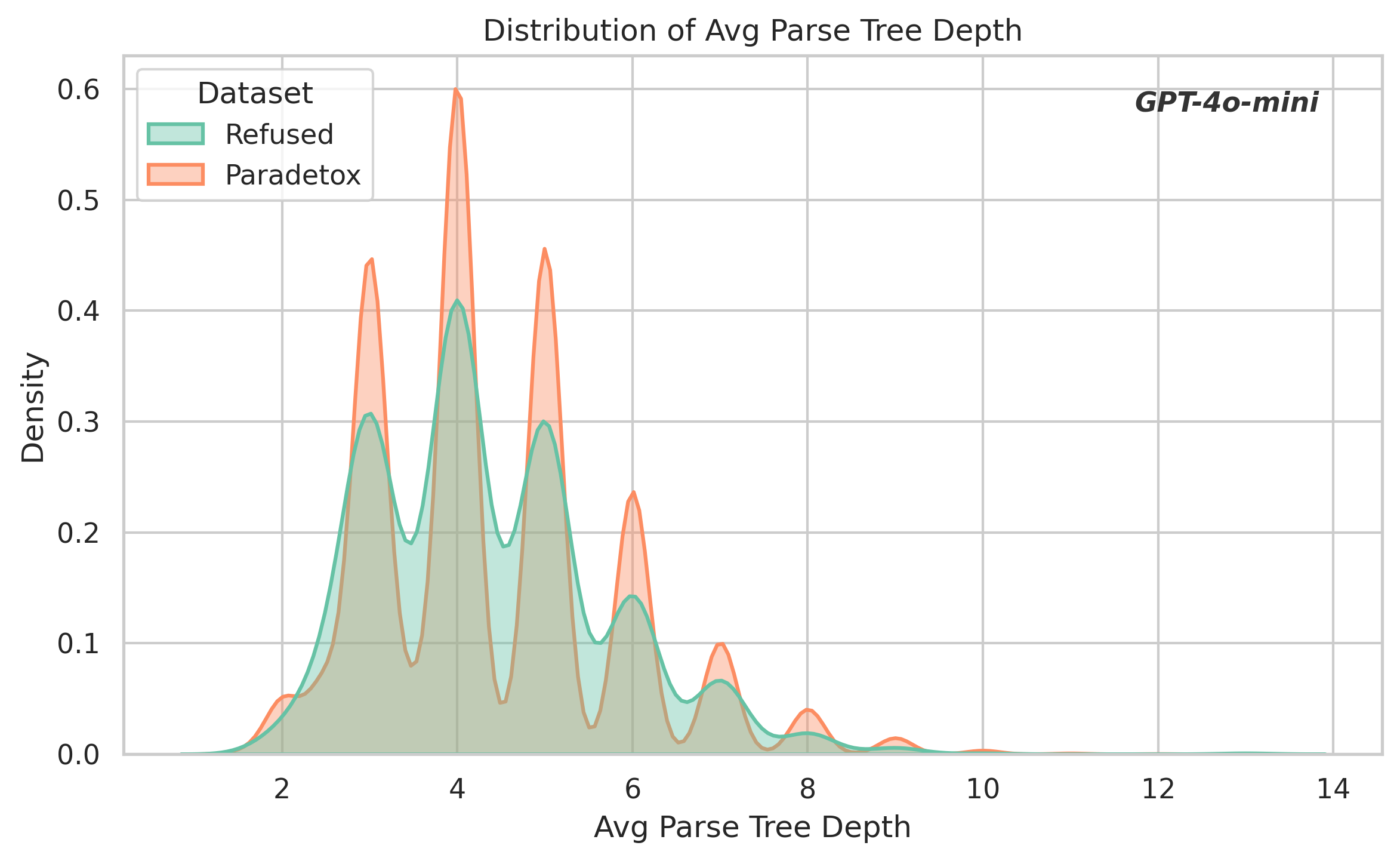}
    \caption{Paradetox}
  \end{subfigure}\hfill
  \begin{subfigure}{0.32\textwidth}
    \centering
    \includegraphics[width=\linewidth]{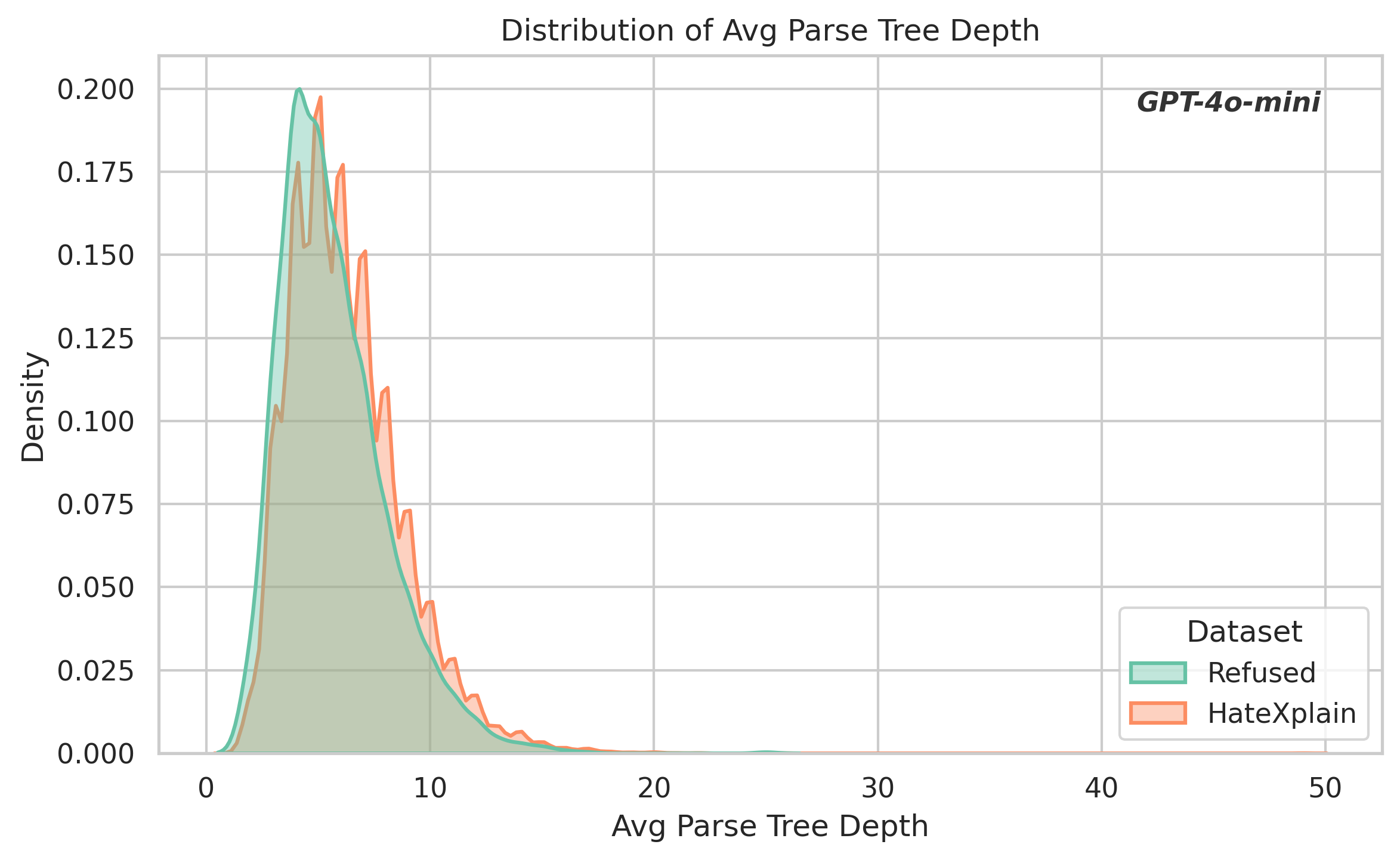}
    \caption{HateXplain}
  \end{subfigure}

  \caption{Parse tree depth distributions across datasets for GPT-3.5-Turbo and GPT-4o-mini.}
  \label{fig:parse_tree_gpt}
\end{figure*}

\begin{figure*}[htbp]
  \centering
  \begin{subfigure}{0.32\textwidth}
    \centering
    \includegraphics[width=\linewidth]{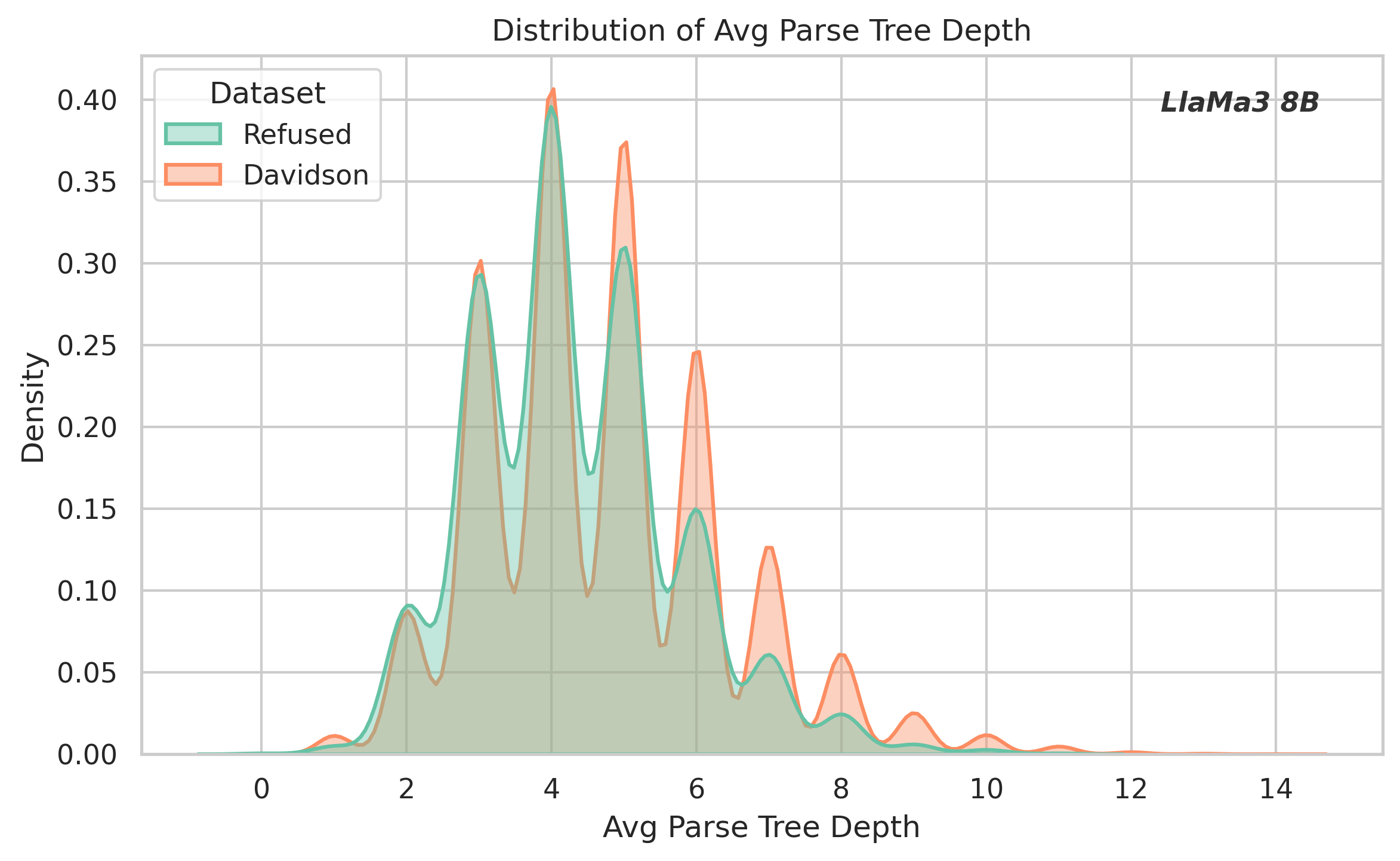}
    \caption{Davidson}
  \end{subfigure}\hfill
  \begin{subfigure}{0.32\textwidth}
    \centering
    \includegraphics[width=\linewidth]{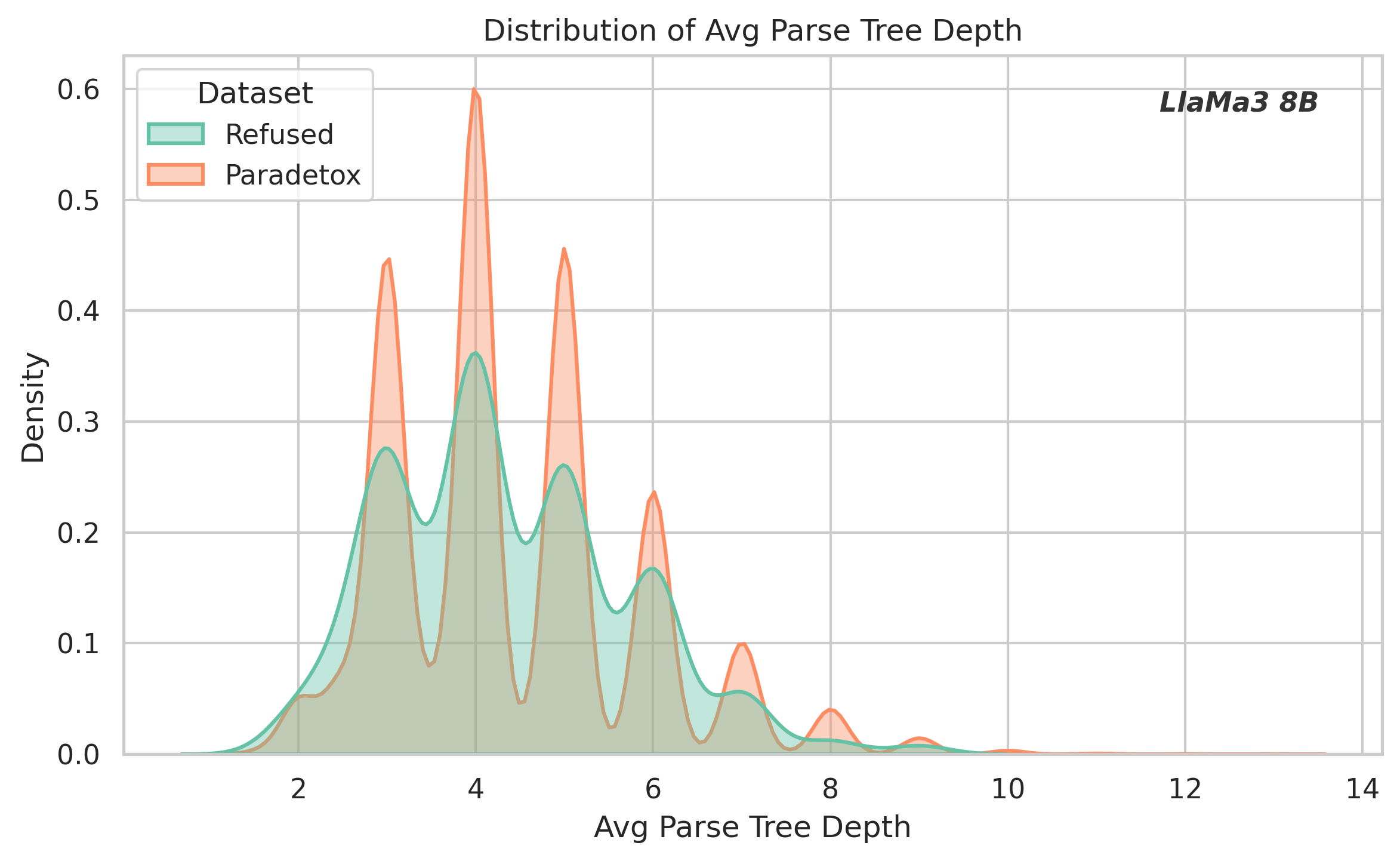}
    \caption{Paradetox}
  \end{subfigure}\hfill
  \begin{subfigure}{0.32\textwidth}
    \centering
    \includegraphics[width=\linewidth]{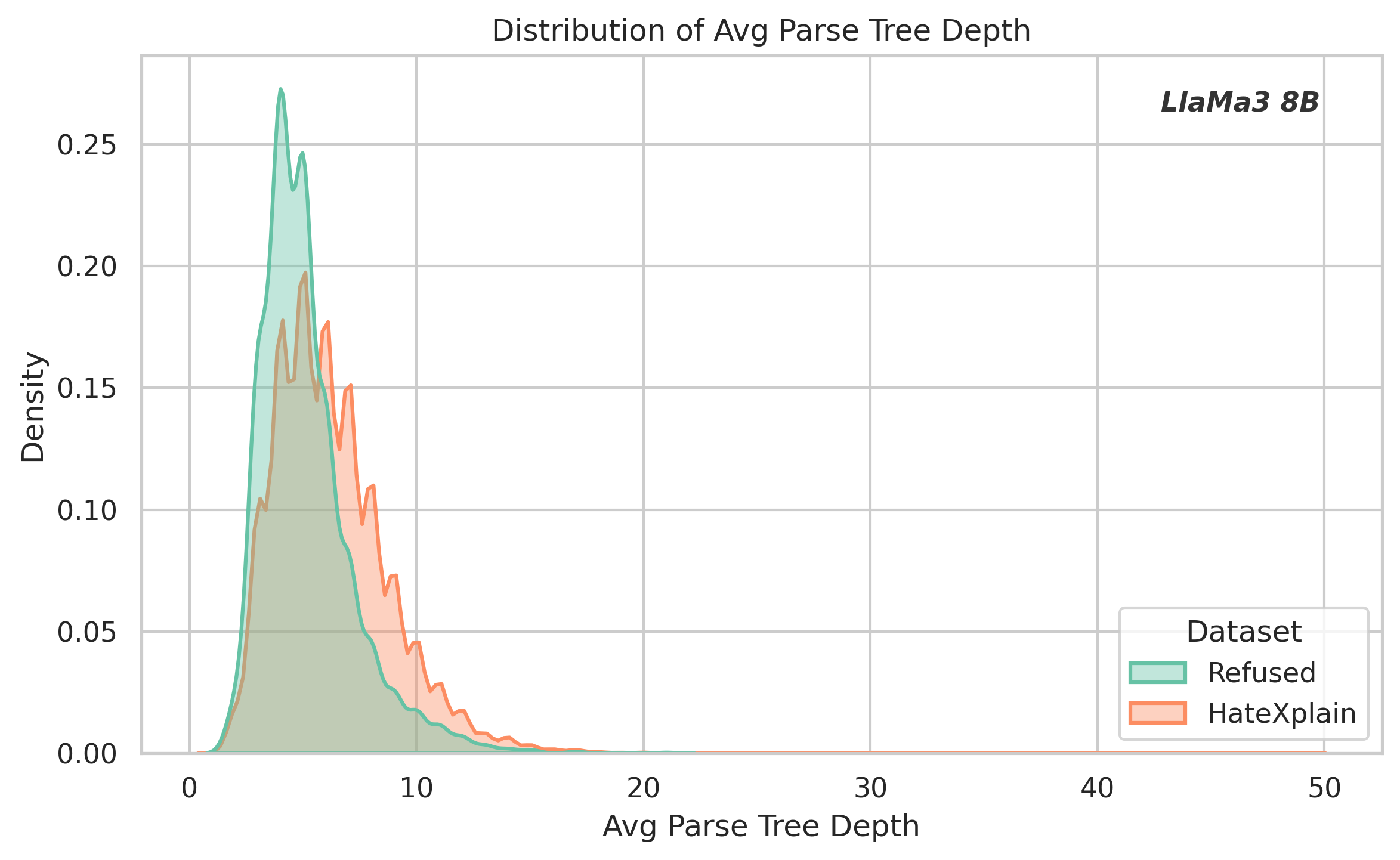}
    \caption{HateXplain}
  \end{subfigure}

  \caption{Parse tree depth distributions across datasets for Llama3 8B.}
  \label{fig:parse_tree_llama}
\end{figure*}

\begin{figure*}[htbp]
  \centering
  \begin{subfigure}{0.32\textwidth}
    \centering
    \includegraphics[width=\linewidth]{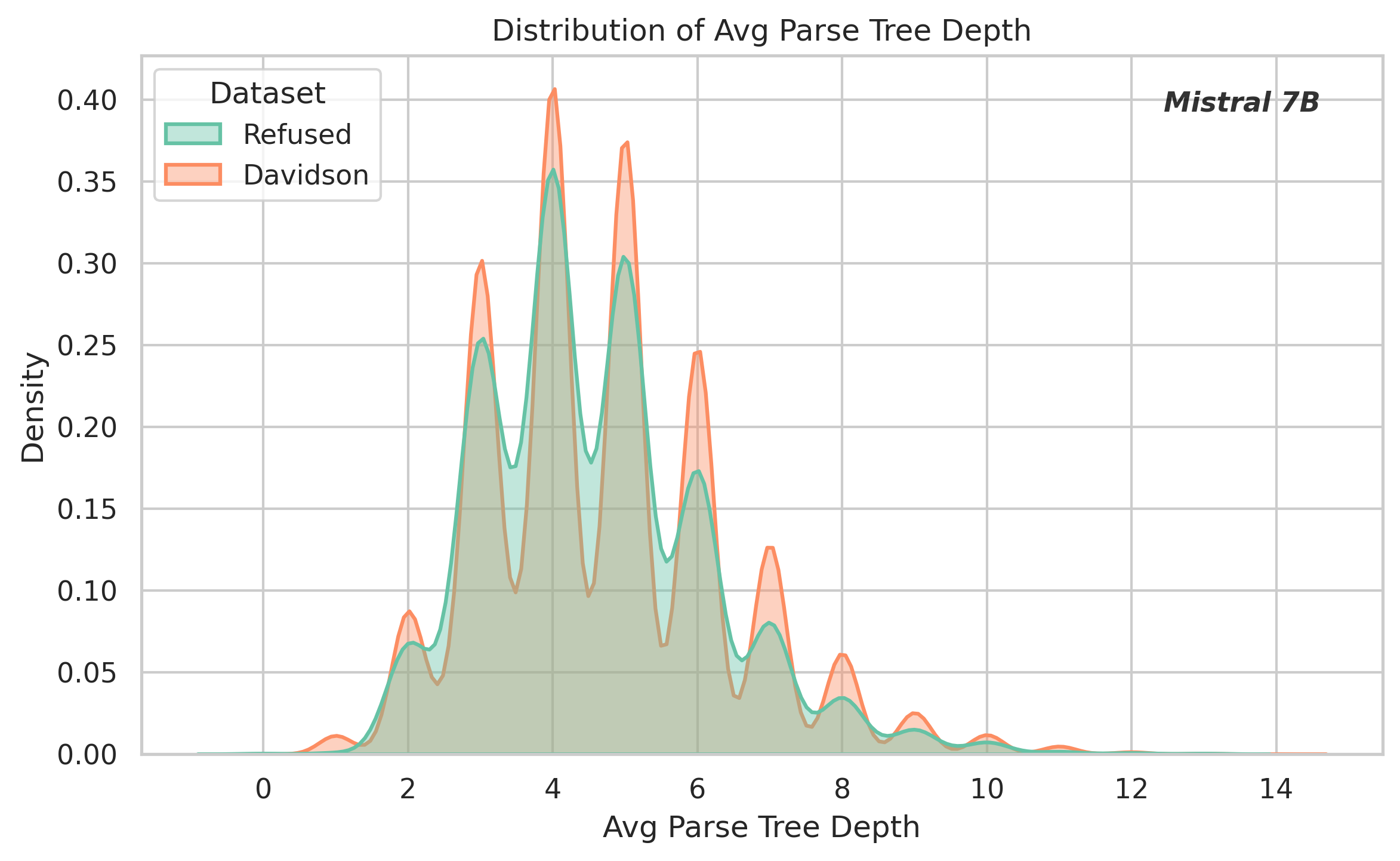}
    \caption{Davidson}
  \end{subfigure}\hfill
  \begin{subfigure}{0.32\textwidth}
    \centering
    \includegraphics[width=\linewidth]{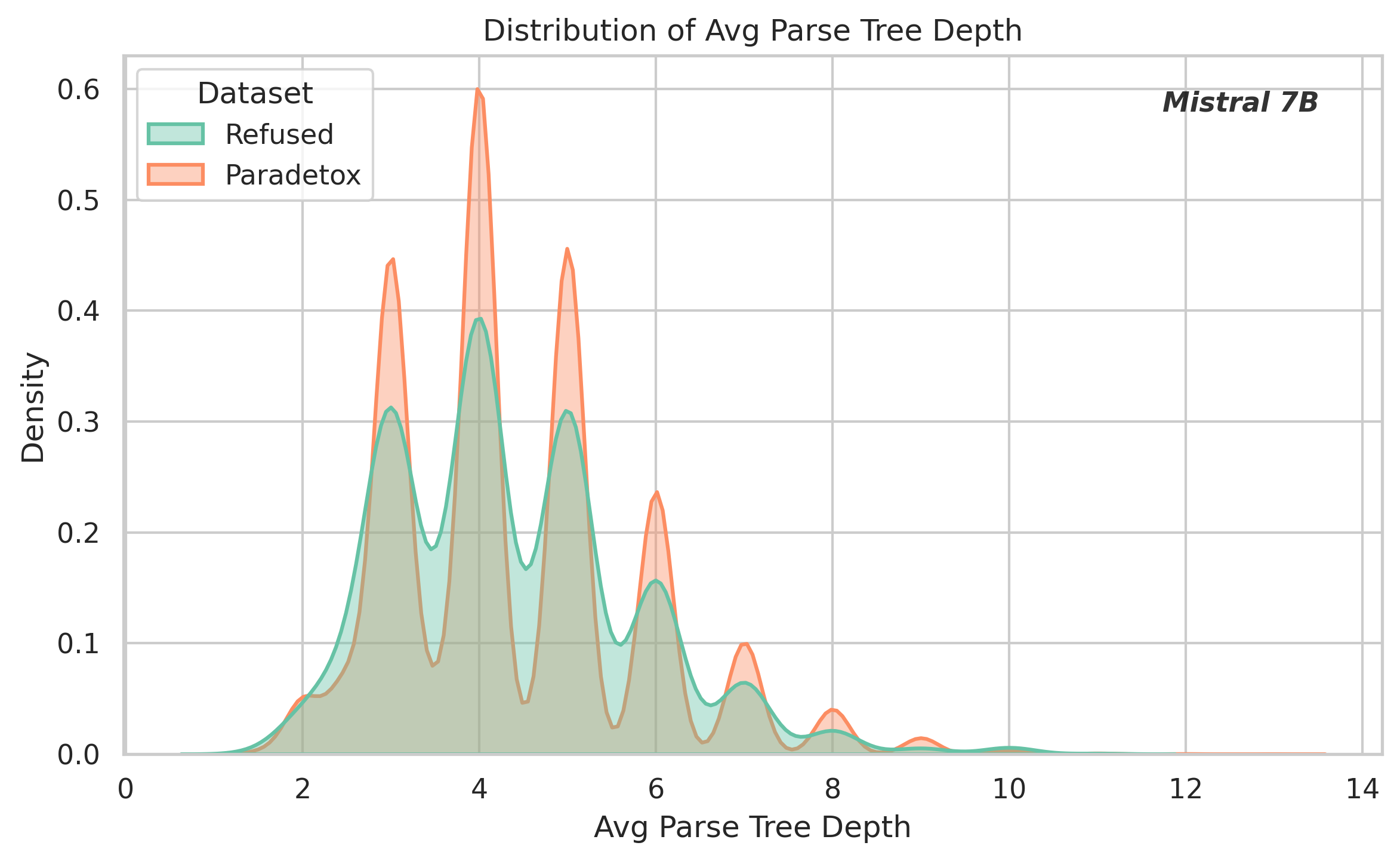}
    \caption{Paradetox}
  \end{subfigure}\hfill
  \begin{subfigure}{0.32\textwidth}
    \centering
    \includegraphics[width=\linewidth]{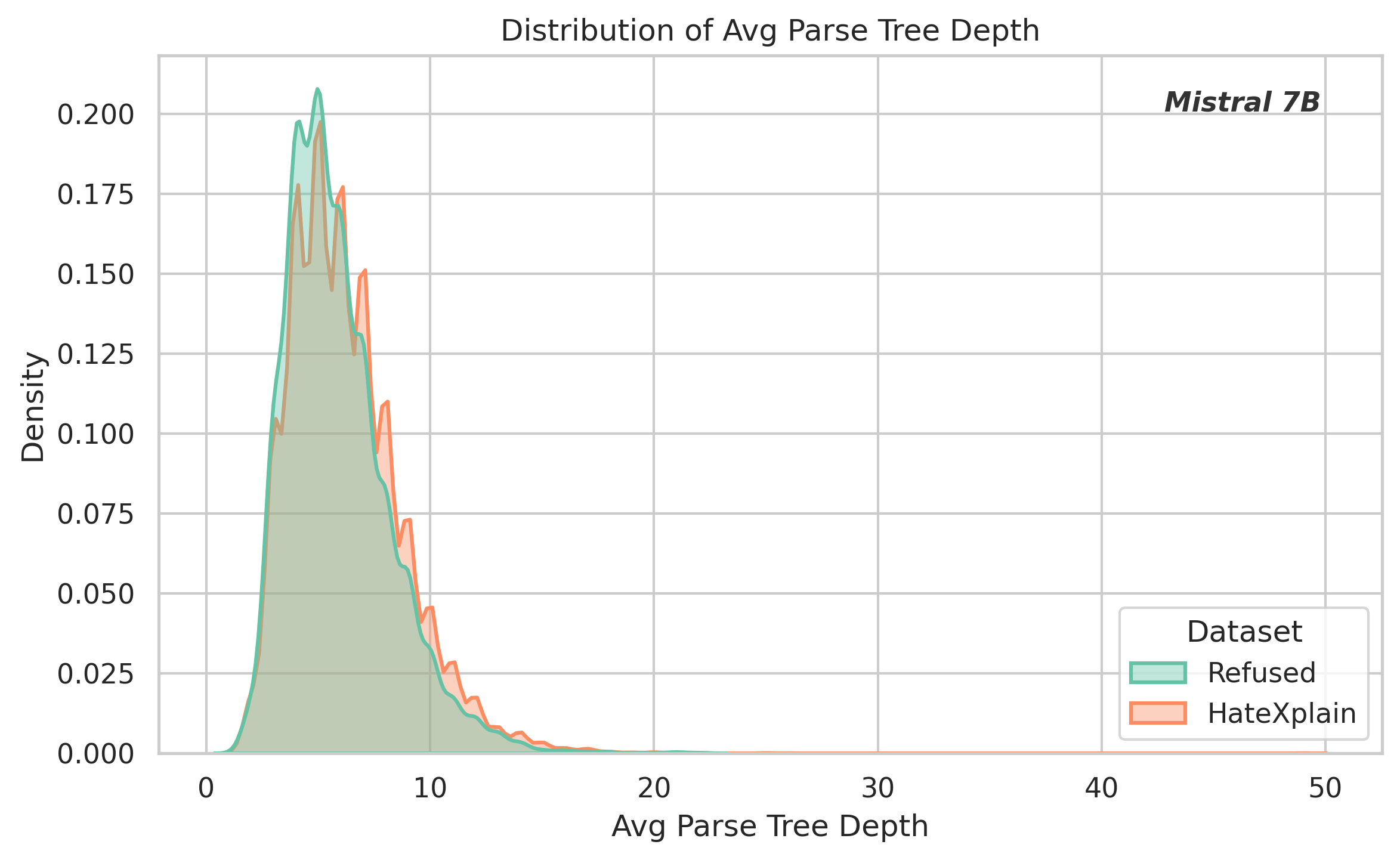}
    \caption{HateXplain}
  \end{subfigure}

  \vspace{0.4em}

  \begin{subfigure}{0.32\textwidth}
    \centering
    \includegraphics[width=\linewidth]{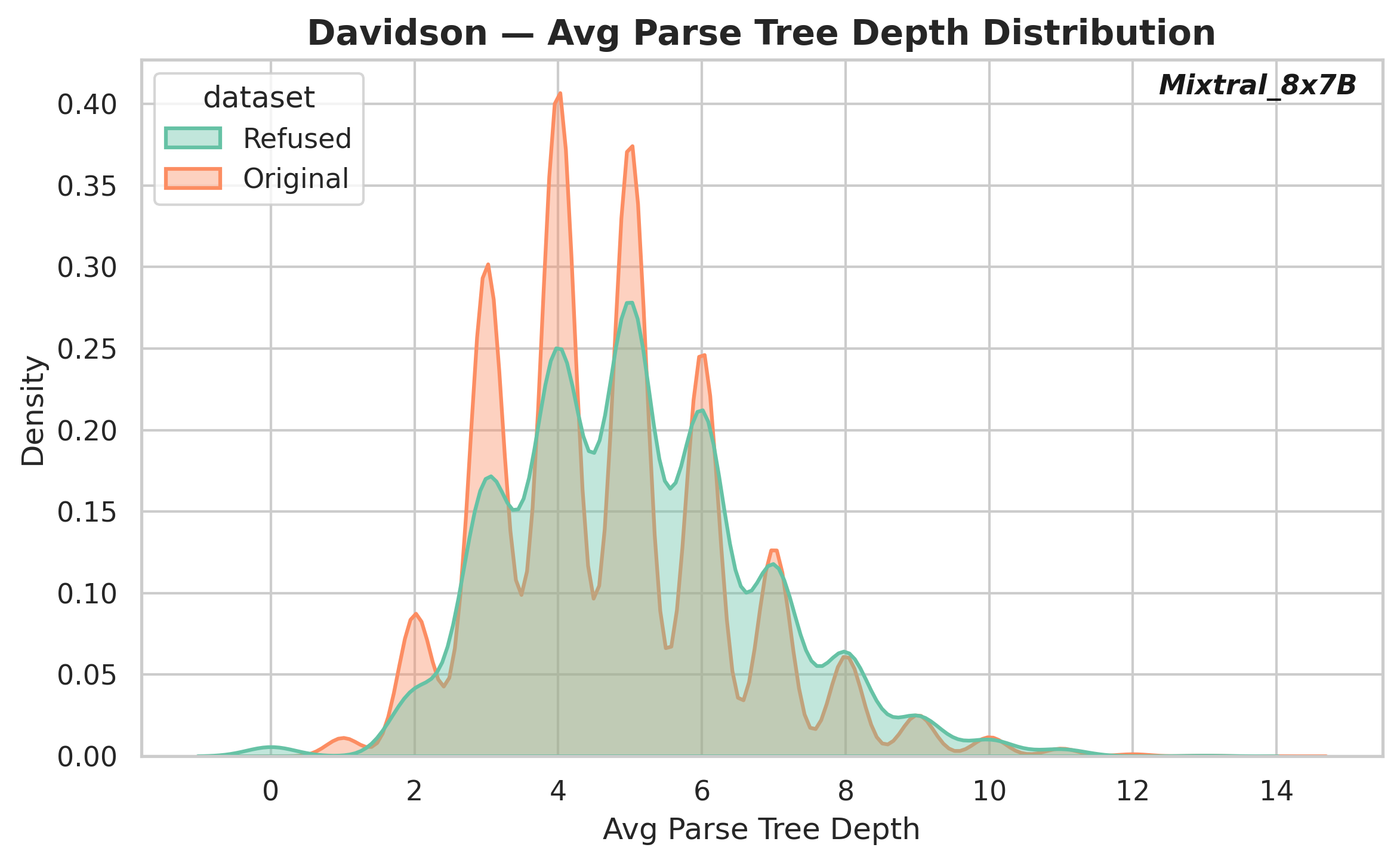}
    \caption{Davidson}
  \end{subfigure}\hfill
  \begin{subfigure}{0.32\textwidth}
    \centering
    \includegraphics[width=\linewidth]{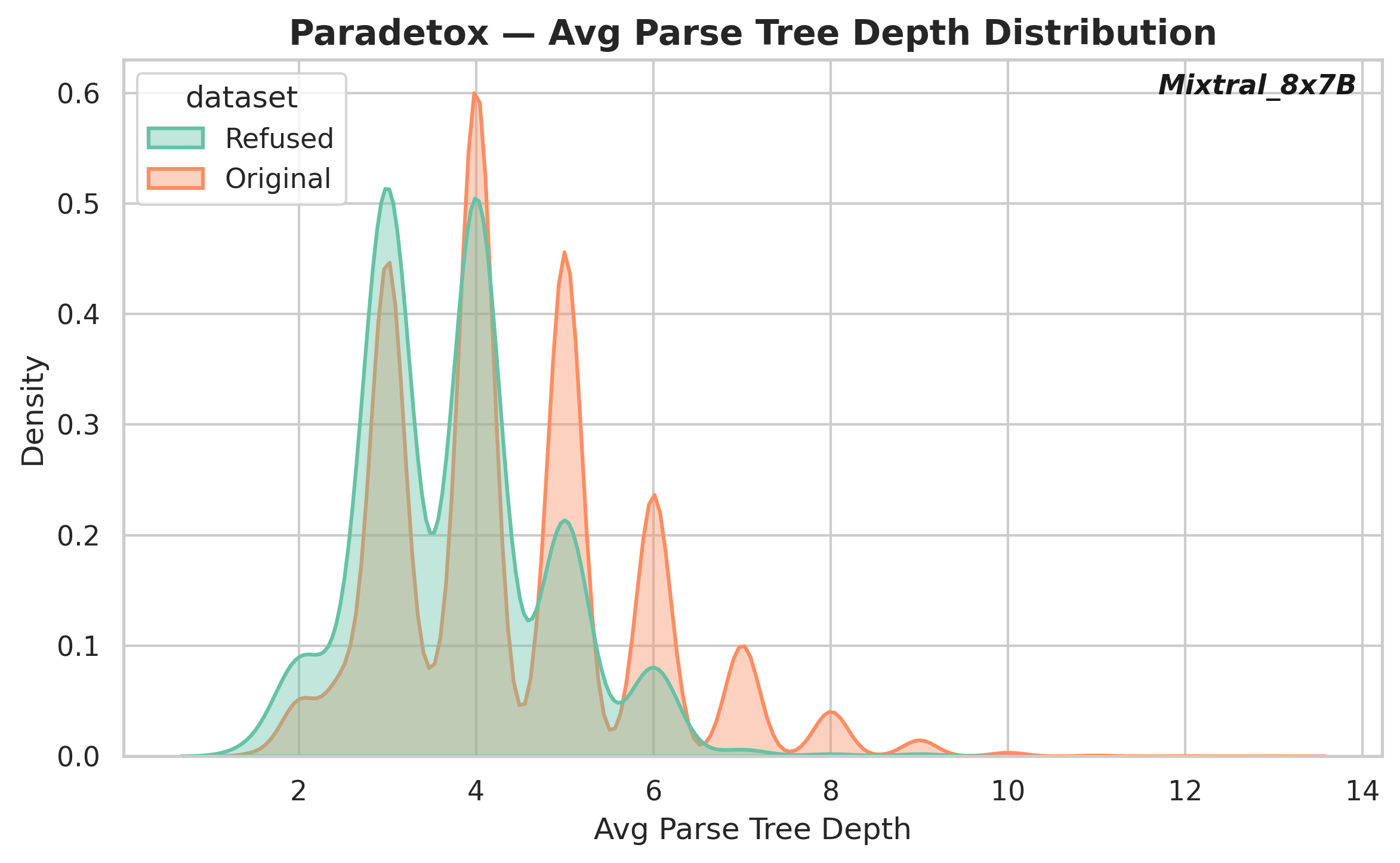}
    \caption{Paradetox}
  \end{subfigure}\hfill
  \begin{subfigure}{0.32\textwidth}
    \centering
    \includegraphics[width=\linewidth]{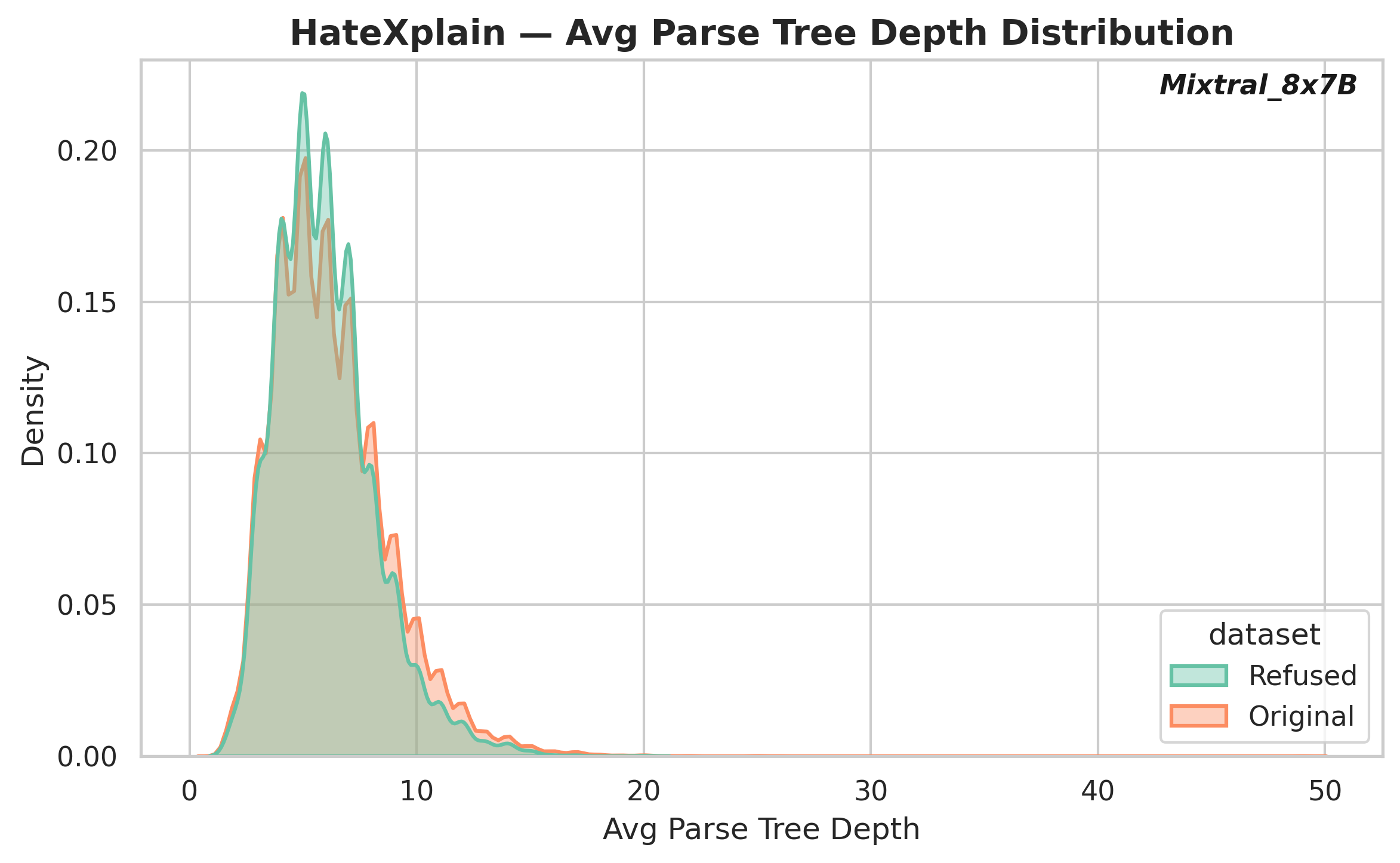}
    \caption{HateXplain}
  \end{subfigure}

  \caption{Parse tree depth distributions across datasets for Mistral 7B and Mixtral 8$\times$7B.}
  \label{fig:parse_tree_mistral}
\end{figure*}

\begin{figure*}[htbp]
  \centering
  \begin{subfigure}{0.32\textwidth}
    \centering
    \includegraphics[width=\linewidth]{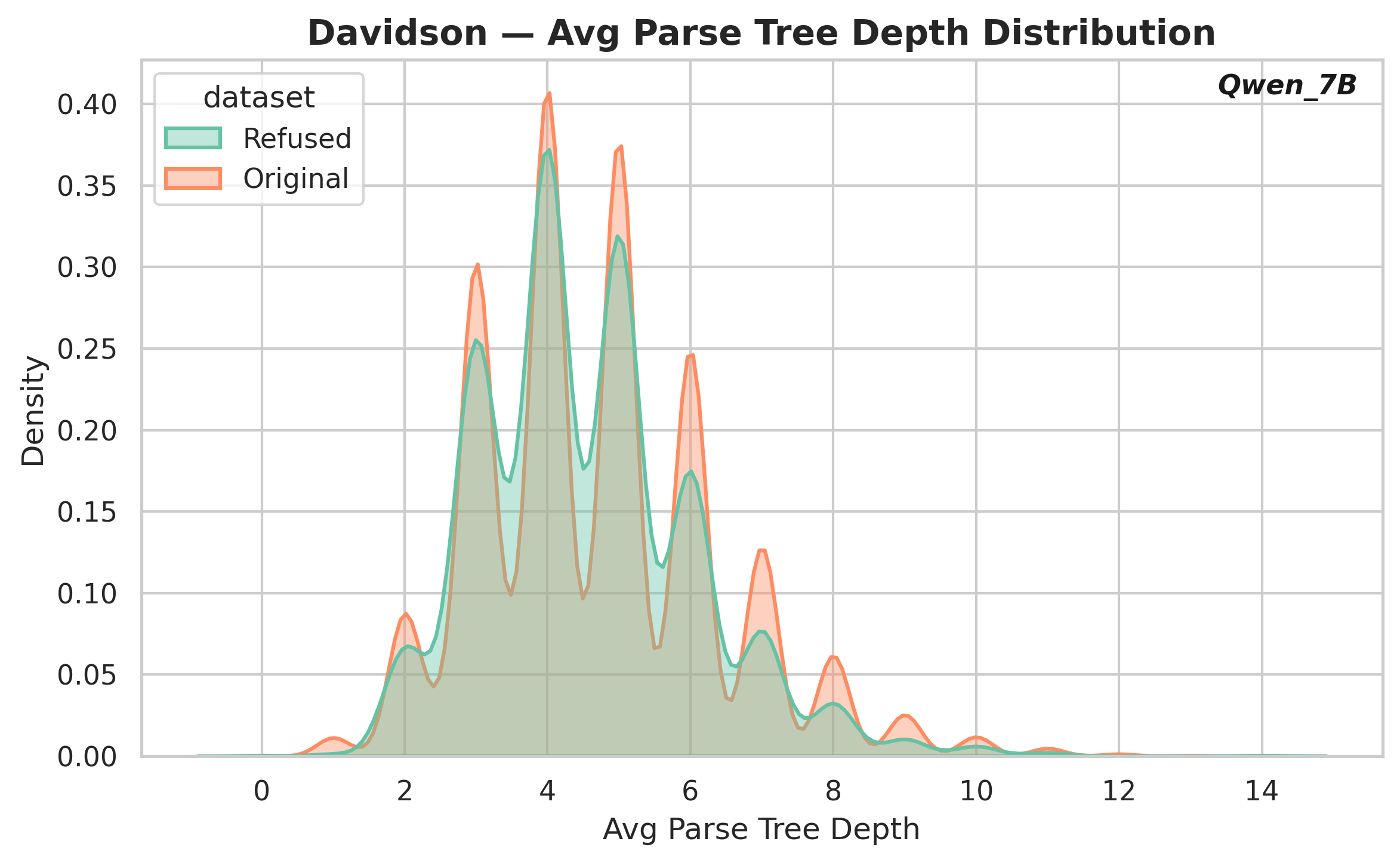}
    \caption{Davidson}
  \end{subfigure}\hfill
  \begin{subfigure}{0.32\textwidth}
    \centering
    \includegraphics[width=\linewidth]{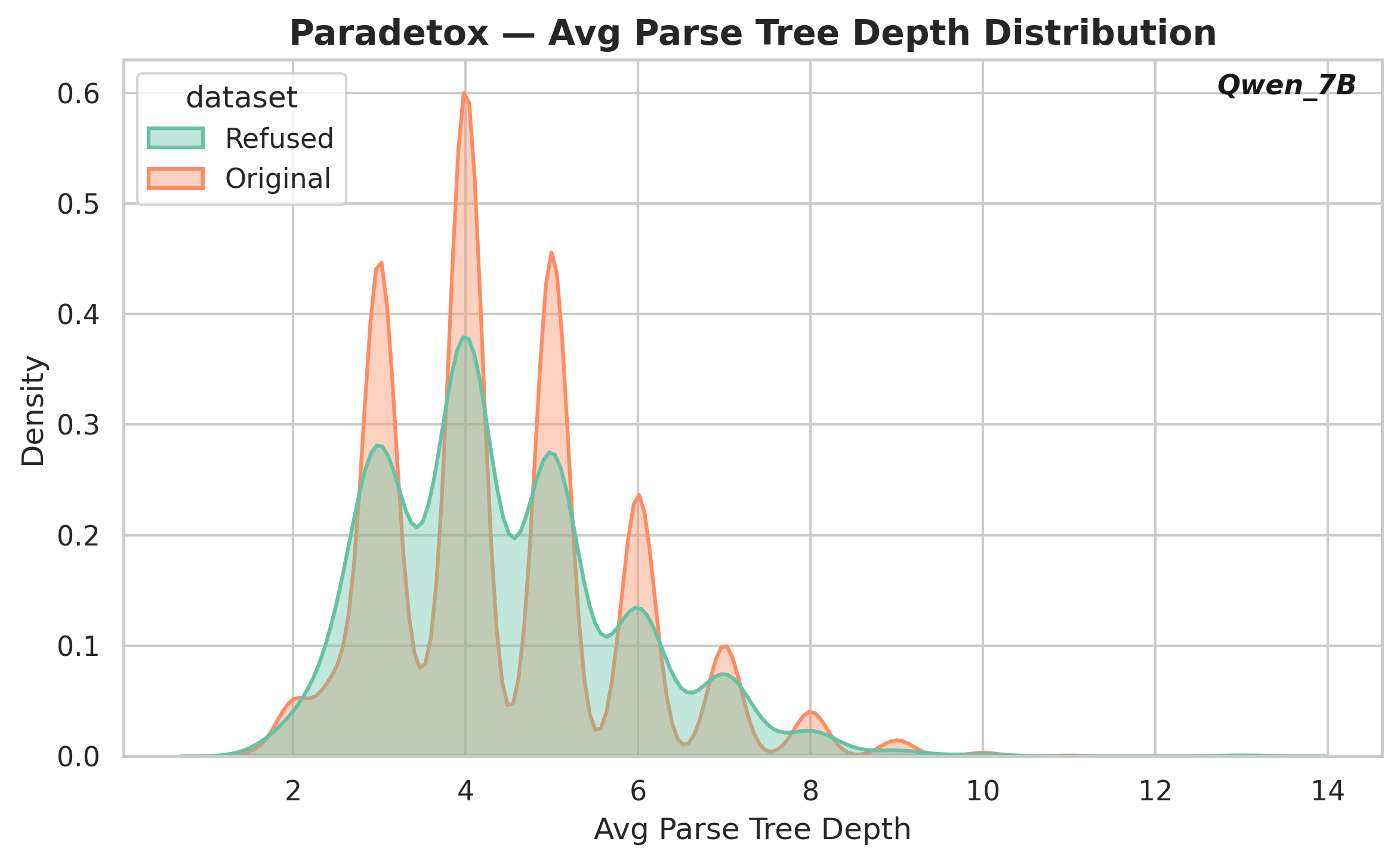}
    \caption{Paradetox}
  \end{subfigure}\hfill
  \begin{subfigure}{0.32\textwidth}
    \centering
    \includegraphics[width=\linewidth]{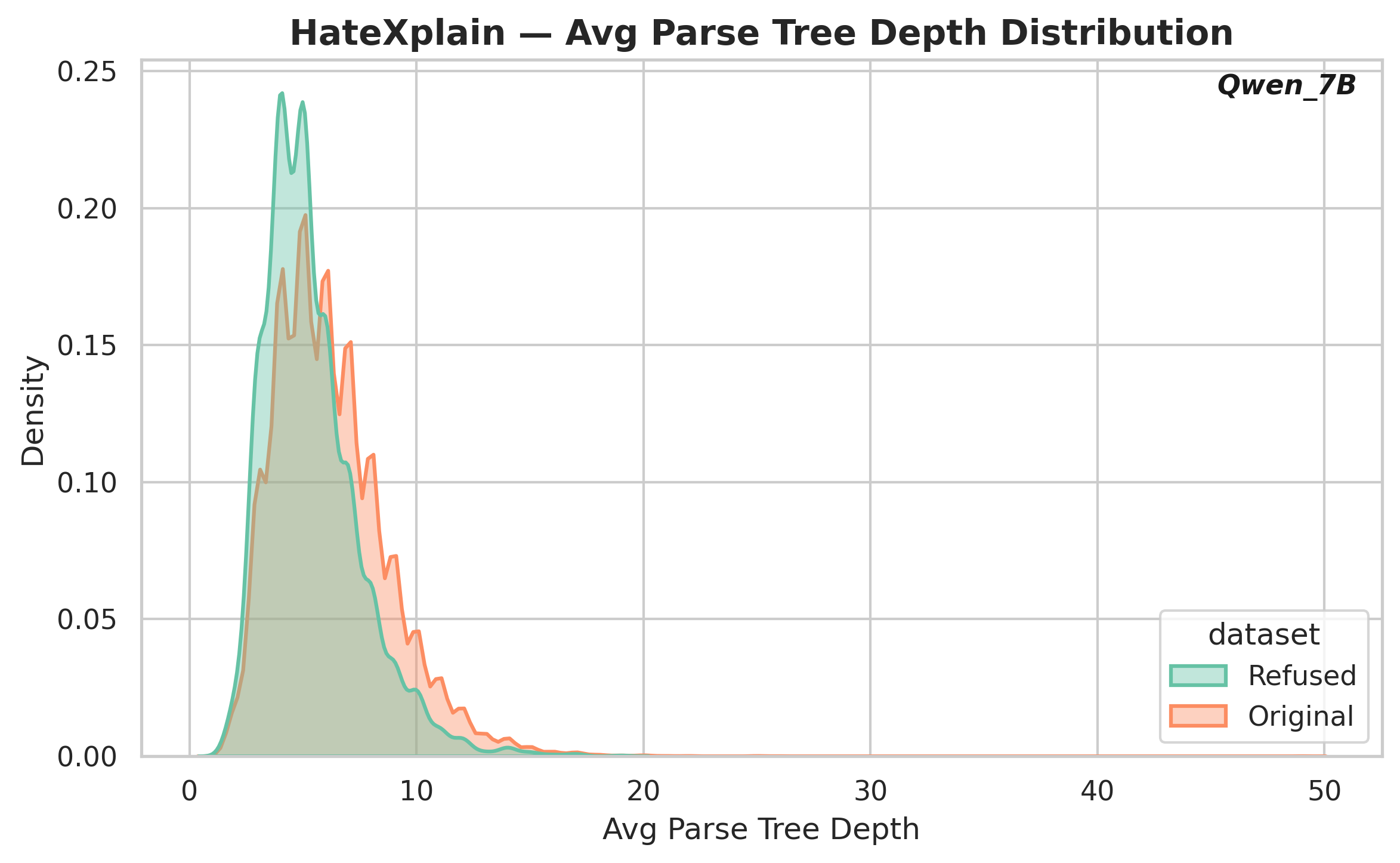}
    \caption{HateXplain}
  \end{subfigure}

  \vspace{0.4em}

  \begin{subfigure}{0.32\textwidth}
    \centering
    \includegraphics[width=\linewidth]{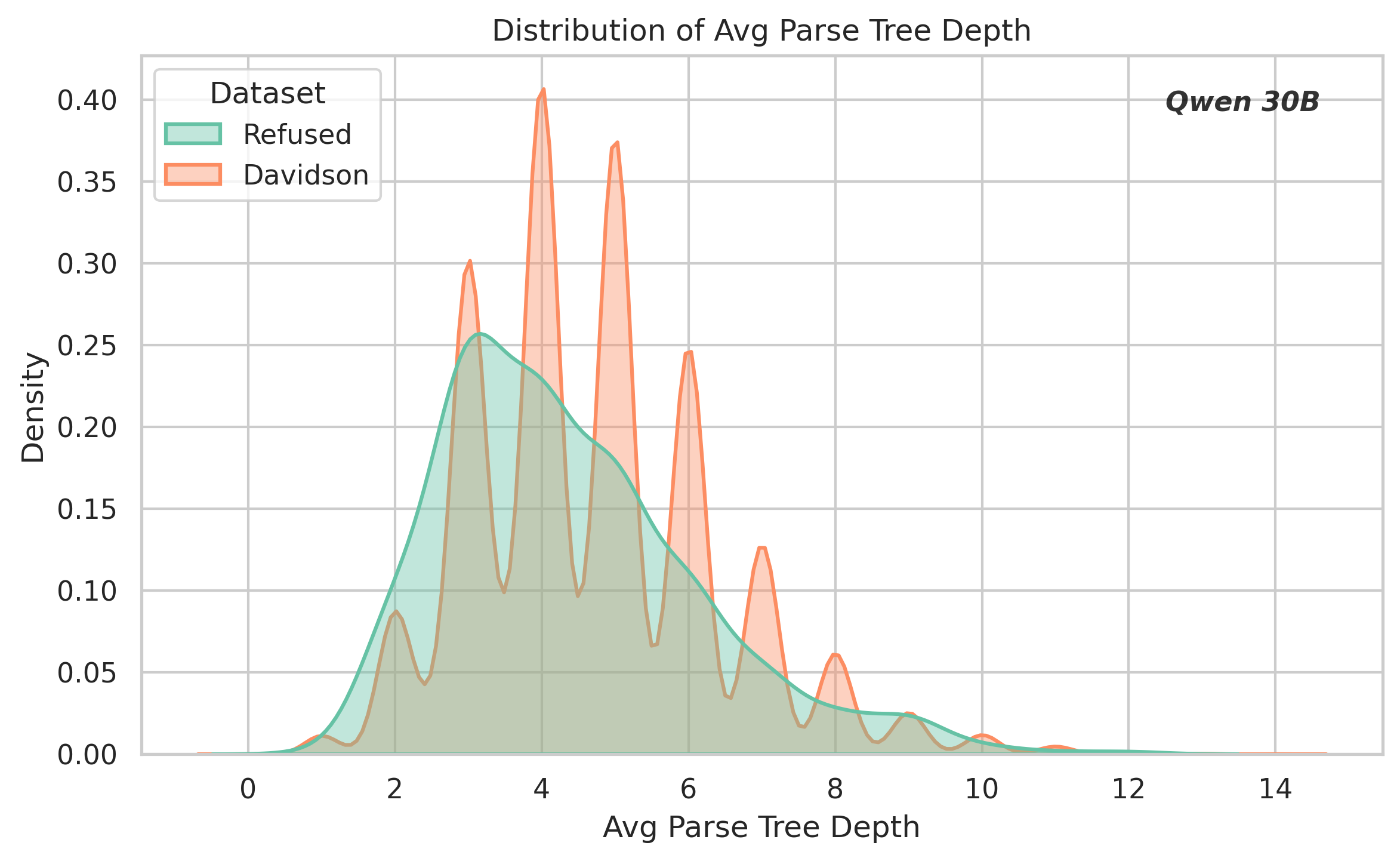}
    \caption{Davidson}
  \end{subfigure}\hfill
  \begin{subfigure}{0.32\textwidth}
    \centering
    \includegraphics[width=\linewidth]{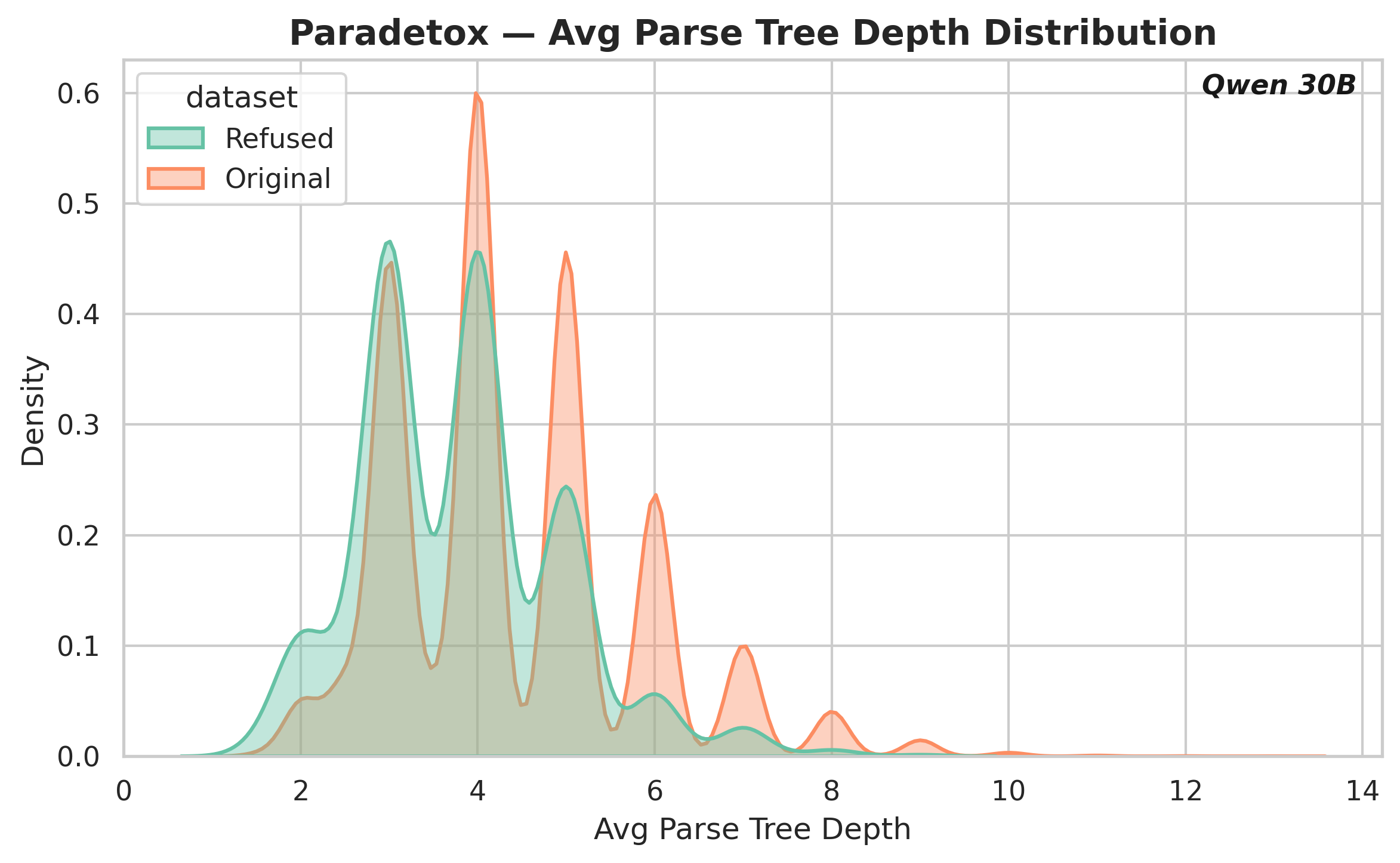}
    \caption{Paradetox}
  \end{subfigure}\hfill
  \begin{subfigure}{0.32\textwidth}
    \centering
    \includegraphics[width=\linewidth]{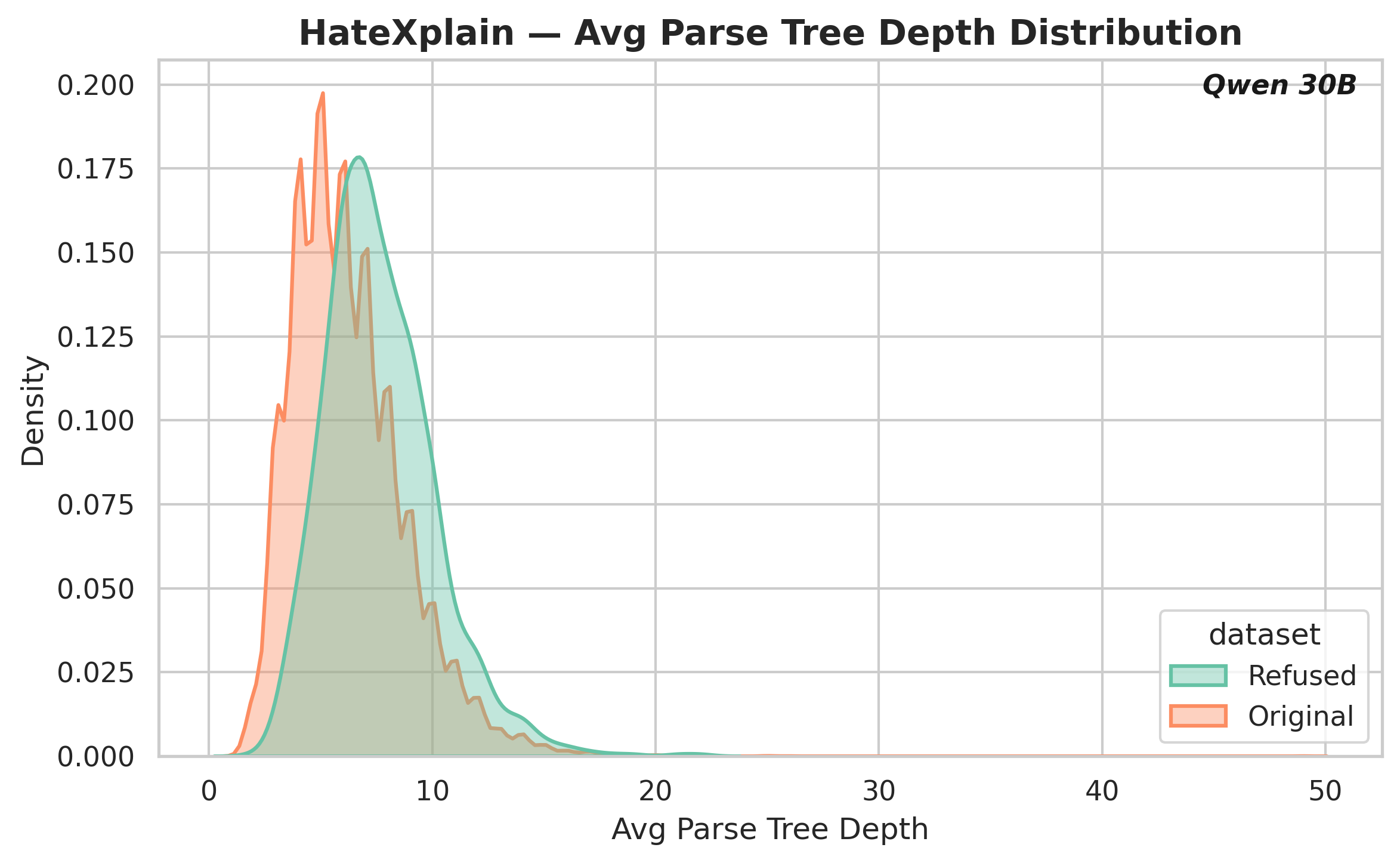}
    \caption{HateXplain}
  \end{subfigure}

  \caption{Parse tree depth distributions across datasets for Qwen2.5 7B and Qwen3 30B.}
  \label{fig:parse_tree_qwen}
\end{figure*}

\begin{figure*}[htbp]
  \centering
  \begin{subfigure}{0.32\textwidth}
    \centering
    \includegraphics[width=\linewidth]{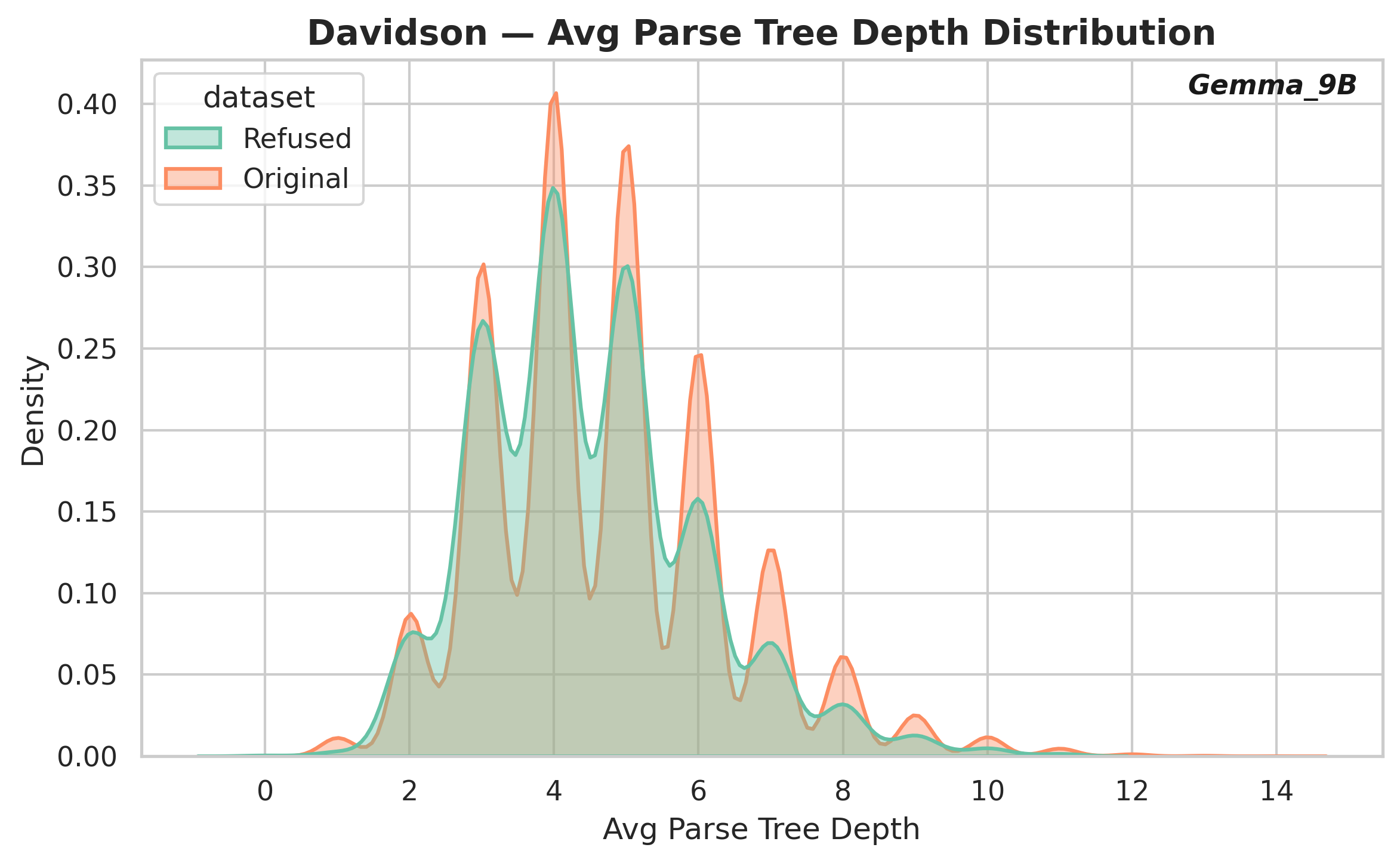}
    \caption{Davidson}
  \end{subfigure}\hfill
  \begin{subfigure}{0.32\textwidth}
    \centering
    \includegraphics[width=\linewidth]{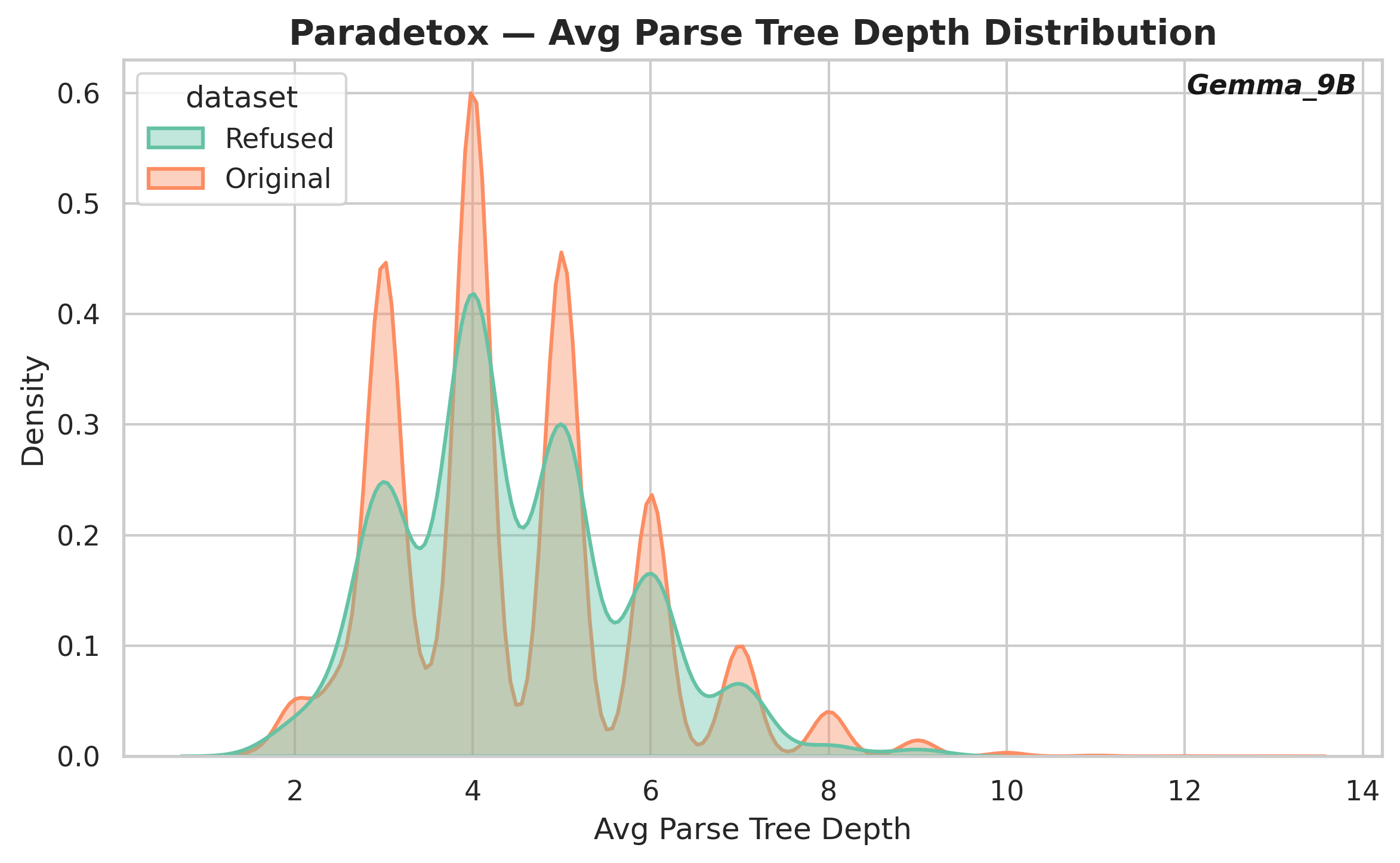}
    \caption{Paradetox}
  \end{subfigure}\hfill
  \begin{subfigure}{0.32\textwidth}
    \centering
    \includegraphics[width=\linewidth]{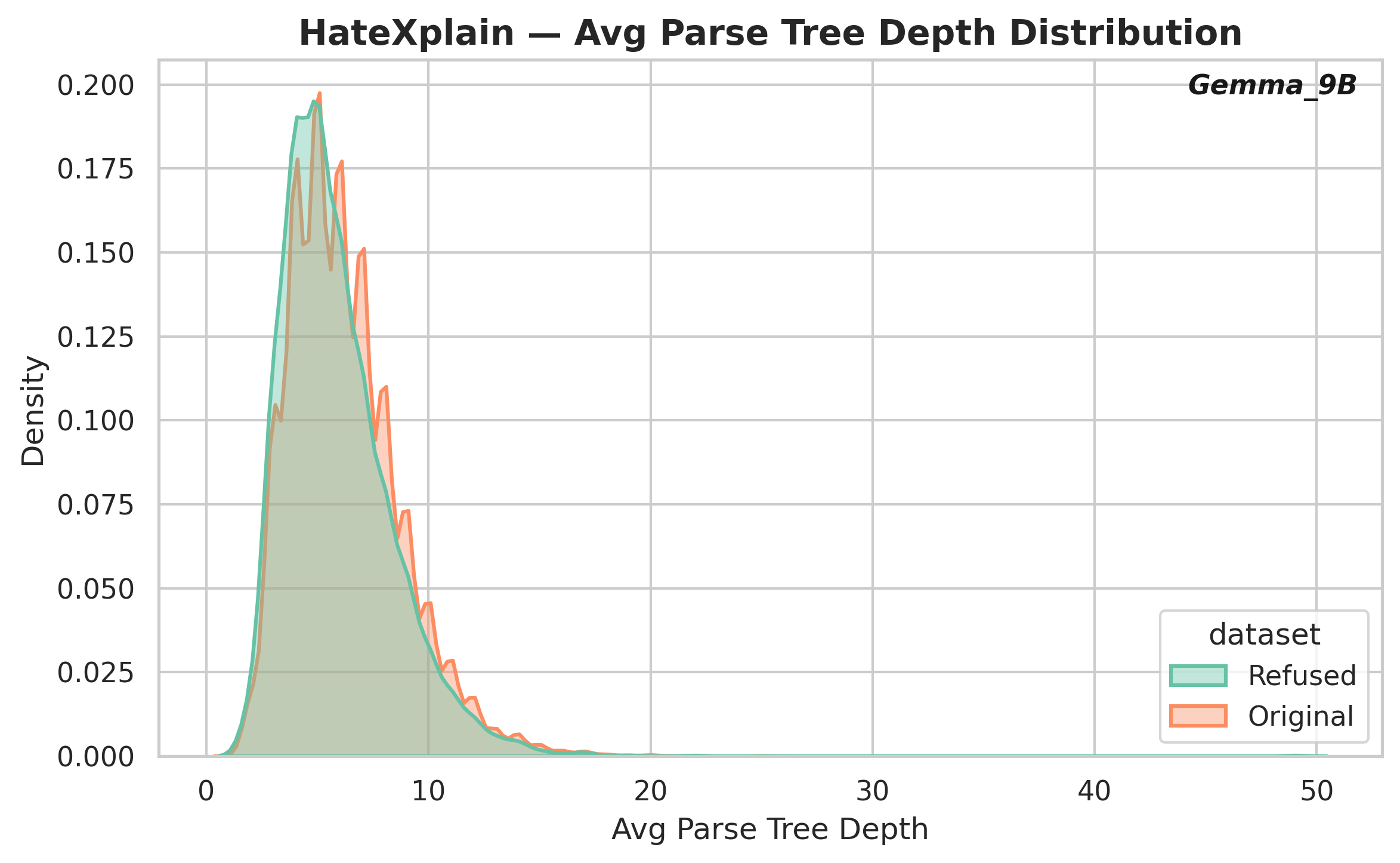}
    \caption{HateXplain}
  \end{subfigure}

  \vspace{0.4em}

  \begin{subfigure}{0.32\textwidth}
    \centering
    \includegraphics[width=\linewidth]{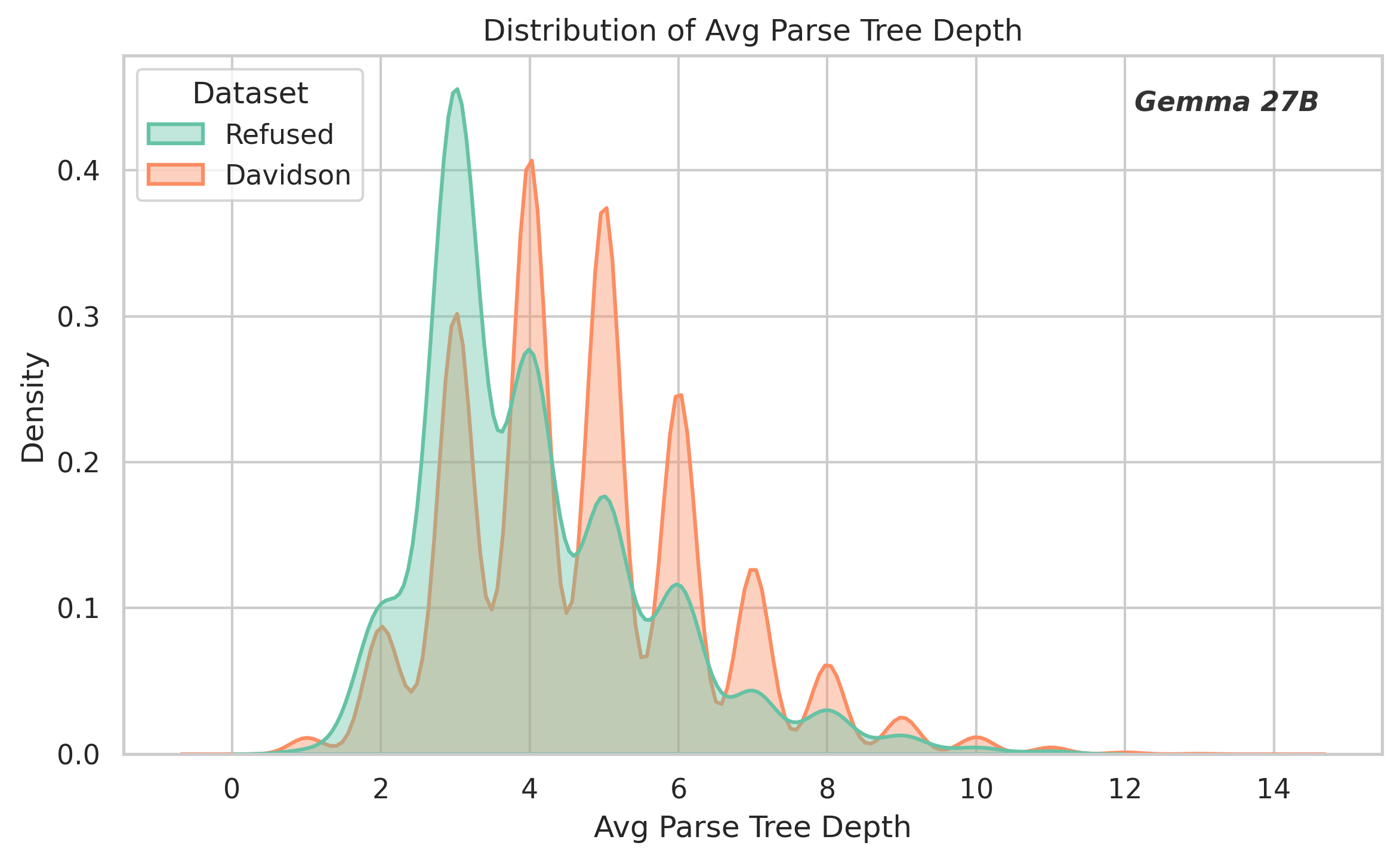}
    \caption{Davidson}
  \end{subfigure}\hfill
  \begin{subfigure}{0.32\textwidth}
    \centering
    \includegraphics[width=\linewidth]{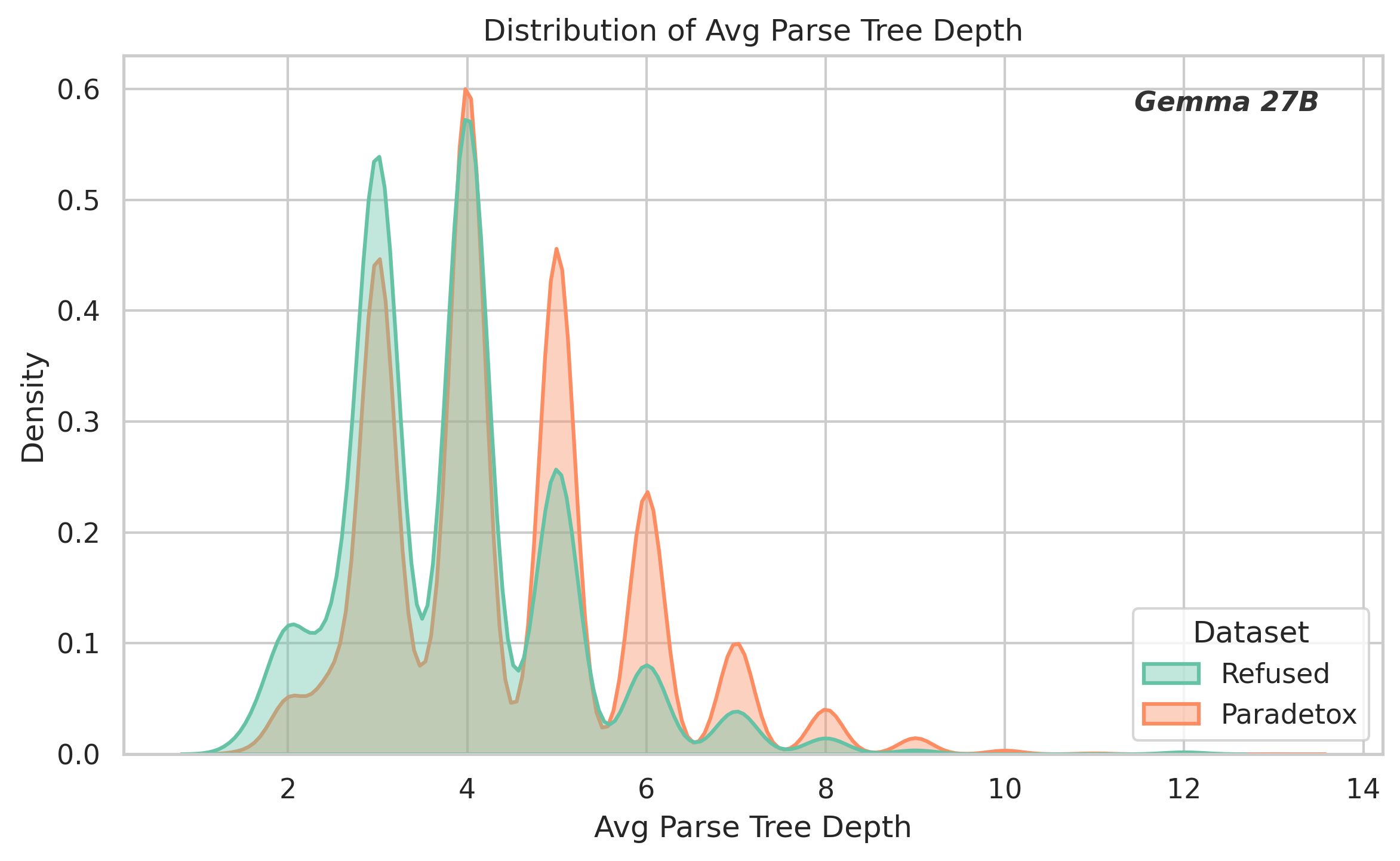}
    \caption{Paradetox}
  \end{subfigure}\hfill
  \begin{subfigure}{0.32\textwidth}
    \centering
    \includegraphics[width=\linewidth]{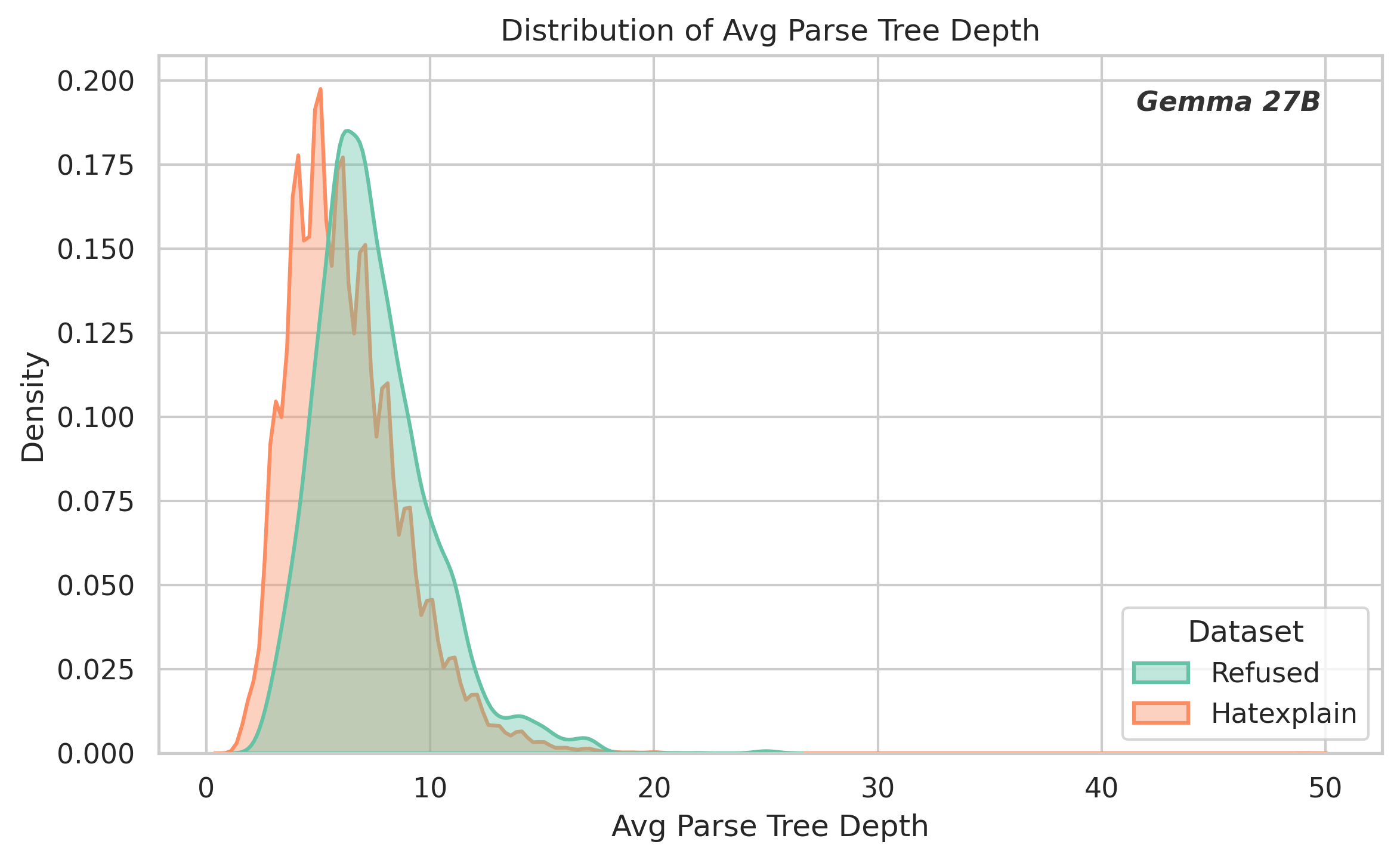}
    \caption{HateXplain}
  \end{subfigure}

  \caption{Parse tree depth distributions across datasets for Gemma2 9B and Gemma3 27B.}
  \label{fig:parse_tree_gemma}
\end{figure*}

\paragraph{Parse Tree Depth.}
Figures~\ref{fig:parse_tree_gpt}–\ref{fig:parse_tree_gemma} report the distributions of parse tree depth, capturing syntactic complexity across false refusals.

Across all three linguistic dimensions, the distributions of falsely refused samples closely overlap with those of the original datasets. These results indicate that token length, clause count, and syntactic complexity do not exhibit consistent or systematic patterns that trigger false refusal behavior in LLM-based hate speech detoxification.

\FloatBarrier

\section{Bias Analysis on Multilingual Datasets}\label{app:analysis_multilingual}

\subsection{Toxicity Level}

\begin{figure}[H]
    \centering
    \includegraphics[width=\linewidth]{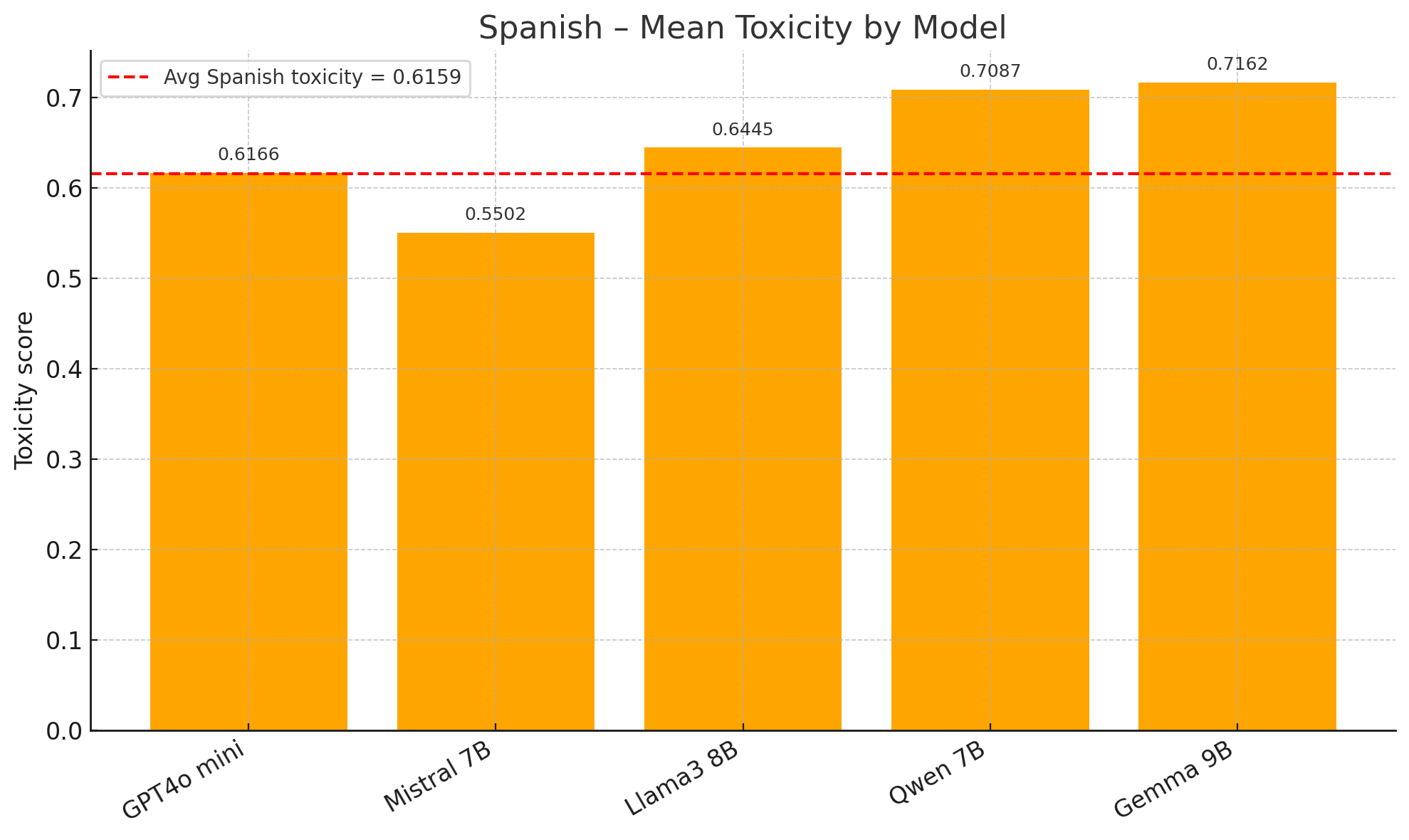}
    \caption{Toxicity Score of French Dataset.}
    \label{fig:mult_toxicity}
\end{figure}

As Detoxify~\cite{Detoxify} doesn't support all multilingual datasets,
Figure~\ref{fig:mult_toxicity} shows the toxicity score in French dataset. It suggests that false refusals in French are associated with higher toxicity scores than the average in original dataset.

\subsection{Swear Word Presence}

\begin{figure}[H]
    \centering
    \includegraphics[width=\linewidth]{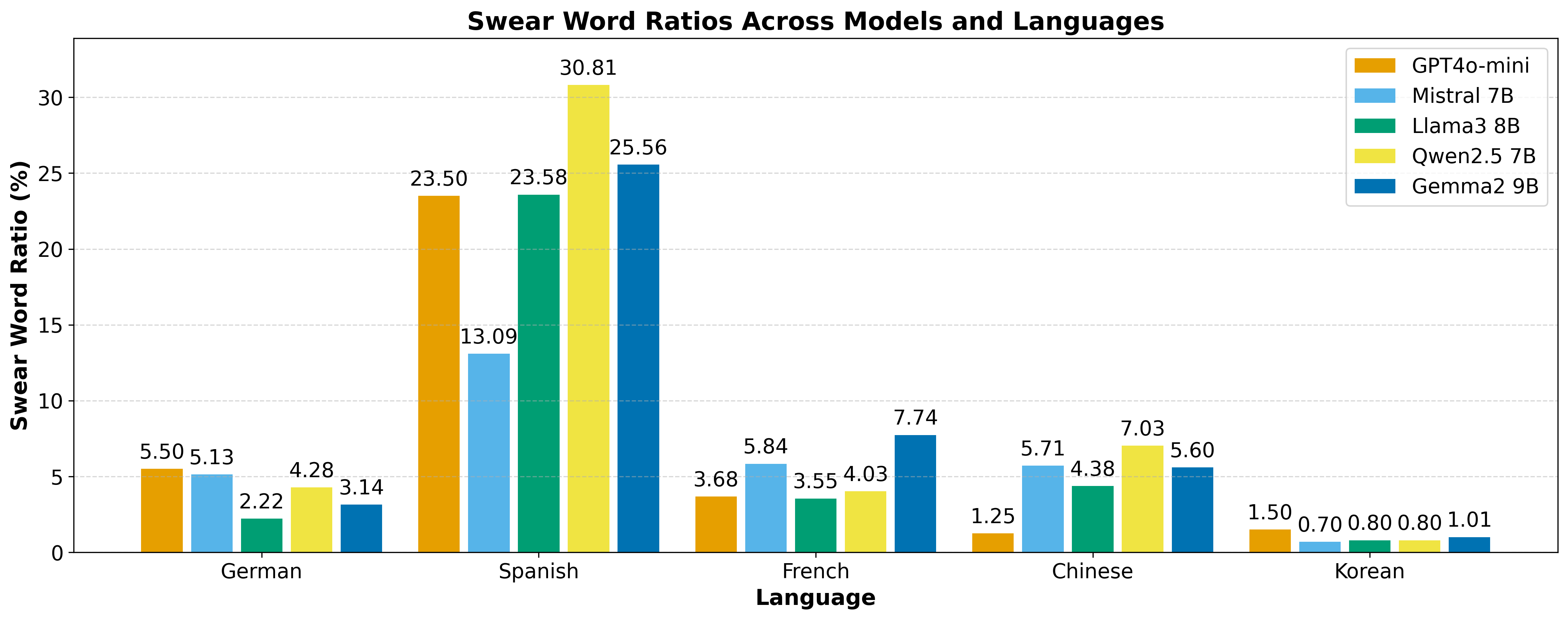}
    \caption{Swear Words Ratio of Multilingual Datasets.}
    \label{fig:mult_swear}
\end{figure}

Swear-word ratios for multilingual datasets are shown in Figure~\ref{fig:mult_swear}. Spanish exhibits the highest ratios across models, exceeding 30\% for Qwen2.5~7B and 25\% for Gemma2~9B, aligning with the elevated false refusal rates observed in Figure~\ref{fig:mult_false_refusal}. This suggests that Spanish false refusals are largely driven by oversensitivity to direct lexical cues. In contrast, French shows substantially lower swear-word ratios (3.68–7.74\%), indicating that false refusals are more influenced by semantic misinterpretation than explicit profanity. German, Korean, and Chinese exhibit consistently low ratios (2–7\%), suggesting that false refusals in these languages are not primarily triggered by profanity but by broader semantic or structural mismatches with English-centric safety alignment. Overall, these results highlight the need for multilingual detoxification mechanisms that jointly calibrate lexical and semantic sensitivity.



\subsection{Sentence Length}

\begin{figure*}[htbp]
  \centering
  \captionsetup{font=small}
  \setlength{\tabcolsep}{3pt}

  \begin{tabular}{cc}
    \subcaptionbox{\textbf{Spanish}\label{fig:spanish_token}}{
      \begin{tabular}{cc}
        \includegraphics[width=0.235\textwidth]{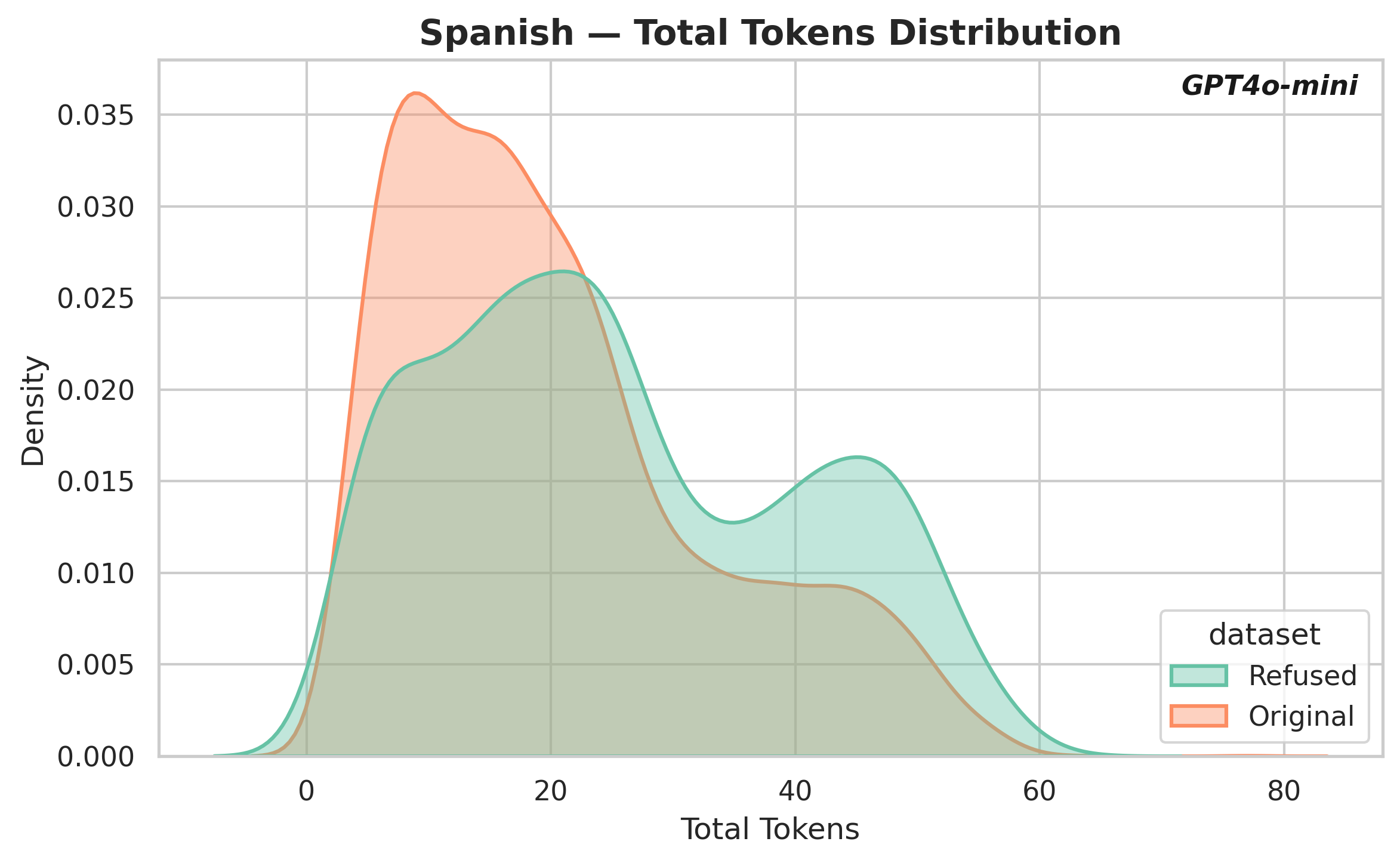} &
        \includegraphics[width=0.235\textwidth]{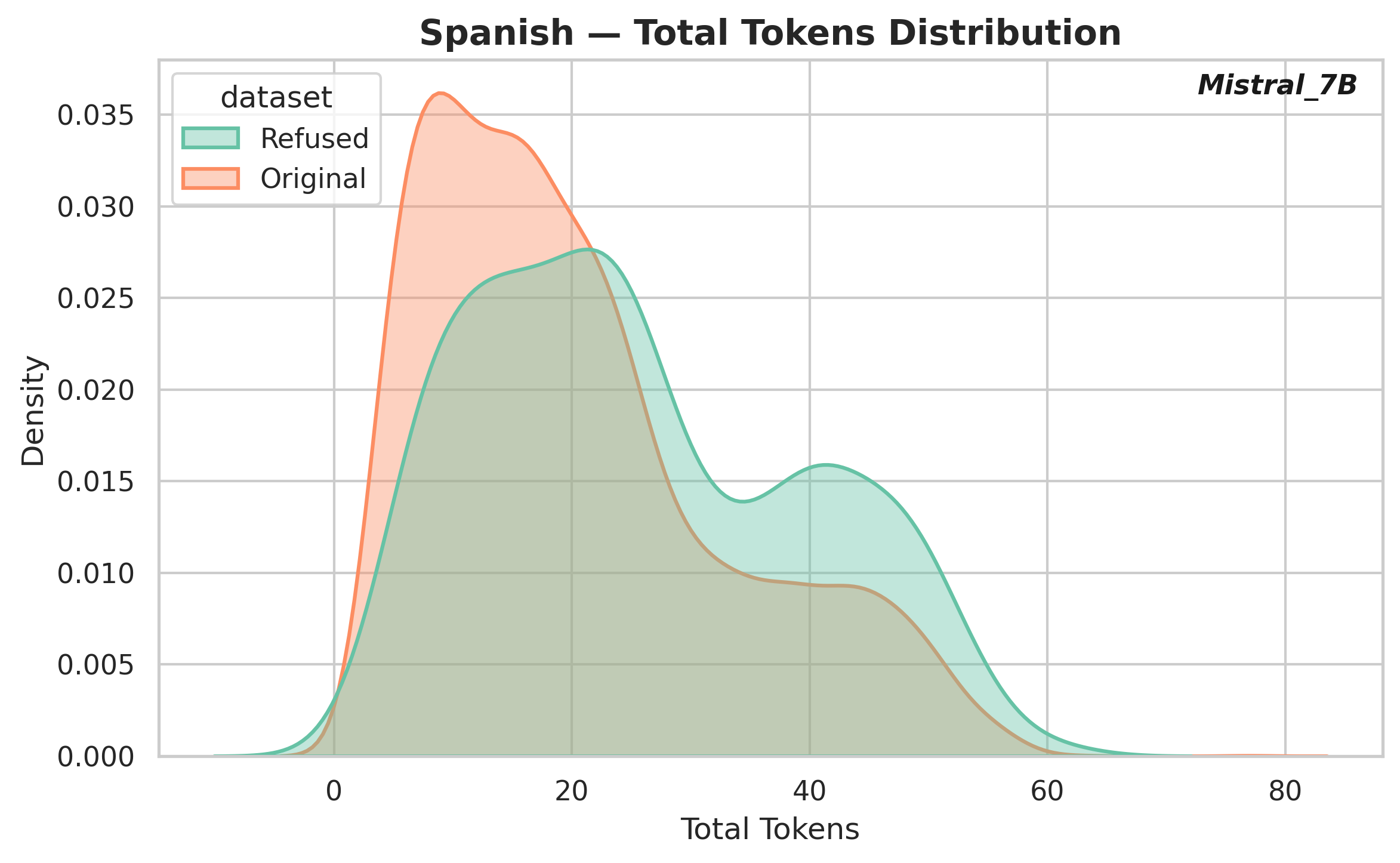} \\
        \includegraphics[width=0.235\textwidth]{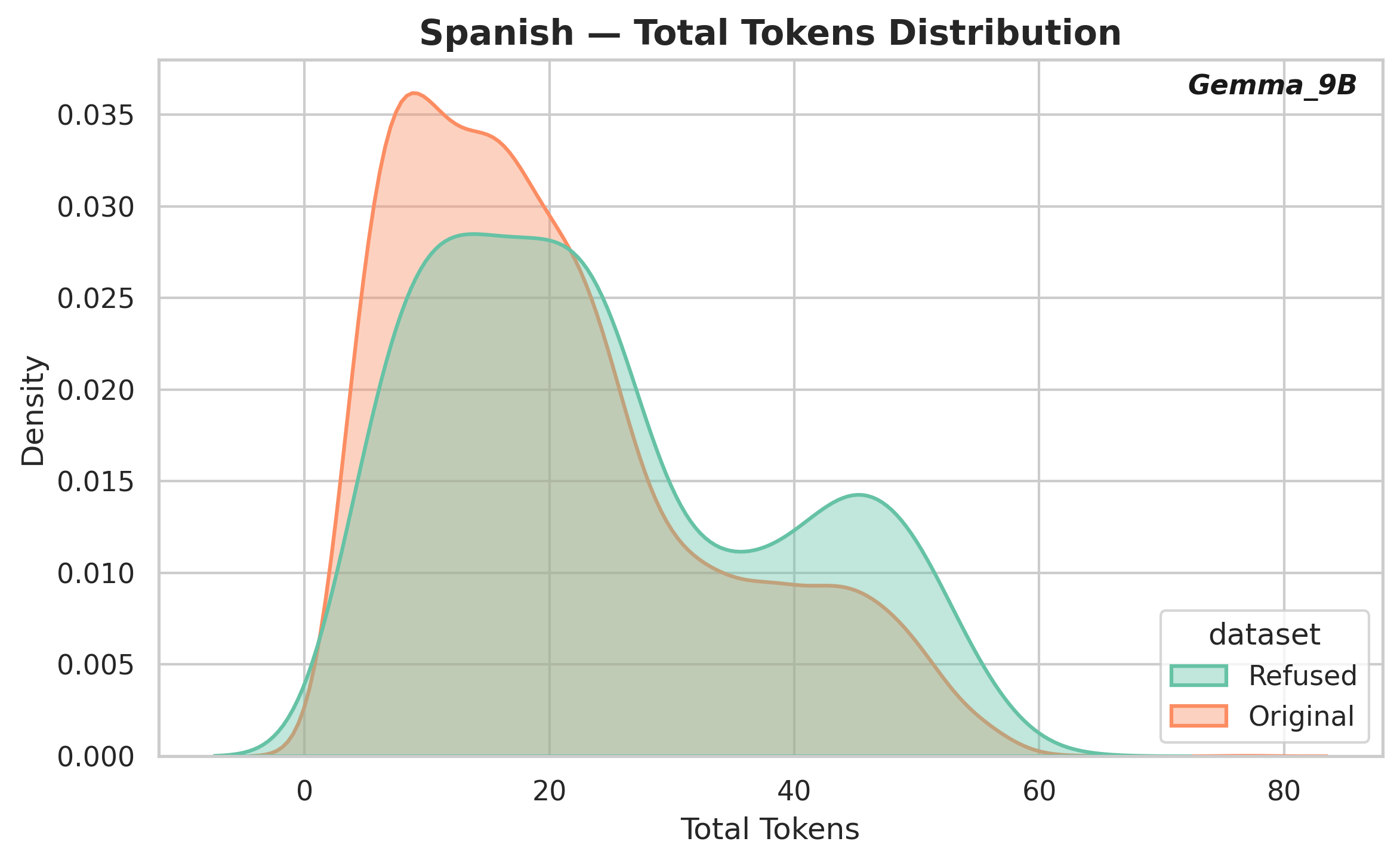} &
        \includegraphics[width=0.235\textwidth]{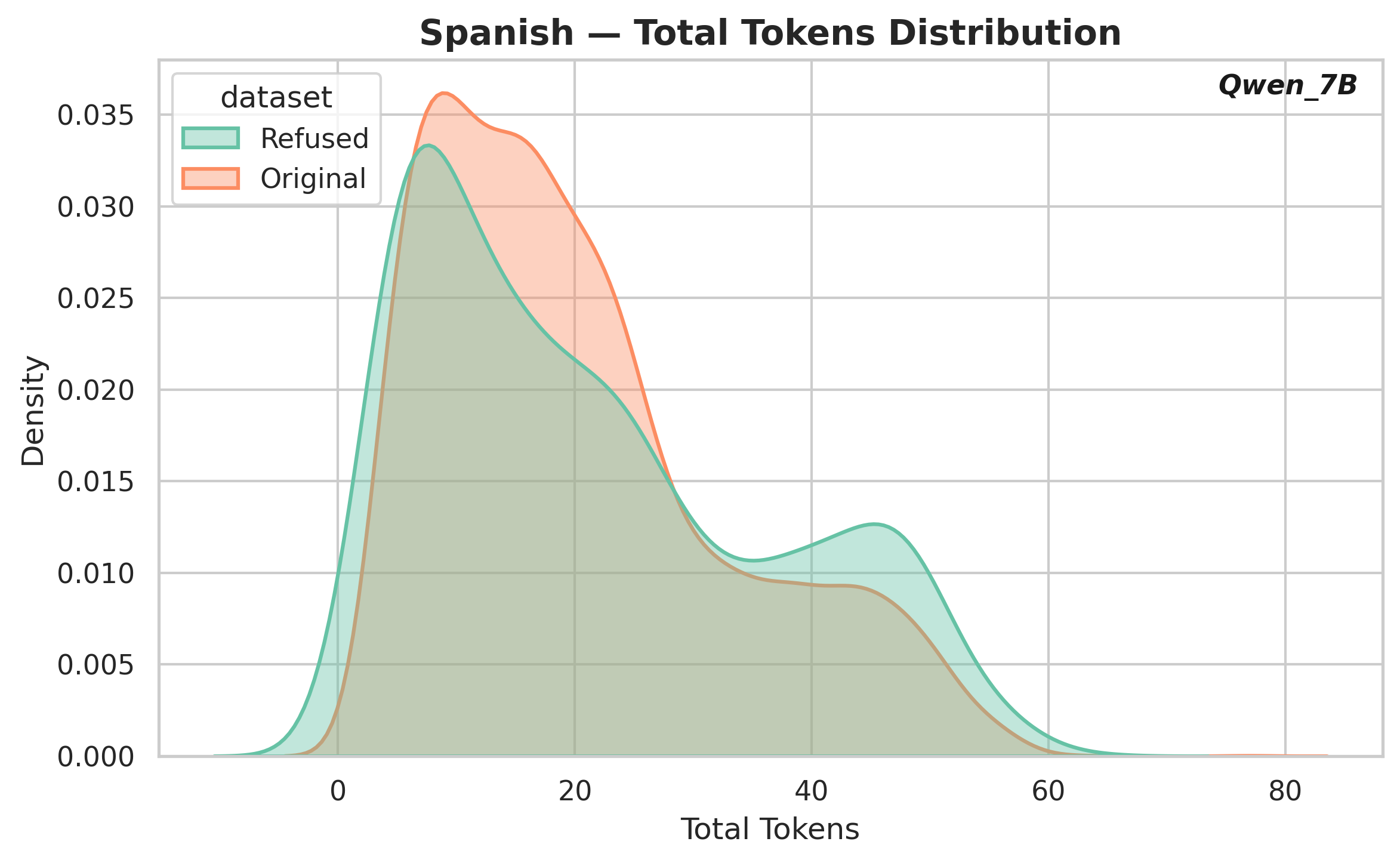} \\
      \end{tabular}
    }
    &
    \subcaptionbox{\textbf{Chinese}\label{fig:chinese_token}}{
      \begin{tabular}{cc}
        \includegraphics[width=0.235\textwidth]{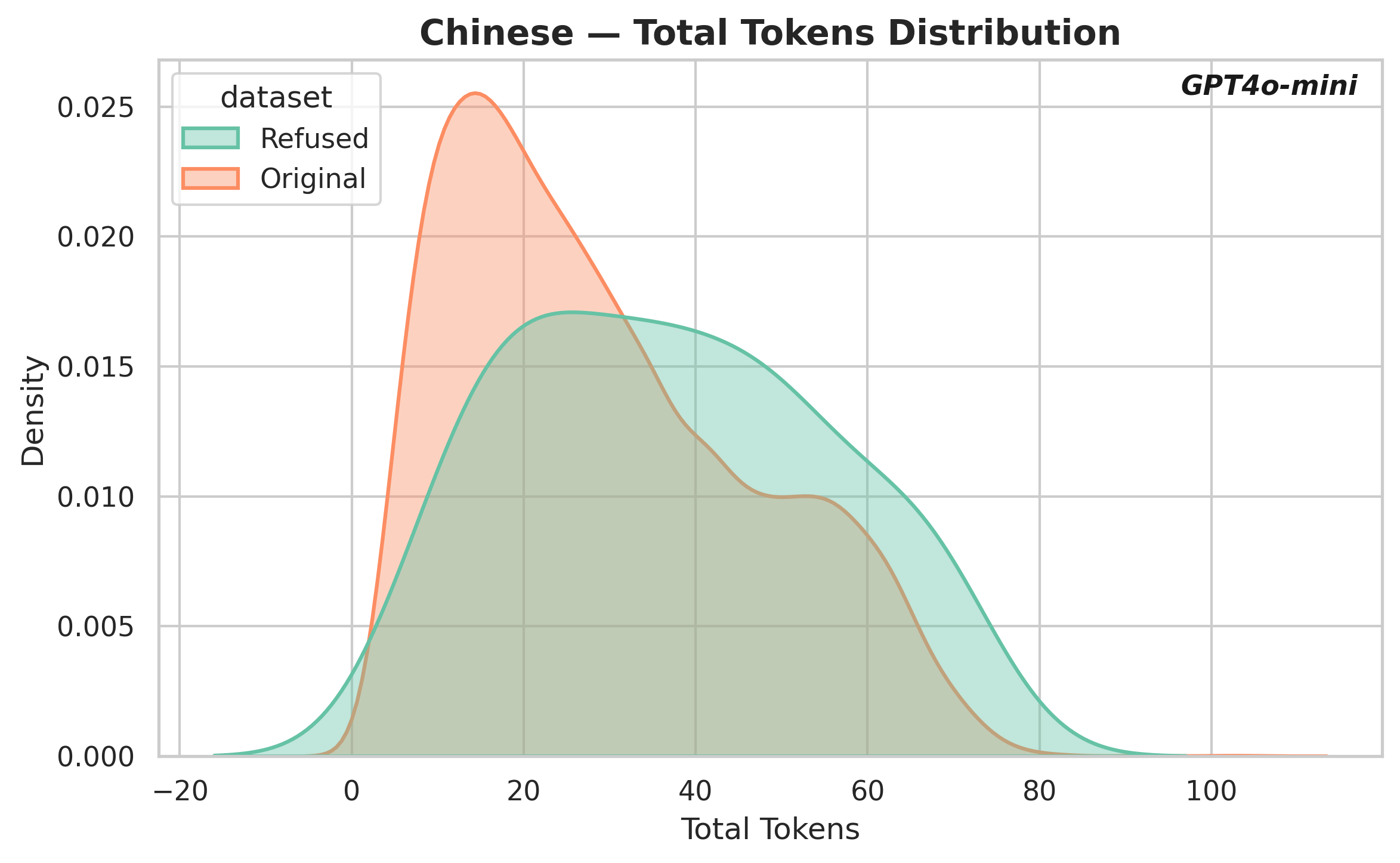} &
        \includegraphics[width=0.235\textwidth]{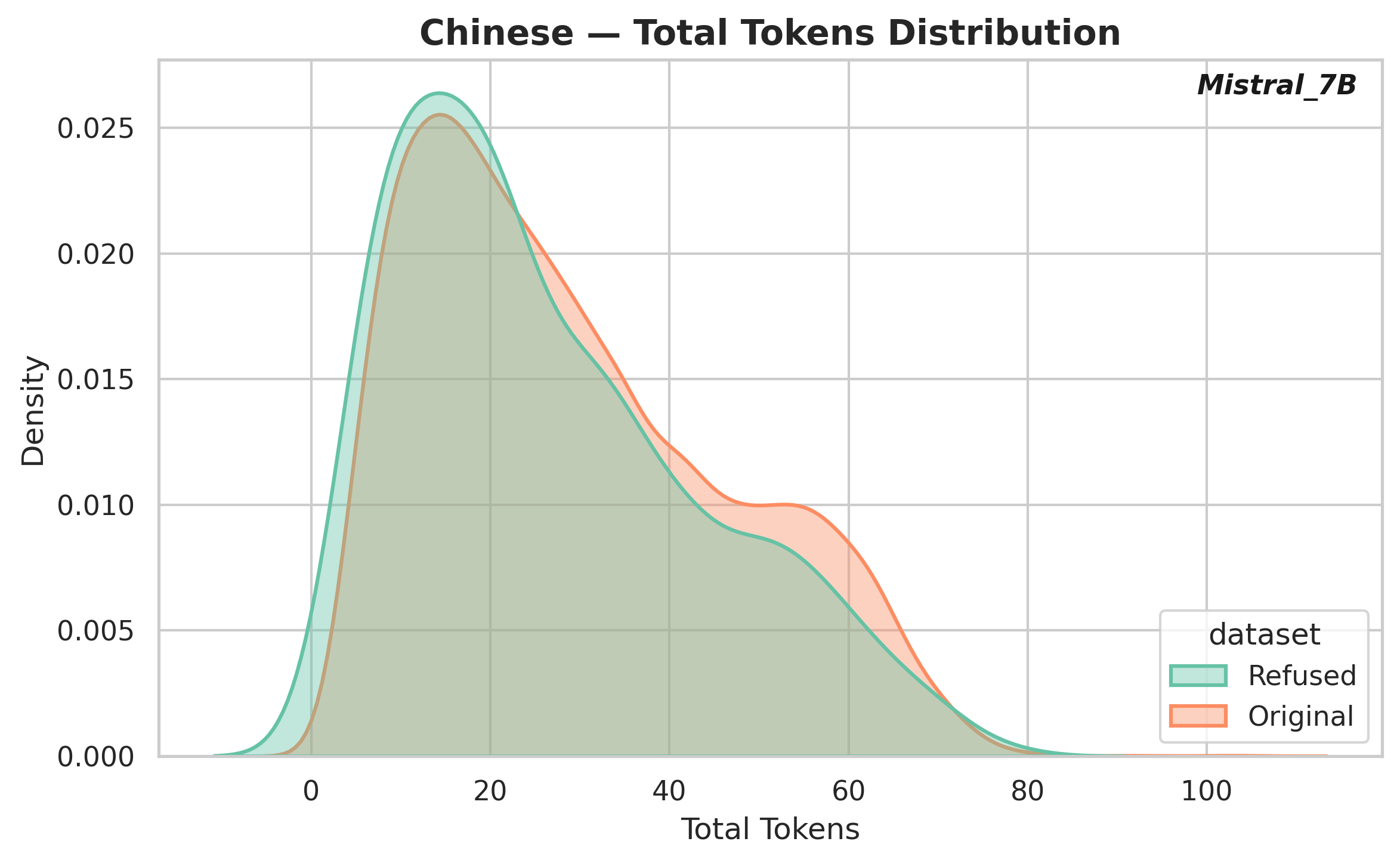} \\
        \includegraphics[width=0.235\textwidth]{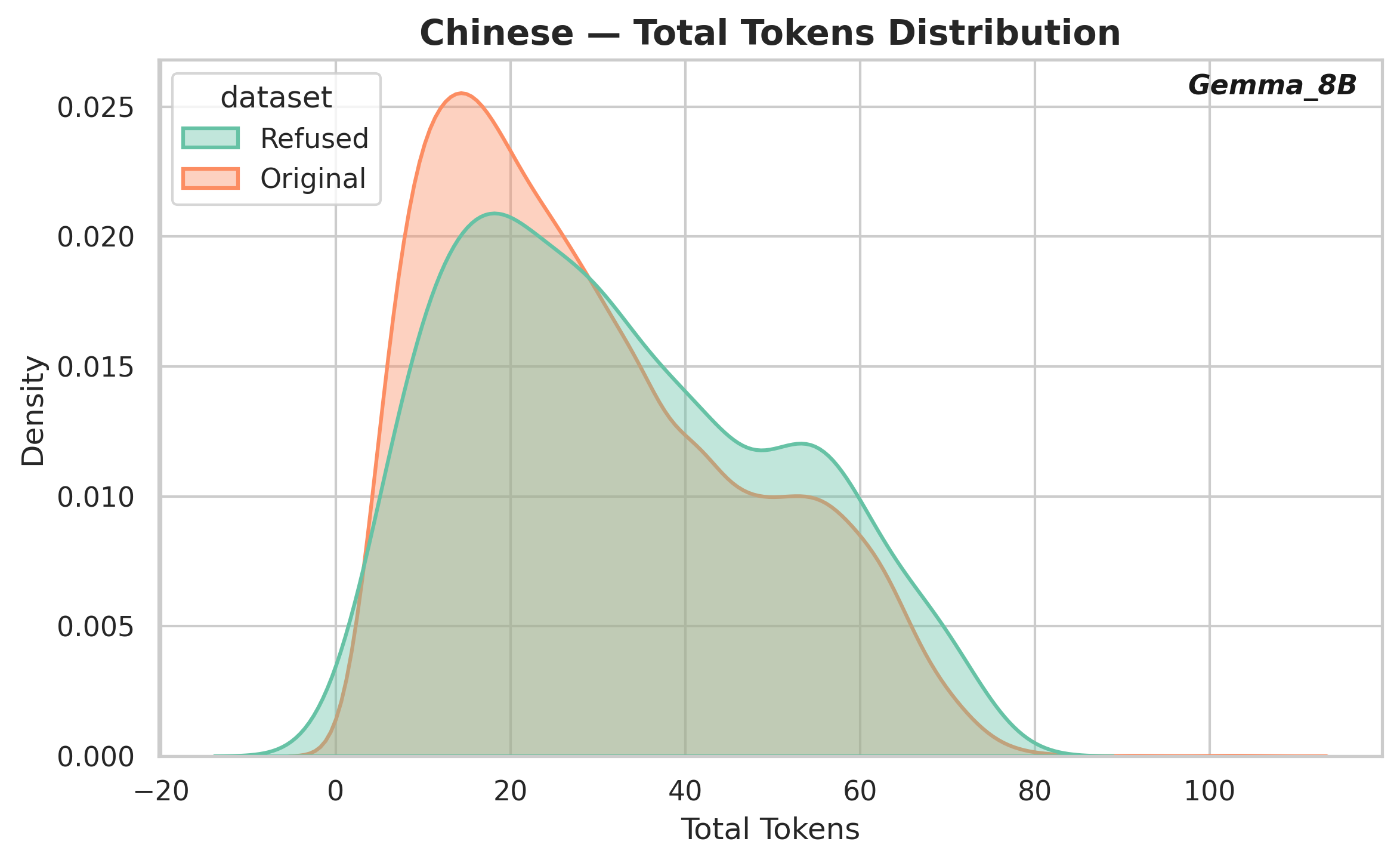} &
        \includegraphics[width=0.235\textwidth]{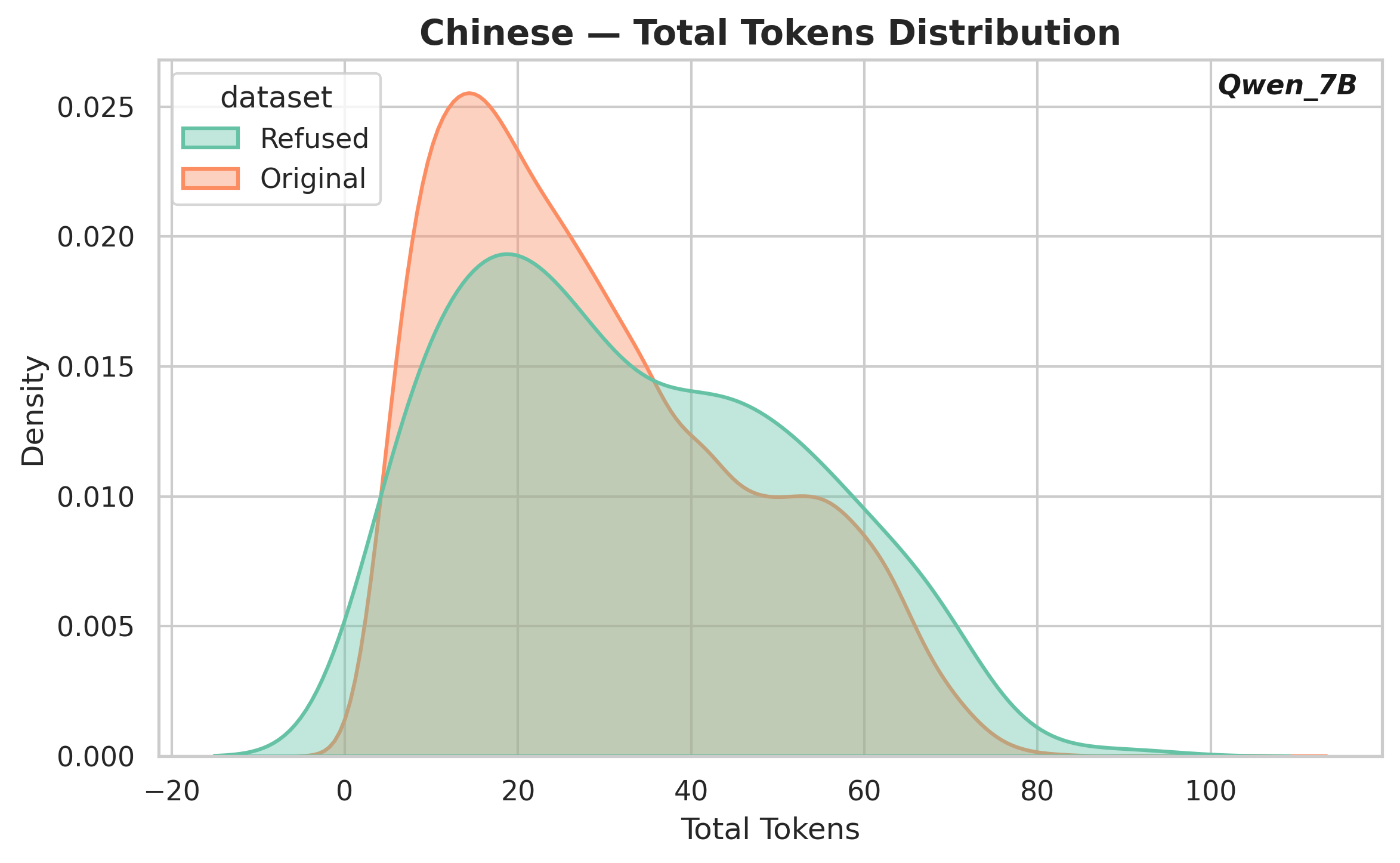} \\
      \end{tabular}
    }
  \end{tabular}
  \caption{Token count distribution for false refusals versus original samples across representative models in Spanish and Chinese dataset.}
  \label{fig:mult_token}
\end{figure*}

\begin{figure*}[htbp]
  \centering
  \setlength{\tabcolsep}{3pt}
  \begin{subfigure}{0.32\textwidth}
    \centering
    \includegraphics[width=\linewidth]{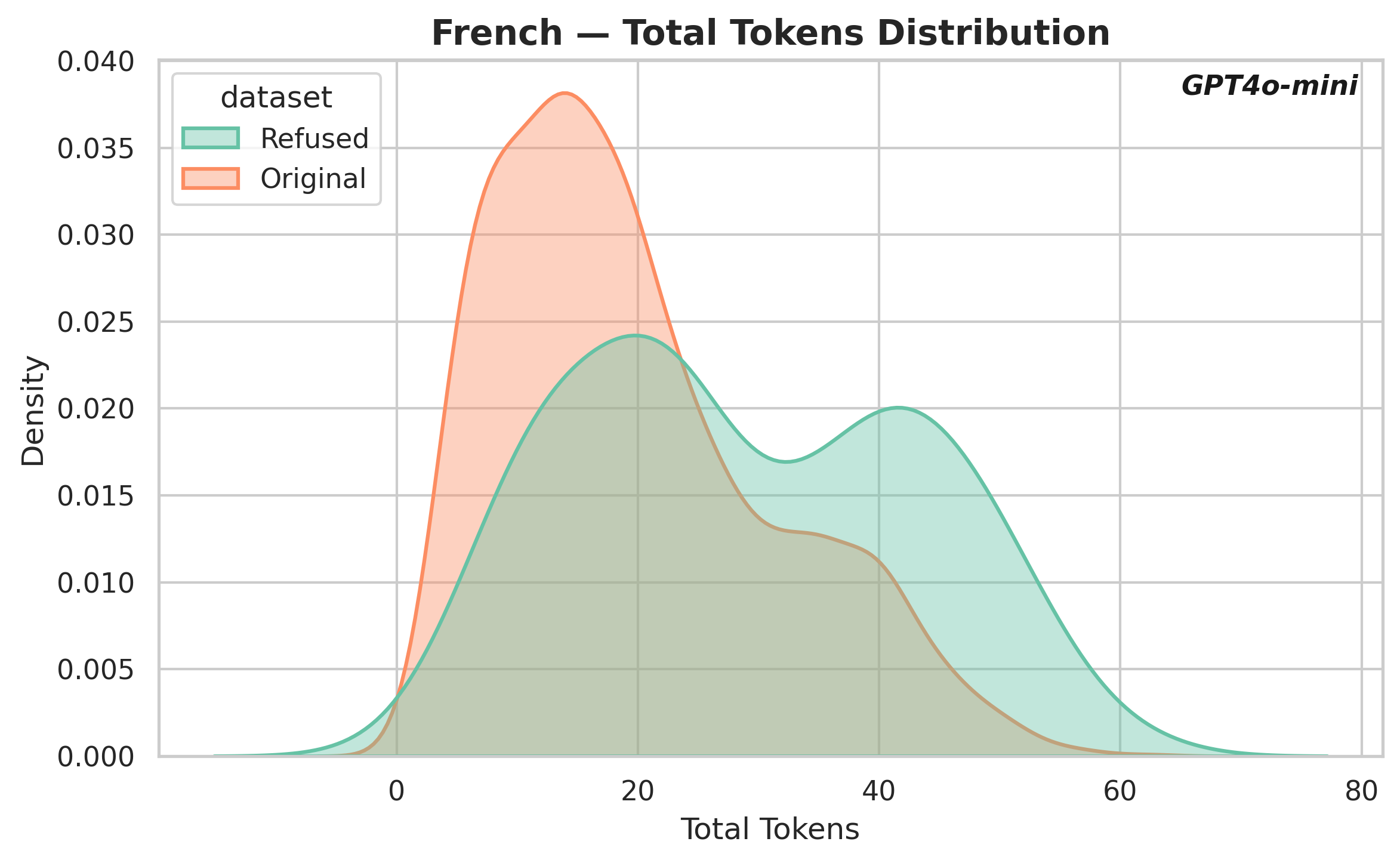}
    \caption{GPT4o-mini}
  \end{subfigure}\hfill
  \begin{subfigure}{0.32\textwidth}
    \centering
    \includegraphics[width=\linewidth]{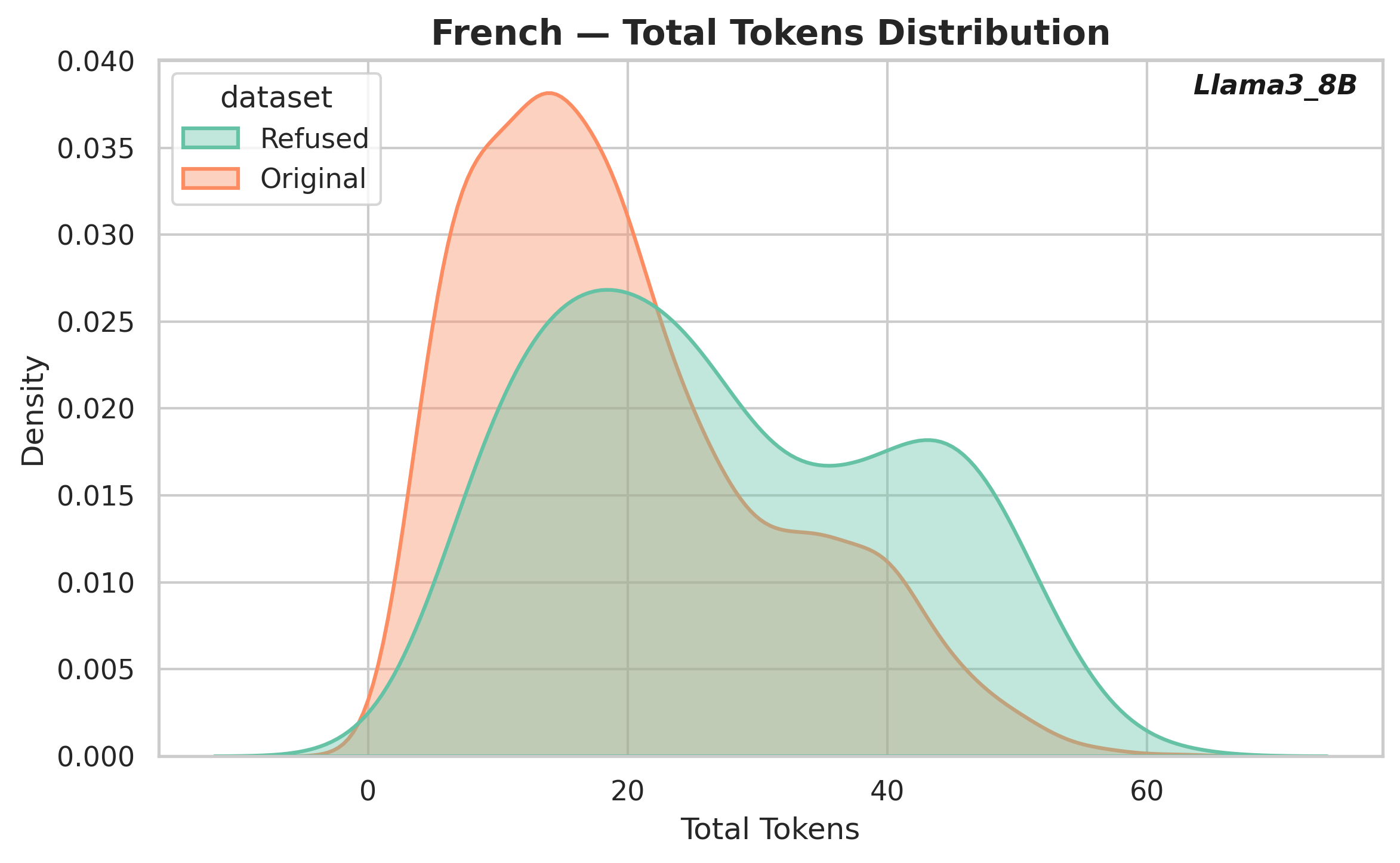}
    \caption{Llama3 8B}
  \end{subfigure}\hfill
  \begin{subfigure}{0.32\textwidth}
    \centering
    \includegraphics[width=\linewidth]{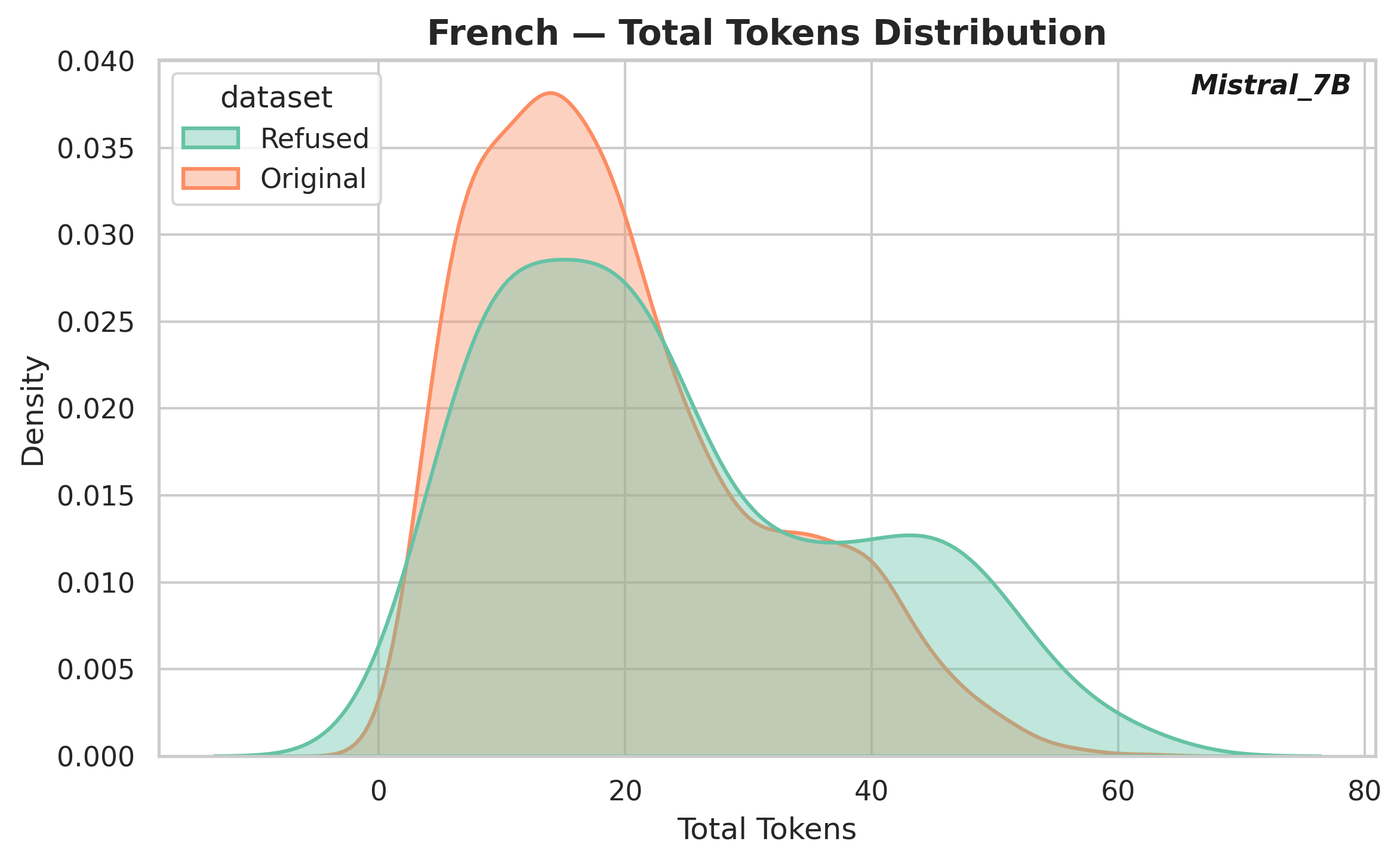}
    \caption{Mistral 7B}
  \end{subfigure}

  \vspace{0.4em}

  \begin{subfigure}{0.32\textwidth}
    \centering
    \includegraphics[width=\linewidth]{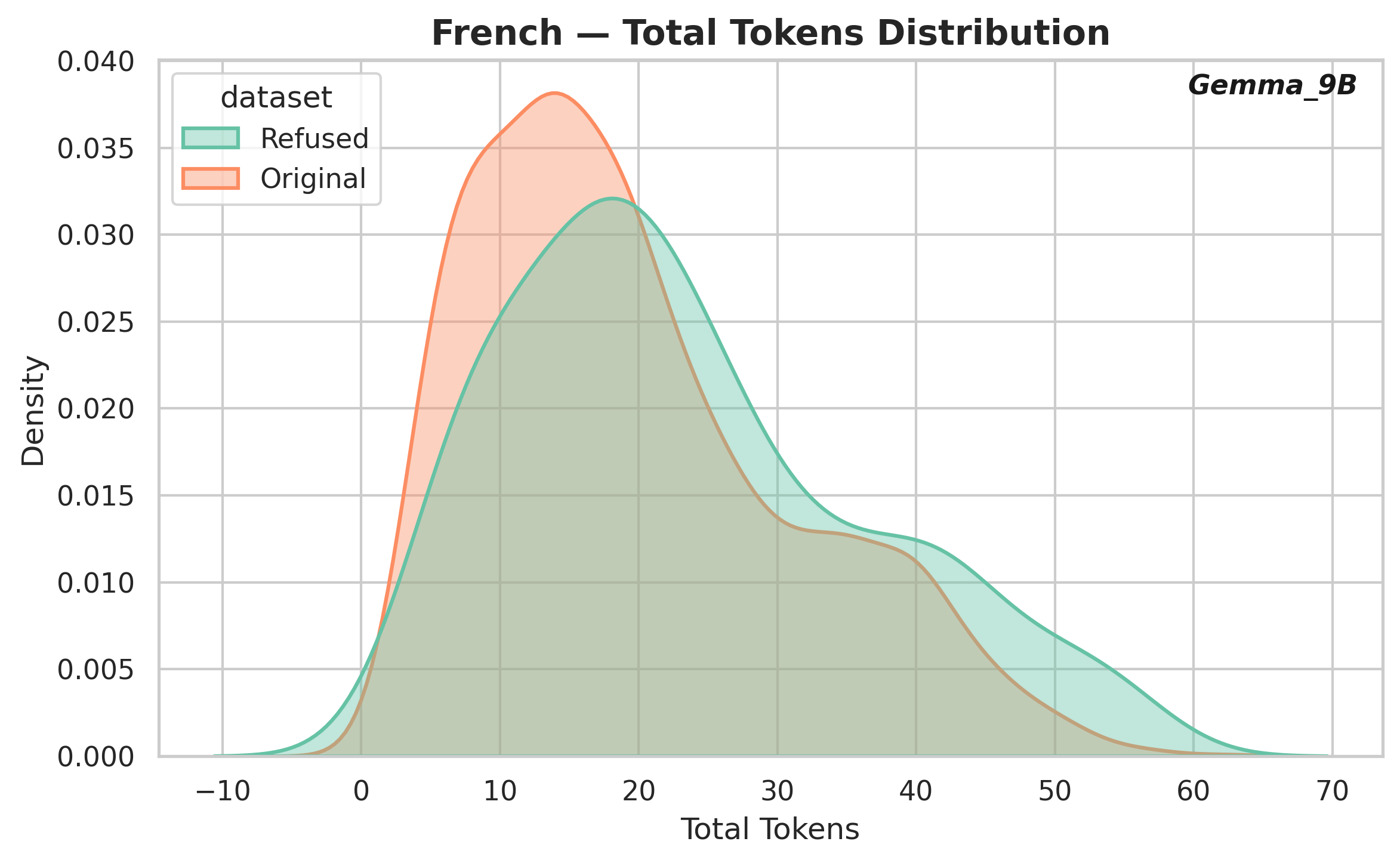}
    \caption{Gemma2 9B}
  \end{subfigure}\hfill
  \begin{subfigure}{0.32\textwidth}
    \centering
    \includegraphics[width=\linewidth]{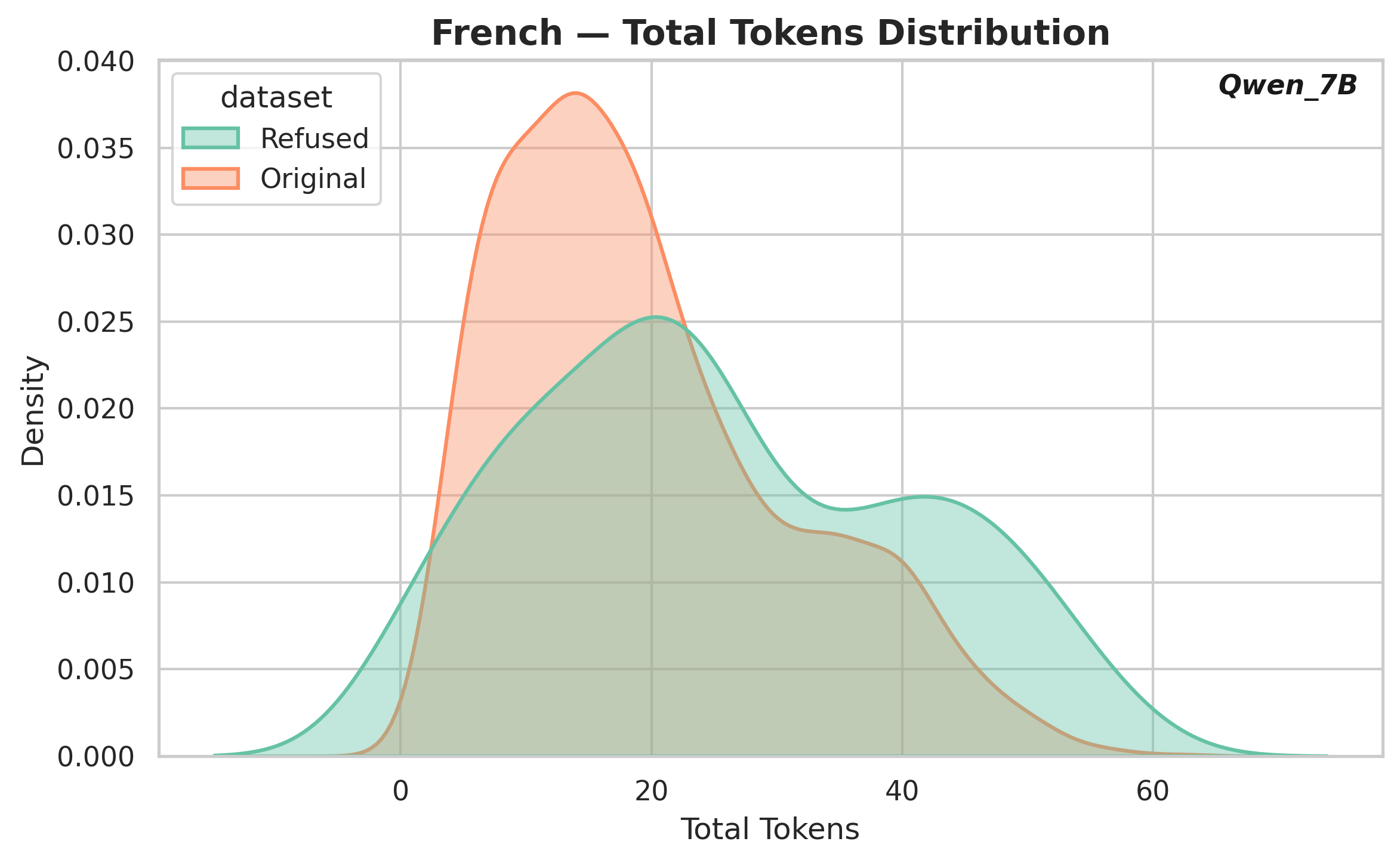}
    \caption{Qwen2.5 7B}
  \end{subfigure}

  \caption{Token count distributions for French dataset.}
  \label{fig:token_french}
\end{figure*}

\begin{figure*}[htbp]
  \centering
  \begin{subfigure}{0.32\textwidth}
    \centering
    \includegraphics[width=\linewidth]{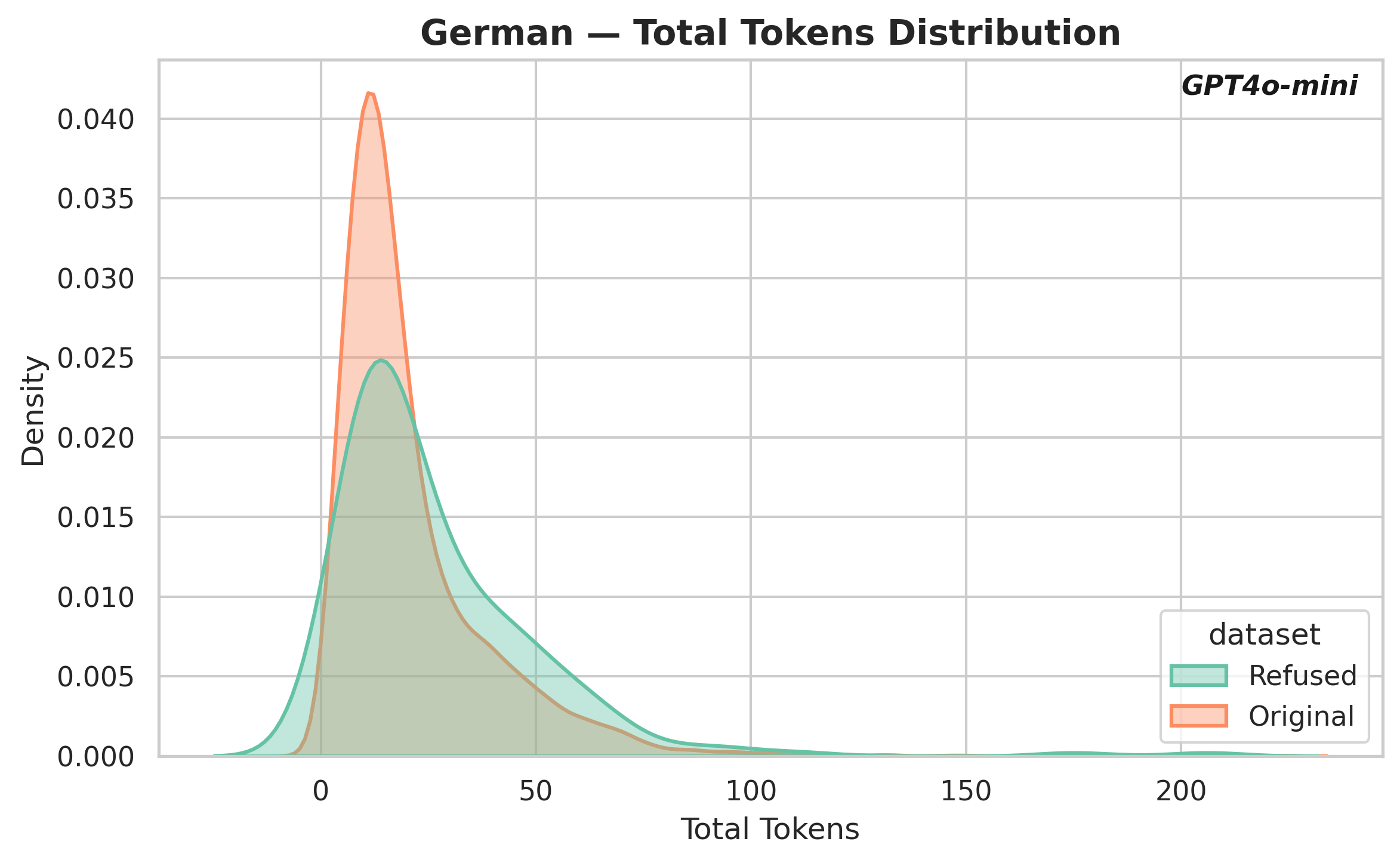}
    \caption{GPT4o-mini}
  \end{subfigure}\hfill
  \begin{subfigure}{0.32\textwidth}
    \centering
    \caption{Llama3 8B (file missing)}
  \end{subfigure}\hfill
  \begin{subfigure}{0.32\textwidth}
    \centering
    \includegraphics[width=\linewidth]{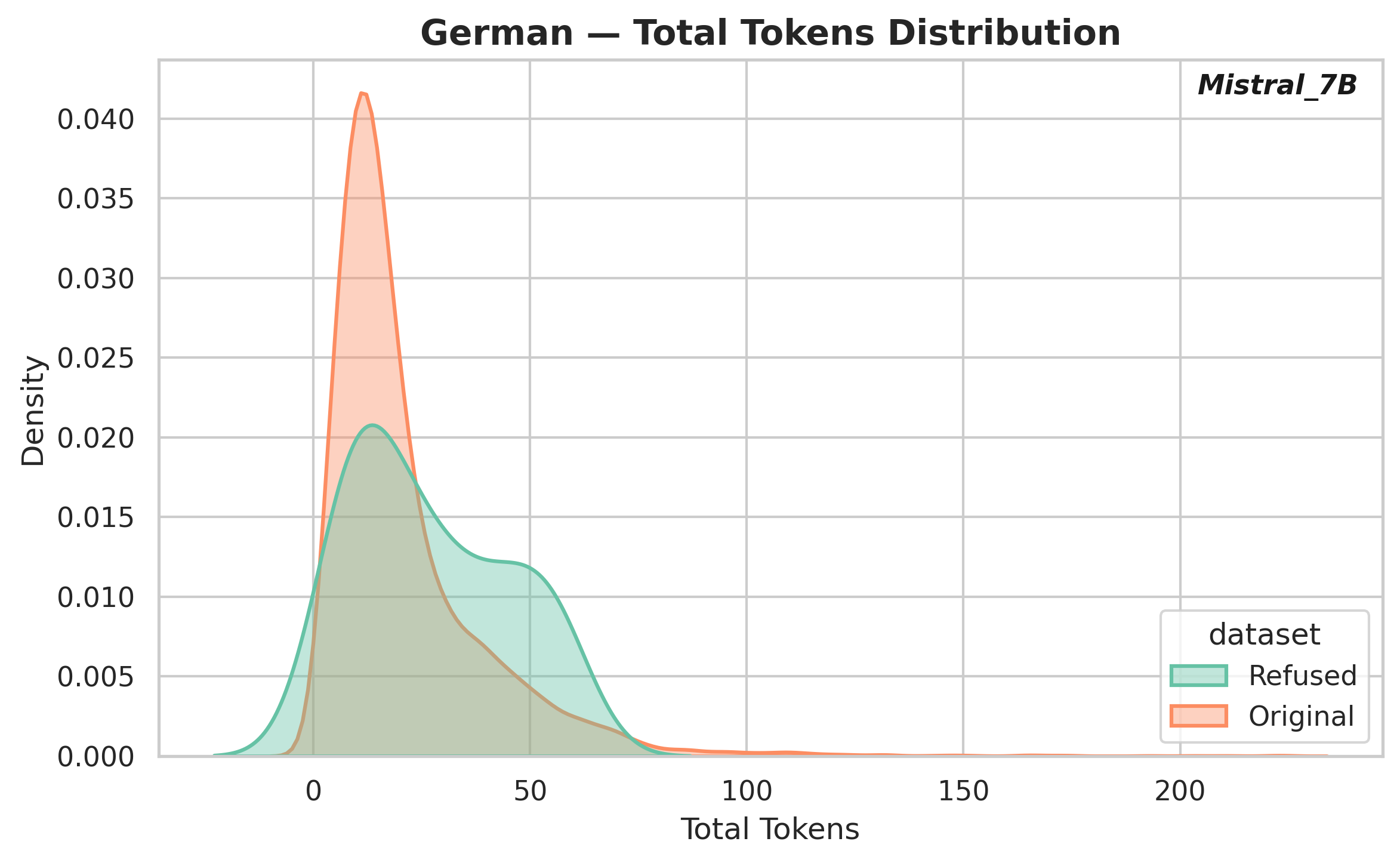}
    \caption{Mistral 7B}
  \end{subfigure}

  \vspace{0.4em}

  \begin{subfigure}{0.32\textwidth}
    \centering
    \includegraphics[width=\linewidth]{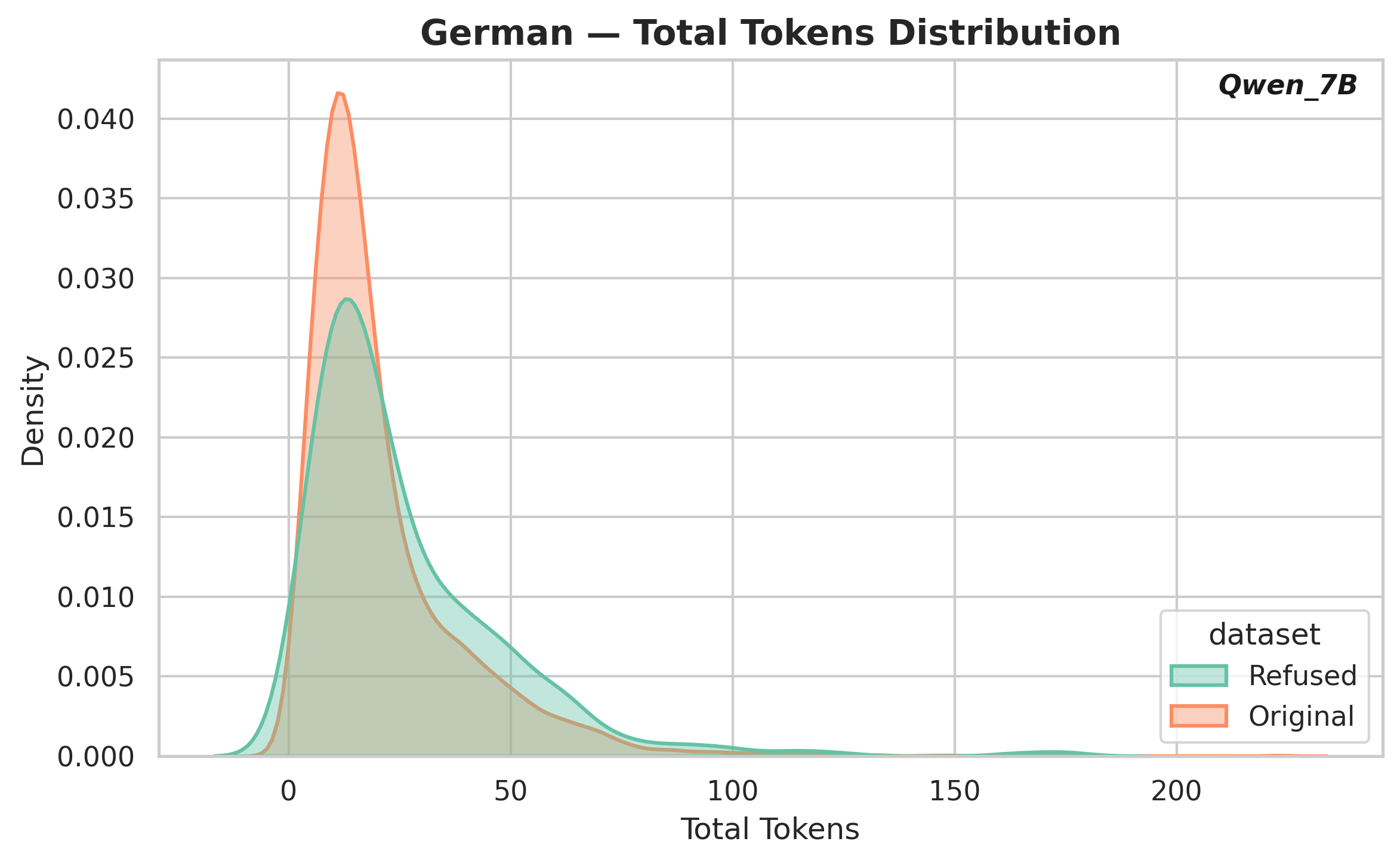}
    \caption{Qwen2.5 7B}
  \end{subfigure}\hfill
  \begin{subfigure}{0.32\textwidth}
    \centering
    \includegraphics[width=\linewidth]{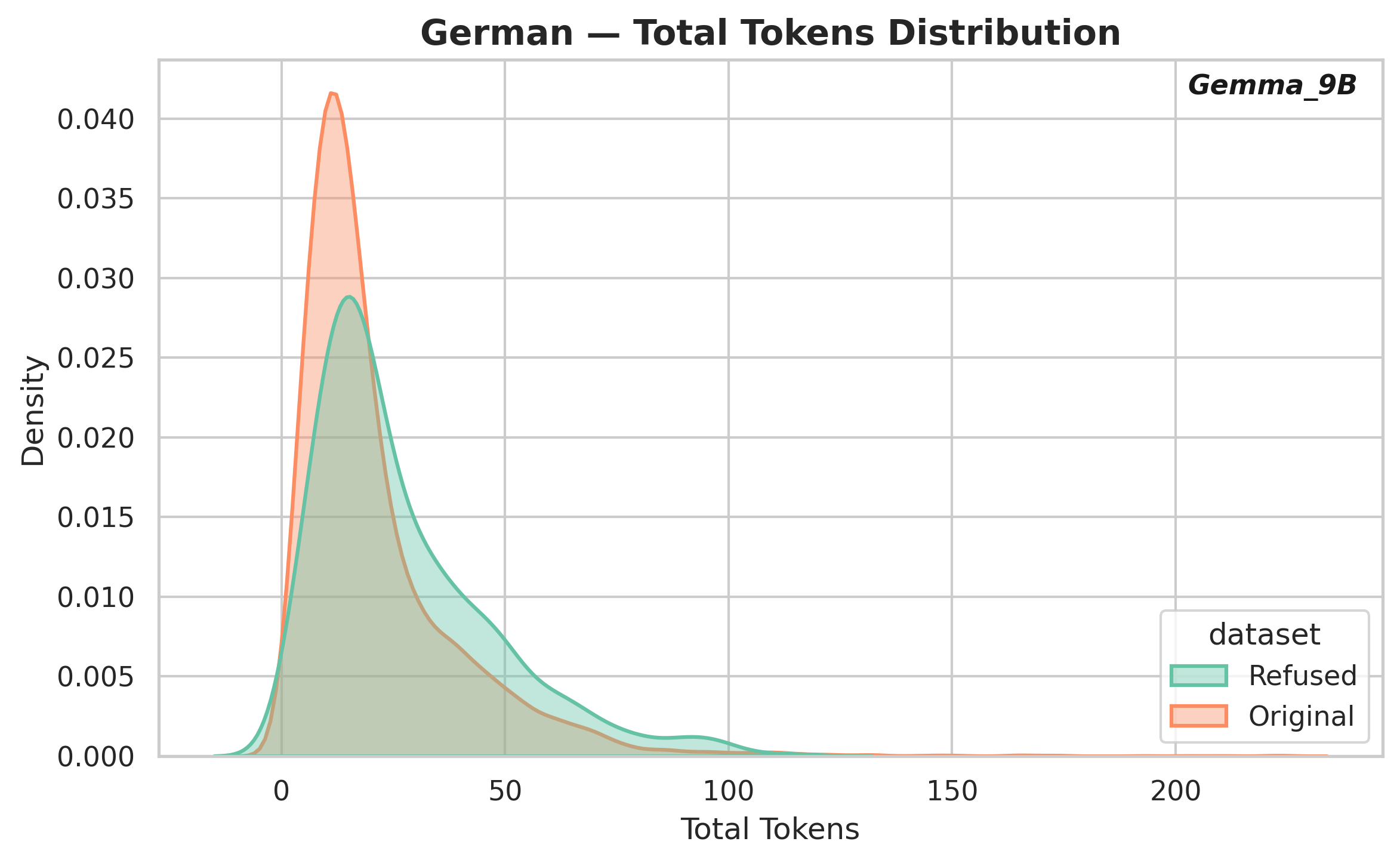}
    \caption{Gemma2 9B}
  \end{subfigure}

  \caption{Token count distributions for German dataset.}
  \label{fig:token_german}
\end{figure*}

\begin{figure*}[htbp]
  \centering
  \begin{subfigure}{0.32\textwidth}
    \centering
    \includegraphics[width=\linewidth]{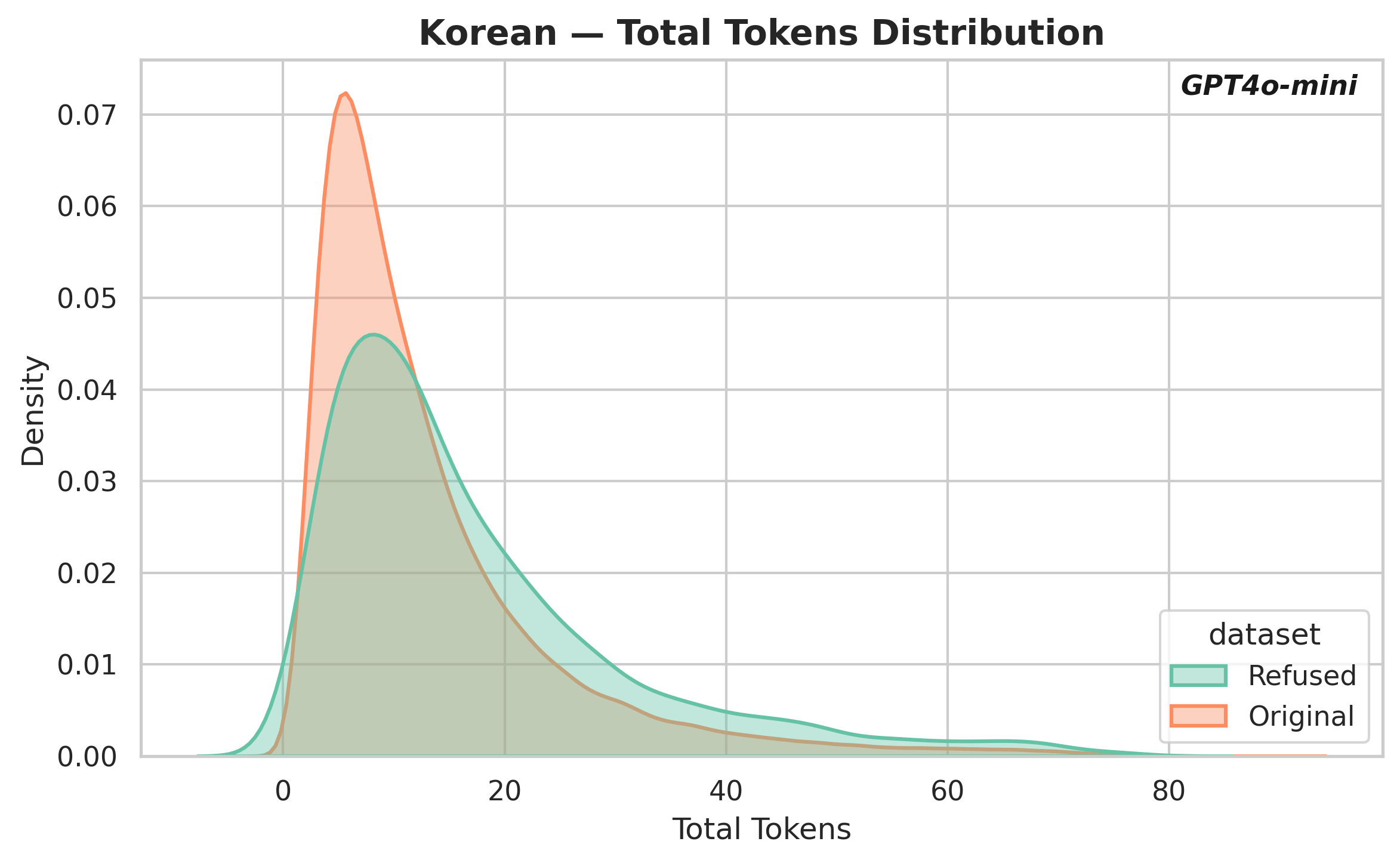}
    \caption{GPT4o-mini}
  \end{subfigure}\hfill
  \begin{subfigure}{0.32\textwidth}
    \centering
    \includegraphics[width=\linewidth]{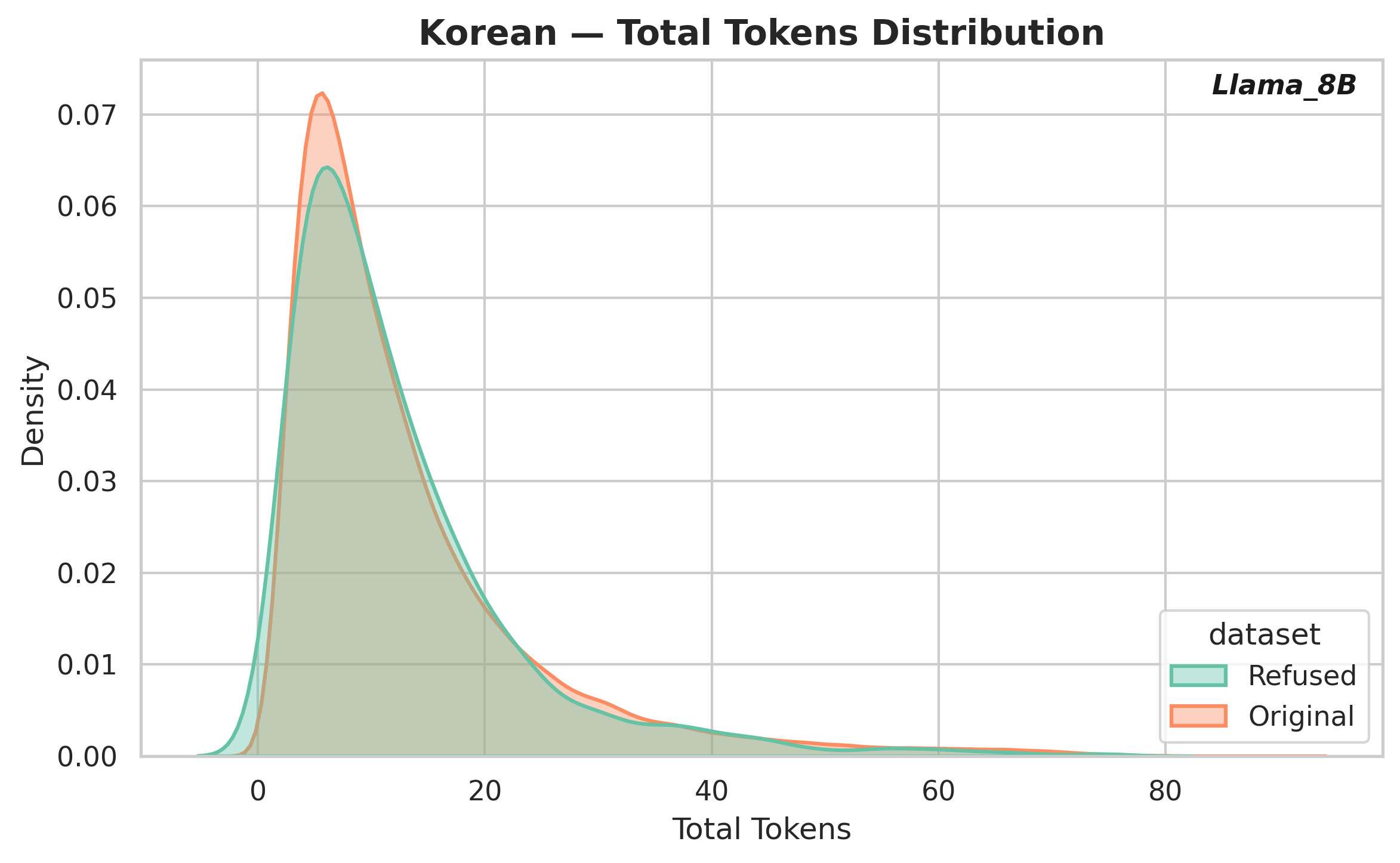}
    \caption{Llama3 8B}
  \end{subfigure}\hfill
  \begin{subfigure}{0.32\textwidth}
    \centering
    \includegraphics[width=\linewidth]{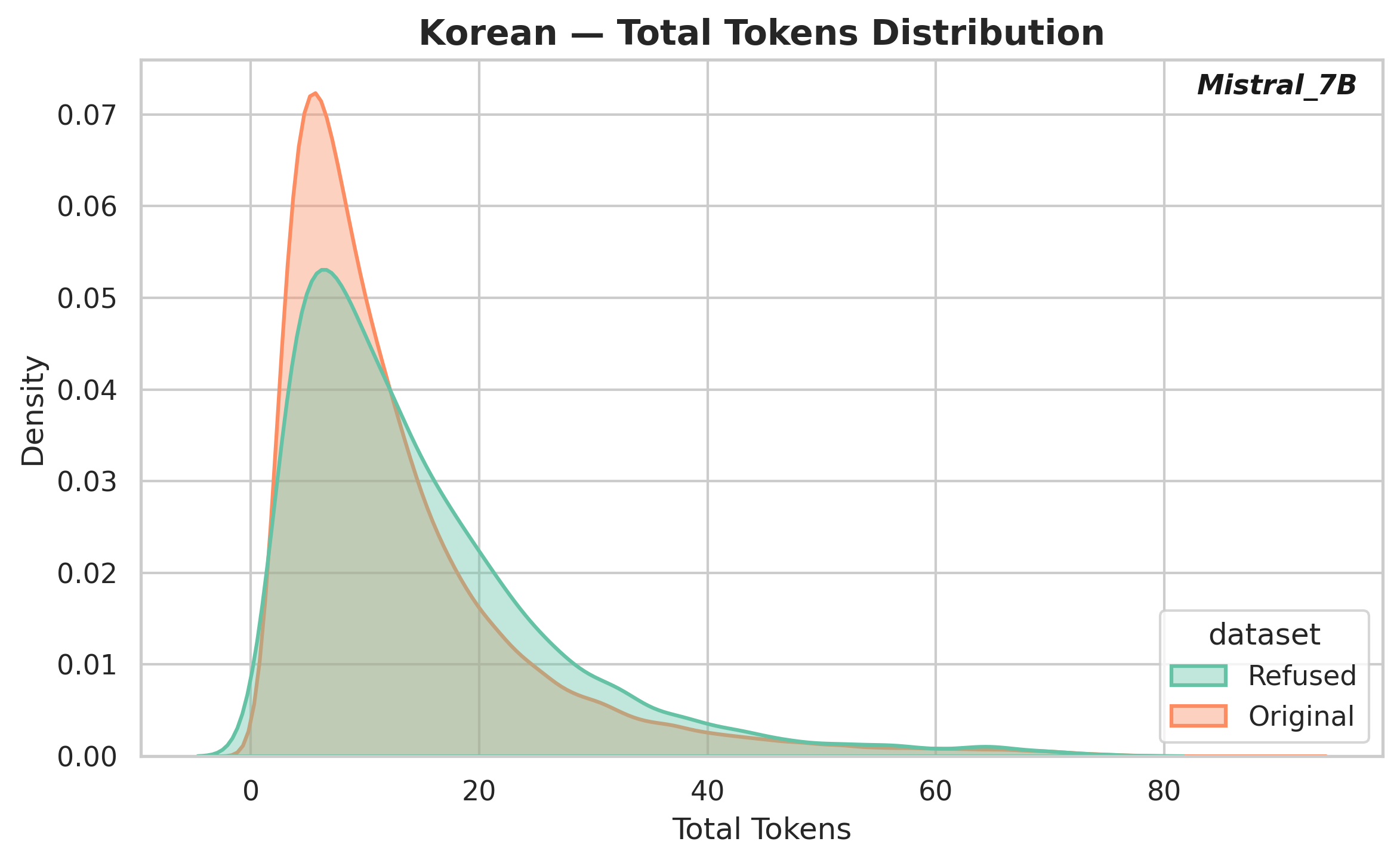}
    \caption{Mistral 7B}
  \end{subfigure}

  \vspace{0.4em}

  \begin{subfigure}{0.32\textwidth}
    \centering
    \includegraphics[width=\linewidth]{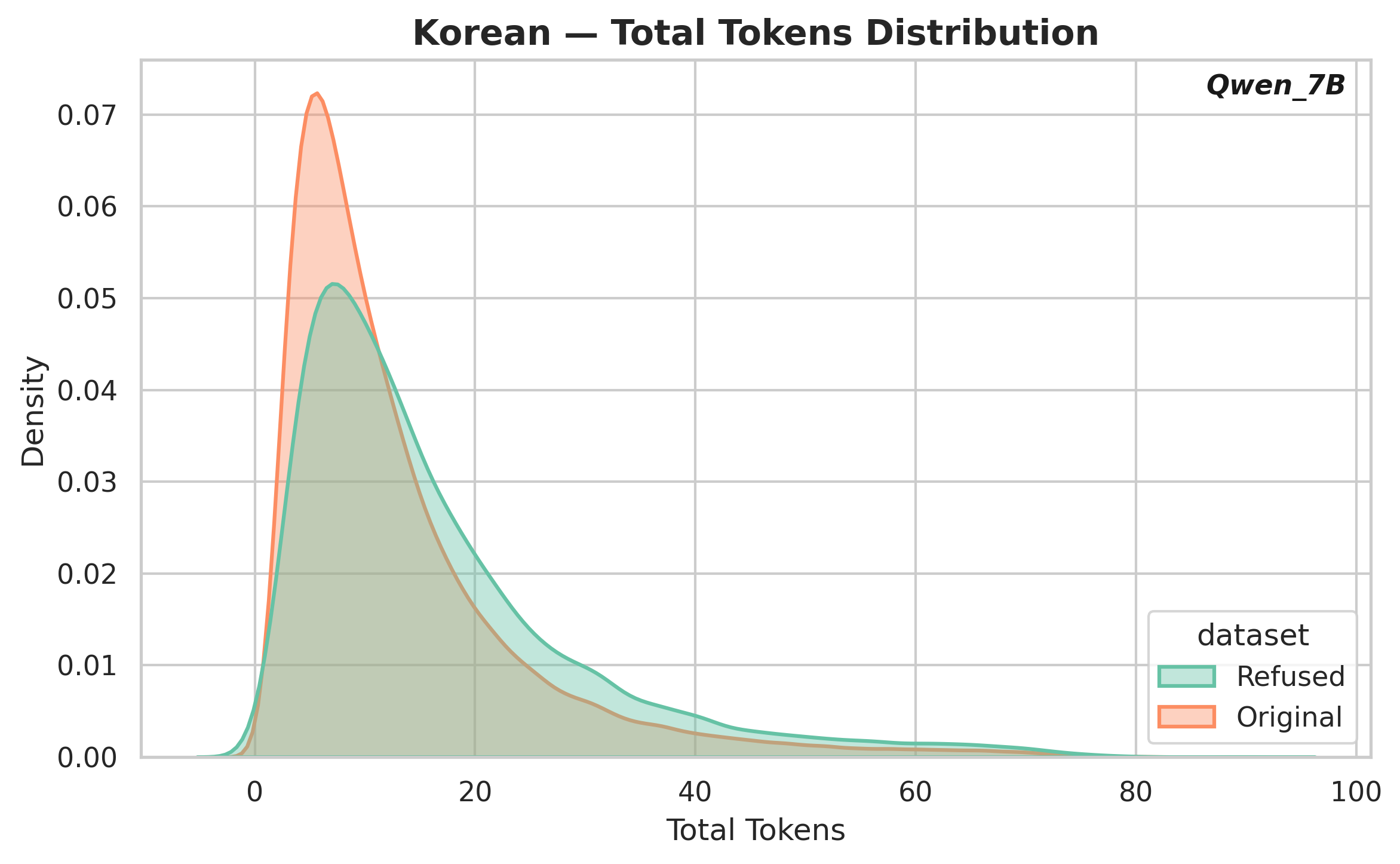}
    \caption{Qwen2.5 7B}
  \end{subfigure}\hfill
  \begin{subfigure}{0.32\textwidth}
    \centering
    \includegraphics[width=\linewidth]{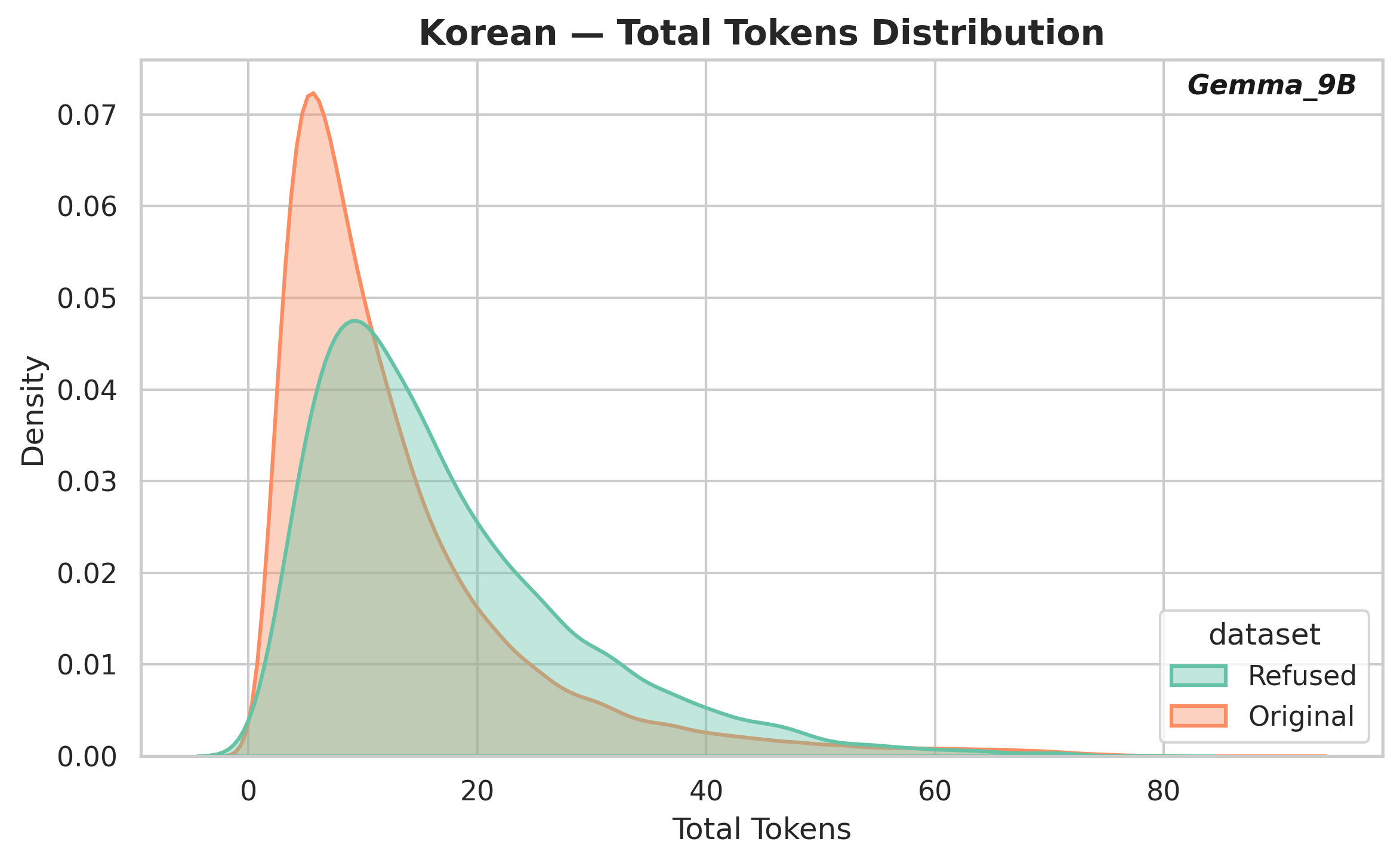}
    \caption{Gemma2 9B}
  \end{subfigure}

  \caption{Token count distribution of Korean dataset.}
  \label{fig:token_korean}
\end{figure*}

Figure~\ref{fig:mult_token} presents the token-length distributions of false refusals and original samples for Spanish and Chinese across all evaluated models. The corresponding distributions for French, German, and Korean are shown in Figures~\ref{fig:token_french}, \ref{fig:token_german}, and \ref{fig:token_korean}, respectively.

Across all languages, the token-length distributions of falsely refused samples largely overlap with those of the original datasets, indicating that sentence length and surface-level verbosity do not systematically trigger refusal behavior. In Spanish, despite substantially higher false-refusal ratios, only minor shifts toward shorter inputs are observed for certain models, such as Gemma2~9B and Mistral~7B. This suggests that over-refusals are more likely driven by semantic or cultural factors, including informal registers or idiomatic expressions, rather than lexical complexity.

By contrast, Chinese exhibits similarly overlapping length distributions alongside consistently low refusal rates, implying under-sensitivity to toxic content despite comparable input lengths. Taken together, these results indicate that multilingual refusal behavior is shaped primarily by semantic and alignment-related factors, rather than by token-level properties.

\subsection{Sentence Complexity}

\begin{figure*}[htbp]
  \centering
  \captionsetup{font=small}
  \setlength{\tabcolsep}{3pt}

  \begin{tabular}{cc}
    \subcaptionbox{\textbf{Spanish}\label{fig:spanish_clause}}{
      \begin{tabular}{cc}
        \includegraphics[width=0.235\textwidth]{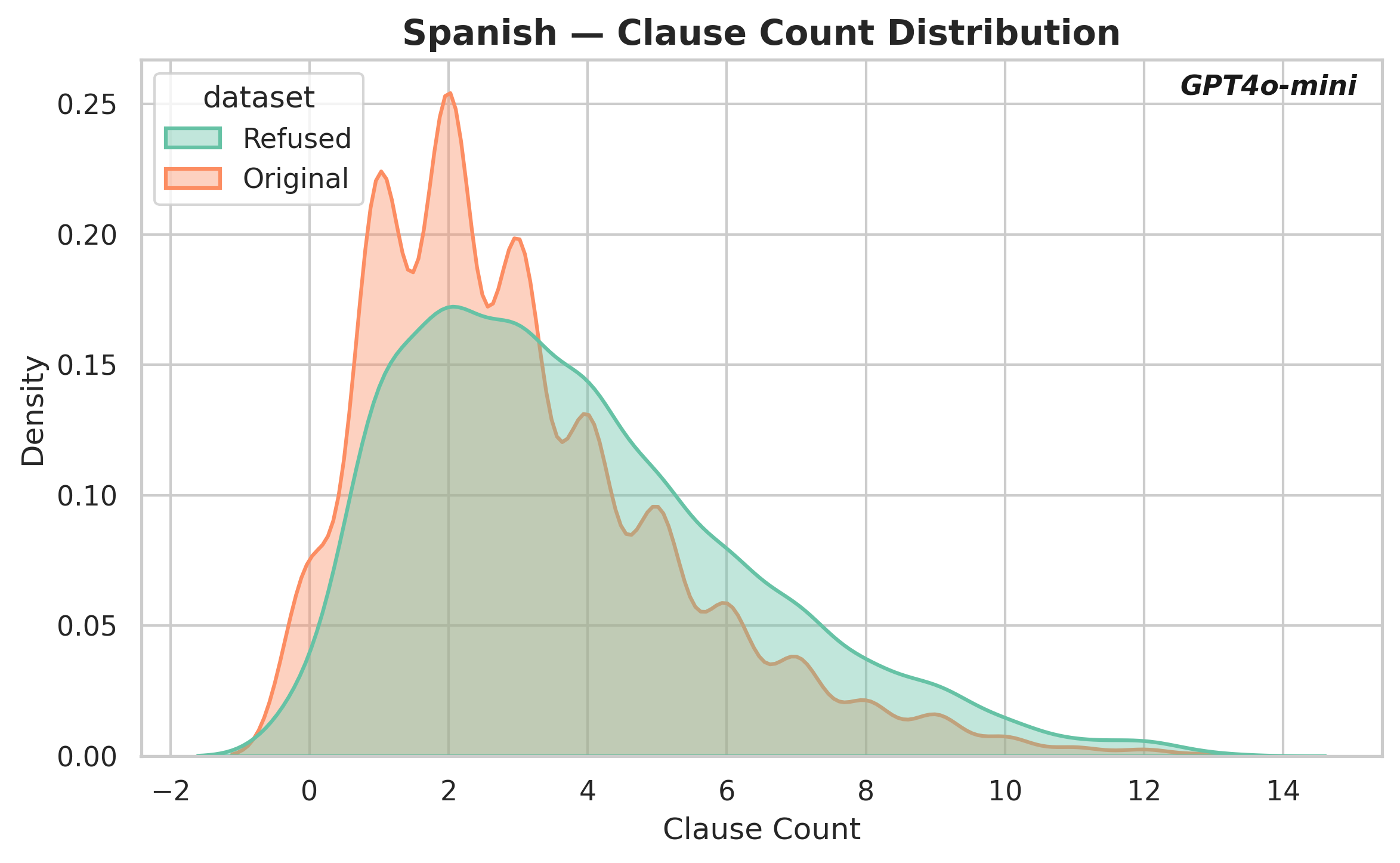} &
        \includegraphics[width=0.235\textwidth]{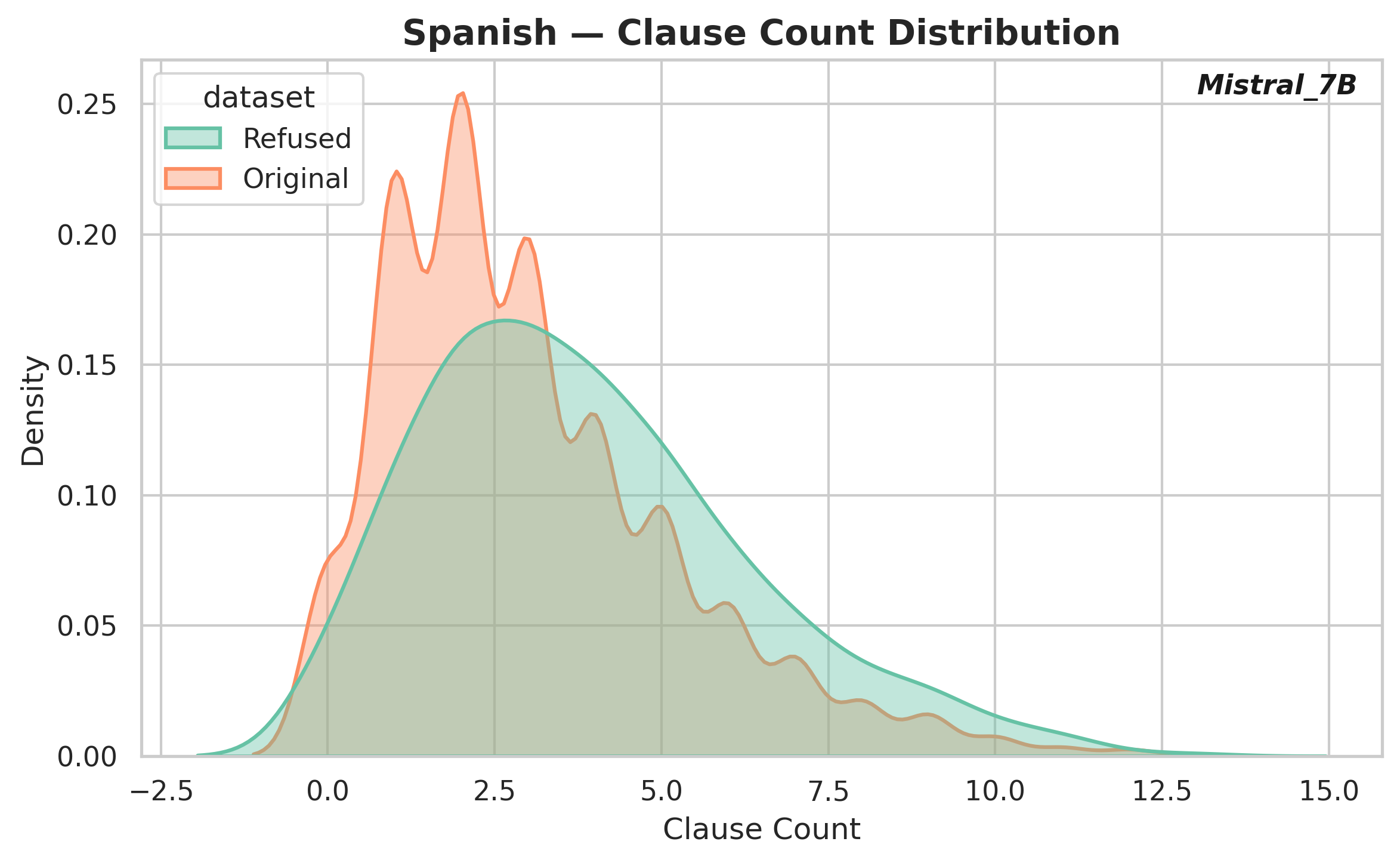} \\
        \includegraphics[width=0.235\textwidth]{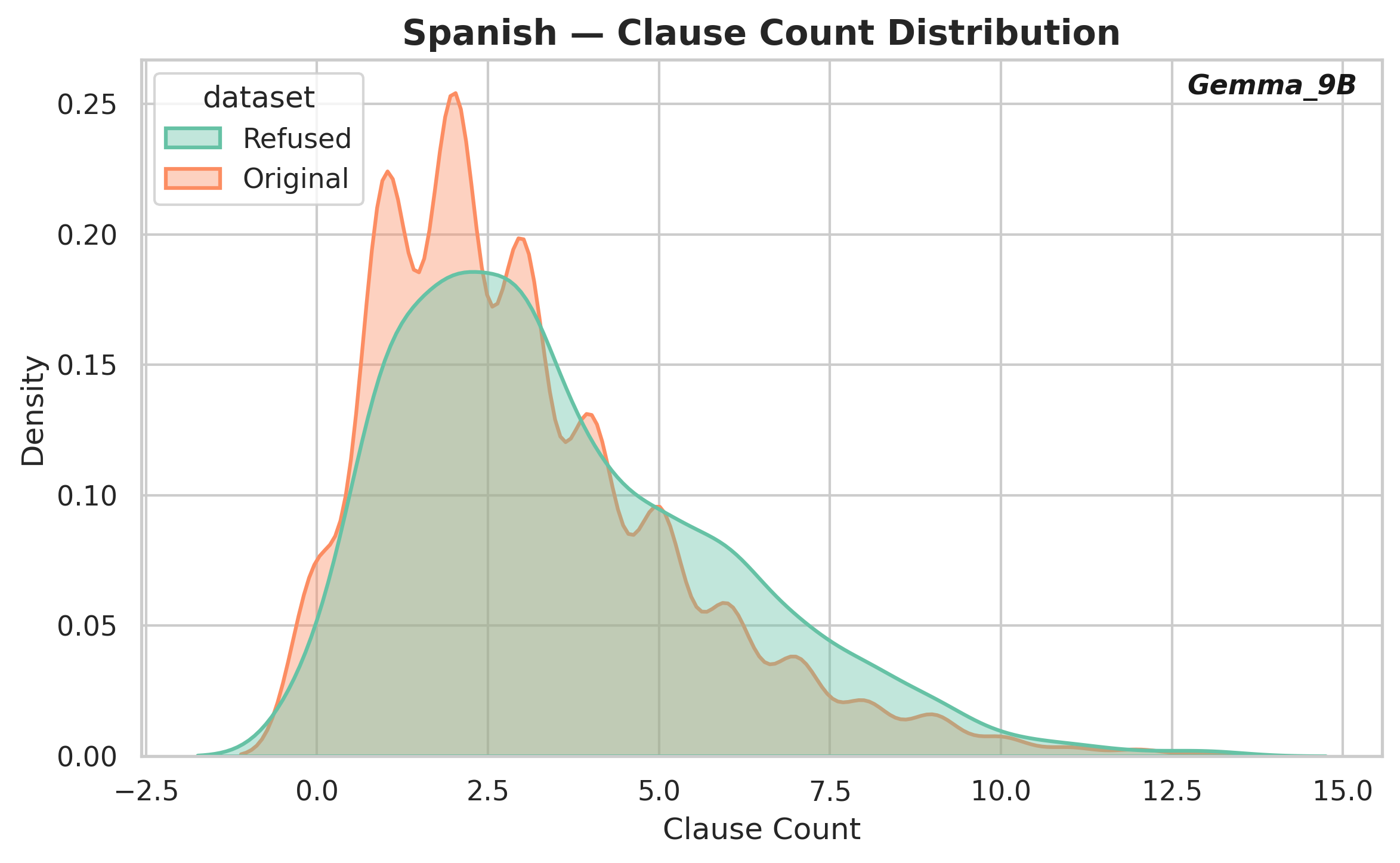} &
        \includegraphics[width=0.235\textwidth]{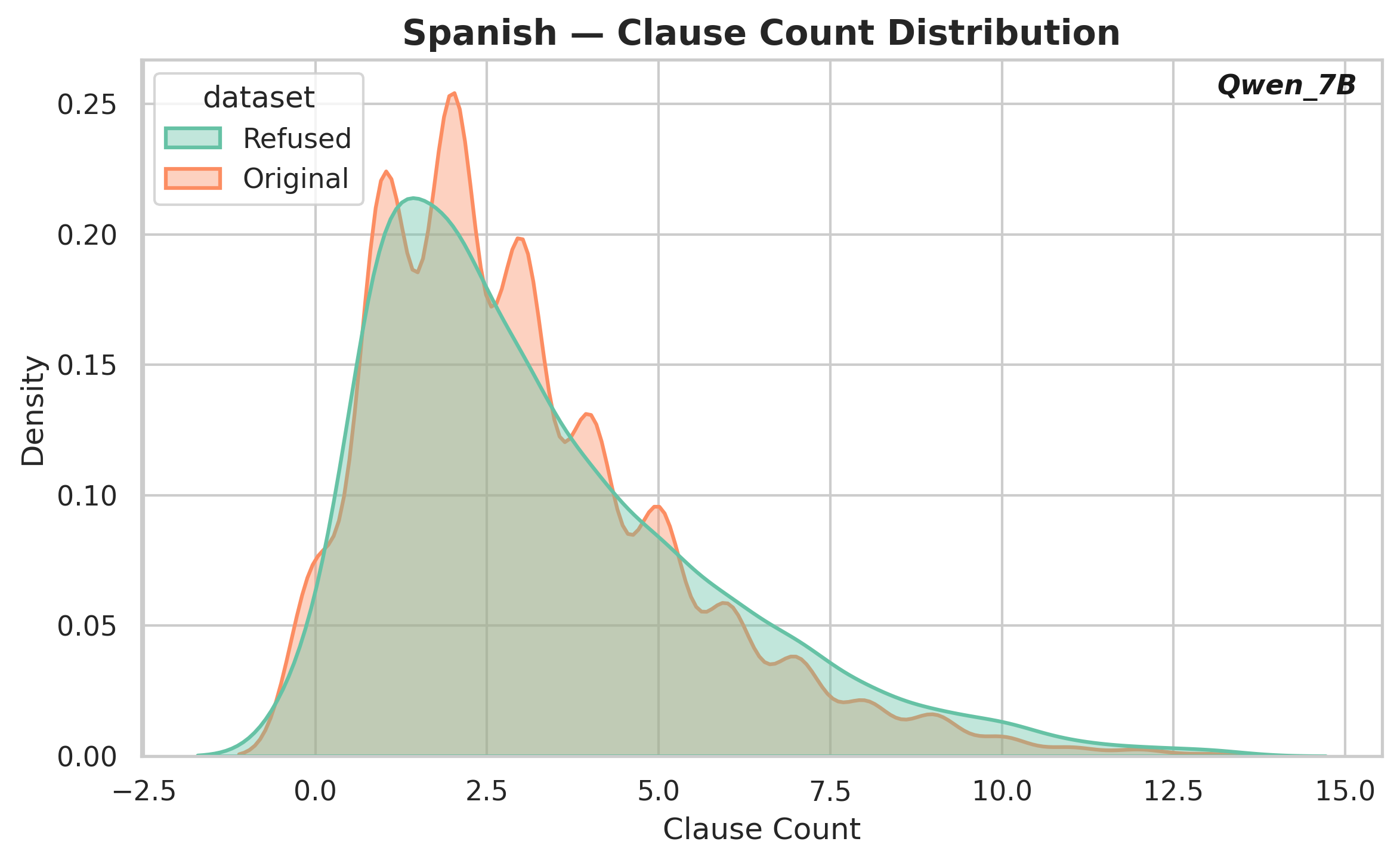} \\
      \end{tabular}
    }
    &
    \subcaptionbox{\textbf{Chinese}\label{fig:chinese_clause}}{
      \begin{tabular}{cc}
        \includegraphics[width=0.235\textwidth]{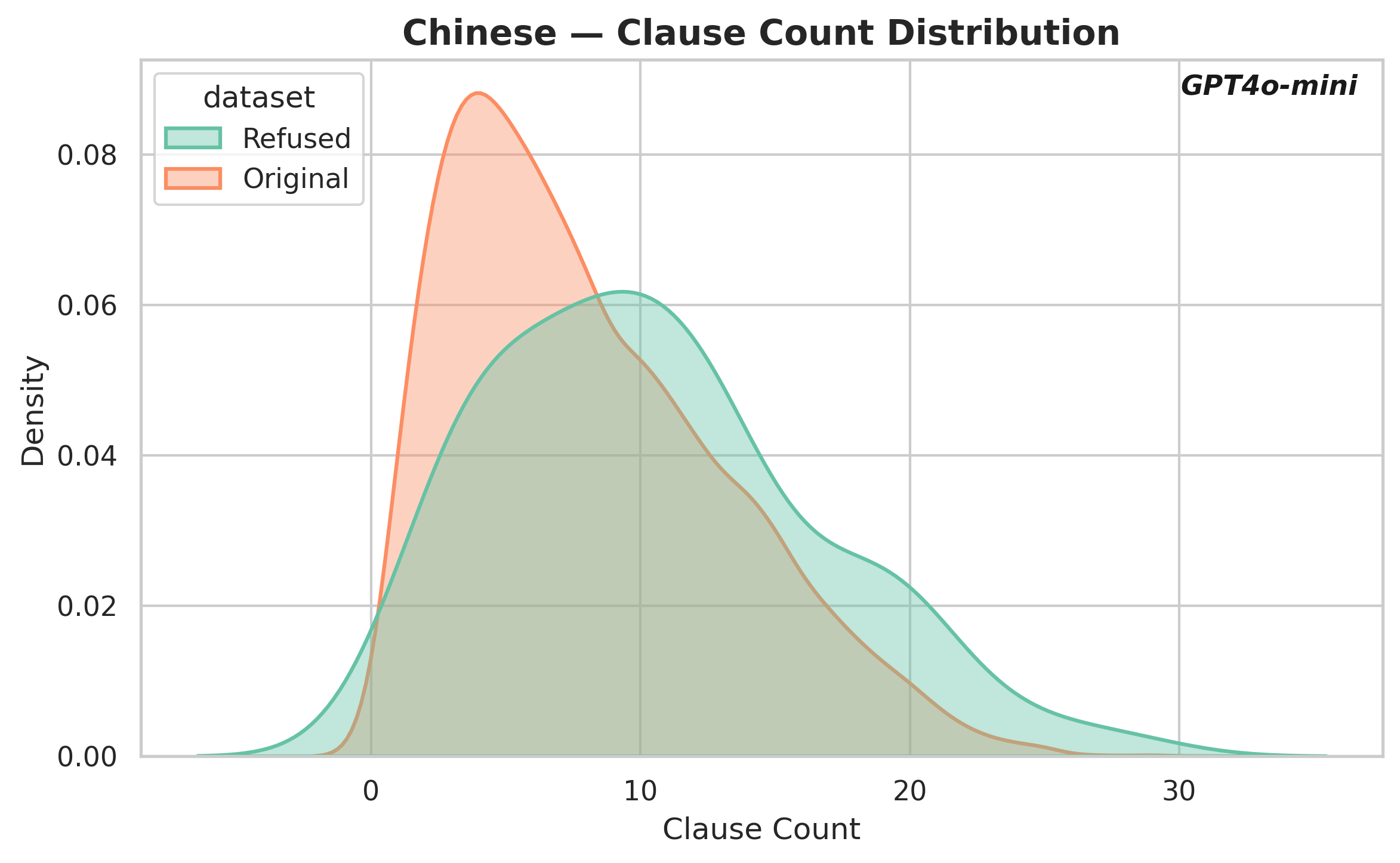} &
        \includegraphics[width=0.235\textwidth]{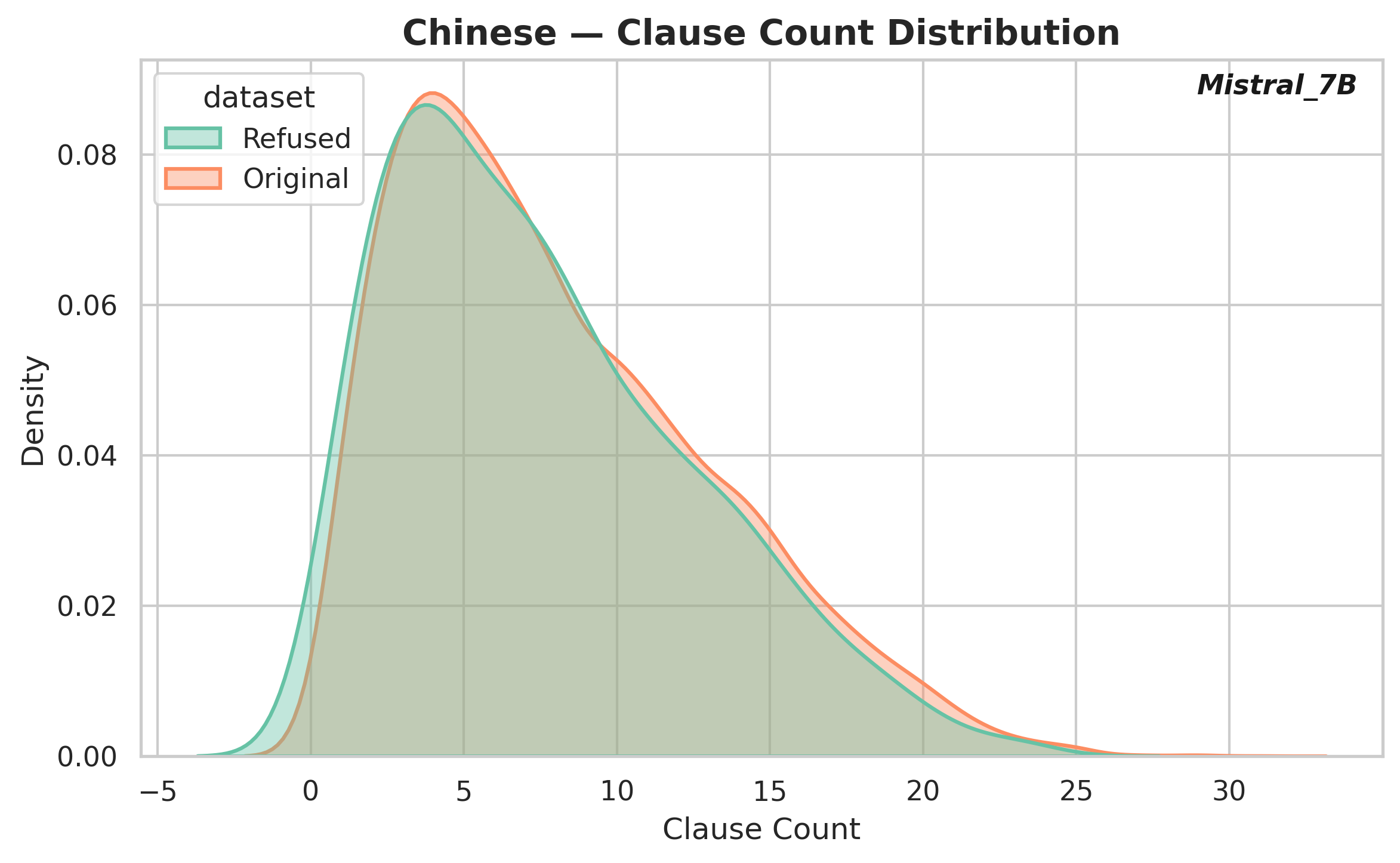} \\
        \includegraphics[width=0.235\textwidth]{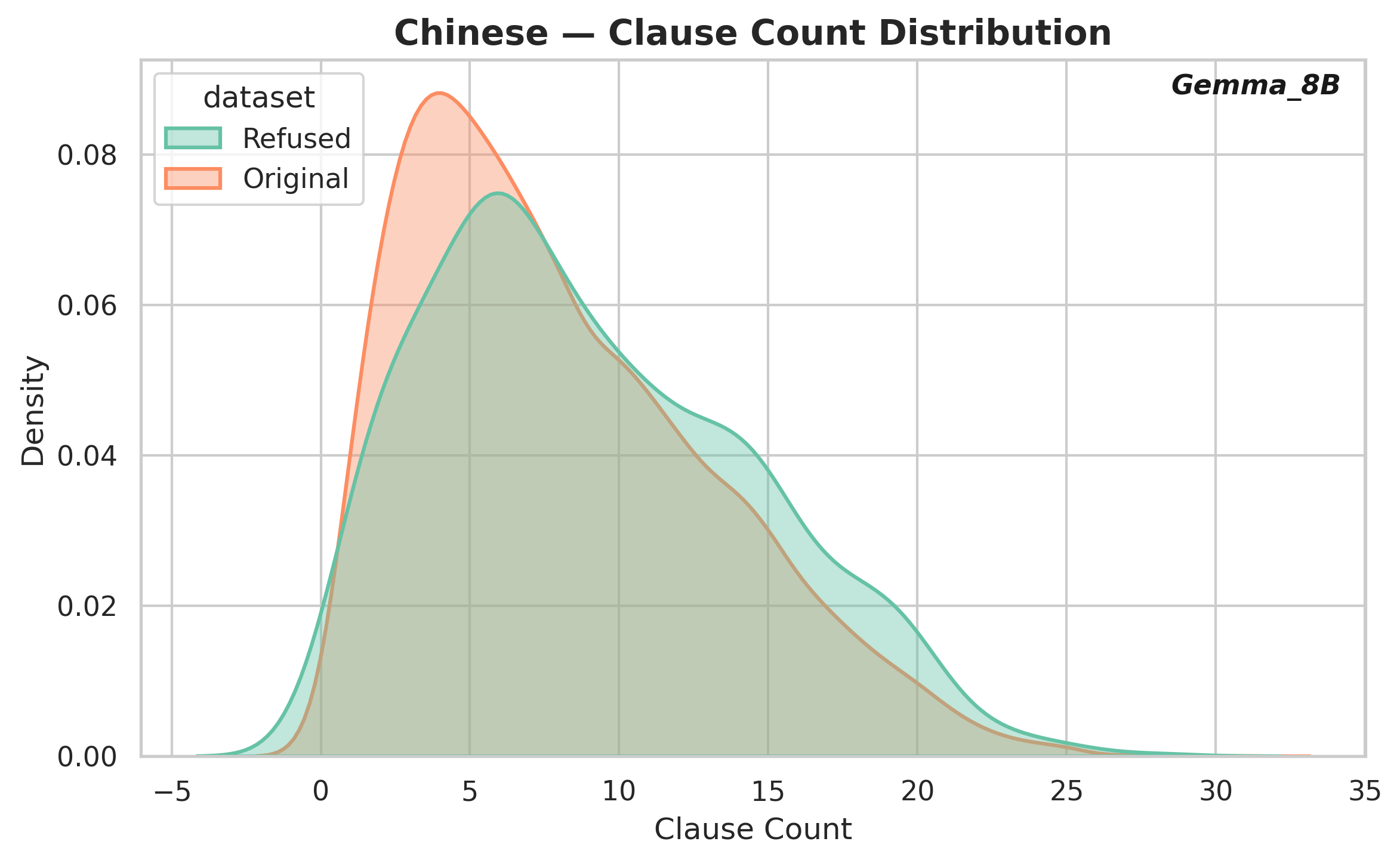} &
        \includegraphics[width=0.235\textwidth]{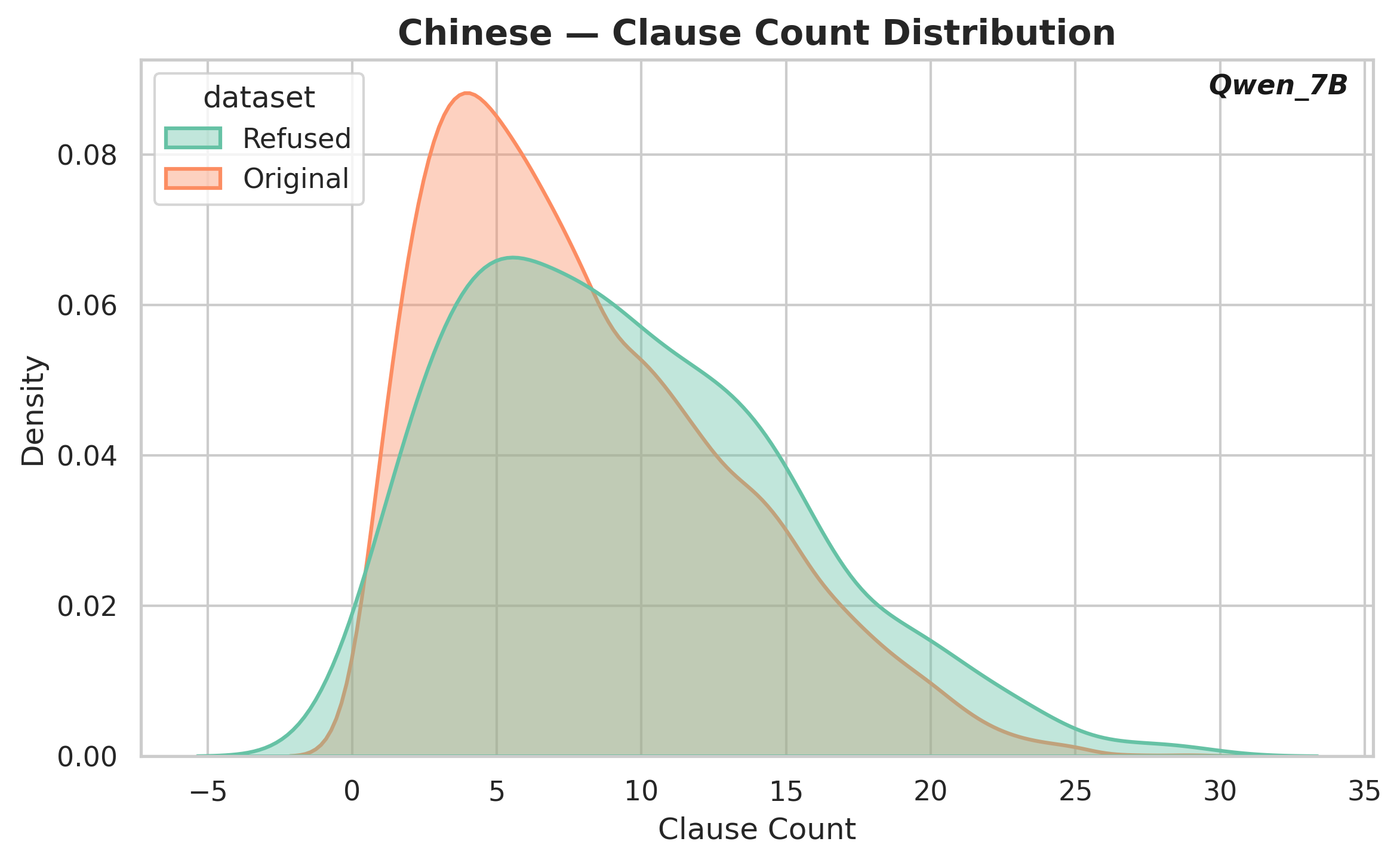} \\
      \end{tabular}
    }
  \end{tabular}
  \caption{Clause count distribution for false refusals versus original samples across representative models in Spanish and Chinese datasets.}
  \label{fig:mult_clause}
\end{figure*}

\begin{figure*}[htbp]
  \centering
  \setlength{\tabcolsep}{3pt}
  \begin{subfigure}{0.32\textwidth}
    \centering
    \includegraphics[width=\linewidth]{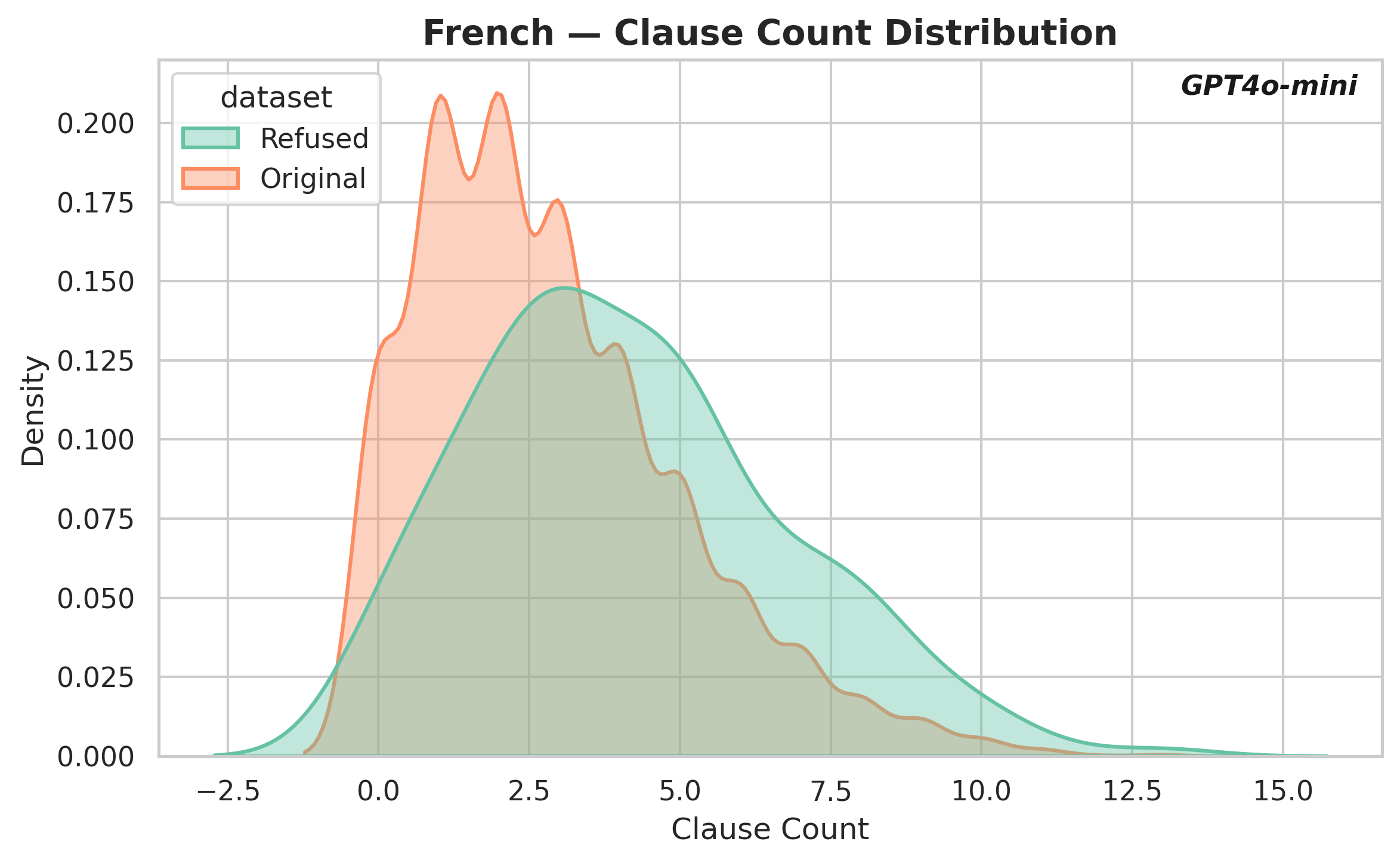}
    \caption{GPT4o-mini}
  \end{subfigure}\hfill
  \begin{subfigure}{0.32\textwidth}
    \centering
    \includegraphics[width=\linewidth]{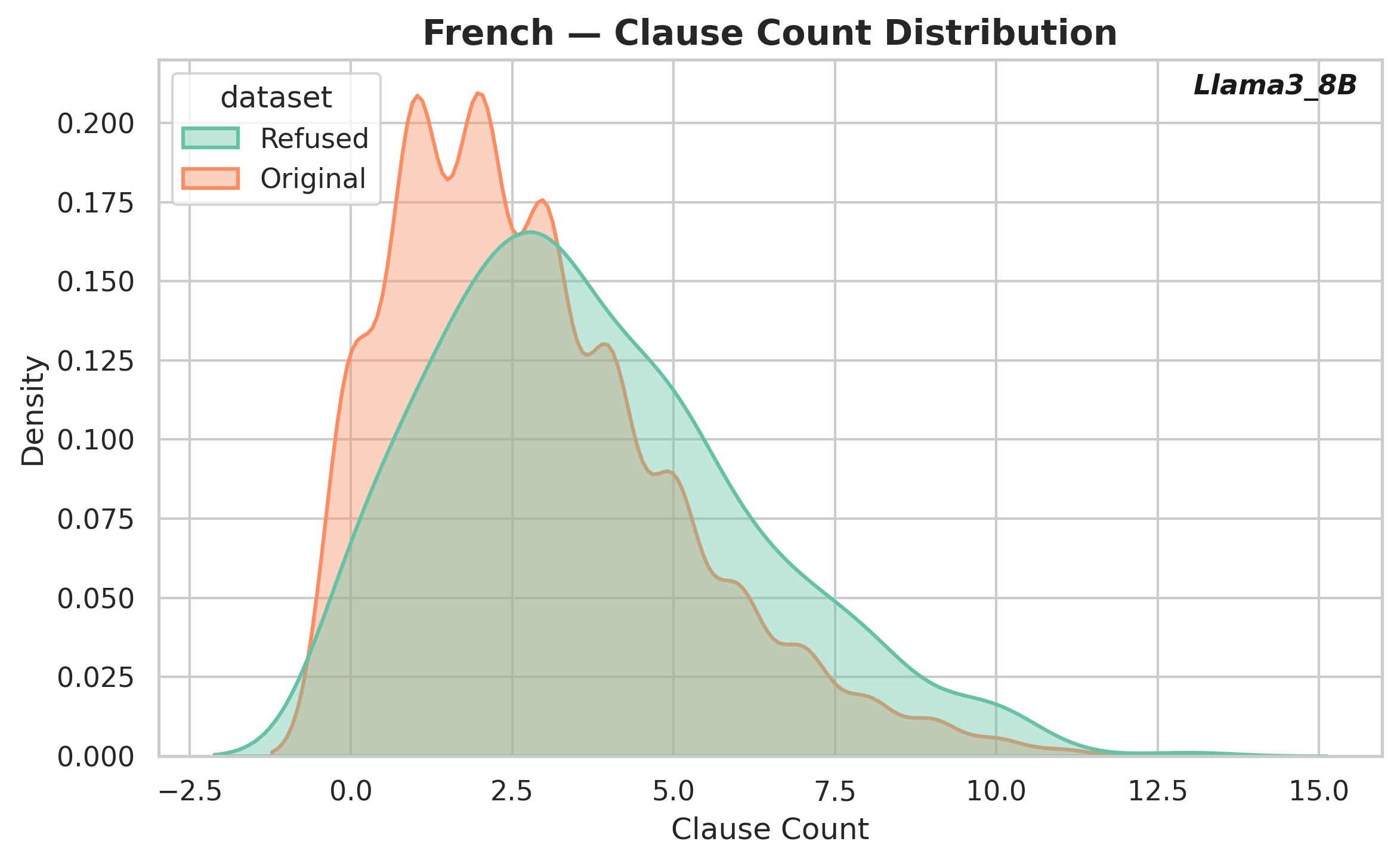}
    \caption{Llama3 8B}
  \end{subfigure}\hfill
  \begin{subfigure}{0.32\textwidth}
    \centering
    \includegraphics[width=\linewidth]{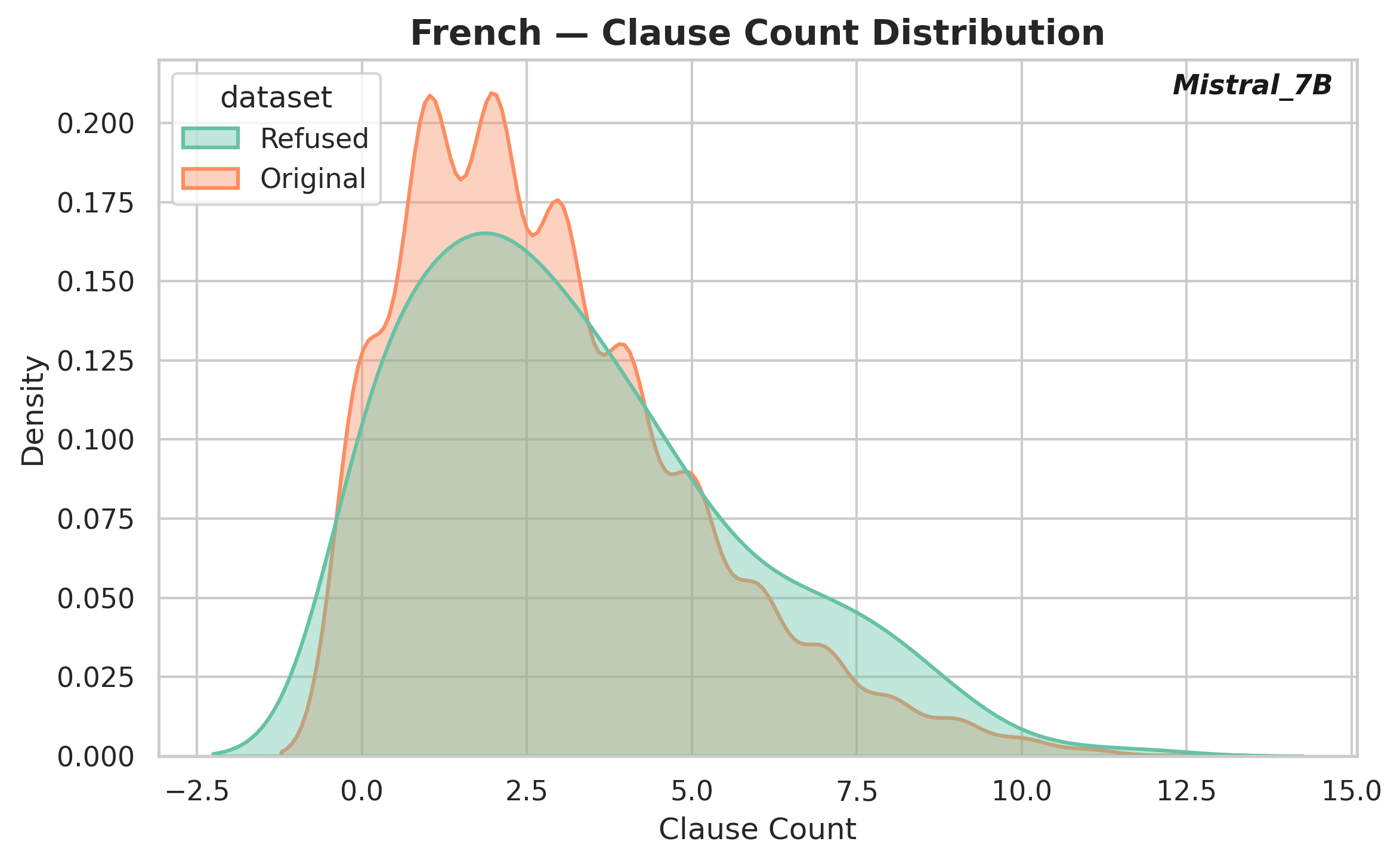}
    \caption{Mistral 7B}
  \end{subfigure}

  \vspace{0.4em}

  \begin{subfigure}{0.32\textwidth}
    \centering
    \includegraphics[width=\linewidth]{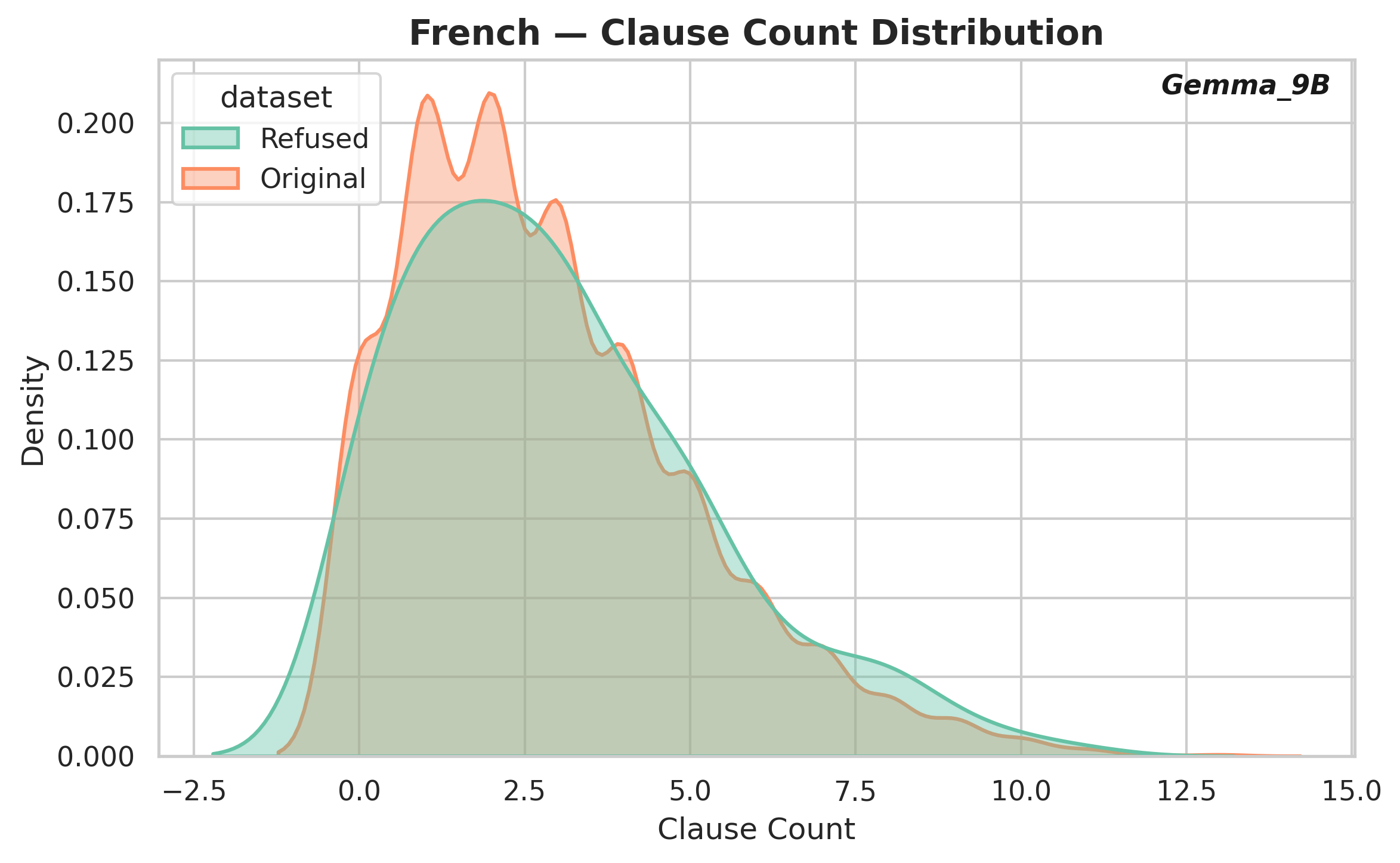}
    \caption{Gemma2 9B}
  \end{subfigure}\hfill
  \begin{subfigure}{0.32\textwidth}
    \centering
    \includegraphics[width=\linewidth]{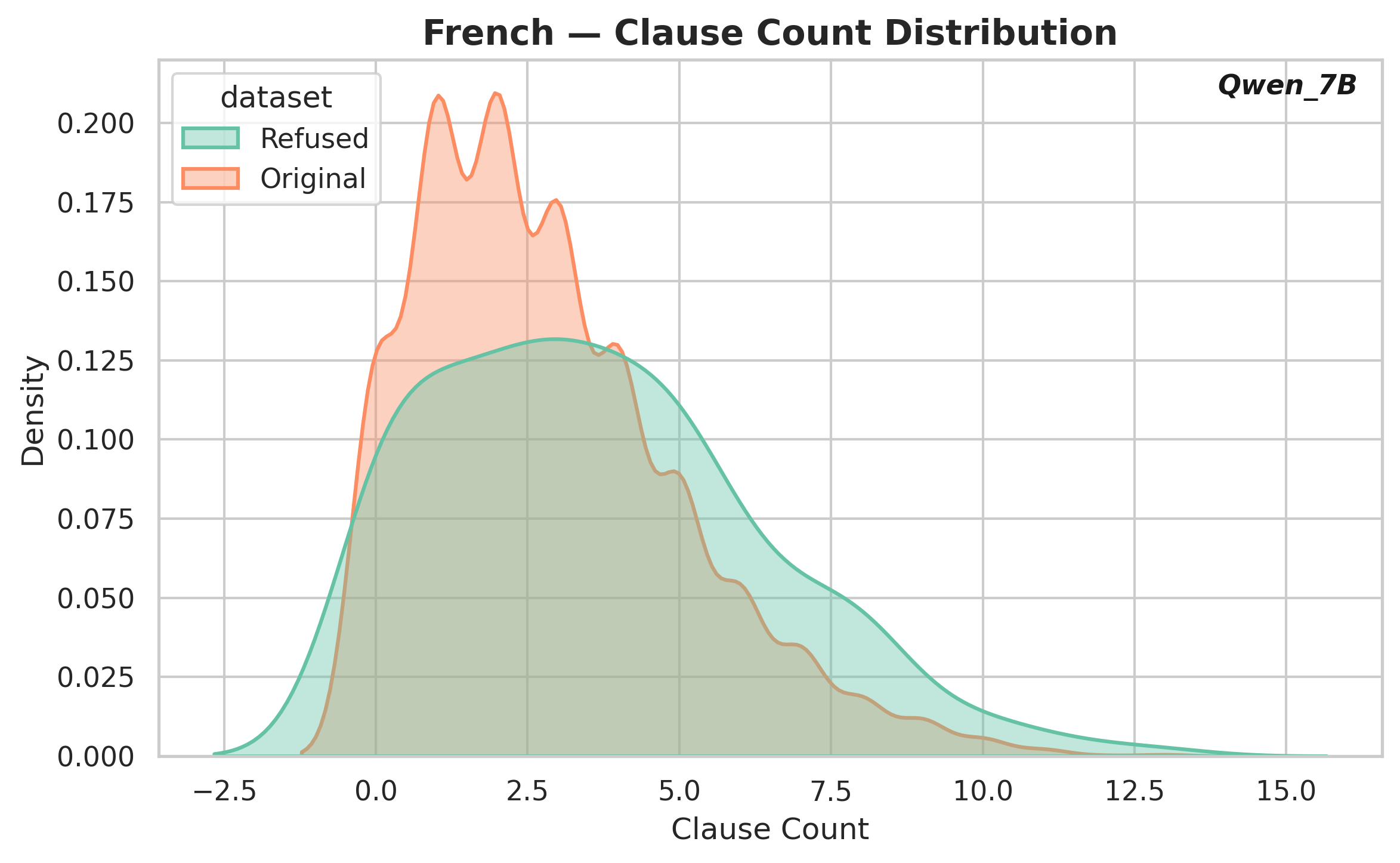}
    \caption{Qwen2.5 7B}
  \end{subfigure}

  \caption{Clause count distributions for French dataset.}
  \label{fig:clause_french}
\end{figure*}

\begin{figure*}[htbp]
  \centering
  \begin{subfigure}{0.32\textwidth}
    \centering
    \includegraphics[width=\linewidth]{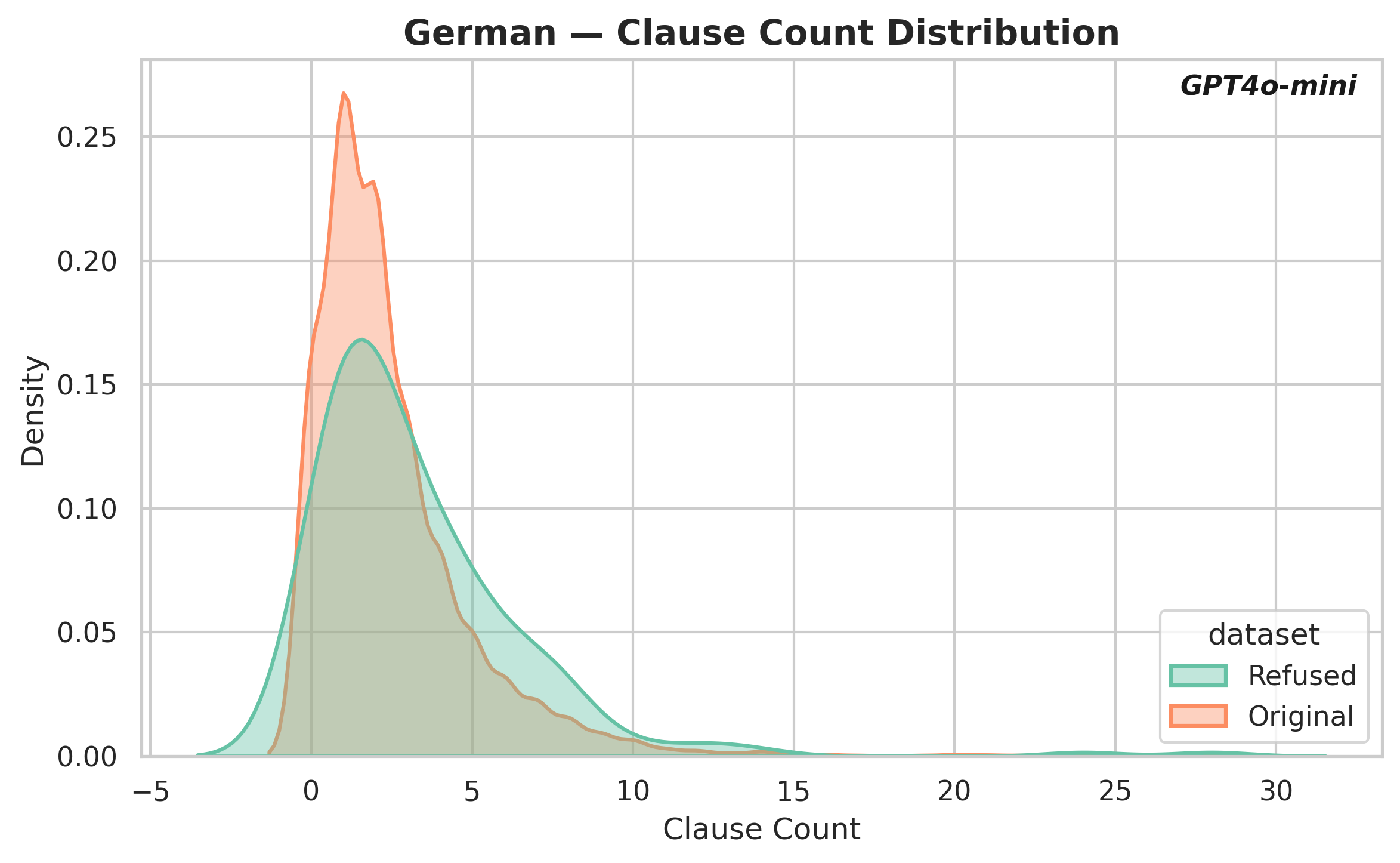}
    \caption{GPT4o-mini}
  \end{subfigure}\hfill
  \begin{subfigure}{0.32\textwidth}
    \centering
    \includegraphics[width=\linewidth]{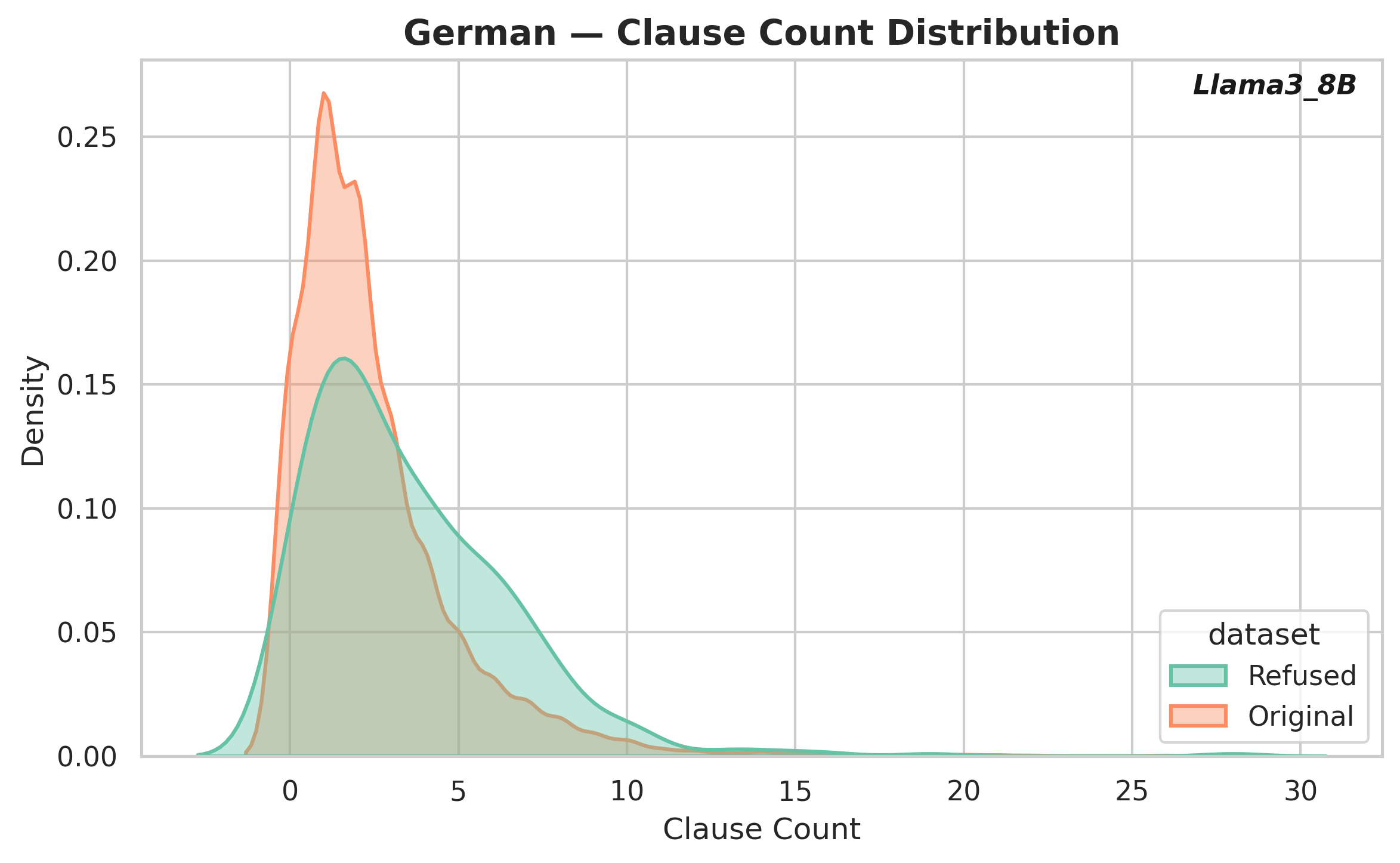}
    \caption{Llama3 8B}
  \end{subfigure}\hfill
  \begin{subfigure}{0.32\textwidth}
    \centering
    \includegraphics[width=\linewidth]{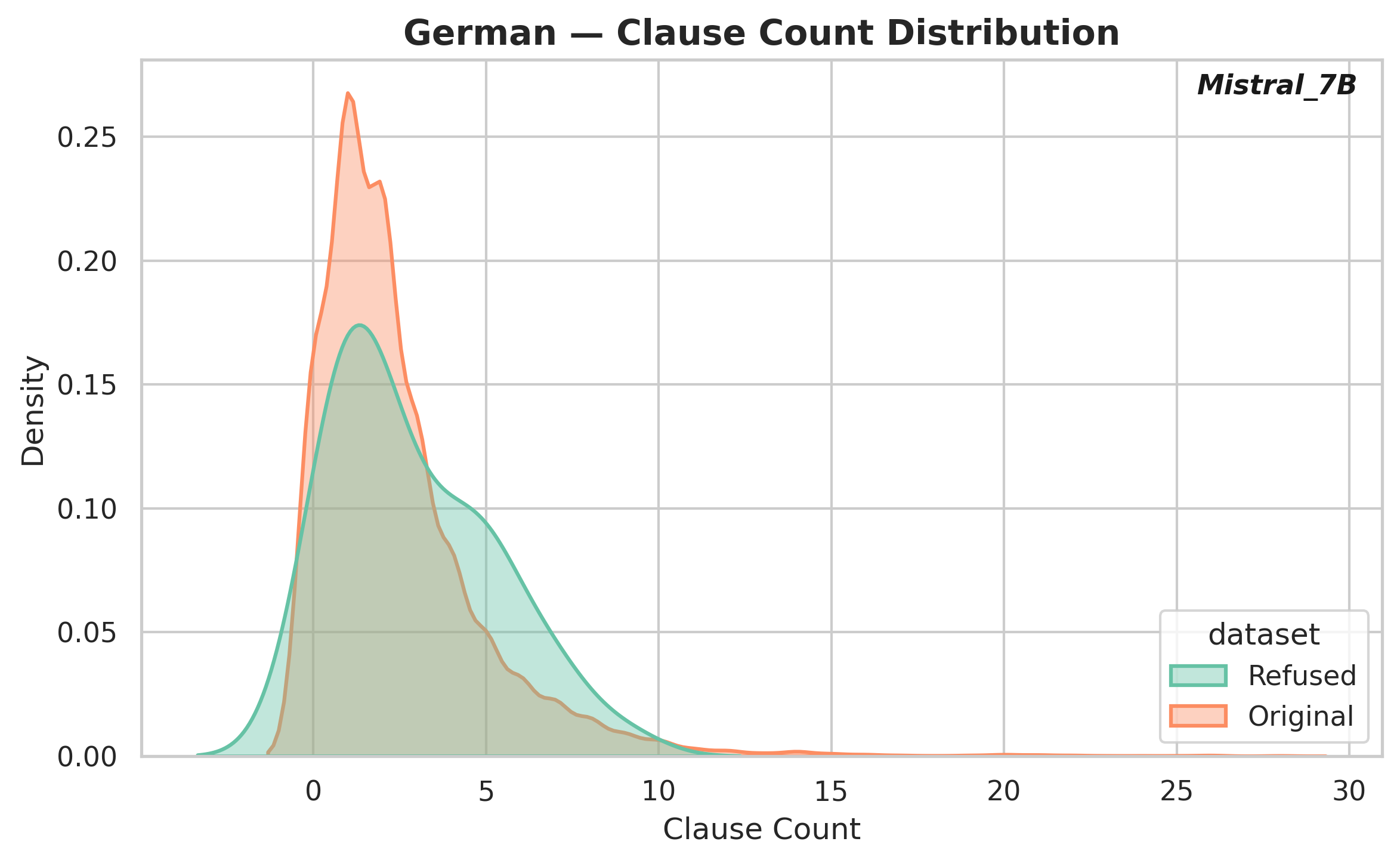}
    \caption{Mistral 7B}
  \end{subfigure}

  \vspace{0.4em}

  \begin{subfigure}{0.32\textwidth}
    \centering
    \includegraphics[width=\linewidth]{figure/Appendix/Linguistic/qwen_7B/German_total_tokens_distribution.png}
    \caption{Qwen2.5 7B}
  \end{subfigure}\hfill
  \begin{subfigure}{0.32\textwidth}
    \centering
    \includegraphics[width=\linewidth]{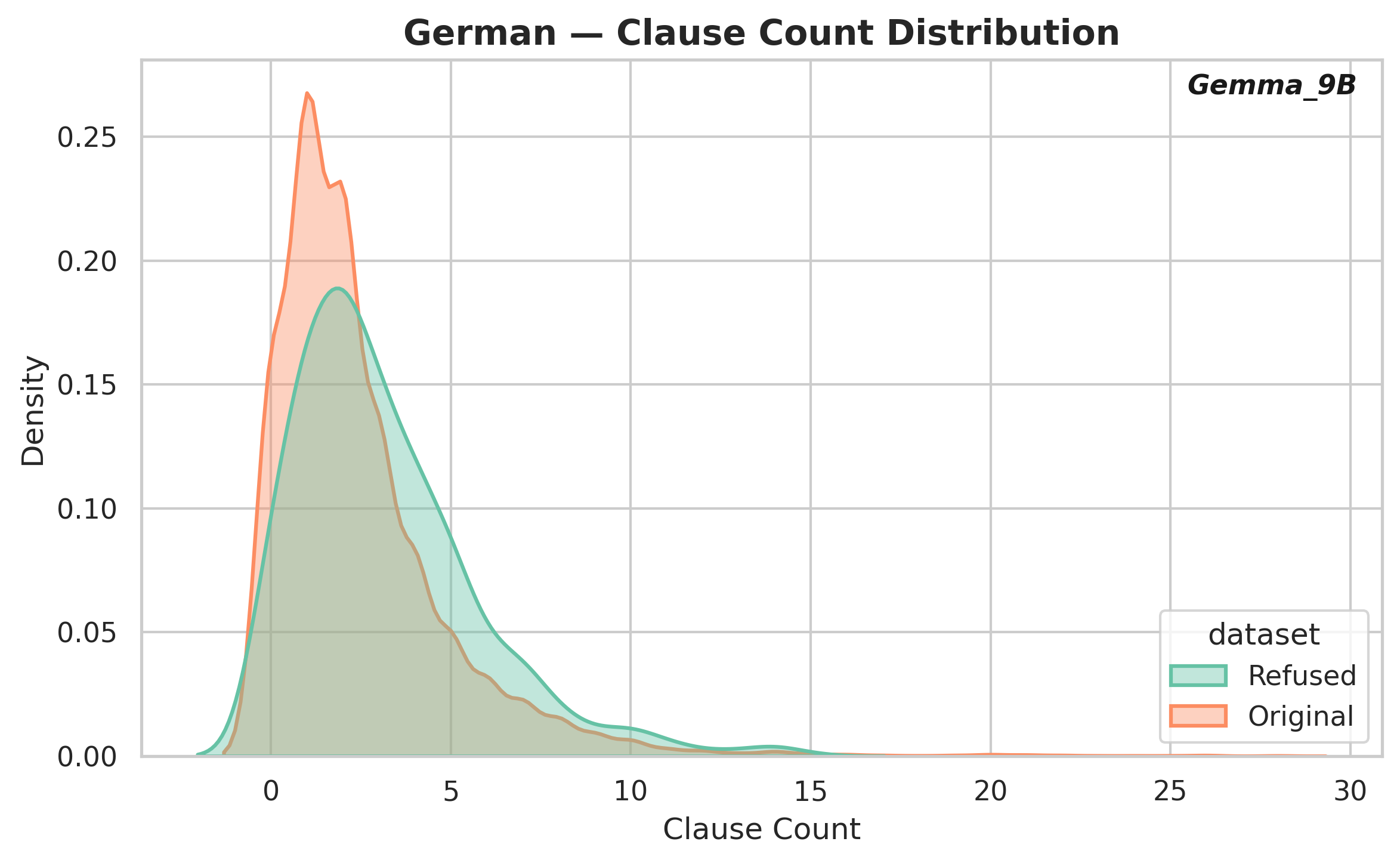}
    \caption{Gemma2 9B}
  \end{subfigure}

  \caption{Clause count distributions for German dataset.}
  \label{fig:clause_german}
\end{figure*}

\begin{figure*}[htbp]
  \centering
  \begin{subfigure}{0.32\textwidth}
    \centering
    \includegraphics[width=\linewidth]{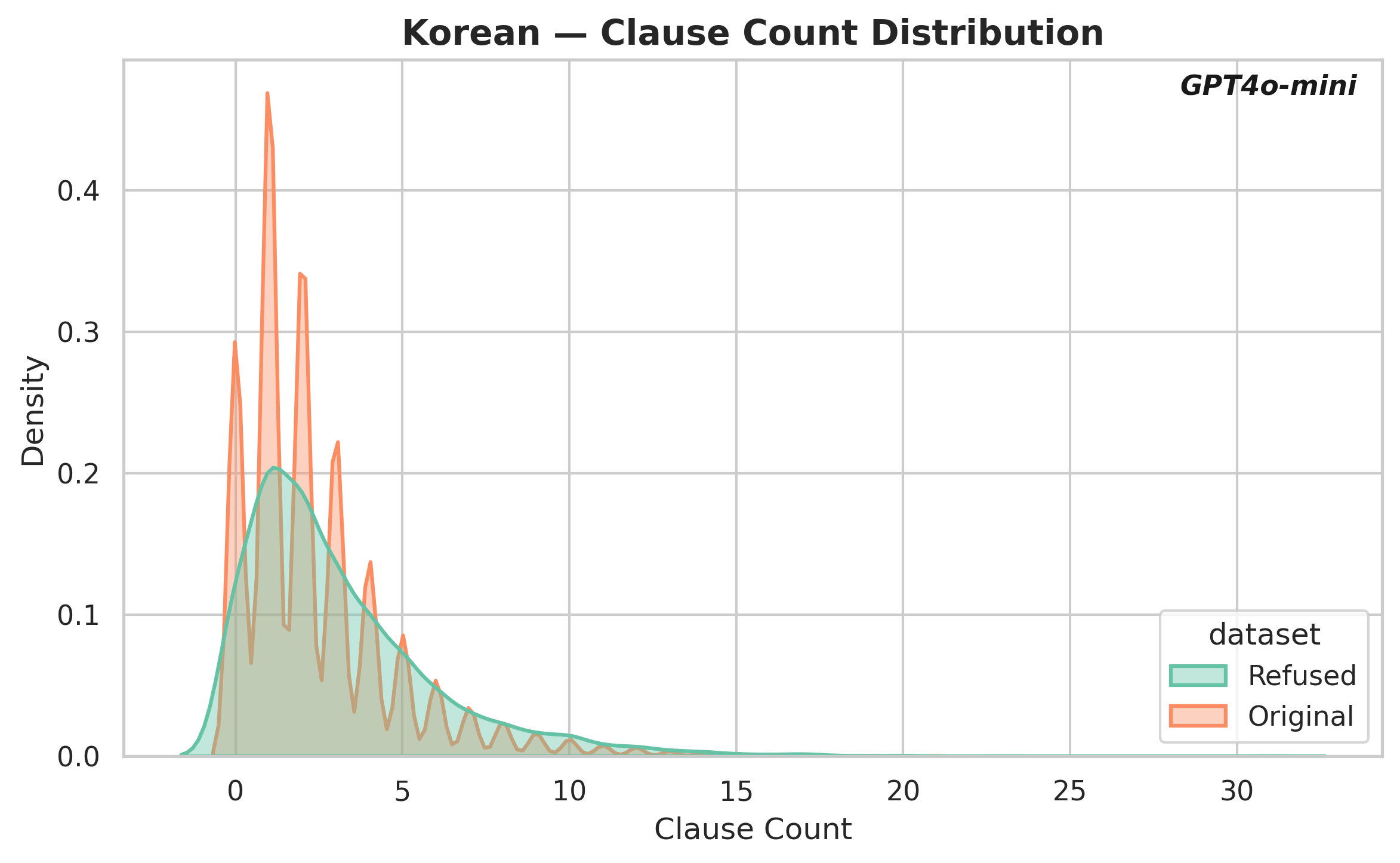}
    \caption{GPT4o-mini}
  \end{subfigure}\hfill
  \begin{subfigure}{0.32\textwidth}
    \centering
    \includegraphics[width=\linewidth]{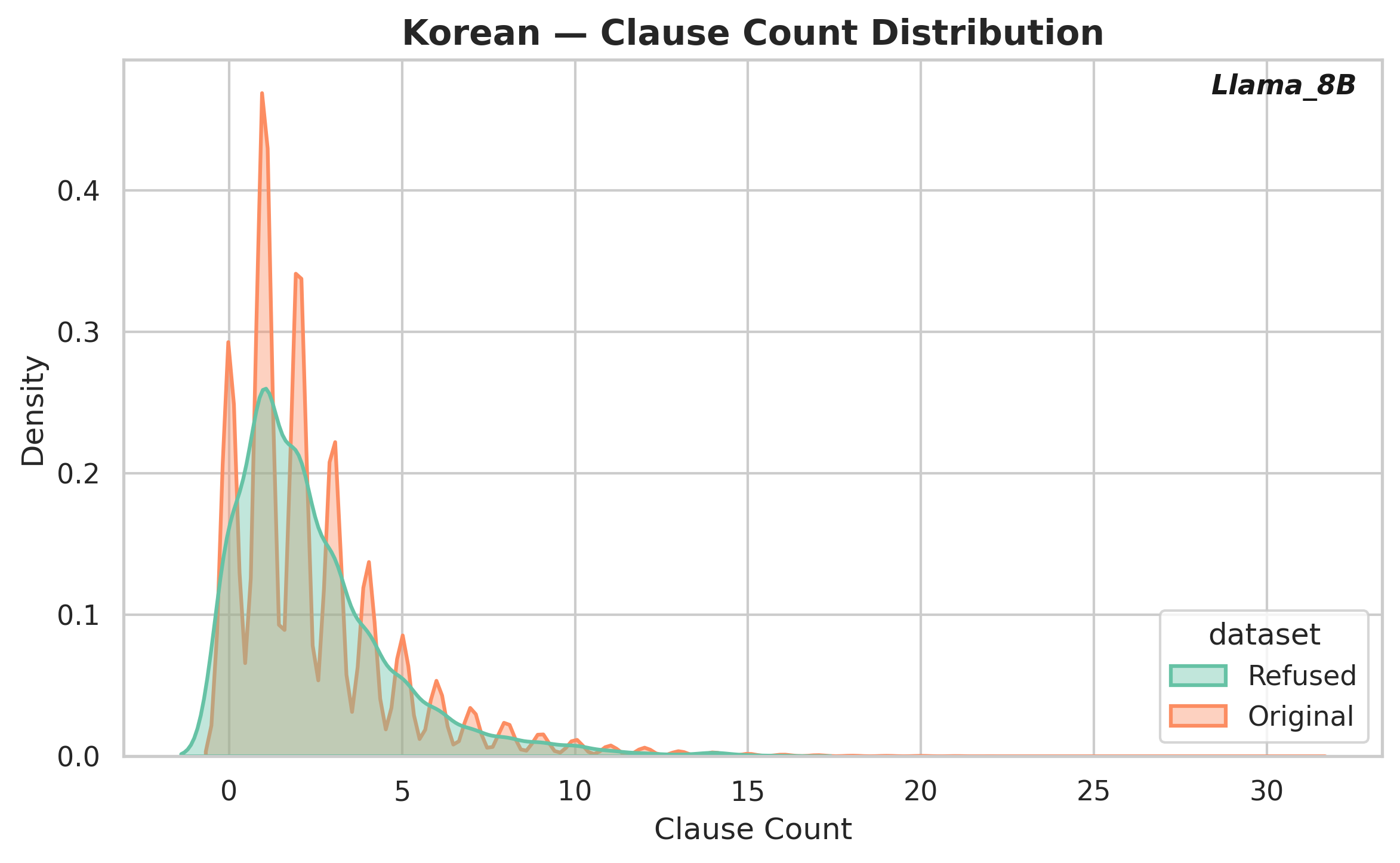}
    \caption{Llama3 8B}
  \end{subfigure}\hfill
  \begin{subfigure}{0.32\textwidth}
    \centering
    \includegraphics[width=\linewidth]{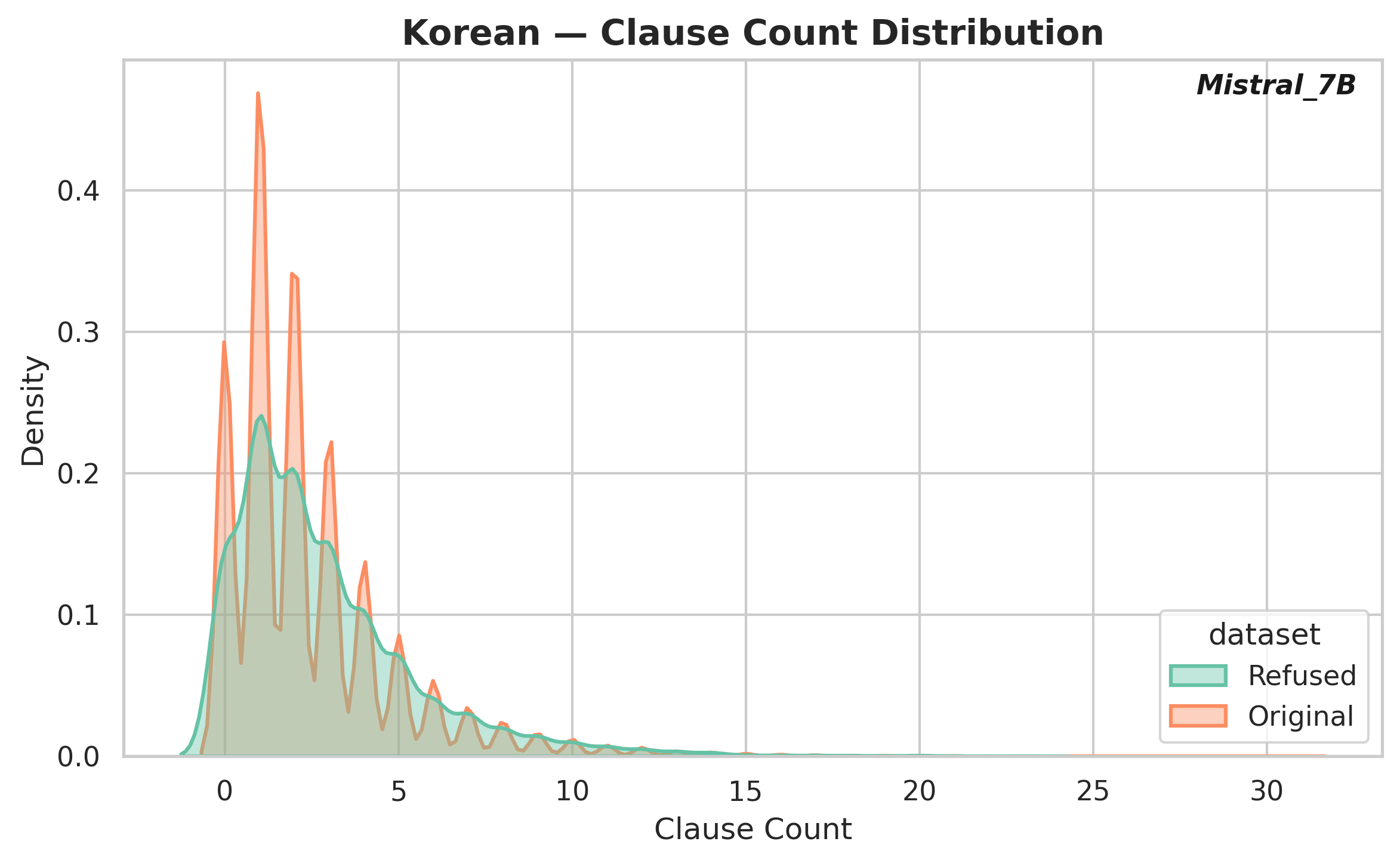}
    \caption{Mistral 7B}
  \end{subfigure}

  \vspace{0.4em}

  \begin{subfigure}{0.32\textwidth}
    \centering
    \includegraphics[width=\linewidth]{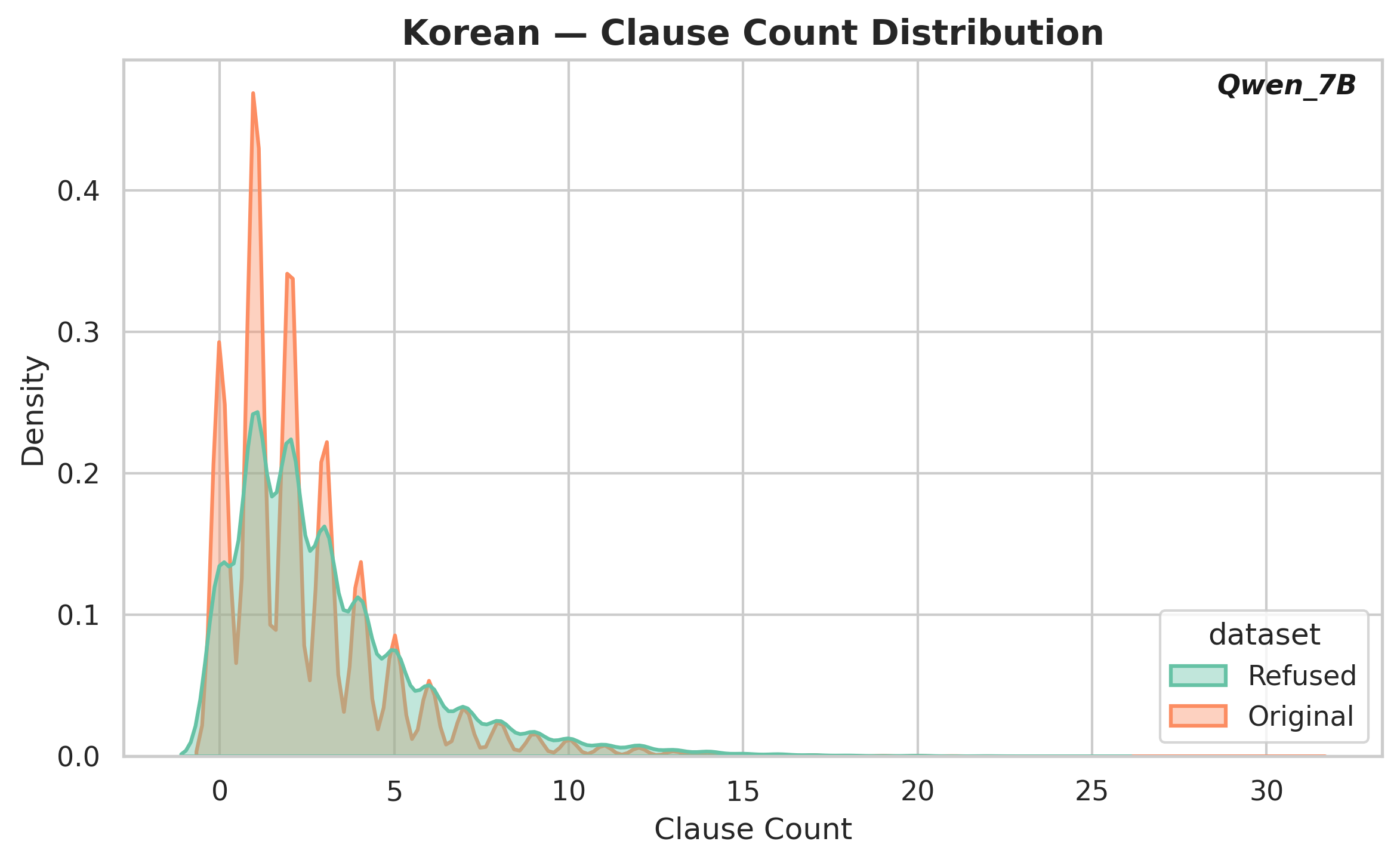}
    \caption{Qwen2.5 7B}
  \end{subfigure}\hfill
  \begin{subfigure}{0.32\textwidth}
    \centering
    \includegraphics[width=\linewidth]{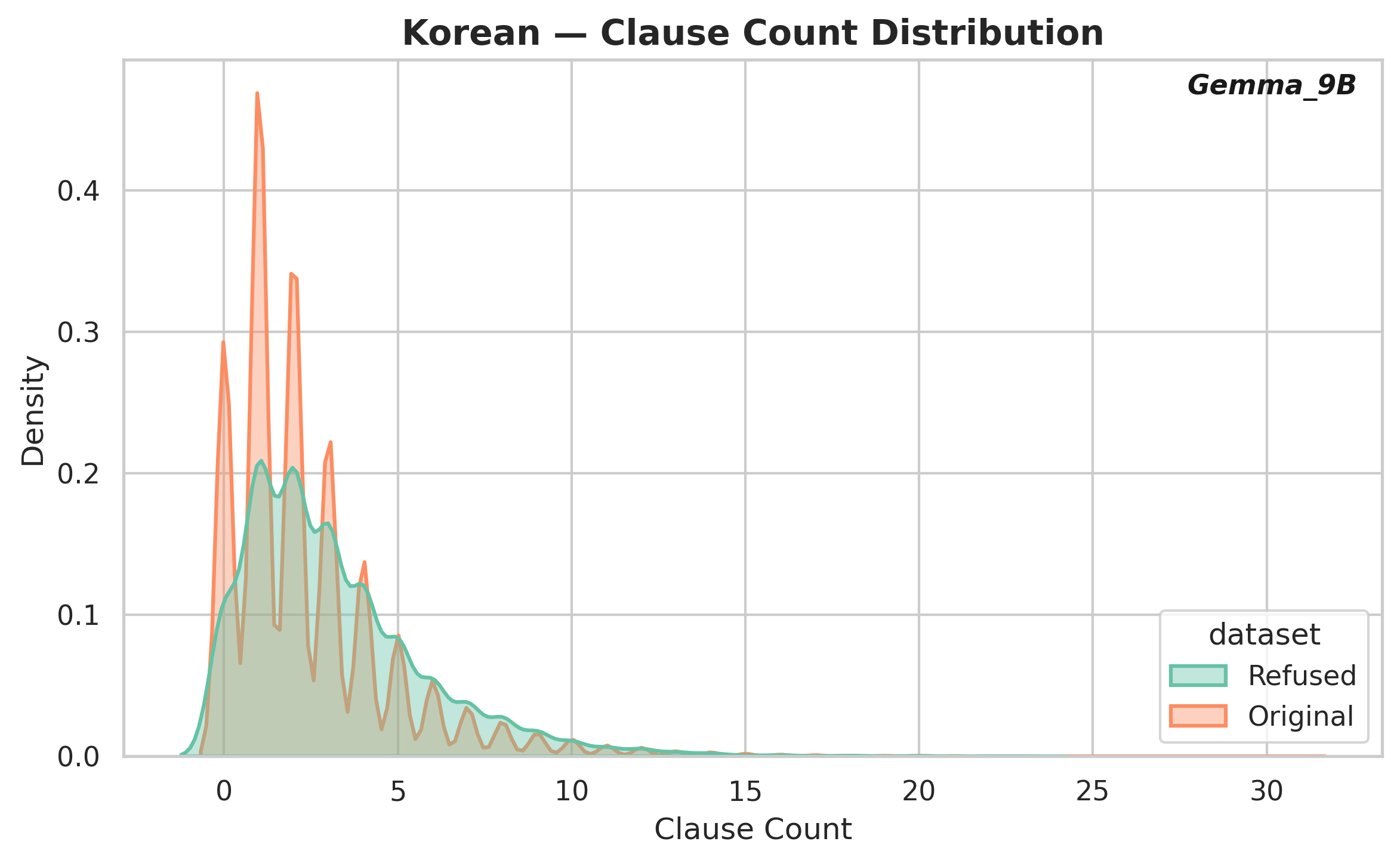}
    \caption{Gemma2 9B}
  \end{subfigure}

  \caption{Clause count distribution of the Korean dataset.}
  \label{fig:clause_korean}
\end{figure*}

\begin{figure*}[htbp]
  \centering
  \setlength{\tabcolsep}{3pt}
  \begin{subfigure}{0.32\textwidth}
    \centering
    \includegraphics[width=\linewidth]{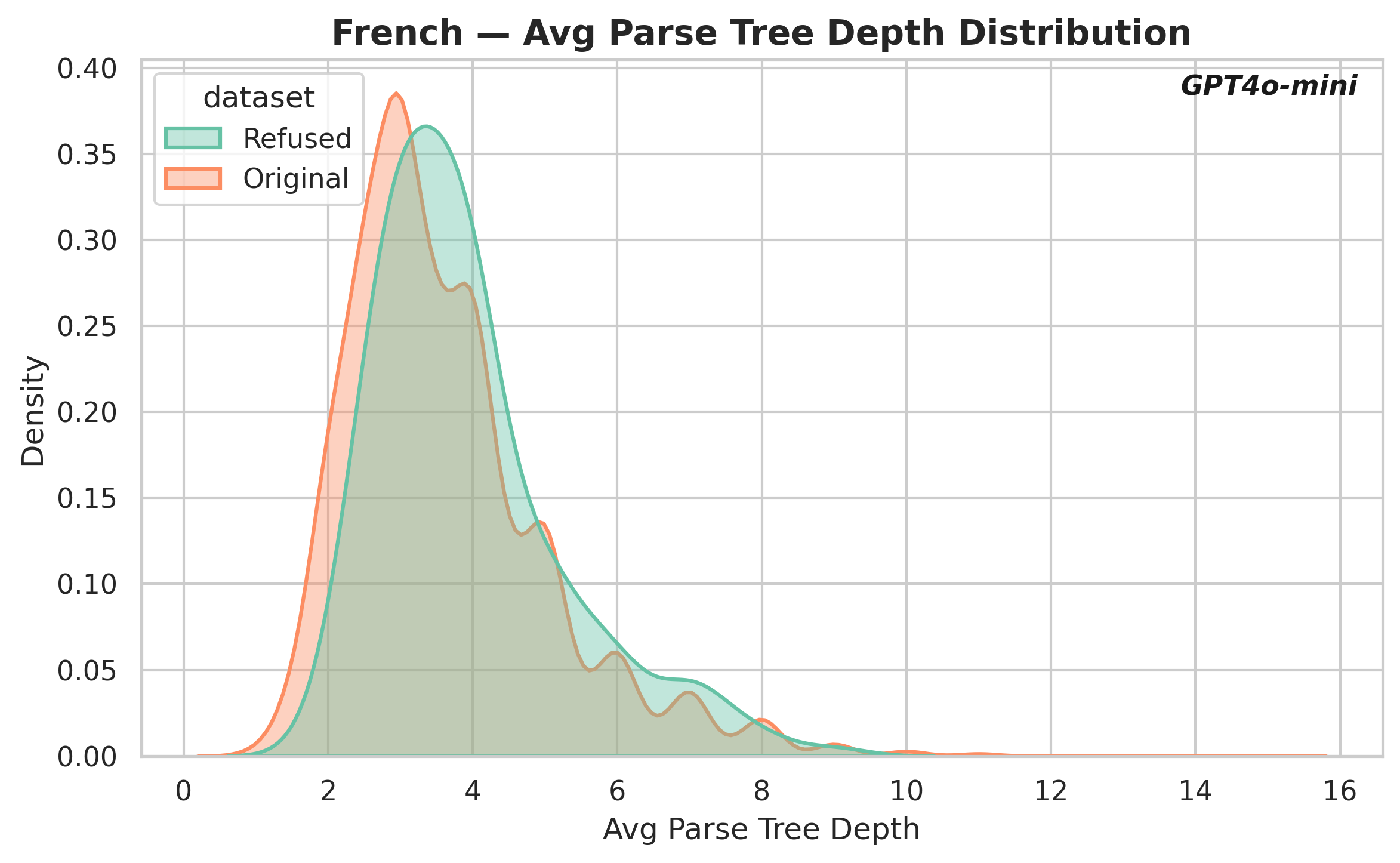}
    \caption{GPT4o-mini}
  \end{subfigure}\hfill
  \begin{subfigure}{0.32\textwidth}
    \centering
    \includegraphics[width=\linewidth]{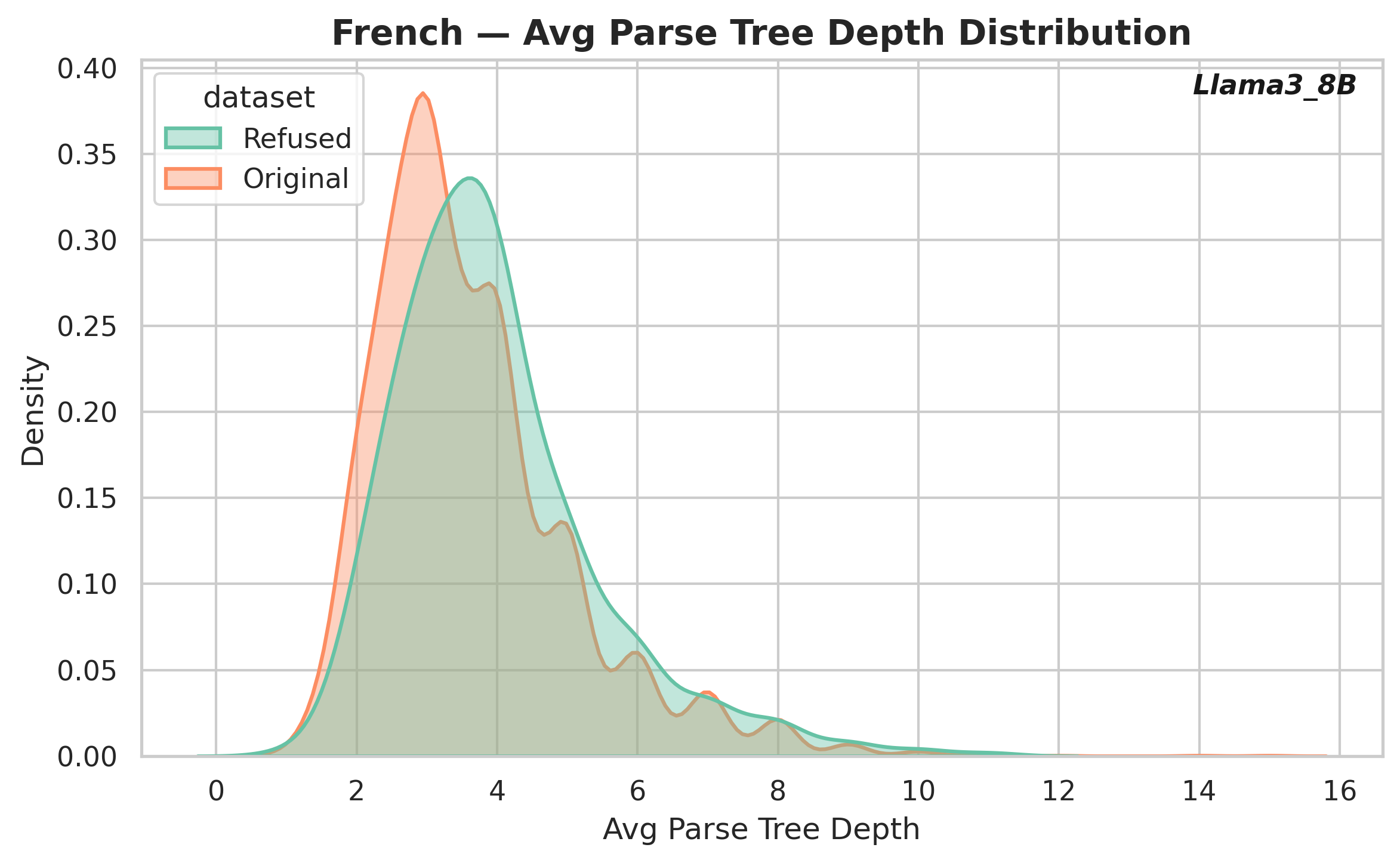}
    \caption{Llama3 8B}
  \end{subfigure}\hfill
  \begin{subfigure}{0.32\textwidth}
    \centering
    \includegraphics[width=\linewidth]{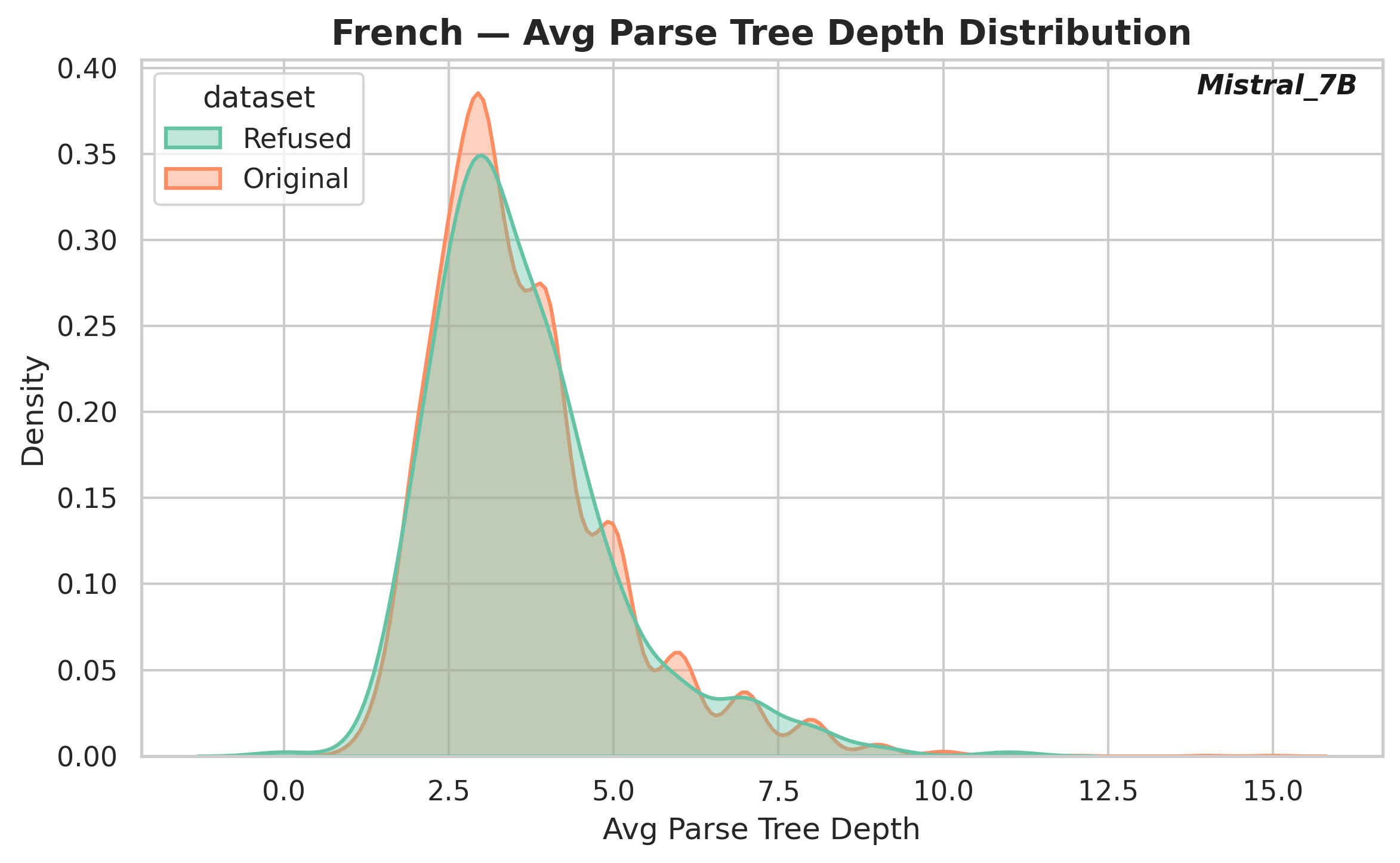}
    \caption{Mistral 7B}
  \end{subfigure}

  \vspace{0.4em}

  \begin{subfigure}{0.32\textwidth}
    \centering
    \includegraphics[width=\linewidth]{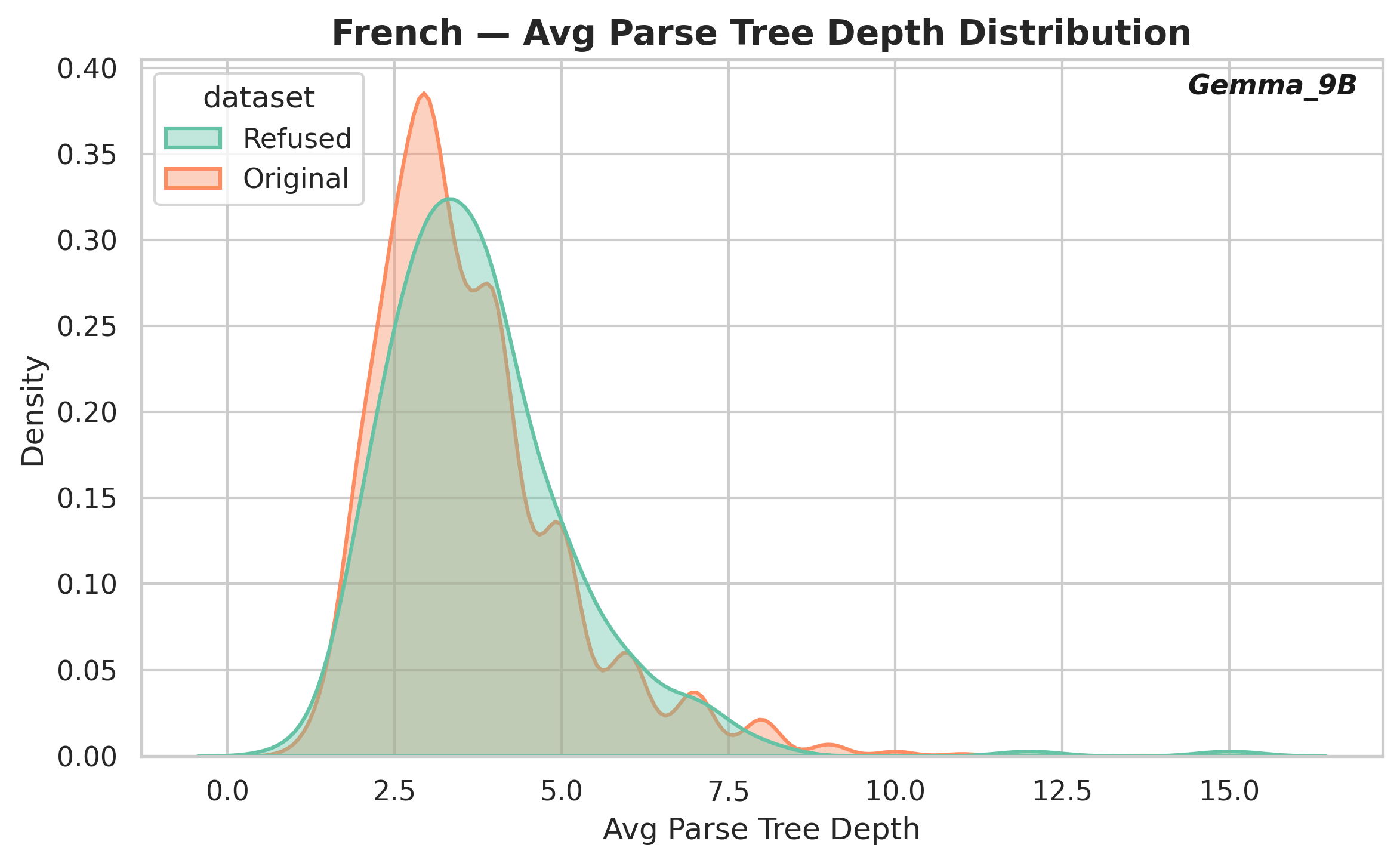}
    \caption{Gemma2 9B}
  \end{subfigure}\hfill
  \begin{subfigure}{0.32\textwidth}
    \centering
    \includegraphics[width=\linewidth]{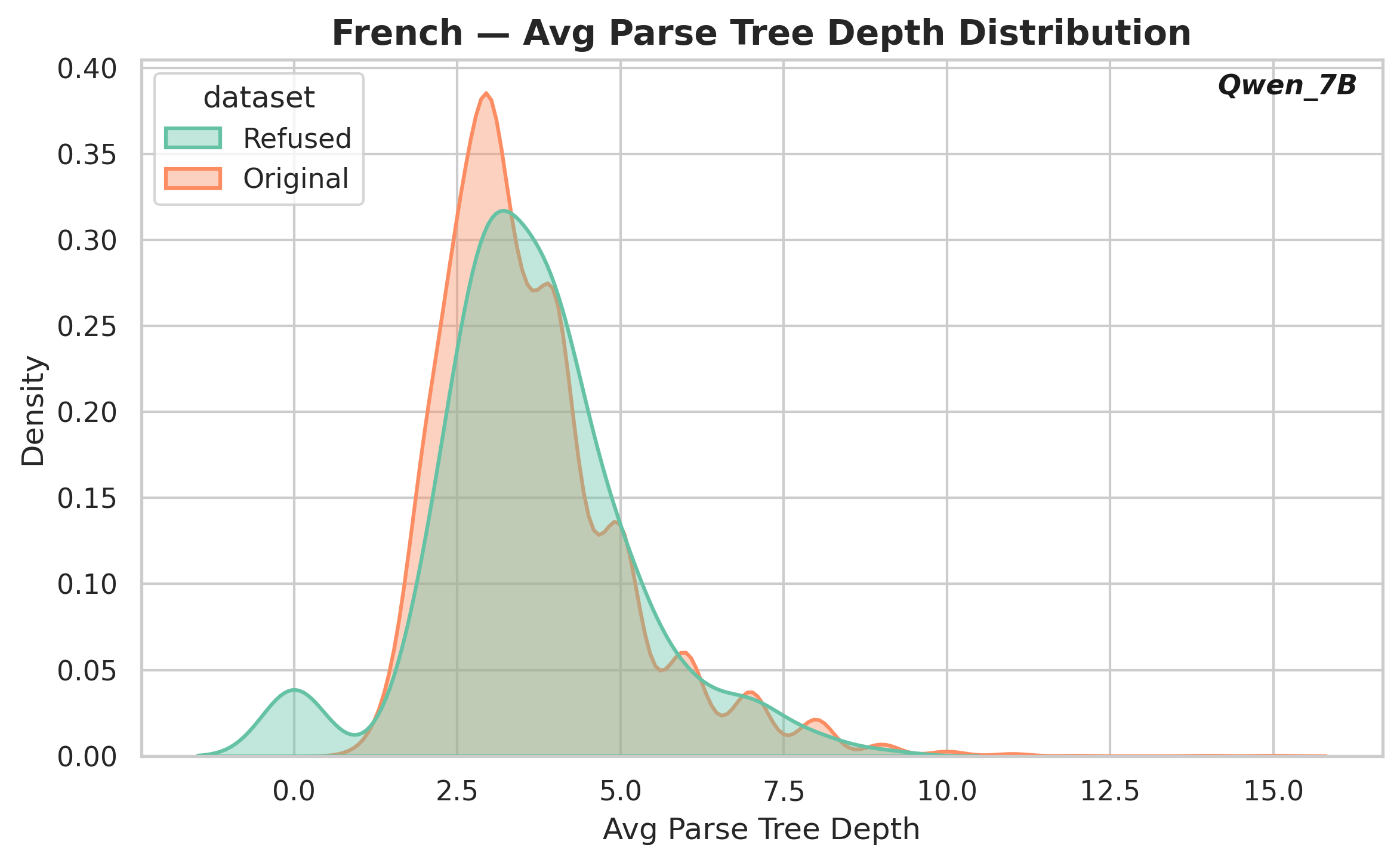}
    \caption{Qwen2.5 7B}
  \end{subfigure}

  \caption{Parse tree depth distributions for French dataset.}
  \label{fig:parse_french}
\end{figure*}

\begin{figure*}[htbp]
  \centering
  \begin{subfigure}{0.32\textwidth}
    \centering
    \includegraphics[width=\linewidth]{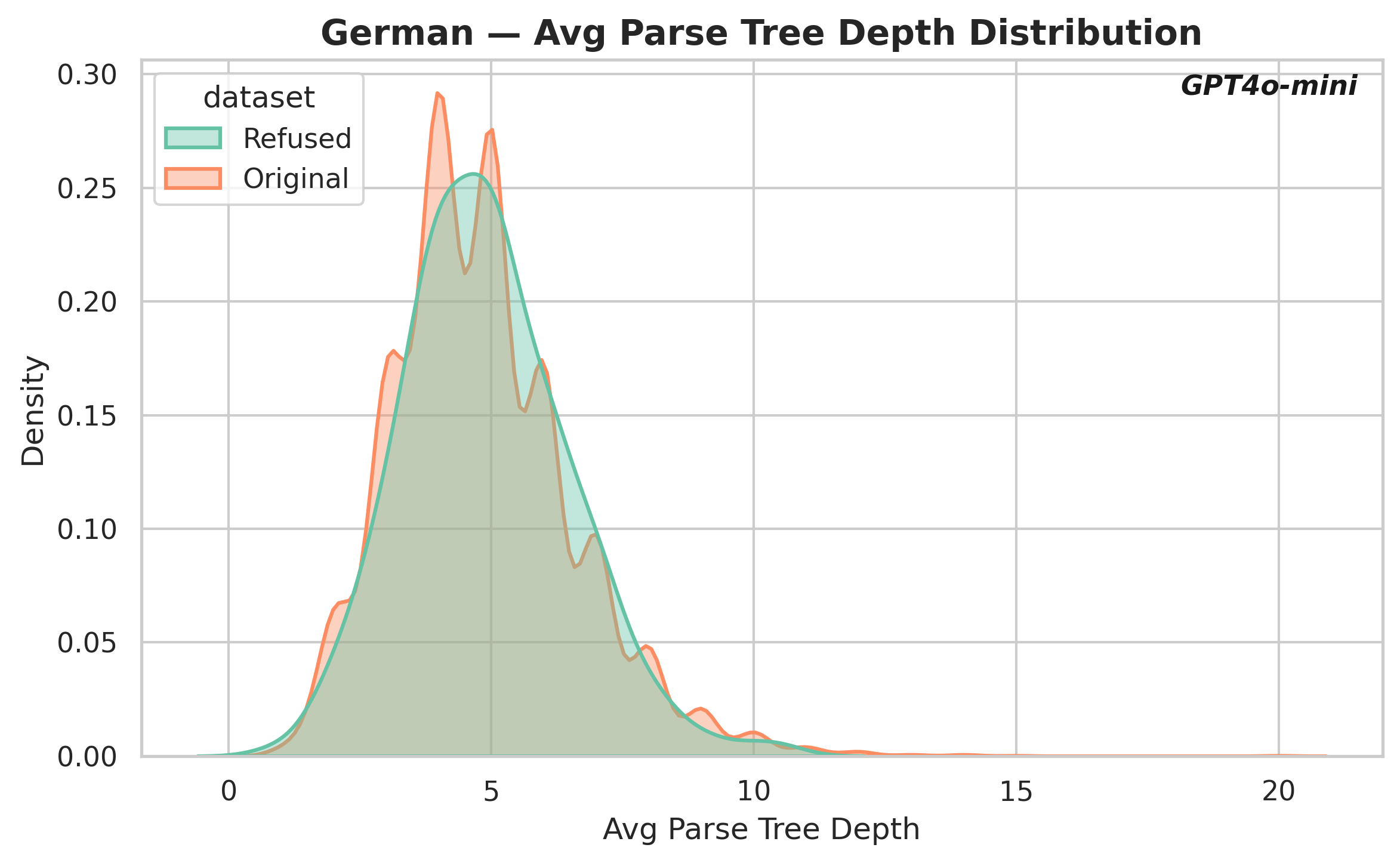}
    \caption{GPT4o-mini}
  \end{subfigure}\hfill
  \begin{subfigure}{0.32\textwidth}
    \centering
    \includegraphics[width=\linewidth]{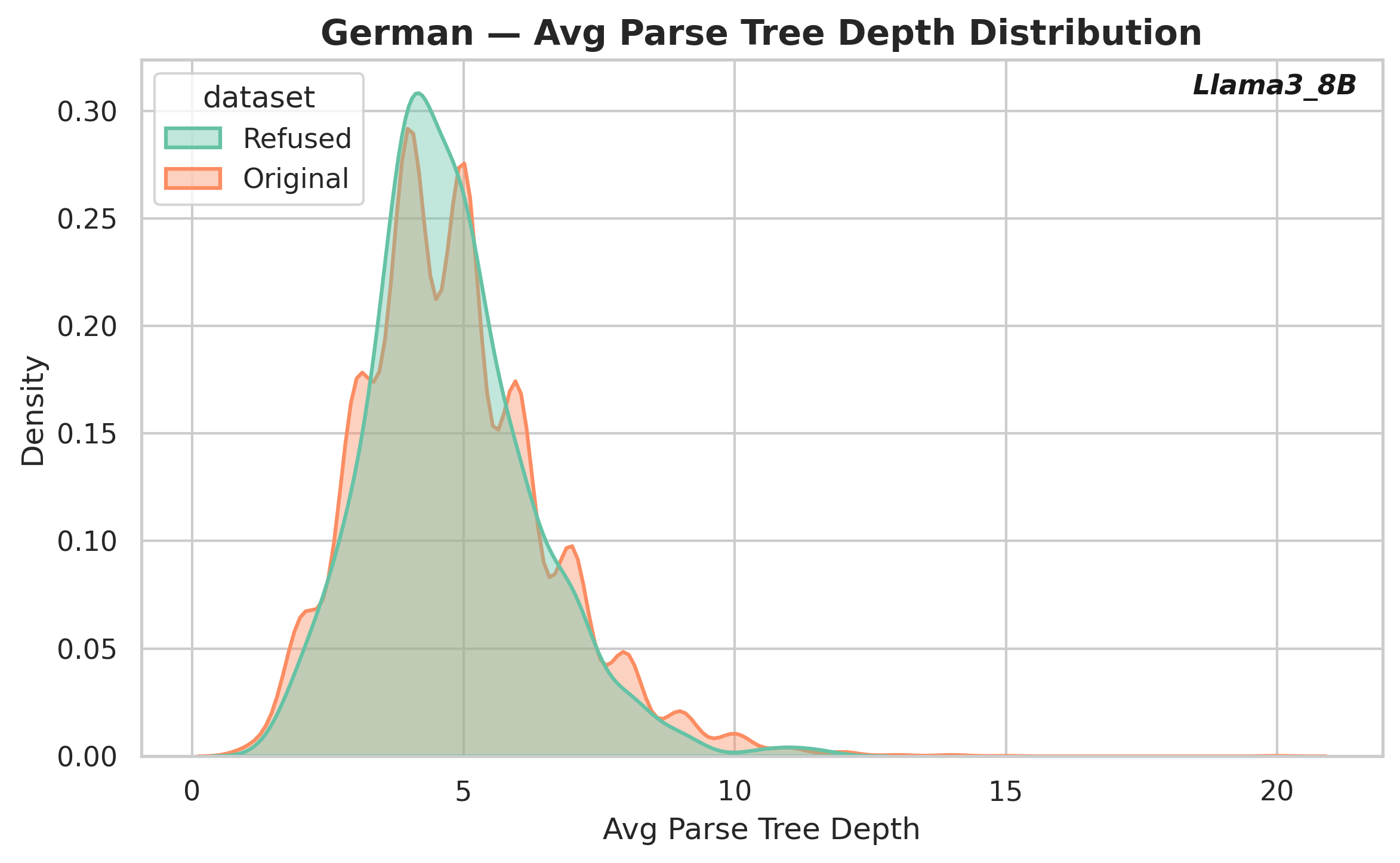}
    \caption{Llama3 8B}
  \end{subfigure}\hfill
  \begin{subfigure}{0.32\textwidth}
    \centering
    \includegraphics[width=\linewidth]{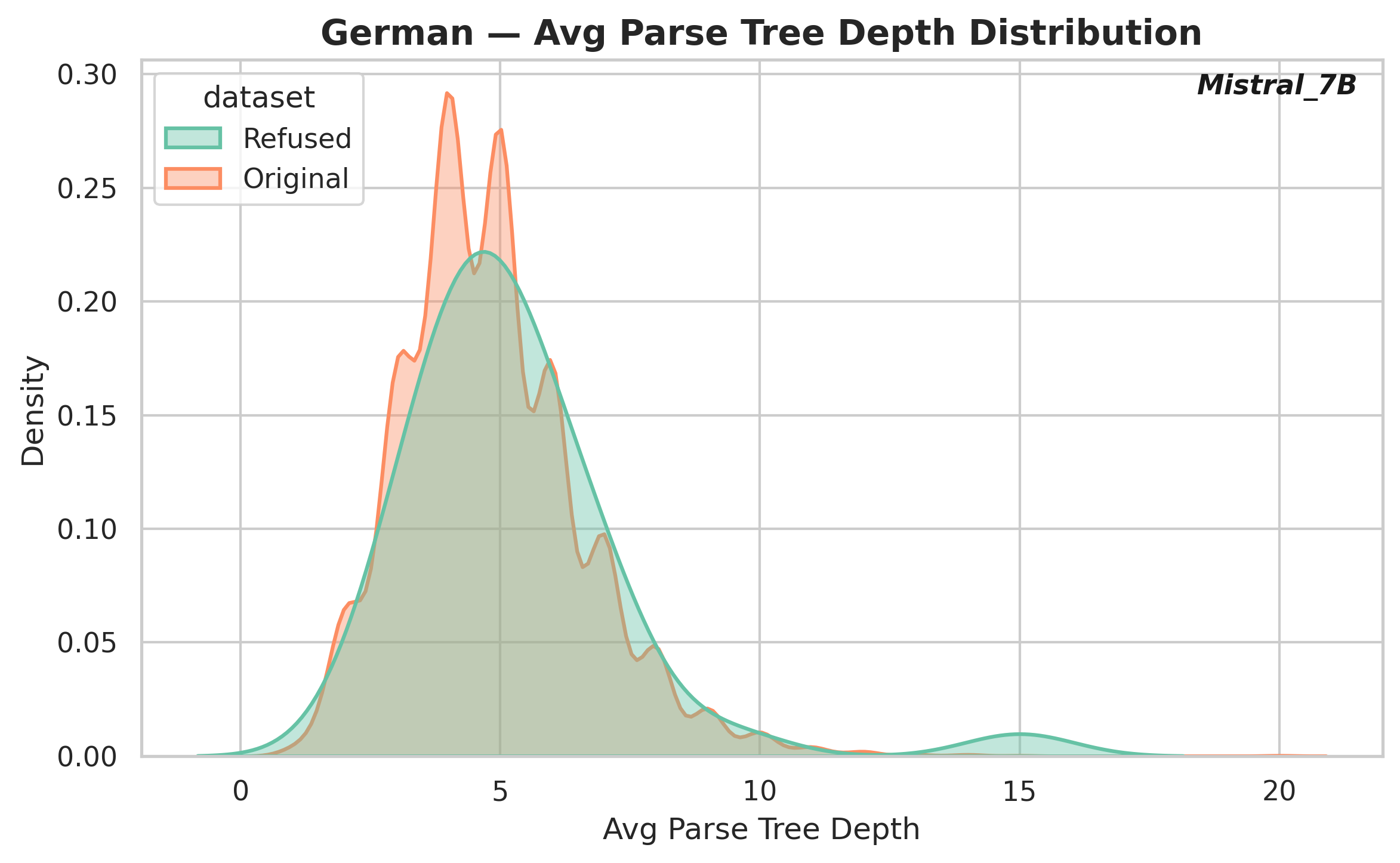}
    \caption{Mistral 7B}
  \end{subfigure}

  \vspace{0.4em}

  \begin{subfigure}{0.32\textwidth}
    \centering
    \includegraphics[width=\linewidth]{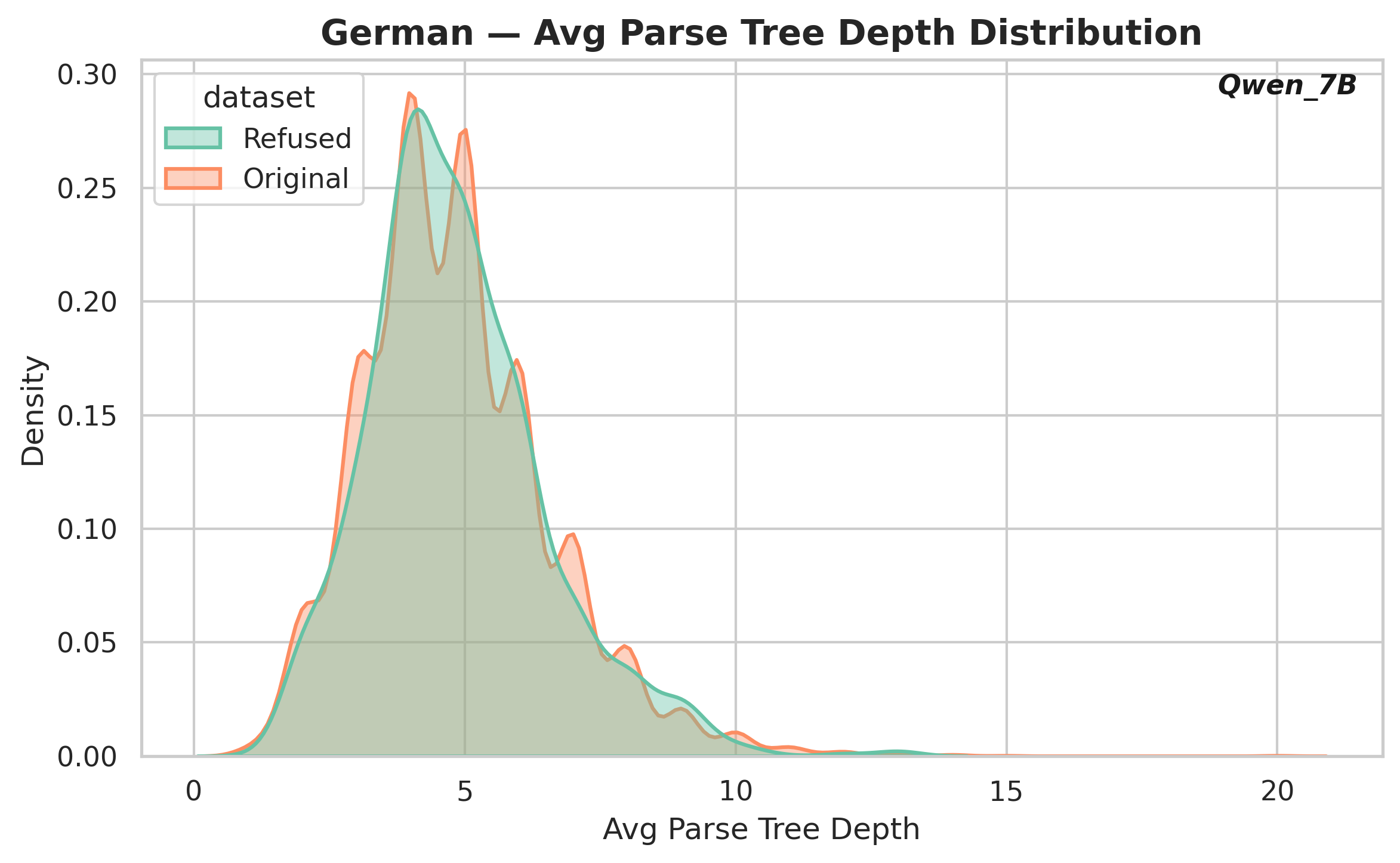}
    \caption{Qwen2.5 7B}
  \end{subfigure}\hfill
  \begin{subfigure}{0.32\textwidth}
    \centering
    \includegraphics[width=\linewidth]{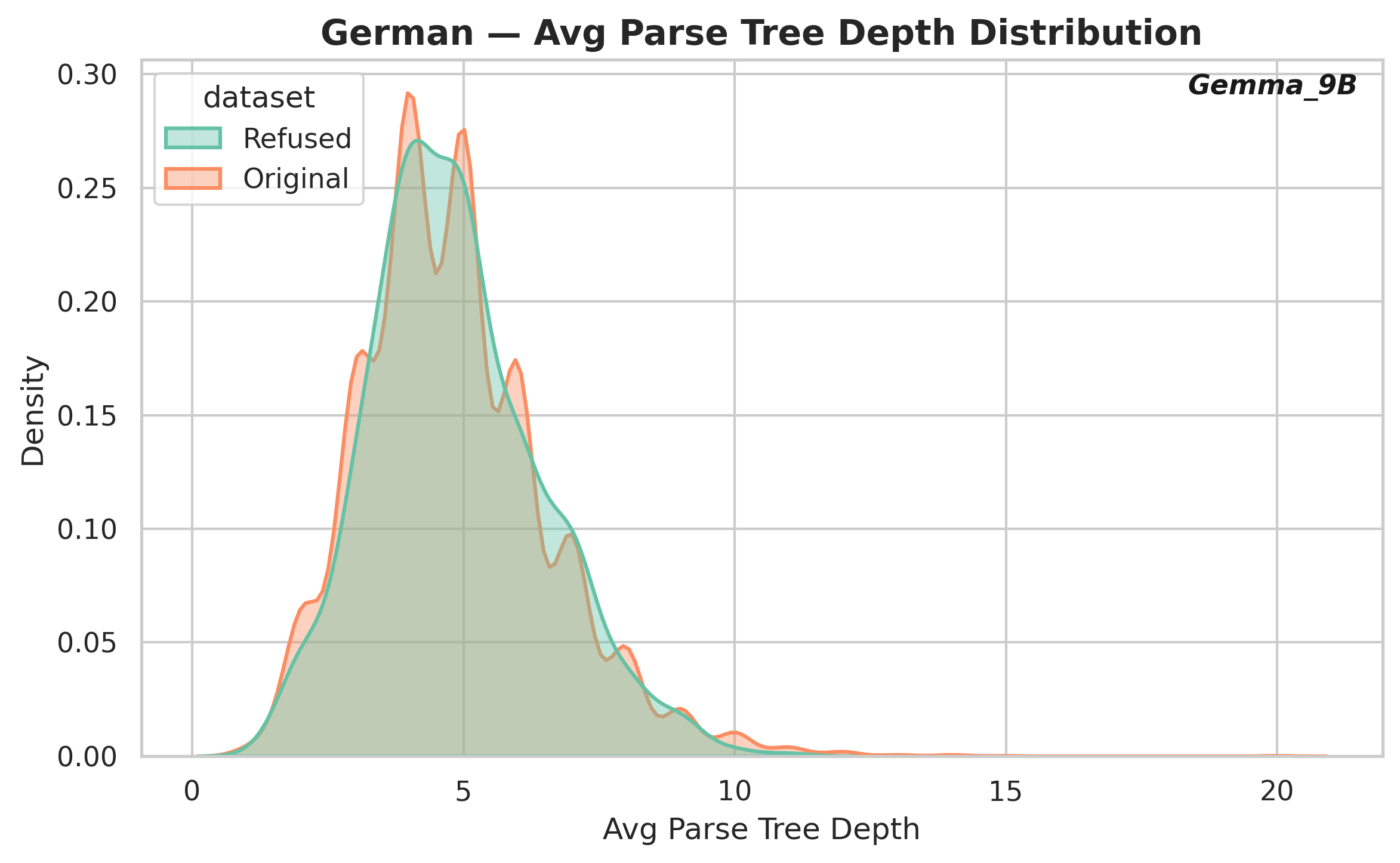}
    \caption{Gemma2 9B}
  \end{subfigure}

  \caption{Parse tree depth distributions for German dataset.}
  \label{fig:parse_german}
\end{figure*}

\begin{figure*}[htbp]
  \centering
  \begin{subfigure}{0.32\textwidth}
    \centering
    \includegraphics[width=\linewidth]{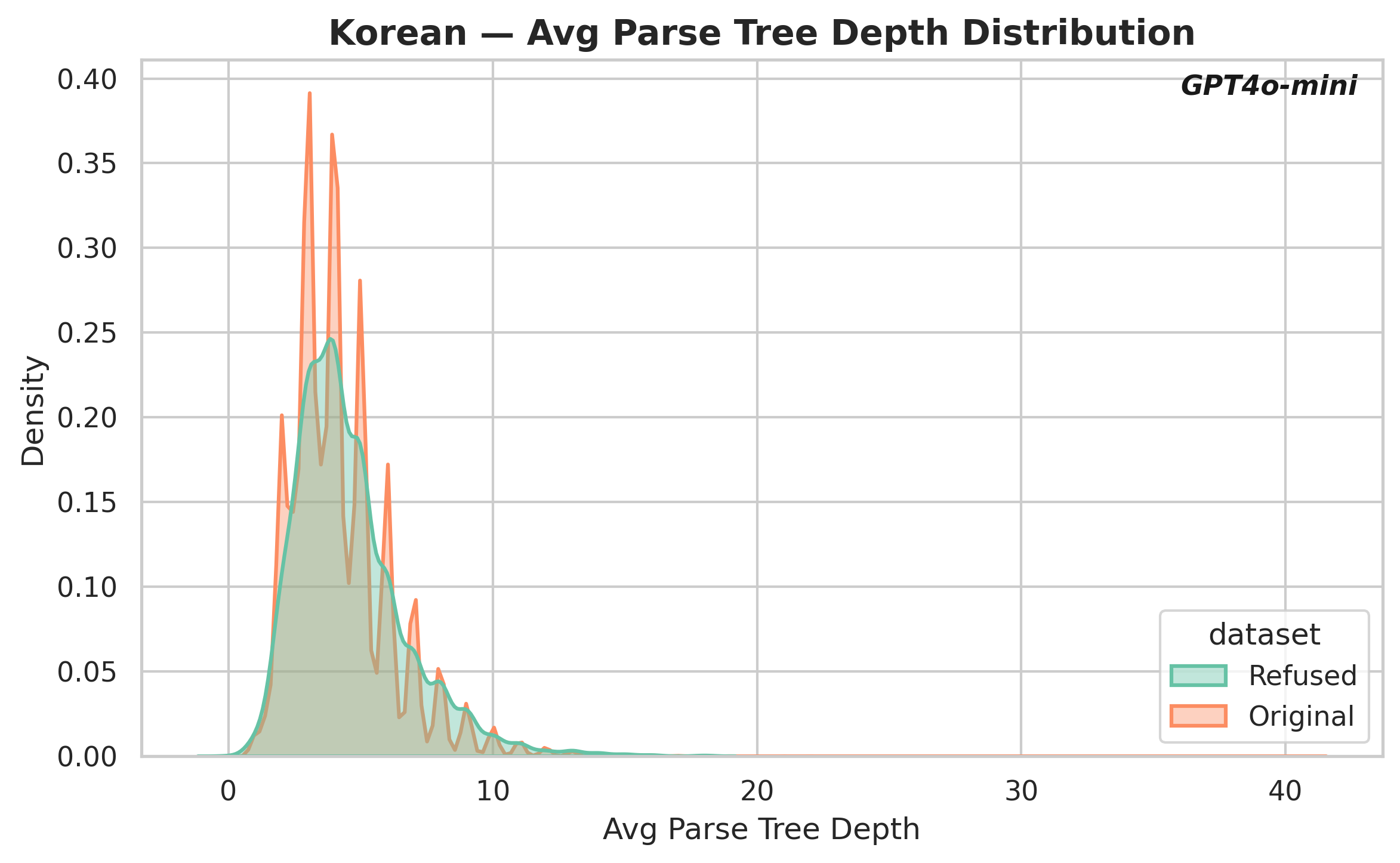}
    \caption{GPT4o-mini}
  \end{subfigure}\hfill
  \begin{subfigure}{0.32\textwidth}
    \centering
    \includegraphics[width=\linewidth]{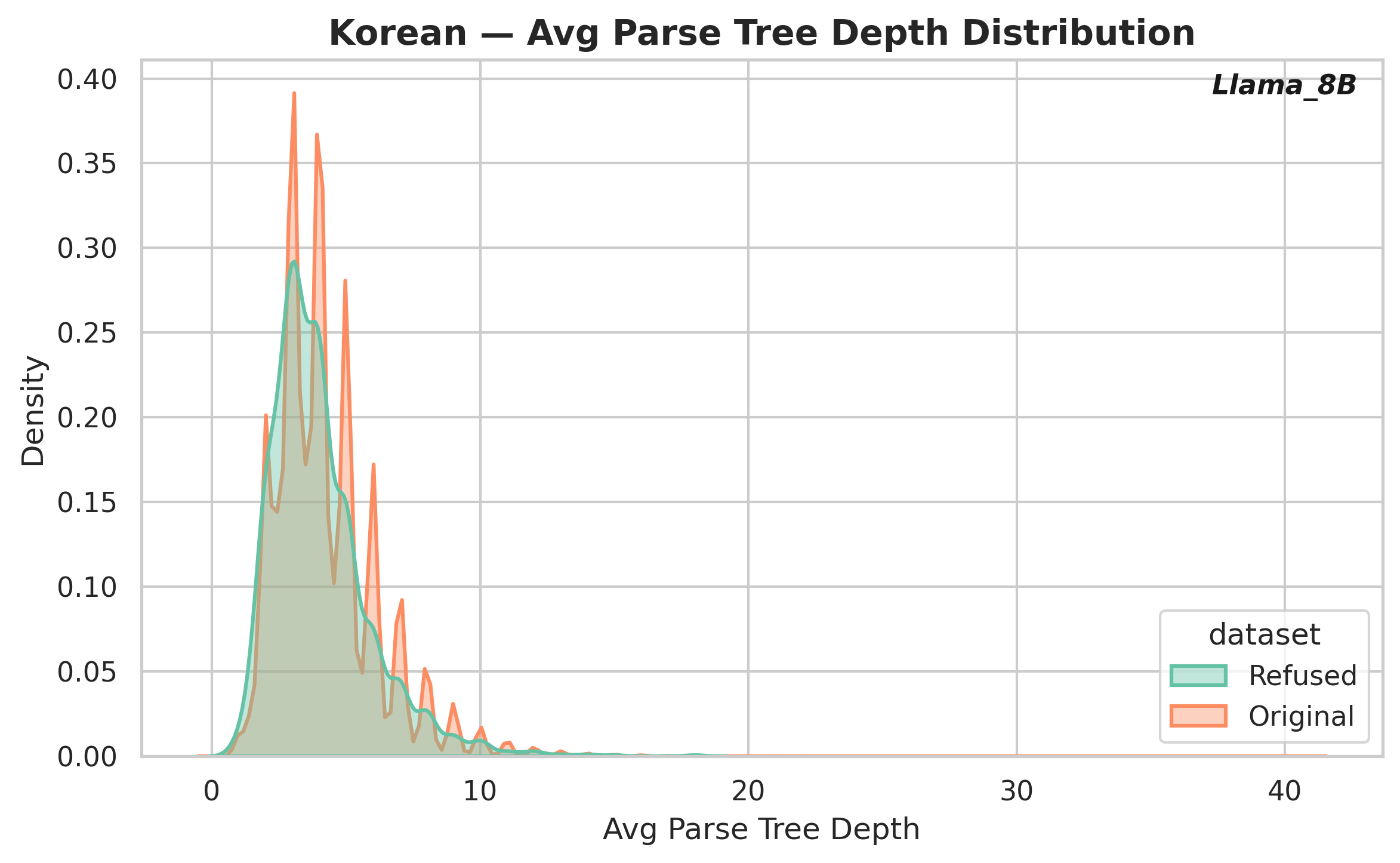}
    \caption{Llama3 8B}
  \end{subfigure}\hfill
  \begin{subfigure}{0.32\textwidth}
    \centering
    \includegraphics[width=\linewidth]{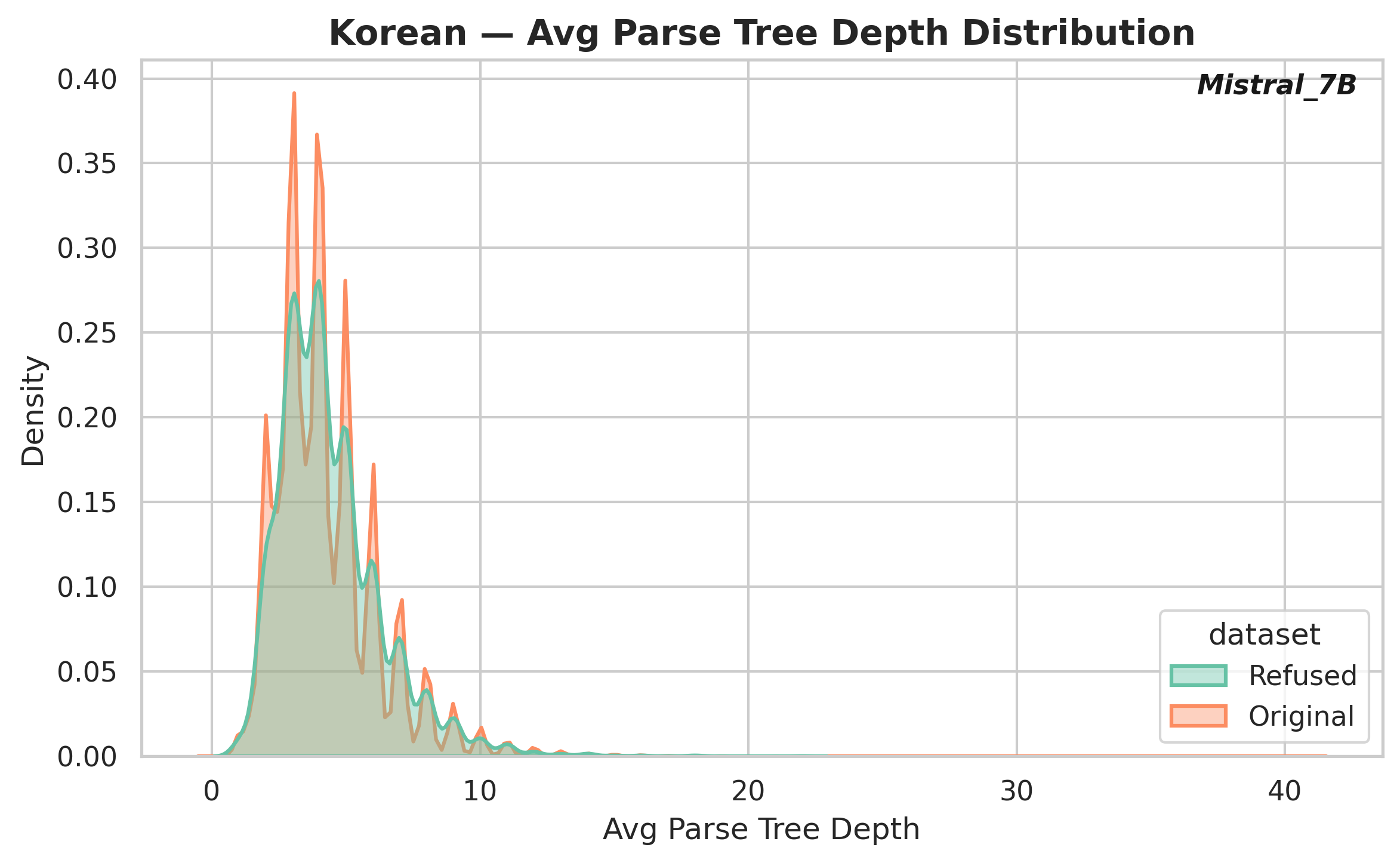}
    \caption{Mistral 7B}
  \end{subfigure}

  \vspace{0.4em}

  \begin{subfigure}{0.32\textwidth}
    \centering
    \includegraphics[width=\linewidth]{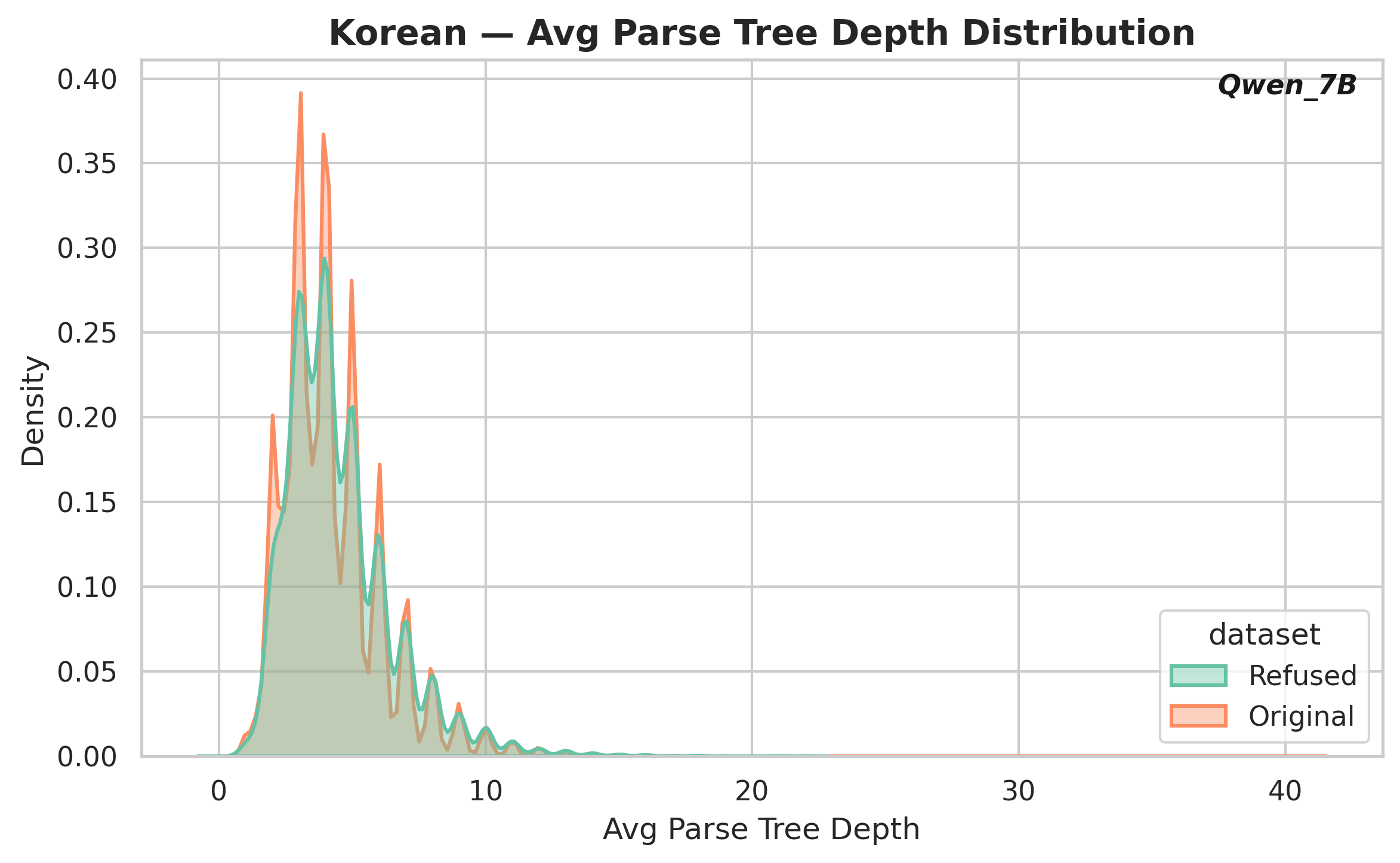}
    \caption{Qwen2.5 7B}
  \end{subfigure}\hfill
  \begin{subfigure}{0.32\textwidth}
    \centering
    \includegraphics[width=\linewidth]{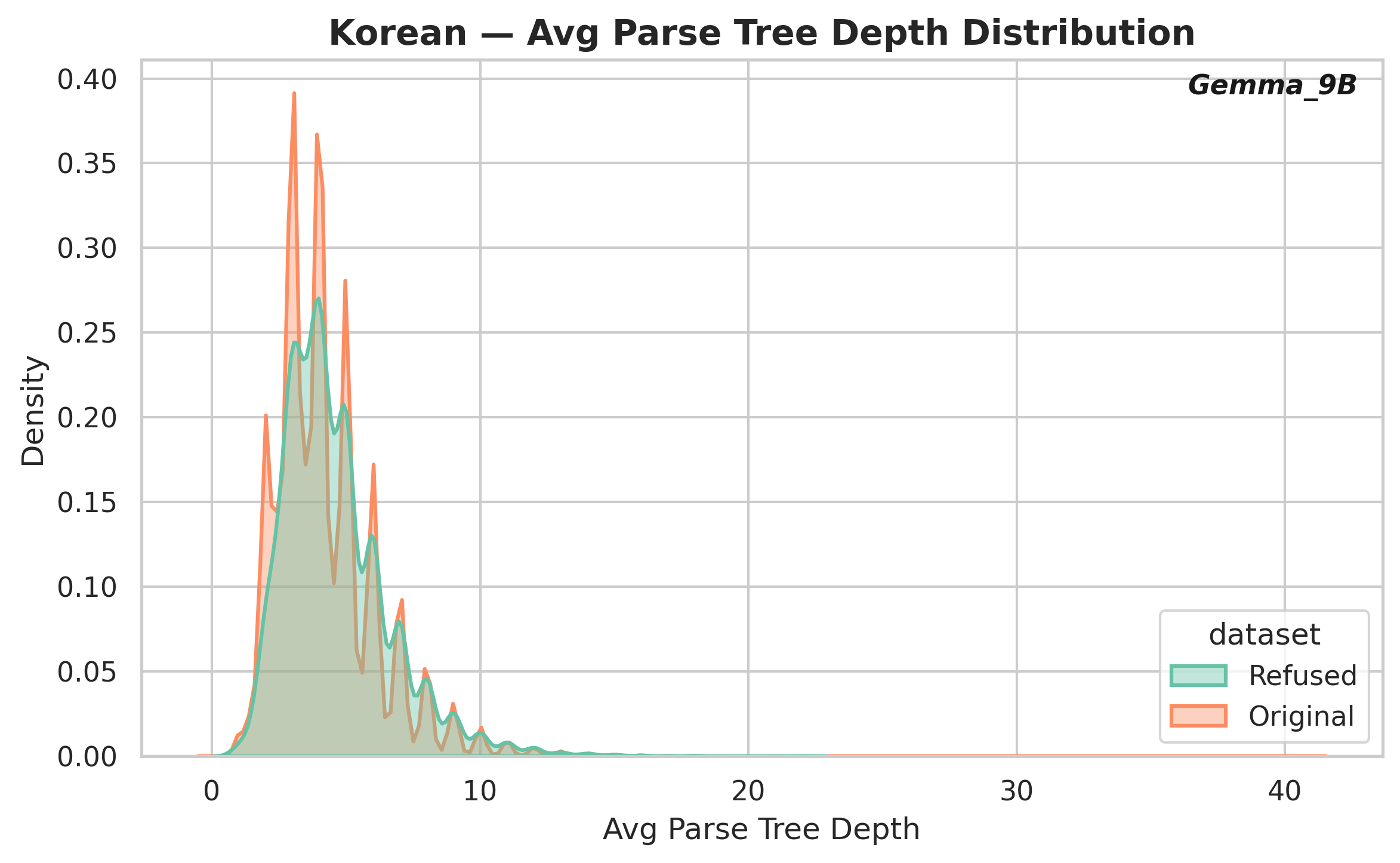}
    \caption{Gemma2 9B}
  \end{subfigure}

  \caption{Parse tree depth distribution of the Korean dataset.}
  \label{fig:parse_korean}
\end{figure*}

Figure~\ref{fig:mult_clause} presents the clause-count distributions for Spanish and Chinese across all evaluated models. Corresponding distributions for French, German, and Korean are shown in Figures~\ref{fig:clause_french}, \ref{fig:clause_german}, and \ref{fig:clause_korean}, respectively.  
Parse tree depth distributions are shown in Figures~\ref{fig:parse_french}, \ref{fig:parse_german}, and \ref{fig:parse_korean}.

Similar to the token-length patterns, the distributions of falsely refused samples largely overlap with those of the original datasets across all languages. Only minor shifts toward simpler structures are observed in a few models. This indicates that clause-level and parse-tree syntactic complexity do not meaningfully contribute to false refusal behavior in multilingual hate speech detoxification.  

Despite substantial differences in overall refusal rates between Spanish and Chinese, the comparable distributions of clause counts and parse-tree depths suggest that refusals are primarily driven by semantic and contextual factors, rather than by structural complexity.

\end{document}